\documentclass[10pt,twocolumn,letterpaper]{article}
\usepackage{iftex}
\usepackage[T1]{fontenc}
\DeclareTextSymbolDefault{\textquotedbl}{T1}

\usepackage[pagenumbers]{cvpr}

\usepackage{cuted} %

\usepackage{currfile} %

\usepackage{caption} %
\newcommand{\TODO}[1]{\textbf{\color{red}[TODO: #1]}}

\usepackage{xspace}
\usepackage{booktabs}
\usepackage{multirow}
\usepackage{amsmath,amssymb}
\usepackage{enumitem}
\newcommand{\bench}{EgoSPARK\xspace}                 %
\newcommand{\method}{SPARK-G\xspace}

\usepackage{pifont}
\newcommand{\cmark}{\ding{51}}\newcommand{\xmark}{\ding{55}}
\DeclareRobustCommand{\hmark}{\tikz[baseline=-0.55ex]{\fill (0,0) -- (90:0.62ex) arc (90:270:0.62ex) -- cycle; \draw[line width=0.4pt] (0,0) circle (0.62ex);}}  %

\usepackage{tikz}
\usetikzlibrary{positioning,arrows.meta,fit,calc,shapes.misc}
\newcommand{\Lone}{\text{Loc-F1}\xspace}\newcommand{\Ltwo}{\text{Role-F1}\xspace}\newcommand{\Lthree}{\text{Int-F1}\xspace}\newcommand{\depacc}{\text{Dep-Acc}$_{\text{found}}$\xspace}%

\providecommand{\nolinenumbers}{}\providecommand{\linenumbers}{}

\usepackage{placeins}
\usepackage{array}
\usepackage{listings}
\definecolor{cvprblue}{rgb}{0.21,0.49,0.74}
\usepackage[pagebackref,breaklinks,colorlinks,allcolors=cvprblue]{hyperref}

\title{EgoSPARK: Do Video Models Find the Right Evidence\\for Egocentric Causal Question Answering?}

\author{
Junghye Kim$^{1}$ \quad Sehun Lee$^{2}$ \quad Soohyun Oh$^{1}$ \quad Yoonjin Oh$^{1}$\\
Eunseo Lee$^{2}$ \quad Sungbin Lim$^{2,\dagger}$ \quad Sungwoong Kim$^{1,\dagger}$\\[4pt]
$^{1}$Department of Artificial Intelligence, Korea University\\
$^{2}$Department of Statistics, Korea University\\[2pt]
{\tt\small \{jh0912, sungbin, swkim01\}@korea.ac.kr}\\[2pt]
{\small $^{\dagger}$Corresponding authors}
}

\newif\ifauxmethod\auxmethodfalse

\begin{document}
\maketitle
\def\TabMainRows{%
\multicolumn{19}{@{}l}{\textit{Open-source MLLMs}} \\
Qwen3.5-2B & 4 & .117 & .145 & .208 & .489 & .099 & .003 & .000 & .000 & .072 & .012 & .57 & .027 & .001 & .001 & .77 & .99 & .99 \\   %
Qwen3.5-9B & 4 & .160 & .211 & .297 & .568 & .232 & .008 & .005 & .005 & .065 & .018 & .51 & .052 & .002 & .002 & .67 & .99 & .99 \\   %
Qwen3.6-27B & 4 & .221 & .198 & .234 & \textbf{.730} & .342 & .075 & .061 & .059 & \underline{.215} & .040 & .56 & .052 & .010 & .005 & .77 & .95 & .98 \\   %
Qwen3.8-27B & 4 & .209 & .202 & .238 & \underline{.686} & .339 & .107 & .091 & .088 & .113 & .039 & .51 & .057 & .016 & .006 & .73 & .92 & .97 \\   %
Qwen3.5-397B-A17B & 4 & .196 & .206 & .253 & .664 & .299 & .018 & .012 & .012 & .163 & .026 & .51 & .049 & .002 & .002 & .75 & .99 & .99 \\   %
GLM-4.6V-Flash (9B)~\cite{hong2025glm45v} & 4 & .218 & .204 & .284 & .624 & .341 & .062 & .046 & .044 & .157 & .030 & .54 & .054 & .004 & .001 & .75 & .98 & .99 \\   %
Gemma 4 (31B)~\cite{gemma2026gemma4} & 4 & .188 & .185 & .274 & .544 & .300 & .052 & .039 & .038 & .195 & .036 & .58 & .054 & .003 & .003 & .71 & .98 & .98 \\   %
InternVL3.5 (38B)~\cite{wang2025internvl35} & $\leq$112 fr. & .167 & .098 & .201 & .327 & .182 & .000 & .000 & .000 & .062 & .010 & \textbf{.67} & .023 & .000 & .000 & .86 & 1.00 & 1.00 \\   %
InternVL3.5 (241B-A28B)~\cite{wang2025internvl35} & $\leq$112 fr. & .160 & .065 & .155 & .294 & .096 & .000 & .000 & .000 & .079 & .013 & .62 & .014 & .000 & .000 & .91 & 1.00 & 1.00 \\   %
\midrule
\multicolumn{19}{@{}l}{\textit{Video reasoning specialists}} \\
EgoThinker~\cite{egothinker2025} & 1 & .234 & .049 & .141 & .103 & .080 & -- & -- & -- & -- & -- & -- & .008 & -- & -- & .97 & -- & -- \\   %
Open-o3 Video~\cite{meng2026openo3video} & 16 fr. & .161 & .139 & .367 & .278 & .218 & -- & -- & -- & -- & -- & -- & .027 & -- & -- & .83 & -- & -- \\   %
VisionCoach~\cite{visioncoach2026} & 16 fr. & .167 & .151 & .363 & .301 & .244 & -- & -- & -- & -- & -- & -- & .032 & -- & -- & .81 & -- & -- \\   %
VideoChat-R1.5~\cite{videochatr15_2025} & 2 & .121 & .155 & .234 & .430 & .250 & -- & -- & -- & -- & -- & -- & .020 & -- & -- & .84 & -- & -- \\   %
VideoMind~\cite{videomind2025} & 1 / 2 & .142 & .167 & .272 & .340 & .258 & -- & -- & -- & -- & -- & -- & .027 & -- & -- & .81 & -- & -- \\   %
LLaVA-ST (7B)~\cite{li2025llavast} & 100 fr. & .088 & .106 & .216 & .176 & .174 & .018 & .012 & .012 & -- & -- & -- & .013 & .002 & -- & .85 & .98 & -- \\   %
GroundVQA~\cite{di2024groundvqa} & 1.87 & .034 & .123 & .228 & .230 & .208 & -- & -- & -- & -- & -- & -- & .003 & -- & -- & .91 & -- & -- \\   %
\midrule
\multicolumn{19}{@{}l}{\textit{Causal reasoning / grounding specialists}} \\
Cosmos-Reason2 (8B)~\cite{nvidia2025cosmosreason2} & 4 & .143 & .164 & .274 & .457 & .257 & .008 & .005 & .005 & .023 & .014 & .48 & .027 & .000 & .000 & .81 & 1.00 & 1.00 \\   %
Causal-VLM~\cite{nadeem2026causalvlm} & 64 fr. & -- & .116 & .164 & .663 & .208 & -- & -- & -- & .043 & .010 & .34 & -- & -- & -- & -- & -- & -- \\   %
MECD+~\cite{chen2025mecdplus} & 2 & -- & -- & -- & -- & -- & -- & -- & -- & .052 & .012 & .54 & -- & -- & -- & -- & -- & -- \\   %
\midrule
\multicolumn{19}{@{}l}{\textit{Proprietary models}} \\
Gemini 3.8 Flash & 4 & \underline{.262} & .213 & .355 & .448 & .361 & .159 & .144 & .141 & .158 & .052 & .56 & \underline{.085} & \underline{.035} & \underline{.018} & .68 & .87 & .93 \\   %
Gemini Robotics-ER 2 & 4$^\ast$ & .206 & .165 & .296 & .329 & .283 & .093 & .084 & .083 & .107 & .038 & .54 & .052 & .021 & .007 & .75 & .90 & .96 \\   %
Gemini 3.1 Pro & 4 & .205 & .220 & .351 & .443 & .358 & .153 & .140 & .137 & .118 & .056 & .60 & .082 & .031 & .016 & \underline{.60} & .85 & .92 \\   %
GPT-6.1 Sol & 4 & \textbf{.330} & \underline{.225} & .347 & .517 & .399 & \textbf{.235} & \textbf{.221} & \textbf{.213} & \textbf{.235} & .058 & .61 & \textbf{.120} & \textbf{.068} & \textbf{.027} & .64 & \textbf{.79} & .92 \\   %
\midrule
\multicolumn{19}{@{}l}{\textit{Fine-tuned on our training data}} \\
Joint SFT (2B) & 4 & .135 & .216 & \underline{.406} & .345 & .256 & .145 & .132 & .129 & .055 & .030 & \underline{.64} & .044 & .021 & \underline{.018} & .68 & .85 & \textbf{.87} \\   %
\method (2B) & 4 / $\leq$8 & .123 & \textbf{.271} & .400 & .455 & \textbf{.443} & \underline{.205} & \underline{.187} & \underline{.184} & .101 & \textbf{.116} & .60 & .056 & .025 & .015 & \textbf{.54} & \underline{.80} & \underline{.88} \\   %
\method (9B) & 4 / $\leq$8 & .184 & \textbf{.271} & \textbf{.425} & .452 & \underline{.432} & .199 & .184 & .178 & .070 & \underline{.081} & .56 & .084 & .033 & \underline{.018} & \textbf{.54} & .82 & .90 \\   %
}
\def\TabMainRowsCompact{%
\multicolumn{13}{@{}l}{\textit{Open-source MLLMs}} \\
Qwen3.5-2B & .117 & .145 & .099 & .003 & .000 & .000 & .072 & .012 & .57 & .027 & .001 & .001 \\
Qwen3.5-9B & .160 & .211 & .232 & .008 & .005 & .005 & .065 & .018 & .51 & .052 & .002 & .002 \\
Qwen3.6-27B & .221 & .198 & .342 & .075 & .061 & .059 & \underline{.215} & .040 & .56 & .052 & .010 & .005 \\
Qwen3.8-27B & .209 & .202 & .339 & .107 & .091 & .088 & .113 & .039 & .51 & .057 & .016 & .006 \\
Qwen3.5-397B-A17B & .196 & .206 & .299 & .018 & .012 & .012 & .163 & .026 & .51 & .049 & .002 & .002 \\
GLM-4.6V-Flash (9B)~\cite{hong2025glm45v} & .218 & .204 & .341 & .062 & .046 & .044 & .157 & .030 & .54 & .054 & .004 & .001 \\
InternVL3.5 (38B)~\cite{wang2025internvl35} & .167 & .098 & .182 & .000 & .000 & .000 & .062 & .010 & \textbf{.67} & .023 & .000 & .000 \\
InternVL3.5 (241B-A28B)~\cite{wang2025internvl35} & .160 & .065 & .096 & .000 & .000 & .000 & .079 & .013 & .62 & .014 & .000 & .000 \\
\midrule
\multicolumn{13}{@{}l}{\textit{Video reasoning specialists}} \\
EgoThinker~\cite{egothinker2025} & .234 & .049 & .080 & -- & -- & -- & -- & -- & -- & .008 & -- & -- \\
VideoMind~\cite{videomind2025} & .142 & .167 & .258 & -- & -- & -- & -- & -- & -- & .027 & -- & -- \\
LLaVA-ST (7B)~\cite{li2025llavast} & .088 & .106 & .174 & .018 & .012 & .012 & -- & -- & -- & .013 & .002 & -- \\
\midrule
\multicolumn{13}{@{}l}{\textit{Causal reasoning / grounding specialists}} \\
Cosmos-Reason2 (8B)~\cite{nvidia2025cosmosreason2} & .143 & .164 & .257 & .008 & .005 & .005 & .023 & .014 & .48 & .027 & .000 & .000 \\
Causal-VLM~\cite{nadeem2026causalvlm} & -- & .116 & .208 & -- & -- & -- & .043 & .010 & .34 & -- & -- & -- \\
MECD+~\cite{chen2025mecdplus} & -- & -- & -- & -- & -- & -- & .052 & .012 & .54 & -- & -- & -- \\
\midrule
\multicolumn{13}{@{}l}{\textit{Proprietary models}} \\
Gemini 3.8 Flash & \underline{.262} & .213 & .361 & .159 & .144 & .141 & .158 & .052 & .56 & \underline{.085} & \underline{.035} & \underline{.018} \\
Gemini Robotics-ER 2 & .206 & .165 & .283 & .093 & .084 & .083 & .107 & .038 & .54 & .052 & .021 & .007 \\
Gemini 3.1 Pro & .205 & .220 & .358 & .153 & .140 & .137 & .118 & .056 & .60 & .082 & .031 & .016 \\
GPT-6.1 Sol & \textbf{.330} & \underline{.225} & .399 & \textbf{.235} & \textbf{.221} & \textbf{.213} & \textbf{.235} & .058 & .61 & \textbf{.120} & \textbf{.068} & \textbf{.027} \\
\midrule
\multicolumn{13}{@{}l}{\textit{Fine-tuned on our training data}} \\
Joint SFT (2B) & .135 & .216 & .256 & .145 & .132 & .129 & .055 & .030 & \underline{.64} & .044 & .021 & \underline{.018} \\
\method (2B) & .123 & \textbf{.271} & \textbf{.443} & \underline{.205} & \underline{.187} & \underline{.184} & .101 & \textbf{.116} & .60 & .056 & .025 & .015 \\
\method (9B) & .184 & \textbf{.271} & \underline{.432} & .199 & .184 & .178 & .070 & \underline{.081} & .56 & .084 & .033 & \underline{.018} \\
}
\providecommand{\TM}[2]{\ifcsname tm@#1@#2\endcsname\csname tm@#1@#2\endcsname\else\TODO{}\fi}
\expandafter\def\csname tm@Base2B@Acc\endcsname{.117}
\expandafter\def\csname tm@Base2B@mTIoU\endcsname{.145}
\expandafter\def\csname tm@Base2B@mTIoP\endcsname{.208}
\expandafter\def\csname tm@Base2B@mTIoG\endcsname{.489}
\expandafter\def\csname tm@Base2B@F1_30\endcsname{.099}
\expandafter\def\csname tm@Base2B@L1\endcsname{.003}
\expandafter\def\csname tm@Base2B@L2\endcsname{.000}
\expandafter\def\csname tm@Base2B@L3\endcsname{.000}
\expandafter\def\csname tm@Base2B@MacroF1\endcsname{.072}
\expandafter\def\csname tm@Base2B@EdgeF1\endcsname{.012}
\expandafter\def\csname tm@Base2B@DepAcc\endcsname{.57}
\expandafter\def\csname tm@Base2B@GCR_T\endcsname{.027}
\expandafter\def\csname tm@Base2B@GCR_ST\endcsname{.001}
\expandafter\def\csname tm@Base2B@GCR_STD\endcsname{.001}
\expandafter\def\csname tm@Base2B@SS_T\endcsname{.77}
\expandafter\def\csname tm@Base2B@SS_ST\endcsname{.99}
\expandafter\def\csname tm@Base2B@SS_STD\endcsname{.99}
\expandafter\def\csname tm@Base9B@Acc\endcsname{.160}
\expandafter\def\csname tm@Base9B@mTIoU\endcsname{.211}
\expandafter\def\csname tm@Base9B@mTIoP\endcsname{.297}
\expandafter\def\csname tm@Base9B@mTIoG\endcsname{.568}
\expandafter\def\csname tm@Base9B@F1_30\endcsname{.232}
\expandafter\def\csname tm@Base9B@L1\endcsname{.008}
\expandafter\def\csname tm@Base9B@L2\endcsname{.005}
\expandafter\def\csname tm@Base9B@L3\endcsname{.005}
\expandafter\def\csname tm@Base9B@MacroF1\endcsname{.065}
\expandafter\def\csname tm@Base9B@EdgeF1\endcsname{.018}
\expandafter\def\csname tm@Base9B@DepAcc\endcsname{.51}
\expandafter\def\csname tm@Base9B@GCR_T\endcsname{.052}
\expandafter\def\csname tm@Base9B@GCR_ST\endcsname{.002}
\expandafter\def\csname tm@Base9B@GCR_STD\endcsname{.002}
\expandafter\def\csname tm@Base9B@SS_T\endcsname{.67}
\expandafter\def\csname tm@Base9B@SS_ST\endcsname{.99}
\expandafter\def\csname tm@Base9B@SS_STD\endcsname{.99}
\expandafter\def\csname tm@Q36@Acc\endcsname{.221}
\expandafter\def\csname tm@Q36@mTIoU\endcsname{.198}
\expandafter\def\csname tm@Q36@mTIoP\endcsname{.234}
\expandafter\def\csname tm@Q36@mTIoG\endcsname{.730}
\expandafter\def\csname tm@Q36@F1_30\endcsname{.342}
\expandafter\def\csname tm@Q36@L1\endcsname{.075}
\expandafter\def\csname tm@Q36@L2\endcsname{.061}
\expandafter\def\csname tm@Q36@L3\endcsname{.059}
\expandafter\def\csname tm@Q36@MacroF1\endcsname{.215}
\expandafter\def\csname tm@Q36@EdgeF1\endcsname{.040}
\expandafter\def\csname tm@Q36@DepAcc\endcsname{.56}
\expandafter\def\csname tm@Q36@GCR_T\endcsname{.052}
\expandafter\def\csname tm@Q36@GCR_ST\endcsname{.010}
\expandafter\def\csname tm@Q36@GCR_STD\endcsname{.005}
\expandafter\def\csname tm@Q36@SS_T\endcsname{.77}
\expandafter\def\csname tm@Q36@SS_ST\endcsname{.95}
\expandafter\def\csname tm@Q36@SS_STD\endcsname{.98}
\expandafter\def\csname tm@Q38@Acc\endcsname{.209}
\expandafter\def\csname tm@Q38@mTIoU\endcsname{.202}
\expandafter\def\csname tm@Q38@mTIoP\endcsname{.238}
\expandafter\def\csname tm@Q38@mTIoG\endcsname{.686}
\expandafter\def\csname tm@Q38@F1_30\endcsname{.339}
\expandafter\def\csname tm@Q38@L1\endcsname{.107}
\expandafter\def\csname tm@Q38@L2\endcsname{.091}
\expandafter\def\csname tm@Q38@L3\endcsname{.088}
\expandafter\def\csname tm@Q38@MacroF1\endcsname{.113}
\expandafter\def\csname tm@Q38@EdgeF1\endcsname{.039}
\expandafter\def\csname tm@Q38@DepAcc\endcsname{.51}
\expandafter\def\csname tm@Q38@GCR_T\endcsname{.057}
\expandafter\def\csname tm@Q38@GCR_ST\endcsname{.016}
\expandafter\def\csname tm@Q38@GCR_STD\endcsname{.006}
\expandafter\def\csname tm@Q38@SS_T\endcsname{.73}
\expandafter\def\csname tm@Q38@SS_ST\endcsname{.92}
\expandafter\def\csname tm@Q38@SS_STD\endcsname{.97}
\expandafter\def\csname tm@Q35L@Acc\endcsname{.196}
\expandafter\def\csname tm@Q35L@mTIoU\endcsname{.206}
\expandafter\def\csname tm@Q35L@mTIoP\endcsname{.253}
\expandafter\def\csname tm@Q35L@mTIoG\endcsname{.664}
\expandafter\def\csname tm@Q35L@F1_30\endcsname{.299}
\expandafter\def\csname tm@Q35L@L1\endcsname{.018}
\expandafter\def\csname tm@Q35L@L2\endcsname{.012}
\expandafter\def\csname tm@Q35L@L3\endcsname{.012}
\expandafter\def\csname tm@Q35L@MacroF1\endcsname{.163}
\expandafter\def\csname tm@Q35L@EdgeF1\endcsname{.026}
\expandafter\def\csname tm@Q35L@DepAcc\endcsname{.51}
\expandafter\def\csname tm@Q35L@GCR_T\endcsname{.049}
\expandafter\def\csname tm@Q35L@GCR_ST\endcsname{.002}
\expandafter\def\csname tm@Q35L@GCR_STD\endcsname{.002}
\expandafter\def\csname tm@Q35L@SS_T\endcsname{.75}
\expandafter\def\csname tm@Q35L@SS_ST\endcsname{.99}
\expandafter\def\csname tm@Q35L@SS_STD\endcsname{.99}
\expandafter\def\csname tm@GLM46VF@Acc\endcsname{.218}
\expandafter\def\csname tm@GLM46VF@mTIoU\endcsname{.204}
\expandafter\def\csname tm@GLM46VF@mTIoP\endcsname{.284}
\expandafter\def\csname tm@GLM46VF@mTIoG\endcsname{.624}
\expandafter\def\csname tm@GLM46VF@F1_30\endcsname{.341}
\expandafter\def\csname tm@GLM46VF@L1\endcsname{.062}
\expandafter\def\csname tm@GLM46VF@L2\endcsname{.046}
\expandafter\def\csname tm@GLM46VF@L3\endcsname{.044}
\expandafter\def\csname tm@GLM46VF@MacroF1\endcsname{.157}
\expandafter\def\csname tm@GLM46VF@EdgeF1\endcsname{.030}
\expandafter\def\csname tm@GLM46VF@DepAcc\endcsname{.54}
\expandafter\def\csname tm@GLM46VF@GCR_T\endcsname{.054}
\expandafter\def\csname tm@GLM46VF@GCR_ST\endcsname{.004}
\expandafter\def\csname tm@GLM46VF@GCR_STD\endcsname{.001}
\expandafter\def\csname tm@GLM46VF@SS_T\endcsname{.75}
\expandafter\def\csname tm@GLM46VF@SS_ST\endcsname{.98}
\expandafter\def\csname tm@GLM46VF@SS_STD\endcsname{.99}
\expandafter\def\csname tm@Gemma4@Acc\endcsname{.188}
\expandafter\def\csname tm@Gemma4@mTIoU\endcsname{.185}
\expandafter\def\csname tm@Gemma4@mTIoP\endcsname{.274}
\expandafter\def\csname tm@Gemma4@mTIoG\endcsname{.544}
\expandafter\def\csname tm@Gemma4@F1_30\endcsname{.300}
\expandafter\def\csname tm@Gemma4@L1\endcsname{.052}
\expandafter\def\csname tm@Gemma4@L2\endcsname{.039}
\expandafter\def\csname tm@Gemma4@L3\endcsname{.038}
\expandafter\def\csname tm@Gemma4@MacroF1\endcsname{.195}
\expandafter\def\csname tm@Gemma4@EdgeF1\endcsname{.036}
\expandafter\def\csname tm@Gemma4@DepAcc\endcsname{.58}
\expandafter\def\csname tm@Gemma4@GCR_T\endcsname{.054}
\expandafter\def\csname tm@Gemma4@GCR_ST\endcsname{.003}
\expandafter\def\csname tm@Gemma4@GCR_STD\endcsname{.003}
\expandafter\def\csname tm@Gemma4@SS_T\endcsname{.71}
\expandafter\def\csname tm@Gemma4@SS_ST\endcsname{.98}
\expandafter\def\csname tm@Gemma4@SS_STD\endcsname{.98}
\expandafter\def\csname tm@InternVL35@Acc\endcsname{.167}
\expandafter\def\csname tm@InternVL35@mTIoU\endcsname{.098}
\expandafter\def\csname tm@InternVL35@mTIoP\endcsname{.201}
\expandafter\def\csname tm@InternVL35@mTIoG\endcsname{.327}
\expandafter\def\csname tm@InternVL35@F1_30\endcsname{.182}
\expandafter\def\csname tm@InternVL35@L1\endcsname{.000}
\expandafter\def\csname tm@InternVL35@L2\endcsname{.000}
\expandafter\def\csname tm@InternVL35@L3\endcsname{.000}
\expandafter\def\csname tm@InternVL35@MacroF1\endcsname{.062}
\expandafter\def\csname tm@InternVL35@EdgeF1\endcsname{.010}
\expandafter\def\csname tm@InternVL35@DepAcc\endcsname{.67}
\expandafter\def\csname tm@InternVL35@GCR_T\endcsname{.023}
\expandafter\def\csname tm@InternVL35@GCR_ST\endcsname{.000}
\expandafter\def\csname tm@InternVL35@GCR_STD\endcsname{.000}
\expandafter\def\csname tm@InternVL35@SS_T\endcsname{.86}
\expandafter\def\csname tm@InternVL35@SS_ST\endcsname{1.00}
\expandafter\def\csname tm@InternVL35@SS_STD\endcsname{1.00}
\expandafter\def\csname tm@InternVL35L@Acc\endcsname{.160}
\expandafter\def\csname tm@InternVL35L@mTIoU\endcsname{.065}
\expandafter\def\csname tm@InternVL35L@mTIoP\endcsname{.155}
\expandafter\def\csname tm@InternVL35L@mTIoG\endcsname{.294}
\expandafter\def\csname tm@InternVL35L@F1_30\endcsname{.096}
\expandafter\def\csname tm@InternVL35L@L1\endcsname{.000}
\expandafter\def\csname tm@InternVL35L@L2\endcsname{.000}
\expandafter\def\csname tm@InternVL35L@L3\endcsname{.000}
\expandafter\def\csname tm@InternVL35L@MacroF1\endcsname{.079}
\expandafter\def\csname tm@InternVL35L@EdgeF1\endcsname{.013}
\expandafter\def\csname tm@InternVL35L@DepAcc\endcsname{.62}
\expandafter\def\csname tm@InternVL35L@GCR_T\endcsname{.014}
\expandafter\def\csname tm@InternVL35L@GCR_ST\endcsname{.000}
\expandafter\def\csname tm@InternVL35L@GCR_STD\endcsname{.000}
\expandafter\def\csname tm@InternVL35L@SS_T\endcsname{.91}
\expandafter\def\csname tm@InternVL35L@SS_ST\endcsname{1.00}
\expandafter\def\csname tm@InternVL35L@SS_STD\endcsname{1.00}
\expandafter\def\csname tm@EgoThinker@Acc\endcsname{.234}
\expandafter\def\csname tm@EgoThinker@mTIoU\endcsname{.049}
\expandafter\def\csname tm@EgoThinker@mTIoP\endcsname{.141}
\expandafter\def\csname tm@EgoThinker@mTIoG\endcsname{.103}
\expandafter\def\csname tm@EgoThinker@F1_30\endcsname{.080}
\expandafter\def\csname tm@EgoThinker@GCR_T\endcsname{.008}
\expandafter\def\csname tm@EgoThinker@SS_T\endcsname{.97}
\expandafter\def\csname tm@OpenO3Video@Acc\endcsname{.161}
\expandafter\def\csname tm@OpenO3Video@mTIoU\endcsname{.139}
\expandafter\def\csname tm@OpenO3Video@mTIoP\endcsname{.367}
\expandafter\def\csname tm@OpenO3Video@mTIoG\endcsname{.278}
\expandafter\def\csname tm@OpenO3Video@F1_30\endcsname{.218}
\expandafter\def\csname tm@OpenO3Video@GCR_T\endcsname{.027}
\expandafter\def\csname tm@OpenO3Video@SS_T\endcsname{.83}
\expandafter\def\csname tm@VisionCoach@Acc\endcsname{.167}
\expandafter\def\csname tm@VisionCoach@mTIoU\endcsname{.151}
\expandafter\def\csname tm@VisionCoach@mTIoP\endcsname{.363}
\expandafter\def\csname tm@VisionCoach@mTIoG\endcsname{.301}
\expandafter\def\csname tm@VisionCoach@F1_30\endcsname{.244}
\expandafter\def\csname tm@VisionCoach@GCR_T\endcsname{.032}
\expandafter\def\csname tm@VisionCoach@SS_T\endcsname{.81}
\expandafter\def\csname tm@VideoChatR15@Acc\endcsname{.121}
\expandafter\def\csname tm@VideoChatR15@mTIoU\endcsname{.155}
\expandafter\def\csname tm@VideoChatR15@mTIoP\endcsname{.234}
\expandafter\def\csname tm@VideoChatR15@mTIoG\endcsname{.430}
\expandafter\def\csname tm@VideoChatR15@F1_30\endcsname{.250}
\expandafter\def\csname tm@VideoChatR15@GCR_T\endcsname{.020}
\expandafter\def\csname tm@VideoChatR15@SS_T\endcsname{.84}
\expandafter\def\csname tm@VideoMind@Acc\endcsname{.142}
\expandafter\def\csname tm@VideoMind@mTIoU\endcsname{.167}
\expandafter\def\csname tm@VideoMind@mTIoP\endcsname{.272}
\expandafter\def\csname tm@VideoMind@mTIoG\endcsname{.340}
\expandafter\def\csname tm@VideoMind@F1_30\endcsname{.258}
\expandafter\def\csname tm@VideoMind@GCR_T\endcsname{.027}
\expandafter\def\csname tm@VideoMind@SS_T\endcsname{.81}
\expandafter\def\csname tm@LLaVAST@Acc\endcsname{.088}
\expandafter\def\csname tm@LLaVAST@mTIoU\endcsname{.106}
\expandafter\def\csname tm@LLaVAST@mTIoP\endcsname{.216}
\expandafter\def\csname tm@LLaVAST@mTIoG\endcsname{.176}
\expandafter\def\csname tm@LLaVAST@F1_30\endcsname{.174}
\expandafter\def\csname tm@LLaVAST@L1\endcsname{.018}
\expandafter\def\csname tm@LLaVAST@L2\endcsname{.012}
\expandafter\def\csname tm@LLaVAST@L3\endcsname{.012}
\expandafter\def\csname tm@LLaVAST@GCR_T\endcsname{.013}
\expandafter\def\csname tm@LLaVAST@GCR_ST\endcsname{.002}
\expandafter\def\csname tm@LLaVAST@SS_T\endcsname{.85}
\expandafter\def\csname tm@LLaVAST@SS_ST\endcsname{.98}
\expandafter\def\csname tm@GroundVQA@Acc\endcsname{.034}
\expandafter\def\csname tm@GroundVQA@mTIoU\endcsname{.123}
\expandafter\def\csname tm@GroundVQA@mTIoP\endcsname{.228}
\expandafter\def\csname tm@GroundVQA@mTIoG\endcsname{.230}
\expandafter\def\csname tm@GroundVQA@F1_30\endcsname{.208}
\expandafter\def\csname tm@GroundVQA@GCR_T\endcsname{.003}
\expandafter\def\csname tm@GroundVQA@SS_T\endcsname{.91}
\expandafter\def\csname tm@CosmosR2@Acc\endcsname{.143}
\expandafter\def\csname tm@CosmosR2@mTIoU\endcsname{.164}
\expandafter\def\csname tm@CosmosR2@mTIoP\endcsname{.274}
\expandafter\def\csname tm@CosmosR2@mTIoG\endcsname{.457}
\expandafter\def\csname tm@CosmosR2@F1_30\endcsname{.257}
\expandafter\def\csname tm@CosmosR2@L1\endcsname{.008}
\expandafter\def\csname tm@CosmosR2@L2\endcsname{.005}
\expandafter\def\csname tm@CosmosR2@L3\endcsname{.005}
\expandafter\def\csname tm@CosmosR2@MacroF1\endcsname{.023}
\expandafter\def\csname tm@CosmosR2@EdgeF1\endcsname{.014}
\expandafter\def\csname tm@CosmosR2@DepAcc\endcsname{.48}
\expandafter\def\csname tm@CosmosR2@GCR_T\endcsname{.027}
\expandafter\def\csname tm@CosmosR2@GCR_ST\endcsname{.000}
\expandafter\def\csname tm@CosmosR2@GCR_STD\endcsname{.000}
\expandafter\def\csname tm@CosmosR2@SS_T\endcsname{.81}
\expandafter\def\csname tm@CosmosR2@SS_ST\endcsname{1.00}
\expandafter\def\csname tm@CosmosR2@SS_STD\endcsname{1.00}
\expandafter\def\csname tm@CausalVLM@mTIoU\endcsname{.116}
\expandafter\def\csname tm@CausalVLM@mTIoP\endcsname{.164}
\expandafter\def\csname tm@CausalVLM@mTIoG\endcsname{.663}
\expandafter\def\csname tm@CausalVLM@F1_30\endcsname{.208}
\expandafter\def\csname tm@CausalVLM@MacroF1\endcsname{.043}
\expandafter\def\csname tm@CausalVLM@EdgeF1\endcsname{.010}
\expandafter\def\csname tm@CausalVLM@DepAcc\endcsname{.34}
\expandafter\def\csname tm@MECDplus@MacroF1\endcsname{.052}
\expandafter\def\csname tm@MECDplus@EdgeF1\endcsname{.012}
\expandafter\def\csname tm@MECDplus@DepAcc\endcsname{.54}
\expandafter\def\csname tm@G38F@Acc\endcsname{.262}
\expandafter\def\csname tm@G38F@mTIoU\endcsname{.213}
\expandafter\def\csname tm@G38F@mTIoP\endcsname{.355}
\expandafter\def\csname tm@G38F@mTIoG\endcsname{.448}
\expandafter\def\csname tm@G38F@F1_30\endcsname{.361}
\expandafter\def\csname tm@G38F@L1\endcsname{.159}
\expandafter\def\csname tm@G38F@L2\endcsname{.144}
\expandafter\def\csname tm@G38F@L3\endcsname{.141}
\expandafter\def\csname tm@G38F@MacroF1\endcsname{.158}
\expandafter\def\csname tm@G38F@EdgeF1\endcsname{.052}
\expandafter\def\csname tm@G38F@DepAcc\endcsname{.56}
\expandafter\def\csname tm@G38F@GCR_T\endcsname{.085}
\expandafter\def\csname tm@G38F@GCR_ST\endcsname{.035}
\expandafter\def\csname tm@G38F@GCR_STD\endcsname{.018}
\expandafter\def\csname tm@G38F@SS_T\endcsname{.68}
\expandafter\def\csname tm@G38F@SS_ST\endcsname{.87}
\expandafter\def\csname tm@G38F@SS_STD\endcsname{.93}
\expandafter\def\csname tm@GER2@Acc\endcsname{.206}
\expandafter\def\csname tm@GER2@mTIoU\endcsname{.165}
\expandafter\def\csname tm@GER2@mTIoP\endcsname{.296}
\expandafter\def\csname tm@GER2@mTIoG\endcsname{.329}
\expandafter\def\csname tm@GER2@F1_30\endcsname{.283}
\expandafter\def\csname tm@GER2@L1\endcsname{.093}
\expandafter\def\csname tm@GER2@L2\endcsname{.084}
\expandafter\def\csname tm@GER2@L3\endcsname{.083}
\expandafter\def\csname tm@GER2@MacroF1\endcsname{.107}
\expandafter\def\csname tm@GER2@EdgeF1\endcsname{.038}
\expandafter\def\csname tm@GER2@DepAcc\endcsname{.54}
\expandafter\def\csname tm@GER2@GCR_T\endcsname{.052}
\expandafter\def\csname tm@GER2@GCR_ST\endcsname{.021}
\expandafter\def\csname tm@GER2@GCR_STD\endcsname{.007}
\expandafter\def\csname tm@GER2@SS_T\endcsname{.75}
\expandafter\def\csname tm@GER2@SS_ST\endcsname{.90}
\expandafter\def\csname tm@GER2@SS_STD\endcsname{.96}
\expandafter\def\csname tm@G31P@Acc\endcsname{.205}
\expandafter\def\csname tm@G31P@mTIoU\endcsname{.220}
\expandafter\def\csname tm@G31P@mTIoP\endcsname{.351}
\expandafter\def\csname tm@G31P@mTIoG\endcsname{.443}
\expandafter\def\csname tm@G31P@F1_30\endcsname{.358}
\expandafter\def\csname tm@G31P@L1\endcsname{.153}
\expandafter\def\csname tm@G31P@L2\endcsname{.140}
\expandafter\def\csname tm@G31P@L3\endcsname{.137}
\expandafter\def\csname tm@G31P@MacroF1\endcsname{.118}
\expandafter\def\csname tm@G31P@EdgeF1\endcsname{.056}
\expandafter\def\csname tm@G31P@DepAcc\endcsname{.60}
\expandafter\def\csname tm@G31P@GCR_T\endcsname{.082}
\expandafter\def\csname tm@G31P@GCR_ST\endcsname{.031}
\expandafter\def\csname tm@G31P@GCR_STD\endcsname{.016}
\expandafter\def\csname tm@G31P@SS_T\endcsname{.60}
\expandafter\def\csname tm@G31P@SS_ST\endcsname{.85}
\expandafter\def\csname tm@G31P@SS_STD\endcsname{.92}
\expandafter\def\csname tm@GPT@Acc\endcsname{.330}
\expandafter\def\csname tm@GPT@mTIoU\endcsname{.225}
\expandafter\def\csname tm@GPT@mTIoP\endcsname{.347}
\expandafter\def\csname tm@GPT@mTIoG\endcsname{.517}
\expandafter\def\csname tm@GPT@F1_30\endcsname{.399}
\expandafter\def\csname tm@GPT@L1\endcsname{.235}
\expandafter\def\csname tm@GPT@L2\endcsname{.221}
\expandafter\def\csname tm@GPT@L3\endcsname{.213}
\expandafter\def\csname tm@GPT@MacroF1\endcsname{.235}
\expandafter\def\csname tm@GPT@EdgeF1\endcsname{.058}
\expandafter\def\csname tm@GPT@DepAcc\endcsname{.61}
\expandafter\def\csname tm@GPT@GCR_T\endcsname{.120}
\expandafter\def\csname tm@GPT@GCR_ST\endcsname{.068}
\expandafter\def\csname tm@GPT@GCR_STD\endcsname{.027}
\expandafter\def\csname tm@GPT@SS_T\endcsname{.64}
\expandafter\def\csname tm@GPT@SS_ST\endcsname{.79}
\expandafter\def\csname tm@GPT@SS_STD\endcsname{.92}
\expandafter\def\csname tm@M1F@Acc\endcsname{.135}
\expandafter\def\csname tm@M1F@mTIoU\endcsname{.216}
\expandafter\def\csname tm@M1F@mTIoP\endcsname{.406}
\expandafter\def\csname tm@M1F@mTIoG\endcsname{.345}
\expandafter\def\csname tm@M1F@F1_30\endcsname{.256}
\expandafter\def\csname tm@M1F@L1\endcsname{.145}
\expandafter\def\csname tm@M1F@L2\endcsname{.132}
\expandafter\def\csname tm@M1F@L3\endcsname{.129}
\expandafter\def\csname tm@M1F@MacroF1\endcsname{.055}
\expandafter\def\csname tm@M1F@EdgeF1\endcsname{.030}
\expandafter\def\csname tm@M1F@DepAcc\endcsname{.64}
\expandafter\def\csname tm@M1F@GCR_T\endcsname{.044}
\expandafter\def\csname tm@M1F@GCR_ST\endcsname{.021}
\expandafter\def\csname tm@M1F@GCR_STD\endcsname{.018}
\expandafter\def\csname tm@M1F@SS_T\endcsname{.68}
\expandafter\def\csname tm@M1F@SS_ST\endcsname{.85}
\expandafter\def\csname tm@M1F@SS_STD\endcsname{.87}
\expandafter\def\csname tm@A@Acc\endcsname{.123}
\expandafter\def\csname tm@A@mTIoU\endcsname{.271}
\expandafter\def\csname tm@A@mTIoP\endcsname{.400}
\expandafter\def\csname tm@A@mTIoG\endcsname{.455}
\expandafter\def\csname tm@A@F1_30\endcsname{.443}
\expandafter\def\csname tm@A@L1\endcsname{.205}
\expandafter\def\csname tm@A@L2\endcsname{.187}
\expandafter\def\csname tm@A@L3\endcsname{.184}
\expandafter\def\csname tm@A@MacroF1\endcsname{.101}
\expandafter\def\csname tm@A@EdgeF1\endcsname{.116}
\expandafter\def\csname tm@A@DepAcc\endcsname{.60}
\expandafter\def\csname tm@A@GCR_T\endcsname{.056}
\expandafter\def\csname tm@A@GCR_ST\endcsname{.025}
\expandafter\def\csname tm@A@GCR_STD\endcsname{.015}
\expandafter\def\csname tm@A@SS_T\endcsname{.54}
\expandafter\def\csname tm@A@SS_ST\endcsname{.80}
\expandafter\def\csname tm@A@SS_STD\endcsname{.88}
\expandafter\def\csname tm@A9@Acc\endcsname{.184}
\expandafter\def\csname tm@A9@mTIoU\endcsname{.271}
\expandafter\def\csname tm@A9@mTIoP\endcsname{.425}
\expandafter\def\csname tm@A9@mTIoG\endcsname{.452}
\expandafter\def\csname tm@A9@F1_30\endcsname{.432}
\expandafter\def\csname tm@A9@L1\endcsname{.199}
\expandafter\def\csname tm@A9@L2\endcsname{.184}
\expandafter\def\csname tm@A9@L3\endcsname{.178}
\expandafter\def\csname tm@A9@MacroF1\endcsname{.070}
\expandafter\def\csname tm@A9@EdgeF1\endcsname{.081}
\expandafter\def\csname tm@A9@DepAcc\endcsname{.56}
\expandafter\def\csname tm@A9@GCR_T\endcsname{.084}
\expandafter\def\csname tm@A9@GCR_ST\endcsname{.033}
\expandafter\def\csname tm@A9@GCR_STD\endcsname{.018}
\expandafter\def\csname tm@A9@SS_T\endcsname{.54}
\expandafter\def\csname tm@A9@SS_ST\endcsname{.82}
\expandafter\def\csname tm@A9@SS_STD\endcsname{.90}

\def\EvQK{3.9K}
\def\EvVideosK{2.4K}
\def\EvQ{3,901}
\def\EvGold{1,639}
\def\EvDatasets{4}
\def\EvVideos{2,444}
\def\EvDur{44.1}
\def\EvKpq{1.59}
\def\EvKdur{4.6}
\def\EvMultiK{48\%}
\def\EvIpq{1.98}
\def\EvIpk{1.25}
\def\EvPpi{2.49}
\def\EvTool{20\%}
\def\EvMultiDep{38\%}
\def\EvCauses{19}
\def\EvEnables{348}
\def\EvAllQ{6,028}
\def\EvCOneKept{5,118}
\def\EvCapQ{5,103}
\def\EvReviewed{2,485}
\def\EvVerified{1,975}
\def\EvGoldEvents{2,605}
\def\EvGoldInter{3,253}
\def\EvGoldPart{8,115}
\def\EvGoldPairs{1,180}
\def\EvTotalK{29.8K}
\def\EvGoldK{1.6K}
\def\EvRmHuman{320}
\def\EvRmNoEv{111}
\def\EvRmCtwo{743}
\def\EvRmCfour{28}
\def\EvSpKeptHuman{1,639}   %
\def\EvSpKeptAuto{2,262}   %
\def\EvSpCtwoHuman{338}
\def\EvSpCtwoAuto{405}
\def\EvSpCfourHuman{7}
\def\EvSpCfourAuto{21}
\def\EvSpNoEvHuman{8}
\def\EvSpNoEvAuto{103}
\def\EvSpHumanRej{320}
\def\EvCapHx{47.94}
\def\EvCapHw{3.18}
\def\EvCapHpct{3}
\def\EvCapBars{0/9/{29.43,18.51,3.18,12.38,28.35,8.15},1/0/{12.10,8.53,2.70,15.63,40.62,20.43}}   %

\input{fig/teaser}
\def\FAgptCorrect{541}
\def\FAgptT{196}
\def\FAgptLoc{138}
\def\FAgptRole{127}
\def\FAgptAssoc{119}
\def\FAgptS{112}
\def\FAgptAll{45}
\def\FAgptAllPctCorrect{8.3}
\def\FAgptAllPctN{2.7}
\def\FApoolN{12}
\def\FApoolLoc{587}
\def\FApoolRA{96}
\def\FApoolRAPct{16}
\def\FApoolRole{74}
\def\FApoolAssoc{22}
\def\FCnDep{302}
\def\FCqD{$-$2.3}
\def\FCqLo{$-$6.0}
\def\FCqHi{+1.3}
\def\FCgD{$-$0.3}
\def\FCgLo{$-$3.3}
\def\FCgHi{+2.6}
\def\FDqVEdgeKept{.531}
\def\FDqVPairsKept{335}
\def\FDqVDepPairsKept{124}
\def\FDgVEdgeKept{.463}
\def\FDgVPairsKept{253}
\def\FDgVDepPairsKept{87}
\def\FDPairsAll{1,180}
\def\FDDepPairsAll{367}
\begin{abstract}
Answering causal questions about egocentric videos requires more than localizing visual evidence: a model must identify which events matter, which objects are manipulated and with which tools, and how earlier events cause or enable later ones.
Existing benchmarks evaluate answer correctness, evidence localization or reasoning traces, but do not jointly assess the functional roles (hand, tool or target) and associations of interaction participants, and typed event dependencies.
We introduce \textbf{\bench}, a benchmark for grounded causal question answering comprising \EvTotalK{} question--video pairs from 13 egocentric datasets, spanning six causal capabilities.
Each question is annotated with key-event intervals, hand--object interactions specifying participant bounding boxes, functional roles and associations, and typed event dependencies.
It is divided into a training split of 25.9K questions and an evaluation split of \EvQK{} questions, including \EvGoldK{} with human-verified evidence.
We conduct an in-depth analysis of multimodal large language models (MLLMs) and causal video reasoning models on \bench, revealing incomplete event recovery, functional role confusion despite correct object localization, and incorrect participant association at moments with multiple interactions.
Dependency recovery further suffers from coarse event predictions and spurious links between adjacent events or events sharing a tool or target.
Even for the frontier model GPT-6.1 Sol, only \FAgptAllPctCorrect\% of correct answers are supported by correct key events, interactions and dependencies.
Motivated by these failures, we explore \textbf{\method}, which is trained on the \bench training split and explicitly predicts structured causal evidence before using it to answer.
It improves causal evidence recovery over a baseline using the same backbone and training data, achieving the best key-event and dependency-edge recovery among the evaluated models.
\end{abstract}

\section{Introduction}
\label{sec:intro}

Wearable assistants and robots that learn from human demonstrations must understand not only what people do but also how their actions affect objects and subsequent activities~\cite{chen2023egoplanbench,grauman2022ego4d,grauman2024egoexo4d}.
In egocentric videos, causal outcomes depend on how hands manipulate objects, which object serves as a tool and which is acted on, and how manipulations depend on one another.
In \cref{fig:teaser}, answering what would happen if the tortilla had not been unfolded requires identifying the unfolding and cutting events, how the hand uses a knife on the tortilla, and that unfolding enables the cutting as performed in the video.

Existing benchmarks evaluate this evidence only partially (\cref{tab:bench_compare}).
Causal video question-answering benchmarks score answers or textual explanations~\cite{foss2025causalvqa,jia2022egotaskqa,li2022causalvidqa,qiu2024egoplanbench2,xiao2021nextqa}, and grounded ones add temporal intervals, spatial regions or reasoning traces~\cite{gelm2025multihop,cheng2025vstar,dai2026egocotbench,guo2026procobject,nagrani2026minervaego,xiao2024nextgqa,zhang2026castbench}, with CaST-Bench~\cite{zhang2026castbench} separating grounded from spurious answers through spatio-temporal localization.
These benchmarks do not jointly evaluate functional roles, interaction associations and typed event dependencies in the visual evidence predicted for an answer, so a model can answer correctly while misassigning roles and merging the participants of different manipulations (\cref{fig:spurious}).
\textbf{Can video models recover the relevant events, identify how their participants interact, and determine how the events depend on one another, and do their correct answers come with such evidence?}

\input{fig/fig_spurious_assoc}

To address these questions, we introduce \textbf{\bench} (\textbf{Ego}centric \textbf{S}patio-temporal \textbf{P}inpointing of \textbf{A}nswer-relevant events, their \textbf{R}elations, and \textbf{K}ey hand--object interactions), a benchmark for grounded causal question answering (GCQA) in egocentric videos.
Given a video and a causal question, a model predicts an answer with three kinds of causal evidence (\cref{fig:teaser}): \emph{key events} (when), each with a temporal interval; \emph{interactions} (what and where), with the bounding box, functional role (hand, tool or target) and association of each participant; and \emph{dependencies} (how), whether an earlier event \emph{causes} or \emph{enables} a later one.

\bench comprises \EvTotalK{} questions from 13 datasets, covering six causal capabilities: explanation, counterfactual, hypothetical, anticipation, planning and goal attribution (\cref{sec:bench_filter}).
It is divided into a training split of 25,894 questions and an evaluation split of \EvQK{} questions on \EvVideos{} videos, of which \EvGold{} have human-verified evidence, and nearly half of the gold-labeled questions require multiple key events.
To build it, we manually reclassify the source question templates by the causal inference they require, and a semi-automatic pipeline preserves hand and object identities across frames, selects the answer-relevant events, organizes their participants into role-aware interactions and labels typed dependencies.
Model-based integrity checks remove questions answerable from text alone, answerable with the key events masked, or unanswerable from the key events alone, and twelve annotators reviewed 2,500 evaluation questions, correcting their evidence and rejecting invalid ones.
Beyond evidence recovery, its structured causal evidence, nested grounding conditions and controlled experiments let us attribute failures to localization, roles or association, to event recovery or dependency prediction, and distinguish grounded from spurious correct answers.

Evaluating multimodal large language models (MLLMs), video reasoning models and causal reasoning models on \bench reveals three limitations (\cref{sec:exp_results,sec:exp_findings}).
\textbf{First}, grounded video reasoning models~\cite{videomind2025,meng2026openo3video,visioncoach2026} return a single interval in their released pipelines, yet their low recall even on single-event questions shows failures beyond this output format.
\textbf{Second}, grounding methods~\cite{li2025llavast,egothinker2025} focus on individual entities rather than on how they form manipulations: models confuse functional roles despite correct object localization and associate participants with the wrong interactions, even when boxes and roles are given.
\textbf{Third}, a dense causal captioning model~\cite{nadeem2026causalvlm} merges dependent events into coarse intervals, and causal discovery~\cite{chen2025mecdplus} favors temporally adjacent events even given the ground-truth events.
MLLMs also over-predict dependencies between events involving the same tool or target.
Overall, GPT-6.1 Sol answers only 33\% of the questions correctly, and only 8.3\% of these answers pass all grounding criteria.

Motivated by these failures, we explore \textbf{\method} (\textbf{S}tructured \textbf{P}rediction of \textbf{A}ction-Role Interactions, \textbf{R}elations, and \textbf{K}ey Events for \textbf{G}rounded Causal Answering, \cref{sec:method}), a ground-then-answer model trained on the automatically annotated training split of \bench.
It first predicts the key events, the role-aware interactions within each event and the typed dependencies between events, each with a module targeting one of the observed failures, and then answers from this evidence alone.
It improves event, interaction and dependency F1 over a joint supervised fine-tuning (SFT) baseline, which generates the evidence and answer in a single output with the same backbone and training data, yet grounded correct rates remain low.

Our contributions are threefold:
\begin{itemize}[leftmargin=*,itemsep=1pt,topsep=2pt]
\item \bench, a large-scale benchmark of \EvTotalK{} questions from 13 egocentric datasets with spatio-temporal, role-aware causal evidence, which jointly evaluates causal answers, key events, hand--object interactions and event dependencies.
\item An in-depth analysis of existing MLLMs and video reasoning models on \bench, which identifies failures in event recovery, functional roles, participant association and dependency prediction through model outputs and controlled experiments with human-verified evidence.
\item \method, a structured prediction approach trained on the automatically annotated training split of \bench that improves evidence recovery over joint generation.
\end{itemize}

\section{Related Work}
\label{sec:related}

\def\EvQK{3.9K}
\def\EvVideosK{2.4K}
\def\EvQ{3,901}
\def\EvGold{1,639}
\def\EvDatasets{4}
\def\EvVideos{2,444}
\def\EvDur{44.1}
\def\EvKpq{1.59}
\def\EvKdur{4.6}
\def\EvMultiK{48\%}
\def\EvIpq{1.98}
\def\EvIpk{1.25}
\def\EvPpi{2.49}
\def\EvTool{20\%}
\def\EvMultiDep{38\%}
\def\EvCauses{19}
\def\EvEnables{348}
\def\EvAllQ{6,028}
\def\EvCOneKept{5,118}
\def\EvCapQ{5,103}
\def\EvReviewed{2,485}
\def\EvVerified{1,975}
\def\EvGoldEvents{2,605}
\def\EvGoldInter{3,253}
\def\EvGoldPart{8,115}
\def\EvGoldPairs{1,180}
\def\EvTotalK{29.8K}
\def\EvGoldK{1.6K}
\def\EvRmHuman{320}
\def\EvRmNoEv{111}
\def\EvRmCtwo{743}
\def\EvRmCfour{28}
\def\EvSpKeptHuman{1,639}   %
\def\EvSpKeptAuto{2,262}   %
\def\EvSpCtwoHuman{338}
\def\EvSpCtwoAuto{405}
\def\EvSpCfourHuman{7}
\def\EvSpCfourAuto{21}
\def\EvSpNoEvHuman{8}
\def\EvSpNoEvAuto{103}
\def\EvSpHumanRej{320}
\def\EvCapHx{47.94}
\def\EvCapHw{3.18}
\def\EvCapHpct{3}
\def\EvCapBars{0/9/{29.43,18.51,3.18,12.38,28.35,8.15},1/0/{12.10,8.53,2.70,15.63,40.62,20.43}}   %

\begin{table*}[t]
\centering
\footnotesize
\setlength{\tabcolsep}{2.6pt}
\resizebox{\textwidth}{!}{%
\begin{tabular}{@{}l c l ccccc ccc@{}}
\toprule
& & & \multicolumn{5}{c}{Causal evidence} & \multicolumn{3}{c}{Error diagnosis} \\
\cmidrule(lr){4-8}\cmidrule(lr){9-11}
Benchmark & Ego & Causal capabilities & \shortstack{Key-event\\intervals} & \shortstack{Participant\\boxes} & \shortstack{Participant\\roles} & \shortstack{Interaction\\association} & \shortstack{Depen-\\dencies} & \shortstack{Box$\,\to\,$role\\$\to\,$association} & \shortstack{Event pair\\$\to\,$dependency} & \shortstack{Grounded vs\\spurious} \\
\midrule
Causal-VidQA~\cite{li2022causalvidqa} & \xmark & Exp, CF, Ant & \xmark & \xmark & \xmark & \xmark & \xmark & \xmark & \xmark & \hmark \\
NExT-GQA~\cite{xiao2024nextgqa} & \xmark & Exp & \cmark & \xmark & \xmark & \xmark & \xmark & \xmark & \xmark & \cmark \\
CaST-Bench~\cite{zhang2026castbench} & \xmark & Exp, CF, Ant, Goal & \cmark & \cmark & \xmark & \xmark & \xmark & \xmark & \xmark & \cmark \\
V-STaR~\cite{cheng2025vstar} & \xmark & -- & \cmark & \cmark & \xmark & \xmark & \xmark & \xmark & \xmark & \cmark \\
\midrule
EgoTaskQA~\cite{jia2022egotaskqa} & \cmark & Exp, CF, Hyp, Ant, Goal & \xmark & \xmark & \xmark & \xmark & \hmark$^{a}$ & \xmark & \xmark & \xmark \\
HourVideo~\cite{chandrasegaran2024hourvideo} & \cmark & Exp, CF, Ant & \xmark & \xmark & \xmark & \xmark & \xmark & \xmark & \xmark & \xmark \\
HD-EPIC~\cite{perrett2025hdepic} & \cmark & Goal, Ant & \xmark$^{b}$ & \xmark & \xmark & \xmark & \xmark & \xmark & \xmark & \xmark \\
EgoPlan-Bench2~\cite{qiu2024egoplanbench2} & \cmark & Pln & \xmark & \xmark & \xmark & \xmark & \xmark & \xmark & \xmark & \xmark \\
CausalVQA~\cite{foss2025causalvqa} & \cmark & CF, Hyp, Ant, Pln & \xmark & \xmark & \xmark & \xmark & \xmark & \xmark & \xmark & \xmark \\
MultiHop-EgoQA~\cite{gelm2025multihop} & \cmark & -- & \cmark & \xmark & \xmark & \xmark & \xmark & \xmark & \xmark & \xmark \\
ProcObject-10K~\cite{guo2026procobject} & \cmark$^{c}$ & Exp, CF & \cmark & \xmark & \xmark & \xmark & \xmark & \xmark & \xmark & \xmark \\
Minerva-Ego~\cite{nagrani2026minervaego} & \cmark & Exp, CF, Goal & \hmark & \hmark$^{d}$ & \xmark & \xmark & \xmark & \xmark & \xmark & \xmark \\
EgoCoT-Bench~\cite{dai2026egocotbench} & \cmark & Exp, Ant & \hmark & \hmark$^{e}$ & \xmark & \hmark & \xmark & \xmark & \xmark & \hmark \\
\midrule
\bench (ours) & \cmark & Exp, CF, Hyp, Ant, Pln, Goal & \cmark & \cmark & \cmark & \cmark & \cmark & \cmark & \cmark & \cmark \\
\bottomrule
\end{tabular}}
\caption{\textbf{Comparison with related benchmarks.}
Causal capabilities: question types mapped to our six capabilities by their definitions, for our sources by the template classification of \cref{sec:bench_filter} (Exp: explanation, CF: counterfactual, Hyp: hypothetical, Ant: anticipation, Pln: planning, Goal: goal attribution).
\cmark{}: annotated per question and evaluated on model predictions. \hmark{}: annotated only or evaluated indirectly.
$^{a}$Dependencies used only for question construction. $^{b}$Standalone localization questions. $^{c}$Egocentric and exocentric videos. $^{d}$Spatio-temporal object masks annotated, not scored. $^{e}$Boxes annotated and tested only as multiple-choice options, not predicted or scored by IoU.
Sizes, annotation and QA formats: Appx.~A.}
\label{tab:bench_compare}
\end{table*}

\paragraph{Benchmarks for causal and grounded video QA.}
Causal VideoQA benchmarks assess explanation, anticipation and counterfactual reasoning~\cite{li2022causalvidqa,xiao2021nextqa}, and egocentric benchmarks extend them to task dependencies, object dynamics, goals and planning~\cite{chandrasegaran2024hourvideo,foss2025causalvqa,jia2022egotaskqa,perrett2025hdepic,qiu2024egoplanbench2}, mostly scoring answers even when structured annotations build the questions~\cite{jia2022egotaskqa,perrett2025hdepic}.
Grounded VideoQA benchmarks also assess temporal evidence~\cite{xiao2024nextgqa,gelm2025multihop,guo2026procobject}, spatio-temporal evidence~\cite{cheng2025vstar} or reasoning traces~\cite{nagrani2025minerva,nagrani2026minervaego}.
Closest to ours, CaST-Bench~\cite{zhang2026castbench} distinguishes grounded from spurious answers through causal evidence localization, and EgoCoT-Bench~\cite{dai2026egocotbench} tests hand--object association through multiple-choice questions and rationale judging.
\bench instead scores key events, participant locations, roles and associations, and typed dependencies (\cref{tab:bench_compare}).

\paragraph{Models for causal and grounded video reasoning.}
Video reasoning models improve temporal grounding through joint localization and answering~\cite{di2024groundvqa}, agentic retrieval~\cite{videomind2025} or iterative perception~\cite{videochatr15_2025}, and spatio-temporal grounding through fine-grained representations~\cite{li2025llavast}, grounded reasoning traces~\cite{meng2026openo3video}, object-aware rewards~\cite{visioncoach2026} or hand--object supervision~\cite{egothinker2025}, without modeling how participants form distinct interactions.
Causal reasoning models add physical commonsense~\cite{nvidia2025cosmosreason1} or build event-level causal graphs by dense causal captioning~\cite{nadeem2026causalvlm} and causal discovery~\cite{chen2025mecdplus}.
\method instead predicts multiple key events, interaction-level participants and event dependencies before answering.

\section{Benchmark and Method}
\label{sec:bench}

\newif\iffigtwoframes\figtwoframestrue
\providecommand{\ftiny}{\fontsize{6.0}{7.0}\selectfont}
\providecommand{\ftframe}[3]{\iffigtwoframes\IfFileExists{#3}{\includegraphics[width=#1,height=#2]{#3}}{\tikz\fill[black!25](0,0)rectangle(#1,#2);}\else\tikz\fill[black!25](0,0)rectangle(#1,#2);\fi}
\providecommand{\ftcrop}[4]{\iffigtwoframes\IfFileExists{#3}{\includegraphics[width=#1,height=#2,trim=#4,clip]{#3}}{\tikz\fill[black!25](0,0)rectangle(#1,#2);}\else\tikz\fill[black!25](0,0)rectangle(#1,#2);\fi}
\providecolor{pDk}{HTML}{306692}\providecolor{pMd}{HTML}{7396C6}\providecolor{pE1}{HTML}{7EB2DA}\providecolor{pE2}{HTML}{7E88AF}
\providecolor{pLt}{HTML}{AEBCE0}\providecolor{pBg}{HTML}{DBE0EB}\providecolor{pBg2}{HTML}{D5DEE6}\providecolor{pGy}{HTML}{C0C4C9}\providecolor{pTx}{HTML}{1B2A41}
\providecolor{tHand}{HTML}{E67E22}\providecolor{tTool}{HTML}{C2185B}\providecolor{tTarg}{HTML}{2E8B57}\providecolor{tGrn}{HTML}{1E7B34}\providecolor{tRed}{HTML}{C0392B}
\providecolor{oA}{HTML}{3B6FB6}\providecolor{oB}{HTML}{8E5BB5}\providecolor{oC}{HTML}{2A9DA6}\providecolor{oD}{HTML}{8A8F99}
\providecolor{mDet}{HTML}{B35C1E}\providecolor{mSam}{HTML}{2F7D5B}\providecolor{mGebd}{HTML}{6B4C9A}\providecolor{mVlm}{HTML}{306692}\providecolor{mLlm}{HTML}{8A3B5E}\providecolor{mHum}{HTML}{4A4A4A}
\usetikzlibrary{decorations.pathreplacing}

\begin{figure*}[t]
    \centering
    \includegraphics[width=0.88\textwidth]{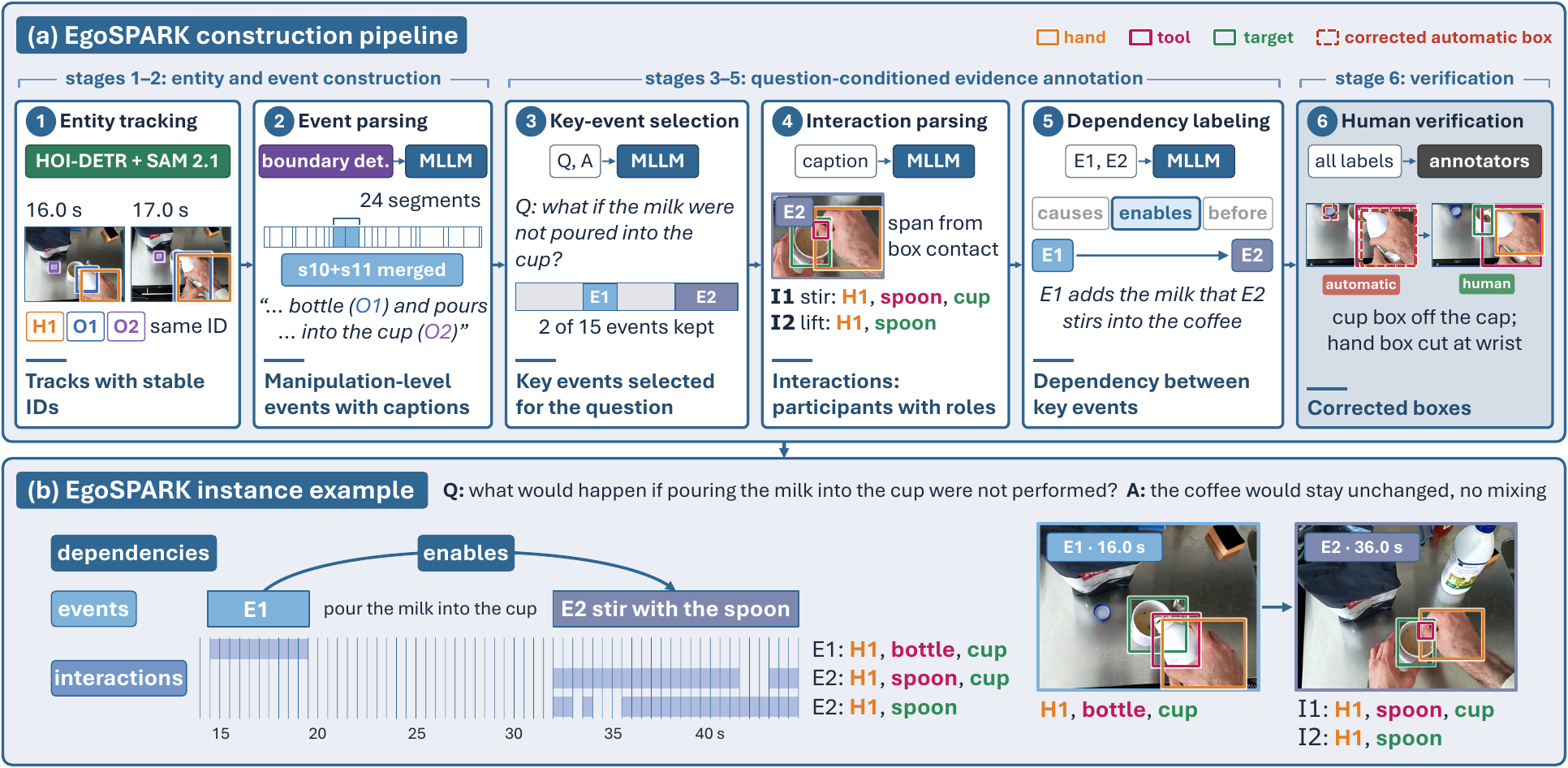}
    \caption{\textbf{\bench construction pipeline (a) and an instance example (b).} The question asks what would happen if the milk were not poured into the cup: pouring the milk (E1) enables stirring it into the coffee (E2), and the annotators (stage~6) correct the cup and hand boxes of the automatic labels. A detailed version is in Appx.~A.}
    \label{fig:pipeline}
\end{figure*}  %
\def\ANNEvents{2,605}\def\ANNInter{3,253}\def\ANConcInter{907}\def\ANConcShare{27.9}\def\ANConcEvents{412}\def\ANNDepPairs{367}\def\ANNoDepQ{81.6}\def\ANNQ{1,639}\def\ANConcPartShare{28.6}
\providecommand{\AN}[2]{\ifcsname an@#1@#2\endcsname\csname an@#1@#2\endcsname\else\TODO{}\fi}
\expandafter\def\csname an@Base2B@hand_loc\endcsname{0.1}
\expandafter\def\csname an@Base2B@tool_loc\endcsname{0.0}
\expandafter\def\csname an@Base2B@target_loc\endcsname{0.7}
\expandafter\def\csname an@Base2B@hand_role\endcsname{--}
\expandafter\def\csname an@Base2B@tool_role\endcsname{--}
\expandafter\def\csname an@Base2B@target_role\endcsname{--}
\expandafter\def\csname an@Base2B@hand_joint\endcsname{0.1}
\expandafter\def\csname an@Base2B@tool_joint\endcsname{0.0}
\expandafter\def\csname an@Base2B@target_joint\endcsname{0.0}
\expandafter\def\csname an@Base2B@tt_conf\endcsname{--}
\expandafter\def\csname an@Base2B@n_hand\endcsname{2}
\expandafter\def\csname an@Base2B@n_tool\endcsname{0}
\expandafter\def\csname an@Base2B@n_target\endcsname{23}
\expandafter\def\csname an@Base2B@n_obj\endcsname{23}
\expandafter\def\csname an@Base2B@conc_found\endcsname{180}
\expandafter\def\csname an@Base2B@conc_multi\endcsname{0.0}
\expandafter\def\csname an@Base2B@conc_multi_same_tok\endcsname{0.0}
\expandafter\def\csname an@Base2B@conc_sep\endcsname{0.0}
\expandafter\def\csname an@Base2B@conc_miss_loc\endcsname{99.0}
\expandafter\def\csname an@Base2B@conc_merged\endcsname{0.0}
\expandafter\def\csname an@Base2B@conc_wrong_role\endcsname{1.0}
\expandafter\def\csname an@Base2B@conc_other\endcsname{0.0}
\expandafter\def\csname an@Base2B@l3_conc_share\endcsname{--}
\expandafter\def\csname an@Base2B@n_notrec\endcsname{390}
\expandafter\def\csname an@Base2B@n_conc_sep\endcsname{0}
\expandafter\def\csname an@Base2B@dep_miss\endcsname{87.7}
\expandafter\def\csname an@Base2B@dep_none\endcsname{8.2}
\expandafter\def\csname an@Base2B@dep_type\endcsname{0.3}
\expandafter\def\csname an@Base2B@dep_ok\endcsname{3.8}
\expandafter\def\csname an@Base2B@std_loss_dep\endcsname{0.0}
\expandafter\def\csname an@Base2B@std_loss_nodep\endcsname{0.0}
\expandafter\def\csname an@Base2B@d_vacuous\endcsname{--}
\expandafter\def\csname an@Base2B@n_std\endcsname{1}
\expandafter\def\csname an@Base2B@n_std_vacuous\endcsname{1}
\expandafter\def\csname an@Base2B@d_rate_vacuous\endcsname{98.5}
\expandafter\def\csname an@Base2B@n_D\endcsname{791}
\expandafter\def\csname an@Base2B@c_tfail\endcsname{77.1}
\expandafter\def\csname an@Base2B@c_tonly\endcsname{22.4}
\expandafter\def\csname an@Base2B@c_tsfail\endcsname{0.0}
\expandafter\def\csname an@Base2B@c_tsd\endcsname{0.5}
\expandafter\def\csname an@Base2B@c_evid_wrong\endcsname{--}
\expandafter\def\csname an@Base2B@n_correct\endcsname{192}
\expandafter\def\csname an@Base2B@n_ts\endcsname{1}
\expandafter\def\csname an@Base2B@c_tfail_CE\endcsname{--}
\expandafter\def\csname an@Base2B@c_tonly_CE\endcsname{--}
\expandafter\def\csname an@Base2B@c_tsfail_CE\endcsname{--}
\expandafter\def\csname an@Base2B@c_tsd_CE\endcsname{--}
\expandafter\def\csname an@Base2B@n_correct_CE\endcsname{37}
\expandafter\def\csname an@Base2B@c_tfail_CF\endcsname{84.8}
\expandafter\def\csname an@Base2B@c_tonly_CF\endcsname{15.2}
\expandafter\def\csname an@Base2B@c_tsfail_CF\endcsname{0.0}
\expandafter\def\csname an@Base2B@c_tsd_CF\endcsname{0.0}
\expandafter\def\csname an@Base2B@n_correct_CF\endcsname{46}
\expandafter\def\csname an@Base2B@c_tfail_H\endcsname{--}
\expandafter\def\csname an@Base2B@c_tonly_H\endcsname{--}
\expandafter\def\csname an@Base2B@c_tsfail_H\endcsname{--}
\expandafter\def\csname an@Base2B@c_tsd_H\endcsname{--}
\expandafter\def\csname an@Base2B@n_correct_H\endcsname{5}
\expandafter\def\csname an@Base2B@c_tfail_A\endcsname{--}
\expandafter\def\csname an@Base2B@c_tonly_A\endcsname{--}
\expandafter\def\csname an@Base2B@c_tsfail_A\endcsname{--}
\expandafter\def\csname an@Base2B@c_tsd_A\endcsname{--}
\expandafter\def\csname an@Base2B@n_correct_A\endcsname{21}
\expandafter\def\csname an@Base2B@c_tfail_P\endcsname{--}
\expandafter\def\csname an@Base2B@c_tonly_P\endcsname{--}
\expandafter\def\csname an@Base2B@c_tsfail_P\endcsname{--}
\expandafter\def\csname an@Base2B@c_tsd_P\endcsname{--}
\expandafter\def\csname an@Base2B@n_correct_P\endcsname{37}
\expandafter\def\csname an@Base2B@c_tfail_G\endcsname{67.4}
\expandafter\def\csname an@Base2B@c_tonly_G\endcsname{30.4}
\expandafter\def\csname an@Base2B@c_tsfail_G\endcsname{0.0}
\expandafter\def\csname an@Base2B@c_tsd_G\endcsname{2.2}
\expandafter\def\csname an@Base2B@n_correct_G\endcsname{46}
\expandafter\def\csname an@Base9B@hand_loc\endcsname{0.5}
\expandafter\def\csname an@Base9B@tool_loc\endcsname{1.5}
\expandafter\def\csname an@Base9B@target_loc\endcsname{1.7}
\expandafter\def\csname an@Base9B@hand_role\endcsname{--}
\expandafter\def\csname an@Base9B@tool_role\endcsname{--}
\expandafter\def\csname an@Base9B@target_role\endcsname{53.7}
\expandafter\def\csname an@Base9B@hand_joint\endcsname{0.4}
\expandafter\def\csname an@Base9B@tool_joint\endcsname{0.7}
\expandafter\def\csname an@Base9B@target_joint\endcsname{0.9}
\expandafter\def\csname an@Base9B@tt_conf\endcsname{25.4}
\expandafter\def\csname an@Base9B@n_hand\endcsname{21}
\expandafter\def\csname an@Base9B@n_tool\endcsname{9}
\expandafter\def\csname an@Base9B@n_target\endcsname{54}
\expandafter\def\csname an@Base9B@n_obj\endcsname{63}
\expandafter\def\csname an@Base9B@conc_found\endcsname{209}
\expandafter\def\csname an@Base9B@conc_multi\endcsname{12.9}
\expandafter\def\csname an@Base9B@conc_multi_same_tok\endcsname{0.0}
\expandafter\def\csname an@Base9B@conc_sep\endcsname{0.5}
\expandafter\def\csname an@Base9B@conc_miss_loc\endcsname{96.4}
\expandafter\def\csname an@Base9B@conc_merged\endcsname{0.9}
\expandafter\def\csname an@Base9B@conc_wrong_role\endcsname{2.7}
\expandafter\def\csname an@Base9B@conc_other\endcsname{0.0}
\expandafter\def\csname an@Base9B@l3_conc_share\endcsname{18.0}
\expandafter\def\csname an@Base9B@n_notrec\endcsname{450}
\expandafter\def\csname an@Base9B@n_conc_sep\endcsname{1}
\expandafter\def\csname an@Base9B@dep_miss\endcsname{89.4}
\expandafter\def\csname an@Base9B@dep_none\endcsname{3.3}
\expandafter\def\csname an@Base9B@dep_type\endcsname{1.1}
\expandafter\def\csname an@Base9B@dep_ok\endcsname{6.3}
\expandafter\def\csname an@Base9B@std_loss_dep\endcsname{0.0}
\expandafter\def\csname an@Base9B@std_loss_nodep\endcsname{0.0}
\expandafter\def\csname an@Base9B@d_vacuous\endcsname{--}
\expandafter\def\csname an@Base9B@n_std\endcsname{3}
\expandafter\def\csname an@Base9B@n_std_vacuous\endcsname{3}
\expandafter\def\csname an@Base9B@d_rate_vacuous\endcsname{96.9}
\expandafter\def\csname an@Base9B@n_D\endcsname{637}
\expandafter\def\csname an@Base9B@c_tfail\endcsname{67.2}
\expandafter\def\csname an@Base9B@c_tonly\endcsname{31.7}
\expandafter\def\csname an@Base9B@c_tsfail\endcsname{0.0}
\expandafter\def\csname an@Base9B@c_tsd\endcsname{1.1}
\expandafter\def\csname an@Base9B@c_evid_wrong\endcsname{--}
\expandafter\def\csname an@Base9B@n_correct\endcsname{262}
\expandafter\def\csname an@Base9B@n_ts\endcsname{5}
\expandafter\def\csname an@Base9B@c_tfail_CE\endcsname{63.3}
\expandafter\def\csname an@Base9B@c_tonly_CE\endcsname{35.0}
\expandafter\def\csname an@Base9B@c_tsfail_CE\endcsname{0.0}
\expandafter\def\csname an@Base9B@c_tsd_CE\endcsname{1.7}
\expandafter\def\csname an@Base9B@n_correct_CE\endcsname{60}
\expandafter\def\csname an@Base9B@c_tfail_CF\endcsname{73.3}
\expandafter\def\csname an@Base9B@c_tonly_CF\endcsname{25.0}
\expandafter\def\csname an@Base9B@c_tsfail_CF\endcsname{0.0}
\expandafter\def\csname an@Base9B@c_tsd_CF\endcsname{1.7}
\expandafter\def\csname an@Base9B@n_correct_CF\endcsname{60}
\expandafter\def\csname an@Base9B@c_tfail_H\endcsname{--}
\expandafter\def\csname an@Base9B@c_tonly_H\endcsname{--}
\expandafter\def\csname an@Base9B@c_tsfail_H\endcsname{--}
\expandafter\def\csname an@Base9B@c_tsd_H\endcsname{--}
\expandafter\def\csname an@Base9B@n_correct_H\endcsname{13}
\expandafter\def\csname an@Base9B@c_tfail_A\endcsname{--}
\expandafter\def\csname an@Base9B@c_tonly_A\endcsname{--}
\expandafter\def\csname an@Base9B@c_tsfail_A\endcsname{--}
\expandafter\def\csname an@Base9B@c_tsd_A\endcsname{--}
\expandafter\def\csname an@Base9B@n_correct_A\endcsname{32}
\expandafter\def\csname an@Base9B@c_tfail_P\endcsname{76.5}
\expandafter\def\csname an@Base9B@c_tonly_P\endcsname{23.5}
\expandafter\def\csname an@Base9B@c_tsfail_P\endcsname{0.0}
\expandafter\def\csname an@Base9B@c_tsd_P\endcsname{0.0}
\expandafter\def\csname an@Base9B@n_correct_P\endcsname{51}
\expandafter\def\csname an@Base9B@c_tfail_G\endcsname{60.9}
\expandafter\def\csname an@Base9B@c_tonly_G\endcsname{39.1}
\expandafter\def\csname an@Base9B@c_tsfail_G\endcsname{0.0}
\expandafter\def\csname an@Base9B@c_tsd_G\endcsname{0.0}
\expandafter\def\csname an@Base9B@n_correct_G\endcsname{46}
\expandafter\def\csname an@Q36@hand_loc\endcsname{10.3}
\expandafter\def\csname an@Q36@tool_loc\endcsname{16.6}
\expandafter\def\csname an@Q36@target_loc\endcsname{13.6}
\expandafter\def\csname an@Q36@hand_role\endcsname{91.1}
\expandafter\def\csname an@Q36@tool_role\endcsname{78.0}
\expandafter\def\csname an@Q36@target_role\endcsname{73.8}
\expandafter\def\csname an@Q36@hand_joint\endcsname{9.4}
\expandafter\def\csname an@Q36@tool_joint\endcsname{12.9}
\expandafter\def\csname an@Q36@target_joint\endcsname{10.1}
\expandafter\def\csname an@Q36@tt_conf\endcsname{12.4}
\expandafter\def\csname an@Q36@n_hand\endcsname{404}
\expandafter\def\csname an@Q36@n_tool\endcsname{100}
\expandafter\def\csname an@Q36@n_target\endcsname{442}
\expandafter\def\csname an@Q36@n_obj\endcsname{542}
\expandafter\def\csname an@Q36@conc_found\endcsname{290}
\expandafter\def\csname an@Q36@conc_multi\endcsname{18.6}
\expandafter\def\csname an@Q36@conc_multi_same_tok\endcsname{0.0}
\expandafter\def\csname an@Q36@conc_sep\endcsname{2.8}
\expandafter\def\csname an@Q36@conc_miss_loc\endcsname{79.9}
\expandafter\def\csname an@Q36@conc_merged\endcsname{12.1}
\expandafter\def\csname an@Q36@conc_wrong_role\endcsname{8.0}
\expandafter\def\csname an@Q36@conc_other\endcsname{0.0}
\expandafter\def\csname an@Q36@l3_conc_share\endcsname{24.3}
\expandafter\def\csname an@Q36@n_notrec\endcsname{513}
\expandafter\def\csname an@Q36@n_conc_sep\endcsname{8}
\expandafter\def\csname an@Q36@dep_miss\endcsname{66.2}
\expandafter\def\csname an@Q36@dep_none\endcsname{9.3}
\expandafter\def\csname an@Q36@dep_type\endcsname{1.4}
\expandafter\def\csname an@Q36@dep_ok\endcsname{23.2}
\expandafter\def\csname an@Q36@std_loss_dep\endcsname{0.2}
\expandafter\def\csname an@Q36@std_loss_nodep\endcsname{0.4}
\expandafter\def\csname an@Q36@d_vacuous\endcsname{--}
\expandafter\def\csname an@Q36@n_std\endcsname{8}
\expandafter\def\csname an@Q36@n_std_vacuous\endcsname{4}
\expandafter\def\csname an@Q36@d_rate_vacuous\endcsname{48.0}
\expandafter\def\csname an@Q36@n_D\endcsname{148}
\expandafter\def\csname an@Q36@c_tfail\endcsname{76.6}
\expandafter\def\csname an@Q36@c_tonly\endcsname{18.7}
\expandafter\def\csname an@Q36@c_tsfail\endcsname{2.5}
\expandafter\def\csname an@Q36@c_tsd\endcsname{2.2}
\expandafter\def\csname an@Q36@c_evid_wrong\endcsname{73.4}
\expandafter\def\csname an@Q36@n_correct\endcsname{363}
\expandafter\def\csname an@Q36@n_ts\endcsname{64}
\expandafter\def\csname an@Q36@c_tfail_CE\endcsname{67.7}
\expandafter\def\csname an@Q36@c_tonly_CE\endcsname{23.7}
\expandafter\def\csname an@Q36@c_tsfail_CE\endcsname{5.4}
\expandafter\def\csname an@Q36@c_tsd_CE\endcsname{3.2}
\expandafter\def\csname an@Q36@n_correct_CE\endcsname{93}
\expandafter\def\csname an@Q36@c_tfail_CF\endcsname{77.8}
\expandafter\def\csname an@Q36@c_tonly_CF\endcsname{19.4}
\expandafter\def\csname an@Q36@c_tsfail_CF\endcsname{0.0}
\expandafter\def\csname an@Q36@c_tsd_CF\endcsname{2.8}
\expandafter\def\csname an@Q36@n_correct_CF\endcsname{72}
\expandafter\def\csname an@Q36@c_tfail_H\endcsname{--}
\expandafter\def\csname an@Q36@c_tonly_H\endcsname{--}
\expandafter\def\csname an@Q36@c_tsfail_H\endcsname{--}
\expandafter\def\csname an@Q36@c_tsd_H\endcsname{--}
\expandafter\def\csname an@Q36@n_correct_H\endcsname{17}
\expandafter\def\csname an@Q36@c_tfail_A\endcsname{85.5}
\expandafter\def\csname an@Q36@c_tonly_A\endcsname{12.7}
\expandafter\def\csname an@Q36@c_tsfail_A\endcsname{0.0}
\expandafter\def\csname an@Q36@c_tsd_A\endcsname{1.8}
\expandafter\def\csname an@Q36@n_correct_A\endcsname{55}
\expandafter\def\csname an@Q36@c_tfail_P\endcsname{85.5}
\expandafter\def\csname an@Q36@c_tonly_P\endcsname{12.9}
\expandafter\def\csname an@Q36@c_tsfail_P\endcsname{1.6}
\expandafter\def\csname an@Q36@c_tsd_P\endcsname{0.0}
\expandafter\def\csname an@Q36@n_correct_P\endcsname{62}
\expandafter\def\csname an@Q36@c_tfail_G\endcsname{68.8}
\expandafter\def\csname an@Q36@c_tonly_G\endcsname{23.4}
\expandafter\def\csname an@Q36@c_tsfail_G\endcsname{4.7}
\expandafter\def\csname an@Q36@c_tsd_G\endcsname{3.1}
\expandafter\def\csname an@Q36@n_correct_G\endcsname{64}
\expandafter\def\csname an@Q38@hand_loc\endcsname{14.2}
\expandafter\def\csname an@Q38@tool_loc\endcsname{22.6}
\expandafter\def\csname an@Q38@target_loc\endcsname{18.2}
\expandafter\def\csname an@Q38@hand_role\endcsname{89.6}
\expandafter\def\csname an@Q38@tool_role\endcsname{83.1}
\expandafter\def\csname an@Q38@target_role\endcsname{81.0}
\expandafter\def\csname an@Q38@hand_joint\endcsname{12.7}
\expandafter\def\csname an@Q38@tool_joint\endcsname{18.7}
\expandafter\def\csname an@Q38@target_joint\endcsname{14.7}
\expandafter\def\csname an@Q38@tt_conf\endcsname{9.8}
\expandafter\def\csname an@Q38@n_hand\endcsname{558}
\expandafter\def\csname an@Q38@n_tool\endcsname{136}
\expandafter\def\csname an@Q38@n_target\endcsname{590}
\expandafter\def\csname an@Q38@n_obj\endcsname{726}
\expandafter\def\csname an@Q38@conc_found\endcsname{252}
\expandafter\def\csname an@Q38@conc_multi\endcsname{34.5}
\expandafter\def\csname an@Q38@conc_multi_same_tok\endcsname{0.0}
\expandafter\def\csname an@Q38@conc_sep\endcsname{7.1}
\expandafter\def\csname an@Q38@conc_miss_loc\endcsname{64.4}
\expandafter\def\csname an@Q38@conc_merged\endcsname{25.5}
\expandafter\def\csname an@Q38@conc_wrong_role\endcsname{10.1}
\expandafter\def\csname an@Q38@conc_other\endcsname{0.0}
\expandafter\def\csname an@Q38@l3_conc_share\endcsname{27.5}
\expandafter\def\csname an@Q38@n_notrec\endcsname{368}
\expandafter\def\csname an@Q38@n_conc_sep\endcsname{18}
\expandafter\def\csname an@Q38@dep_miss\endcsname{77.7}
\expandafter\def\csname an@Q38@dep_none\endcsname{5.7}
\expandafter\def\csname an@Q38@dep_type\endcsname{1.6}
\expandafter\def\csname an@Q38@dep_ok\endcsname{15.0}
\expandafter\def\csname an@Q38@std_loss_dep\endcsname{0.5}
\expandafter\def\csname an@Q38@std_loss_nodep\endcsname{0.5}
\expandafter\def\csname an@Q38@d_vacuous\endcsname{--}
\expandafter\def\csname an@Q38@n_std\endcsname{10}
\expandafter\def\csname an@Q38@n_std_vacuous\endcsname{7}
\expandafter\def\csname an@Q38@d_rate_vacuous\endcsname{74.8}
\expandafter\def\csname an@Q38@n_D\endcsname{210}
\expandafter\def\csname an@Q38@c_tfail\endcsname{72.6}
\expandafter\def\csname an@Q38@c_tonly\endcsname{19.8}
\expandafter\def\csname an@Q38@c_tsfail\endcsname{4.7}
\expandafter\def\csname an@Q38@c_tsd\endcsname{2.9}
\expandafter\def\csname an@Q38@c_evid_wrong\endcsname{74.5}
\expandafter\def\csname an@Q38@n_correct\endcsname{343}
\expandafter\def\csname an@Q38@n_ts\endcsname{102}
\expandafter\def\csname an@Q38@c_tfail_CE\endcsname{66.7}
\expandafter\def\csname an@Q38@c_tonly_CE\endcsname{26.2}
\expandafter\def\csname an@Q38@c_tsfail_CE\endcsname{6.0}
\expandafter\def\csname an@Q38@c_tsd_CE\endcsname{1.2}
\expandafter\def\csname an@Q38@n_correct_CE\endcsname{84}
\expandafter\def\csname an@Q38@c_tfail_CF\endcsname{69.7}
\expandafter\def\csname an@Q38@c_tonly_CF\endcsname{20.2}
\expandafter\def\csname an@Q38@c_tsfail_CF\endcsname{3.4}
\expandafter\def\csname an@Q38@c_tsd_CF\endcsname{6.7}
\expandafter\def\csname an@Q38@n_correct_CF\endcsname{89}
\expandafter\def\csname an@Q38@c_tfail_H\endcsname{--}
\expandafter\def\csname an@Q38@c_tonly_H\endcsname{--}
\expandafter\def\csname an@Q38@c_tsfail_H\endcsname{--}
\expandafter\def\csname an@Q38@c_tsd_H\endcsname{--}
\expandafter\def\csname an@Q38@n_correct_H\endcsname{8}
\expandafter\def\csname an@Q38@c_tfail_A\endcsname{81.5}
\expandafter\def\csname an@Q38@c_tonly_A\endcsname{13.0}
\expandafter\def\csname an@Q38@c_tsfail_A\endcsname{1.9}
\expandafter\def\csname an@Q38@c_tsd_A\endcsname{3.7}
\expandafter\def\csname an@Q38@n_correct_A\endcsname{54}
\expandafter\def\csname an@Q38@c_tfail_P\endcsname{81.0}
\expandafter\def\csname an@Q38@c_tonly_P\endcsname{12.7}
\expandafter\def\csname an@Q38@c_tsfail_P\endcsname{4.8}
\expandafter\def\csname an@Q38@c_tsd_P\endcsname{1.6}
\expandafter\def\csname an@Q38@n_correct_P\endcsname{63}
\expandafter\def\csname an@Q38@c_tfail_G\endcsname{68.9}
\expandafter\def\csname an@Q38@c_tonly_G\endcsname{22.2}
\expandafter\def\csname an@Q38@c_tsfail_G\endcsname{8.9}
\expandafter\def\csname an@Q38@c_tsd_G\endcsname{0.0}
\expandafter\def\csname an@Q38@n_correct_G\endcsname{45}
\expandafter\def\csname an@Q35L@hand_loc\endcsname{1.6}
\expandafter\def\csname an@Q35L@tool_loc\endcsname{2.3}
\expandafter\def\csname an@Q35L@target_loc\endcsname{4.0}
\expandafter\def\csname an@Q35L@hand_role\endcsname{66.7}
\expandafter\def\csname an@Q35L@tool_role\endcsname{--}
\expandafter\def\csname an@Q35L@target_role\endcsname{68.7}
\expandafter\def\csname an@Q35L@hand_joint\endcsname{1.1}
\expandafter\def\csname an@Q35L@tool_joint\endcsname{1.3}
\expandafter\def\csname an@Q35L@target_joint\endcsname{2.8}
\expandafter\def\csname an@Q35L@tt_conf\endcsname{15.2}
\expandafter\def\csname an@Q35L@n_hand\endcsname{63}
\expandafter\def\csname an@Q35L@n_tool\endcsname{14}
\expandafter\def\csname an@Q35L@n_target\endcsname{131}
\expandafter\def\csname an@Q35L@n_obj\endcsname{145}
\expandafter\def\csname an@Q35L@conc_found\endcsname{269}
\expandafter\def\csname an@Q35L@conc_multi\endcsname{6.7}
\expandafter\def\csname an@Q35L@conc_multi_same_tok\endcsname{0.0}
\expandafter\def\csname an@Q35L@conc_sep\endcsname{0.0}
\expandafter\def\csname an@Q35L@conc_miss_loc\endcsname{94.0}
\expandafter\def\csname an@Q35L@conc_merged\endcsname{1.4}
\expandafter\def\csname an@Q35L@conc_wrong_role\endcsname{4.6}
\expandafter\def\csname an@Q35L@conc_other\endcsname{0.0}
\expandafter\def\csname an@Q35L@l3_conc_share\endcsname{21.6}
\expandafter\def\csname an@Q35L@n_notrec\endcsname{568}
\expandafter\def\csname an@Q35L@n_conc_sep\endcsname{0}
\expandafter\def\csname an@Q35L@dep_miss\endcsname{75.5}
\expandafter\def\csname an@Q35L@dep_none\endcsname{7.6}
\expandafter\def\csname an@Q35L@dep_type\endcsname{2.2}
\expandafter\def\csname an@Q35L@dep_ok\endcsname{14.7}
\expandafter\def\csname an@Q35L@std_loss_dep\endcsname{0.0}
\expandafter\def\csname an@Q35L@std_loss_nodep\endcsname{0.0}
\expandafter\def\csname an@Q35L@d_vacuous\endcsname{--}
\expandafter\def\csname an@Q35L@n_std\endcsname{3}
\expandafter\def\csname an@Q35L@n_std_vacuous\endcsname{3}
\expandafter\def\csname an@Q35L@d_rate_vacuous\endcsname{75.7}
\expandafter\def\csname an@Q35L@n_D\endcsname{210}
\expandafter\def\csname an@Q35L@c_tfail\endcsname{74.8}
\expandafter\def\csname an@Q35L@c_tonly\endcsname{24.3}
\expandafter\def\csname an@Q35L@c_tsfail\endcsname{0.0}
\expandafter\def\csname an@Q35L@c_tsd\endcsname{0.9}
\expandafter\def\csname an@Q35L@c_evid_wrong\endcsname{--}
\expandafter\def\csname an@Q35L@n_correct\endcsname{321}
\expandafter\def\csname an@Q35L@n_ts\endcsname{7}
\expandafter\def\csname an@Q35L@c_tfail_CE\endcsname{75.0}
\expandafter\def\csname an@Q35L@c_tonly_CE\endcsname{23.8}
\expandafter\def\csname an@Q35L@c_tsfail_CE\endcsname{0.0}
\expandafter\def\csname an@Q35L@c_tsd_CE\endcsname{1.2}
\expandafter\def\csname an@Q35L@n_correct_CE\endcsname{80}
\expandafter\def\csname an@Q35L@c_tfail_CF\endcsname{81.2}
\expandafter\def\csname an@Q35L@c_tonly_CF\endcsname{17.4}
\expandafter\def\csname an@Q35L@c_tsfail_CF\endcsname{0.0}
\expandafter\def\csname an@Q35L@c_tsd_CF\endcsname{1.4}
\expandafter\def\csname an@Q35L@n_correct_CF\endcsname{69}
\expandafter\def\csname an@Q35L@c_tfail_H\endcsname{--}
\expandafter\def\csname an@Q35L@c_tonly_H\endcsname{--}
\expandafter\def\csname an@Q35L@c_tsfail_H\endcsname{--}
\expandafter\def\csname an@Q35L@c_tsd_H\endcsname{--}
\expandafter\def\csname an@Q35L@n_correct_H\endcsname{19}
\expandafter\def\csname an@Q35L@c_tfail_A\endcsname{--}
\expandafter\def\csname an@Q35L@c_tonly_A\endcsname{--}
\expandafter\def\csname an@Q35L@c_tsfail_A\endcsname{--}
\expandafter\def\csname an@Q35L@c_tsd_A\endcsname{--}
\expandafter\def\csname an@Q35L@n_correct_A\endcsname{37}
\expandafter\def\csname an@Q35L@c_tfail_P\endcsname{80.9}
\expandafter\def\csname an@Q35L@c_tonly_P\endcsname{19.1}
\expandafter\def\csname an@Q35L@c_tsfail_P\endcsname{0.0}
\expandafter\def\csname an@Q35L@c_tsd_P\endcsname{0.0}
\expandafter\def\csname an@Q35L@n_correct_P\endcsname{47}
\expandafter\def\csname an@Q35L@c_tfail_G\endcsname{59.4}
\expandafter\def\csname an@Q35L@c_tonly_G\endcsname{39.1}
\expandafter\def\csname an@Q35L@c_tsfail_G\endcsname{0.0}
\expandafter\def\csname an@Q35L@c_tsd_G\endcsname{1.4}
\expandafter\def\csname an@Q35L@n_correct_G\endcsname{69}
\expandafter\def\csname an@GLM46VF@hand_loc\endcsname{10.4}
\expandafter\def\csname an@GLM46VF@tool_loc\endcsname{13.3}
\expandafter\def\csname an@GLM46VF@target_loc\endcsname{10.7}
\expandafter\def\csname an@GLM46VF@hand_role\endcsname{87.3}
\expandafter\def\csname an@GLM46VF@tool_role\endcsname{56.2}
\expandafter\def\csname an@GLM46VF@target_role\endcsname{64.2}
\expandafter\def\csname an@GLM46VF@hand_joint\endcsname{9.1}
\expandafter\def\csname an@GLM46VF@tool_joint\endcsname{7.5}
\expandafter\def\csname an@GLM46VF@target_joint\endcsname{6.8}
\expandafter\def\csname an@GLM46VF@tt_conf\endcsname{22.3}
\expandafter\def\csname an@GLM46VF@n_hand\endcsname{409}
\expandafter\def\csname an@GLM46VF@n_tool\endcsname{80}
\expandafter\def\csname an@GLM46VF@n_target\endcsname{346}
\expandafter\def\csname an@GLM46VF@n_obj\endcsname{426}
\expandafter\def\csname an@GLM46VF@conc_found\endcsname{281}
\expandafter\def\csname an@GLM46VF@conc_multi\endcsname{21.4}
\expandafter\def\csname an@GLM46VF@conc_multi_same_tok\endcsname{0.0}
\expandafter\def\csname an@GLM46VF@conc_sep\endcsname{3.6}
\expandafter\def\csname an@GLM46VF@conc_miss_loc\endcsname{81.1}
\expandafter\def\csname an@GLM46VF@conc_merged\endcsname{11.9}
\expandafter\def\csname an@GLM46VF@conc_wrong_role\endcsname{7.0}
\expandafter\def\csname an@GLM46VF@conc_other\endcsname{0.0}
\expandafter\def\csname an@GLM46VF@l3_conc_share\endcsname{28.8}
\expandafter\def\csname an@GLM46VF@n_notrec\endcsname{503}
\expandafter\def\csname an@GLM46VF@n_conc_sep\endcsname{10}
\expandafter\def\csname an@GLM46VF@dep_miss\endcsname{76.3}
\expandafter\def\csname an@GLM46VF@dep_none\endcsname{7.6}
\expandafter\def\csname an@GLM46VF@dep_type\endcsname{1.6}
\expandafter\def\csname an@GLM46VF@dep_ok\endcsname{14.4}
\expandafter\def\csname an@GLM46VF@std_loss_dep\endcsname{0.0}
\expandafter\def\csname an@GLM46VF@std_loss_nodep\endcsname{0.2}
\expandafter\def\csname an@GLM46VF@d_vacuous\endcsname{--}
\expandafter\def\csname an@GLM46VF@n_std\endcsname{2}
\expandafter\def\csname an@GLM46VF@n_std_vacuous\endcsname{1}
\expandafter\def\csname an@GLM46VF@d_rate_vacuous\endcsname{86.4}
\expandafter\def\csname an@GLM46VF@n_D\endcsname{360}
\expandafter\def\csname an@GLM46VF@c_tfail\endcsname{75.4}
\expandafter\def\csname an@GLM46VF@c_tonly\endcsname{22.9}
\expandafter\def\csname an@GLM46VF@c_tsfail\endcsname{1.1}
\expandafter\def\csname an@GLM46VF@c_tsd\endcsname{0.6}
\expandafter\def\csname an@GLM46VF@c_evid_wrong\endcsname{--}
\expandafter\def\csname an@GLM46VF@n_correct\endcsname{358}
\expandafter\def\csname an@GLM46VF@n_ts\endcsname{38}
\expandafter\def\csname an@GLM46VF@c_tfail_CE\endcsname{83.6}
\expandafter\def\csname an@GLM46VF@c_tonly_CE\endcsname{16.4}
\expandafter\def\csname an@GLM46VF@c_tsfail_CE\endcsname{0.0}
\expandafter\def\csname an@GLM46VF@c_tsd_CE\endcsname{0.0}
\expandafter\def\csname an@GLM46VF@n_correct_CE\endcsname{61}
\expandafter\def\csname an@GLM46VF@c_tfail_CF\endcsname{73.6}
\expandafter\def\csname an@GLM46VF@c_tonly_CF\endcsname{23.6}
\expandafter\def\csname an@GLM46VF@c_tsfail_CF\endcsname{2.8}
\expandafter\def\csname an@GLM46VF@c_tsd_CF\endcsname{0.0}
\expandafter\def\csname an@GLM46VF@n_correct_CF\endcsname{72}
\expandafter\def\csname an@GLM46VF@c_tfail_H\endcsname{--}
\expandafter\def\csname an@GLM46VF@c_tonly_H\endcsname{--}
\expandafter\def\csname an@GLM46VF@c_tsfail_H\endcsname{--}
\expandafter\def\csname an@GLM46VF@c_tsd_H\endcsname{--}
\expandafter\def\csname an@GLM46VF@n_correct_H\endcsname{26}
\expandafter\def\csname an@GLM46VF@c_tfail_A\endcsname{75.4}
\expandafter\def\csname an@GLM46VF@c_tonly_A\endcsname{23.1}
\expandafter\def\csname an@GLM46VF@c_tsfail_A\endcsname{0.0}
\expandafter\def\csname an@GLM46VF@c_tsd_A\endcsname{1.5}
\expandafter\def\csname an@GLM46VF@n_correct_A\endcsname{65}
\expandafter\def\csname an@GLM46VF@c_tfail_P\endcsname{60.5}
\expandafter\def\csname an@GLM46VF@c_tonly_P\endcsname{37.2}
\expandafter\def\csname an@GLM46VF@c_tsfail_P\endcsname{0.0}
\expandafter\def\csname an@GLM46VF@c_tsd_P\endcsname{2.3}
\expandafter\def\csname an@GLM46VF@n_correct_P\endcsname{43}
\expandafter\def\csname an@GLM46VF@c_tfail_G\endcsname{79.1}
\expandafter\def\csname an@GLM46VF@c_tonly_G\endcsname{19.8}
\expandafter\def\csname an@GLM46VF@c_tsfail_G\endcsname{1.1}
\expandafter\def\csname an@GLM46VF@c_tsd_G\endcsname{0.0}
\expandafter\def\csname an@GLM46VF@n_correct_G\endcsname{91}
\expandafter\def\csname an@Gemma4@hand_loc\endcsname{5.3}
\expandafter\def\csname an@Gemma4@tool_loc\endcsname{13.6}
\expandafter\def\csname an@Gemma4@target_loc\endcsname{9.7}
\expandafter\def\csname an@Gemma4@hand_role\endcsname{78.4}
\expandafter\def\csname an@Gemma4@tool_role\endcsname{64.6}
\expandafter\def\csname an@Gemma4@target_role\endcsname{74.4}
\expandafter\def\csname an@Gemma4@hand_joint\endcsname{4.2}
\expandafter\def\csname an@Gemma4@tool_joint\endcsname{8.8}
\expandafter\def\csname an@Gemma4@target_joint\endcsname{7.2}
\expandafter\def\csname an@Gemma4@tt_conf\endcsname{13.1}
\expandafter\def\csname an@Gemma4@n_hand\endcsname{208}
\expandafter\def\csname an@Gemma4@n_tool\endcsname{82}
\expandafter\def\csname an@Gemma4@n_target\endcsname{316}
\expandafter\def\csname an@Gemma4@n_obj\endcsname{398}
\expandafter\def\csname an@Gemma4@conc_found\endcsname{257}
\expandafter\def\csname an@Gemma4@conc_multi\endcsname{8.9}
\expandafter\def\csname an@Gemma4@conc_multi_same_tok\endcsname{0.0}
\expandafter\def\csname an@Gemma4@conc_sep\endcsname{0.8}
\expandafter\def\csname an@Gemma4@conc_miss_loc\endcsname{84.1}
\expandafter\def\csname an@Gemma4@conc_merged\endcsname{8.7}
\expandafter\def\csname an@Gemma4@conc_wrong_role\endcsname{7.2}
\expandafter\def\csname an@Gemma4@conc_other\endcsname{0.0}
\expandafter\def\csname an@Gemma4@l3_conc_share\endcsname{25.3}
\expandafter\def\csname an@Gemma4@n_notrec\endcsname{471}
\expandafter\def\csname an@Gemma4@n_conc_sep\endcsname{2}
\expandafter\def\csname an@Gemma4@dep_miss\endcsname{73.8}
\expandafter\def\csname an@Gemma4@dep_none\endcsname{8.2}
\expandafter\def\csname an@Gemma4@dep_type\endcsname{0.8}
\expandafter\def\csname an@Gemma4@dep_ok\endcsname{17.2}
\expandafter\def\csname an@Gemma4@std_loss_dep\endcsname{0.0}
\expandafter\def\csname an@Gemma4@std_loss_nodep\endcsname{0.0}
\expandafter\def\csname an@Gemma4@d_vacuous\endcsname{--}
\expandafter\def\csname an@Gemma4@n_std\endcsname{5}
\expandafter\def\csname an@Gemma4@n_std_vacuous\endcsname{2}
\expandafter\def\csname an@Gemma4@d_rate_vacuous\endcsname{82.1}
\expandafter\def\csname an@Gemma4@n_D\endcsname{313}
\expandafter\def\csname an@Gemma4@c_tfail\endcsname{71.1}
\expandafter\def\csname an@Gemma4@c_tonly\endcsname{27.3}
\expandafter\def\csname an@Gemma4@c_tsfail\endcsname{0.0}
\expandafter\def\csname an@Gemma4@c_tsd\endcsname{1.6}
\expandafter\def\csname an@Gemma4@c_evid_wrong\endcsname{--}
\expandafter\def\csname an@Gemma4@n_correct\endcsname{308}
\expandafter\def\csname an@Gemma4@n_ts\endcsname{36}
\expandafter\def\csname an@Gemma4@c_tfail_CE\endcsname{71.6}
\expandafter\def\csname an@Gemma4@c_tonly_CE\endcsname{25.7}
\expandafter\def\csname an@Gemma4@c_tsfail_CE\endcsname{0.0}
\expandafter\def\csname an@Gemma4@c_tsd_CE\endcsname{2.7}
\expandafter\def\csname an@Gemma4@n_correct_CE\endcsname{74}
\expandafter\def\csname an@Gemma4@c_tfail_CF\endcsname{66.3}
\expandafter\def\csname an@Gemma4@c_tonly_CF\endcsname{32.5}
\expandafter\def\csname an@Gemma4@c_tsfail_CF\endcsname{0.0}
\expandafter\def\csname an@Gemma4@c_tsd_CF\endcsname{1.2}
\expandafter\def\csname an@Gemma4@n_correct_CF\endcsname{83}
\expandafter\def\csname an@Gemma4@c_tfail_H\endcsname{--}
\expandafter\def\csname an@Gemma4@c_tonly_H\endcsname{--}
\expandafter\def\csname an@Gemma4@c_tsfail_H\endcsname{--}
\expandafter\def\csname an@Gemma4@c_tsd_H\endcsname{--}
\expandafter\def\csname an@Gemma4@n_correct_H\endcsname{13}
\expandafter\def\csname an@Gemma4@c_tfail_A\endcsname{--}
\expandafter\def\csname an@Gemma4@c_tonly_A\endcsname{--}
\expandafter\def\csname an@Gemma4@c_tsfail_A\endcsname{--}
\expandafter\def\csname an@Gemma4@c_tsd_A\endcsname{--}
\expandafter\def\csname an@Gemma4@n_correct_A\endcsname{38}
\expandafter\def\csname an@Gemma4@c_tfail_P\endcsname{72.0}
\expandafter\def\csname an@Gemma4@c_tonly_P\endcsname{26.0}
\expandafter\def\csname an@Gemma4@c_tsfail_P\endcsname{0.0}
\expandafter\def\csname an@Gemma4@c_tsd_P\endcsname{2.0}
\expandafter\def\csname an@Gemma4@n_correct_P\endcsname{50}
\expandafter\def\csname an@Gemma4@c_tfail_G\endcsname{70.0}
\expandafter\def\csname an@Gemma4@c_tonly_G\endcsname{28.0}
\expandafter\def\csname an@Gemma4@c_tsfail_G\endcsname{0.0}
\expandafter\def\csname an@Gemma4@c_tsd_G\endcsname{2.0}
\expandafter\def\csname an@Gemma4@n_correct_G\endcsname{50}
\expandafter\def\csname an@InternVL35@hand_loc\endcsname{0.0}
\expandafter\def\csname an@InternVL35@tool_loc\endcsname{0.0}
\expandafter\def\csname an@InternVL35@target_loc\endcsname{0.1}
\expandafter\def\csname an@InternVL35@hand_role\endcsname{--}
\expandafter\def\csname an@InternVL35@tool_role\endcsname{--}
\expandafter\def\csname an@InternVL35@target_role\endcsname{--}
\expandafter\def\csname an@InternVL35@hand_joint\endcsname{0.0}
\expandafter\def\csname an@InternVL35@tool_joint\endcsname{0.0}
\expandafter\def\csname an@InternVL35@target_joint\endcsname{0.1}
\expandafter\def\csname an@InternVL35@tt_conf\endcsname{--}
\expandafter\def\csname an@InternVL35@n_hand\endcsname{0}
\expandafter\def\csname an@InternVL35@n_tool\endcsname{0}
\expandafter\def\csname an@InternVL35@n_target\endcsname{2}
\expandafter\def\csname an@InternVL35@n_obj\endcsname{2}
\expandafter\def\csname an@InternVL35@conc_found\endcsname{134}
\expandafter\def\csname an@InternVL35@conc_multi\endcsname{4.5}
\expandafter\def\csname an@InternVL35@conc_multi_same_tok\endcsname{0.0}
\expandafter\def\csname an@InternVL35@conc_sep\endcsname{0.0}
\expandafter\def\csname an@InternVL35@conc_miss_loc\endcsname{100.0}
\expandafter\def\csname an@InternVL35@conc_merged\endcsname{0.0}
\expandafter\def\csname an@InternVL35@conc_wrong_role\endcsname{0.0}
\expandafter\def\csname an@InternVL35@conc_other\endcsname{0.0}
\expandafter\def\csname an@InternVL35@l3_conc_share\endcsname{--}
\expandafter\def\csname an@InternVL35@n_notrec\endcsname{298}
\expandafter\def\csname an@InternVL35@n_conc_sep\endcsname{0}
\expandafter\def\csname an@InternVL35@dep_miss\endcsname{89.1}
\expandafter\def\csname an@InternVL35@dep_none\endcsname{8.2}
\expandafter\def\csname an@InternVL35@dep_type\endcsname{0.3}
\expandafter\def\csname an@InternVL35@dep_ok\endcsname{2.5}
\expandafter\def\csname an@InternVL35@std_loss_dep\endcsname{0.0}
\expandafter\def\csname an@InternVL35@std_loss_nodep\endcsname{0.0}
\expandafter\def\csname an@InternVL35@d_vacuous\endcsname{--}
\expandafter\def\csname an@InternVL35@n_std\endcsname{0}
\expandafter\def\csname an@InternVL35@n_std_vacuous\endcsname{0}
\expandafter\def\csname an@InternVL35@d_rate_vacuous\endcsname{99.0}
\expandafter\def\csname an@InternVL35@n_D\endcsname{858}
\expandafter\def\csname an@InternVL35@c_tfail\endcsname{86.4}
\expandafter\def\csname an@InternVL35@c_tonly\endcsname{13.6}
\expandafter\def\csname an@InternVL35@c_tsfail\endcsname{0.0}
\expandafter\def\csname an@InternVL35@c_tsd\endcsname{0.0}
\expandafter\def\csname an@InternVL35@c_evid_wrong\endcsname{--}
\expandafter\def\csname an@InternVL35@n_correct\endcsname{273}
\expandafter\def\csname an@InternVL35@n_ts\endcsname{0}
\expandafter\def\csname an@InternVL35@c_tfail_CE\endcsname{89.0}
\expandafter\def\csname an@InternVL35@c_tonly_CE\endcsname{11.0}
\expandafter\def\csname an@InternVL35@c_tsfail_CE\endcsname{0.0}
\expandafter\def\csname an@InternVL35@c_tsd_CE\endcsname{0.0}
\expandafter\def\csname an@InternVL35@n_correct_CE\endcsname{73}
\expandafter\def\csname an@InternVL35@c_tfail_CF\endcsname{--}
\expandafter\def\csname an@InternVL35@c_tonly_CF\endcsname{--}
\expandafter\def\csname an@InternVL35@c_tsfail_CF\endcsname{--}
\expandafter\def\csname an@InternVL35@c_tsd_CF\endcsname{--}
\expandafter\def\csname an@InternVL35@n_correct_CF\endcsname{29}
\expandafter\def\csname an@InternVL35@c_tfail_H\endcsname{--}
\expandafter\def\csname an@InternVL35@c_tonly_H\endcsname{--}
\expandafter\def\csname an@InternVL35@c_tsfail_H\endcsname{--}
\expandafter\def\csname an@InternVL35@c_tsd_H\endcsname{--}
\expandafter\def\csname an@InternVL35@n_correct_H\endcsname{25}
\expandafter\def\csname an@InternVL35@c_tfail_A\endcsname{83.9}
\expandafter\def\csname an@InternVL35@c_tonly_A\endcsname{16.1}
\expandafter\def\csname an@InternVL35@c_tsfail_A\endcsname{0.0}
\expandafter\def\csname an@InternVL35@c_tsd_A\endcsname{0.0}
\expandafter\def\csname an@InternVL35@n_correct_A\endcsname{56}
\expandafter\def\csname an@InternVL35@c_tfail_P\endcsname{--}
\expandafter\def\csname an@InternVL35@c_tonly_P\endcsname{--}
\expandafter\def\csname an@InternVL35@c_tsfail_P\endcsname{--}
\expandafter\def\csname an@InternVL35@c_tsd_P\endcsname{--}
\expandafter\def\csname an@InternVL35@n_correct_P\endcsname{38}
\expandafter\def\csname an@InternVL35@c_tfail_G\endcsname{90.4}
\expandafter\def\csname an@InternVL35@c_tonly_G\endcsname{9.6}
\expandafter\def\csname an@InternVL35@c_tsfail_G\endcsname{0.0}
\expandafter\def\csname an@InternVL35@c_tsd_G\endcsname{0.0}
\expandafter\def\csname an@InternVL35@n_correct_G\endcsname{52}
\expandafter\def\csname an@InternVL35L@hand_loc\endcsname{0.0}
\expandafter\def\csname an@InternVL35L@tool_loc\endcsname{0.0}
\expandafter\def\csname an@InternVL35L@target_loc\endcsname{0.1}
\expandafter\def\csname an@InternVL35L@hand_role\endcsname{--}
\expandafter\def\csname an@InternVL35L@tool_role\endcsname{--}
\expandafter\def\csname an@InternVL35L@target_role\endcsname{--}
\expandafter\def\csname an@InternVL35L@hand_joint\endcsname{0.0}
\expandafter\def\csname an@InternVL35L@tool_joint\endcsname{0.0}
\expandafter\def\csname an@InternVL35L@target_joint\endcsname{0.1}
\expandafter\def\csname an@InternVL35L@tt_conf\endcsname{--}
\expandafter\def\csname an@InternVL35L@n_hand\endcsname{1}
\expandafter\def\csname an@InternVL35L@n_tool\endcsname{0}
\expandafter\def\csname an@InternVL35L@n_target\endcsname{3}
\expandafter\def\csname an@InternVL35L@n_obj\endcsname{3}
\expandafter\def\csname an@InternVL35L@conc_found\endcsname{128}
\expandafter\def\csname an@InternVL35L@conc_multi\endcsname{27.3}
\expandafter\def\csname an@InternVL35L@conc_multi_same_tok\endcsname{2.3}
\expandafter\def\csname an@InternVL35L@conc_sep\endcsname{0.0}
\expandafter\def\csname an@InternVL35L@conc_miss_loc\endcsname{100.0}
\expandafter\def\csname an@InternVL35L@conc_merged\endcsname{0.0}
\expandafter\def\csname an@InternVL35L@conc_wrong_role\endcsname{0.0}
\expandafter\def\csname an@InternVL35L@conc_other\endcsname{0.0}
\expandafter\def\csname an@InternVL35L@l3_conc_share\endcsname{--}
\expandafter\def\csname an@InternVL35L@n_notrec\endcsname{279}
\expandafter\def\csname an@InternVL35L@n_conc_sep\endcsname{0}
\expandafter\def\csname an@InternVL35L@dep_miss\endcsname{87.7}
\expandafter\def\csname an@InternVL35L@dep_none\endcsname{7.1}
\expandafter\def\csname an@InternVL35L@dep_type\endcsname{0.5}
\expandafter\def\csname an@InternVL35L@dep_ok\endcsname{4.6}
\expandafter\def\csname an@InternVL35L@std_loss_dep\endcsname{0.0}
\expandafter\def\csname an@InternVL35L@std_loss_nodep\endcsname{0.0}
\expandafter\def\csname an@InternVL35L@d_vacuous\endcsname{--}
\expandafter\def\csname an@InternVL35L@n_std\endcsname{0}
\expandafter\def\csname an@InternVL35L@n_std_vacuous\endcsname{0}
\expandafter\def\csname an@InternVL35L@d_rate_vacuous\endcsname{98.2}
\expandafter\def\csname an@InternVL35L@n_D\endcsname{848}
\expandafter\def\csname an@InternVL35L@c_tfail\endcsname{91.2}
\expandafter\def\csname an@InternVL35L@c_tonly\endcsname{8.8}
\expandafter\def\csname an@InternVL35L@c_tsfail\endcsname{0.0}
\expandafter\def\csname an@InternVL35L@c_tsd\endcsname{0.0}
\expandafter\def\csname an@InternVL35L@c_evid_wrong\endcsname{--}
\expandafter\def\csname an@InternVL35L@n_correct\endcsname{262}
\expandafter\def\csname an@InternVL35L@n_ts\endcsname{0}
\expandafter\def\csname an@InternVL35L@c_tfail_CE\endcsname{88.2}
\expandafter\def\csname an@InternVL35L@c_tonly_CE\endcsname{11.8}
\expandafter\def\csname an@InternVL35L@c_tsfail_CE\endcsname{0.0}
\expandafter\def\csname an@InternVL35L@c_tsd_CE\endcsname{0.0}
\expandafter\def\csname an@InternVL35L@n_correct_CE\endcsname{68}
\expandafter\def\csname an@InternVL35L@c_tfail_CF\endcsname{92.1}
\expandafter\def\csname an@InternVL35L@c_tonly_CF\endcsname{7.9}
\expandafter\def\csname an@InternVL35L@c_tsfail_CF\endcsname{0.0}
\expandafter\def\csname an@InternVL35L@c_tsd_CF\endcsname{0.0}
\expandafter\def\csname an@InternVL35L@n_correct_CF\endcsname{76}
\expandafter\def\csname an@InternVL35L@c_tfail_H\endcsname{--}
\expandafter\def\csname an@InternVL35L@c_tonly_H\endcsname{--}
\expandafter\def\csname an@InternVL35L@c_tsfail_H\endcsname{--}
\expandafter\def\csname an@InternVL35L@c_tsd_H\endcsname{--}
\expandafter\def\csname an@InternVL35L@n_correct_H\endcsname{23}
\expandafter\def\csname an@InternVL35L@c_tfail_A\endcsname{--}
\expandafter\def\csname an@InternVL35L@c_tonly_A\endcsname{--}
\expandafter\def\csname an@InternVL35L@c_tsfail_A\endcsname{--}
\expandafter\def\csname an@InternVL35L@c_tsd_A\endcsname{--}
\expandafter\def\csname an@InternVL35L@n_correct_A\endcsname{30}
\expandafter\def\csname an@InternVL35L@c_tfail_P\endcsname{87.5}
\expandafter\def\csname an@InternVL35L@c_tonly_P\endcsname{12.5}
\expandafter\def\csname an@InternVL35L@c_tsfail_P\endcsname{0.0}
\expandafter\def\csname an@InternVL35L@c_tsd_P\endcsname{0.0}
\expandafter\def\csname an@InternVL35L@n_correct_P\endcsname{40}
\expandafter\def\csname an@InternVL35L@c_tfail_G\endcsname{--}
\expandafter\def\csname an@InternVL35L@c_tonly_G\endcsname{--}
\expandafter\def\csname an@InternVL35L@c_tsfail_G\endcsname{--}
\expandafter\def\csname an@InternVL35L@c_tsd_G\endcsname{--}
\expandafter\def\csname an@InternVL35L@n_correct_G\endcsname{25}
\expandafter\def\csname an@LLaVAST@hand_loc\endcsname{0.9}
\expandafter\def\csname an@LLaVAST@tool_loc\endcsname{0.8}
\expandafter\def\csname an@LLaVAST@target_loc\endcsname{1.8}
\expandafter\def\csname an@LLaVAST@hand_role\endcsname{--}
\expandafter\def\csname an@LLaVAST@tool_role\endcsname{--}
\expandafter\def\csname an@LLaVAST@target_role\endcsname{78.9}
\expandafter\def\csname an@LLaVAST@hand_joint\endcsname{0.6}
\expandafter\def\csname an@LLaVAST@tool_joint\endcsname{0.0}
\expandafter\def\csname an@LLaVAST@target_joint\endcsname{1.4}
\expandafter\def\csname an@LLaVAST@tt_conf\endcsname{4.8}
\expandafter\def\csname an@LLaVAST@n_hand\endcsname{36}
\expandafter\def\csname an@LLaVAST@n_tool\endcsname{5}
\expandafter\def\csname an@LLaVAST@n_target\endcsname{57}
\expandafter\def\csname an@LLaVAST@n_obj\endcsname{62}
\expandafter\def\csname an@LLaVAST@conc_found\endcsname{68}
\expandafter\def\csname an@LLaVAST@conc_multi\endcsname{0.0}
\expandafter\def\csname an@LLaVAST@conc_multi_same_tok\endcsname{0.0}
\expandafter\def\csname an@LLaVAST@conc_sep\endcsname{0.0}
\expandafter\def\csname an@LLaVAST@conc_miss_loc\endcsname{92.0}
\expandafter\def\csname an@LLaVAST@conc_merged\endcsname{2.2}
\expandafter\def\csname an@LLaVAST@conc_wrong_role\endcsname{5.8}
\expandafter\def\csname an@LLaVAST@conc_other\endcsname{0.0}
\expandafter\def\csname an@LLaVAST@l3_conc_share\endcsname{16.2}
\expandafter\def\csname an@LLaVAST@n_notrec\endcsname{137}
\expandafter\def\csname an@LLaVAST@n_conc_sep\endcsname{0}
\expandafter\def\csname an@CosmosR2@hand_loc\endcsname{1.8}
\expandafter\def\csname an@CosmosR2@tool_loc\endcsname{3.2}
\expandafter\def\csname an@CosmosR2@target_loc\endcsname{2.5}
\expandafter\def\csname an@CosmosR2@hand_role\endcsname{69.0}
\expandafter\def\csname an@CosmosR2@tool_role\endcsname{--}
\expandafter\def\csname an@CosmosR2@target_role\endcsname{57.3}
\expandafter\def\csname an@CosmosR2@hand_joint\endcsname{1.2}
\expandafter\def\csname an@CosmosR2@tool_joint\endcsname{1.7}
\expandafter\def\csname an@CosmosR2@target_joint\endcsname{1.4}
\expandafter\def\csname an@CosmosR2@tt_conf\endcsname{11.9}
\expandafter\def\csname an@CosmosR2@n_hand\endcsname{71}
\expandafter\def\csname an@CosmosR2@n_tool\endcsname{19}
\expandafter\def\csname an@CosmosR2@n_target\endcsname{82}
\expandafter\def\csname an@CosmosR2@n_obj\endcsname{101}
\expandafter\def\csname an@CosmosR2@conc_found\endcsname{132}
\expandafter\def\csname an@CosmosR2@conc_multi\endcsname{60.6}
\expandafter\def\csname an@CosmosR2@conc_multi_same_tok\endcsname{6.8}
\expandafter\def\csname an@CosmosR2@conc_sep\endcsname{4.5}
\expandafter\def\csname an@CosmosR2@conc_miss_loc\endcsname{88.4}
\expandafter\def\csname an@CosmosR2@conc_merged\endcsname{4.7}
\expandafter\def\csname an@CosmosR2@conc_wrong_role\endcsname{7.0}
\expandafter\def\csname an@CosmosR2@conc_other\endcsname{0.0}
\expandafter\def\csname an@CosmosR2@l3_conc_share\endcsname{38.5}
\expandafter\def\csname an@CosmosR2@n_notrec\endcsname{258}
\expandafter\def\csname an@CosmosR2@n_conc_sep\endcsname{6}
\expandafter\def\csname an@CosmosR2@dep_miss\endcsname{95.1}
\expandafter\def\csname an@CosmosR2@dep_none\endcsname{3.5}
\expandafter\def\csname an@CosmosR2@dep_type\endcsname{0.0}
\expandafter\def\csname an@CosmosR2@dep_ok\endcsname{1.4}
\expandafter\def\csname an@CosmosR2@std_loss_dep\endcsname{0.0}
\expandafter\def\csname an@CosmosR2@std_loss_nodep\endcsname{0.0}
\expandafter\def\csname an@CosmosR2@d_vacuous\endcsname{--}
\expandafter\def\csname an@CosmosR2@n_std\endcsname{0}
\expandafter\def\csname an@CosmosR2@n_std_vacuous\endcsname{0}
\expandafter\def\csname an@CosmosR2@d_rate_vacuous\endcsname{99.5}
\expandafter\def\csname an@CosmosR2@n_D\endcsname{1,076}
\expandafter\def\csname an@CosmosR2@c_tfail\endcsname{81.2}
\expandafter\def\csname an@CosmosR2@c_tonly\endcsname{18.8}
\expandafter\def\csname an@CosmosR2@c_tsfail\endcsname{0.0}
\expandafter\def\csname an@CosmosR2@c_tsd\endcsname{0.0}
\expandafter\def\csname an@CosmosR2@c_evid_wrong\endcsname{--}
\expandafter\def\csname an@CosmosR2@n_correct\endcsname{234}
\expandafter\def\csname an@CosmosR2@n_ts\endcsname{9}
\expandafter\def\csname an@CosmosR2@c_tfail_CE\endcsname{74.6}
\expandafter\def\csname an@CosmosR2@c_tonly_CE\endcsname{25.4}
\expandafter\def\csname an@CosmosR2@c_tsfail_CE\endcsname{0.0}
\expandafter\def\csname an@CosmosR2@c_tsd_CE\endcsname{0.0}
\expandafter\def\csname an@CosmosR2@n_correct_CE\endcsname{59}
\expandafter\def\csname an@CosmosR2@c_tfail_CF\endcsname{93.1}
\expandafter\def\csname an@CosmosR2@c_tonly_CF\endcsname{6.9}
\expandafter\def\csname an@CosmosR2@c_tsfail_CF\endcsname{0.0}
\expandafter\def\csname an@CosmosR2@c_tsd_CF\endcsname{0.0}
\expandafter\def\csname an@CosmosR2@n_correct_CF\endcsname{58}
\expandafter\def\csname an@CosmosR2@c_tfail_H\endcsname{--}
\expandafter\def\csname an@CosmosR2@c_tonly_H\endcsname{--}
\expandafter\def\csname an@CosmosR2@c_tsfail_H\endcsname{--}
\expandafter\def\csname an@CosmosR2@c_tsd_H\endcsname{--}
\expandafter\def\csname an@CosmosR2@n_correct_H\endcsname{8}
\expandafter\def\csname an@CosmosR2@c_tfail_A\endcsname{--}
\expandafter\def\csname an@CosmosR2@c_tonly_A\endcsname{--}
\expandafter\def\csname an@CosmosR2@c_tsfail_A\endcsname{--}
\expandafter\def\csname an@CosmosR2@c_tsd_A\endcsname{--}
\expandafter\def\csname an@CosmosR2@n_correct_A\endcsname{28}
\expandafter\def\csname an@CosmosR2@c_tfail_P\endcsname{--}
\expandafter\def\csname an@CosmosR2@c_tonly_P\endcsname{--}
\expandafter\def\csname an@CosmosR2@c_tsfail_P\endcsname{--}
\expandafter\def\csname an@CosmosR2@c_tsd_P\endcsname{--}
\expandafter\def\csname an@CosmosR2@n_correct_P\endcsname{32}
\expandafter\def\csname an@CosmosR2@c_tfail_G\endcsname{73.5}
\expandafter\def\csname an@CosmosR2@c_tonly_G\endcsname{26.5}
\expandafter\def\csname an@CosmosR2@c_tsfail_G\endcsname{0.0}
\expandafter\def\csname an@CosmosR2@c_tsd_G\endcsname{0.0}
\expandafter\def\csname an@CosmosR2@n_correct_G\endcsname{49}
\expandafter\def\csname an@CausalVLM@dep_miss\endcsname{91.6}
\expandafter\def\csname an@CausalVLM@dep_none\endcsname{0.5}
\expandafter\def\csname an@CausalVLM@dep_type\endcsname{0.8}
\expandafter\def\csname an@CausalVLM@dep_ok\endcsname{7.1}
\expandafter\def\csname an@CausalVLM@d_rate_vacuous\endcsname{95.6}
\expandafter\def\csname an@CausalVLM@n_D\endcsname{342}
\expandafter\def\csname an@MECDplus@dep_miss\endcsname{91.6}
\expandafter\def\csname an@MECDplus@dep_none\endcsname{3.0}
\expandafter\def\csname an@MECDplus@dep_type\endcsname{0.8}
\expandafter\def\csname an@MECDplus@dep_ok\endcsname{4.6}
\expandafter\def\csname an@MECDplus@d_rate_vacuous\endcsname{97.7}
\expandafter\def\csname an@MECDplus@n_D\endcsname{705}
\expandafter\def\csname an@G38F@hand_loc\endcsname{18.2}
\expandafter\def\csname an@G38F@tool_loc\endcsname{28.9}
\expandafter\def\csname an@G38F@target_loc\endcsname{19.9}
\expandafter\def\csname an@G38F@hand_role\endcsname{94.4}
\expandafter\def\csname an@G38F@tool_role\endcsname{90.8}
\expandafter\def\csname an@G38F@target_role\endcsname{86.7}
\expandafter\def\csname an@G38F@hand_joint\endcsname{17.1}
\expandafter\def\csname an@G38F@tool_joint\endcsname{26.2}
\expandafter\def\csname an@G38F@target_joint\endcsname{17.3}
\expandafter\def\csname an@G38F@tt_conf\endcsname{8.4}
\expandafter\def\csname an@G38F@n_hand\endcsname{712}
\expandafter\def\csname an@G38F@n_tool\endcsname{174}
\expandafter\def\csname an@G38F@n_target\endcsname{646}
\expandafter\def\csname an@G38F@n_obj\endcsname{820}
\expandafter\def\csname an@G38F@conc_found\endcsname{220}
\expandafter\def\csname an@G38F@conc_multi\endcsname{6.4}
\expandafter\def\csname an@G38F@conc_multi_same_tok\endcsname{0.5}
\expandafter\def\csname an@G38F@conc_sep\endcsname{3.2}
\expandafter\def\csname an@G38F@conc_miss_loc\endcsname{63.3}
\expandafter\def\csname an@G38F@conc_merged\endcsname{30.7}
\expandafter\def\csname an@G38F@conc_wrong_role\endcsname{6.1}
\expandafter\def\csname an@G38F@conc_other\endcsname{0.0}
\expandafter\def\csname an@G38F@l3_conc_share\endcsname{20.7}
\expandafter\def\csname an@G38F@n_notrec\endcsname{313}
\expandafter\def\csname an@G38F@n_conc_sep\endcsname{7}
\expandafter\def\csname an@G38F@dep_miss\endcsname{72.8}
\expandafter\def\csname an@G38F@dep_none\endcsname{7.9}
\expandafter\def\csname an@G38F@dep_type\endcsname{2.2}
\expandafter\def\csname an@G38F@dep_ok\endcsname{17.2}
\expandafter\def\csname an@G38F@std_loss_dep\endcsname{0.5}
\expandafter\def\csname an@G38F@std_loss_nodep\endcsname{1.2}
\expandafter\def\csname an@G38F@d_vacuous\endcsname{--}
\expandafter\def\csname an@G38F@n_std\endcsname{29}
\expandafter\def\csname an@G38F@n_std_vacuous\endcsname{18}
\expandafter\def\csname an@G38F@d_rate_vacuous\endcsname{83.6}
\expandafter\def\csname an@G38F@n_D\endcsname{366}
\expandafter\def\csname an@G38F@c_tfail\endcsname{67.7}
\expandafter\def\csname an@G38F@c_tonly\endcsname{18.8}
\expandafter\def\csname an@G38F@c_tsfail\endcsname{6.7}
\expandafter\def\csname an@G38F@c_tsd\endcsname{6.7}
\expandafter\def\csname an@G38F@c_evid_wrong\endcsname{65.3}
\expandafter\def\csname an@G38F@n_correct\endcsname{430}
\expandafter\def\csname an@G38F@n_ts\endcsname{167}
\expandafter\def\csname an@G38F@c_tfail_CE\endcsname{67.3}
\expandafter\def\csname an@G38F@c_tonly_CE\endcsname{16.8}
\expandafter\def\csname an@G38F@c_tsfail_CE\endcsname{10.9}
\expandafter\def\csname an@G38F@c_tsd_CE\endcsname{5.0}
\expandafter\def\csname an@G38F@n_correct_CE\endcsname{101}
\expandafter\def\csname an@G38F@c_tfail_CF\endcsname{62.7}
\expandafter\def\csname an@G38F@c_tonly_CF\endcsname{22.9}
\expandafter\def\csname an@G38F@c_tsfail_CF\endcsname{3.6}
\expandafter\def\csname an@G38F@c_tsd_CF\endcsname{10.8}
\expandafter\def\csname an@G38F@n_correct_CF\endcsname{83}
\expandafter\def\csname an@G38F@c_tfail_H\endcsname{--}
\expandafter\def\csname an@G38F@c_tonly_H\endcsname{--}
\expandafter\def\csname an@G38F@c_tsfail_H\endcsname{--}
\expandafter\def\csname an@G38F@c_tsd_H\endcsname{--}
\expandafter\def\csname an@G38F@n_correct_H\endcsname{17}
\expandafter\def\csname an@G38F@c_tfail_A\endcsname{69.7}
\expandafter\def\csname an@G38F@c_tonly_A\endcsname{21.1}
\expandafter\def\csname an@G38F@c_tsfail_A\endcsname{5.3}
\expandafter\def\csname an@G38F@c_tsd_A\endcsname{3.9}
\expandafter\def\csname an@G38F@n_correct_A\endcsname{76}
\expandafter\def\csname an@G38F@c_tfail_P\endcsname{75.0}
\expandafter\def\csname an@G38F@c_tonly_P\endcsname{14.3}
\expandafter\def\csname an@G38F@c_tsfail_P\endcsname{6.0}
\expandafter\def\csname an@G38F@c_tsd_P\endcsname{4.8}
\expandafter\def\csname an@G38F@n_correct_P\endcsname{84}
\expandafter\def\csname an@G38F@c_tfail_G\endcsname{60.9}
\expandafter\def\csname an@G38F@c_tonly_G\endcsname{18.8}
\expandafter\def\csname an@G38F@c_tsfail_G\endcsname{8.7}
\expandafter\def\csname an@G38F@c_tsd_G\endcsname{11.6}
\expandafter\def\csname an@G38F@n_correct_G\endcsname{69}
\expandafter\def\csname an@GER2@hand_loc\endcsname{9.4}
\expandafter\def\csname an@GER2@tool_loc\endcsname{15.1}
\expandafter\def\csname an@GER2@target_loc\endcsname{10.9}
\expandafter\def\csname an@GER2@hand_role\endcsname{95.4}
\expandafter\def\csname an@GER2@tool_role\endcsname{92.3}
\expandafter\def\csname an@GER2@target_role\endcsname{84.5}
\expandafter\def\csname an@GER2@hand_joint\endcsname{9.0}
\expandafter\def\csname an@GER2@tool_joint\endcsname{13.9}
\expandafter\def\csname an@GER2@target_joint\endcsname{9.3}
\expandafter\def\csname an@GER2@tt_conf\endcsname{8.3}
\expandafter\def\csname an@GER2@n_hand\endcsname{369}
\expandafter\def\csname an@GER2@n_tool\endcsname{91}
\expandafter\def\csname an@GER2@n_target\endcsname{355}
\expandafter\def\csname an@GER2@n_obj\endcsname{446}
\expandafter\def\csname an@GER2@conc_found\endcsname{165}
\expandafter\def\csname an@GER2@conc_multi\endcsname{6.1}
\expandafter\def\csname an@GER2@conc_multi_same_tok\endcsname{0.0}
\expandafter\def\csname an@GER2@conc_sep\endcsname{0.6}
\expandafter\def\csname an@GER2@conc_miss_loc\endcsname{74.4}
\expandafter\def\csname an@GER2@conc_merged\endcsname{21.8}
\expandafter\def\csname an@GER2@conc_wrong_role\endcsname{3.8}
\expandafter\def\csname an@GER2@conc_other\endcsname{0.0}
\expandafter\def\csname an@GER2@l3_conc_share\endcsname{20.3}
\expandafter\def\csname an@GER2@n_notrec\endcsname{266}
\expandafter\def\csname an@GER2@n_conc_sep\endcsname{1}
\expandafter\def\csname an@GER2@dep_miss\endcsname{83.4}
\expandafter\def\csname an@GER2@dep_none\endcsname{4.1}
\expandafter\def\csname an@GER2@dep_type\endcsname{1.1}
\expandafter\def\csname an@GER2@dep_ok\endcsname{11.4}
\expandafter\def\csname an@GER2@std_loss_dep\endcsname{0.3}
\expandafter\def\csname an@GER2@std_loss_nodep\endcsname{1.0}
\expandafter\def\csname an@GER2@d_vacuous\endcsname{--}
\expandafter\def\csname an@GER2@n_std\endcsname{12}
\expandafter\def\csname an@GER2@n_std_vacuous\endcsname{8}
\expandafter\def\csname an@GER2@d_rate_vacuous\endcsname{89.7}
\expandafter\def\csname an@GER2@n_D\endcsname{398}
\expandafter\def\csname an@GER2@c_tfail\endcsname{74.9}
\expandafter\def\csname an@GER2@c_tonly\endcsname{15.1}
\expandafter\def\csname an@GER2@c_tsfail\endcsname{6.5}
\expandafter\def\csname an@GER2@c_tsd\endcsname{3.6}
\expandafter\def\csname an@GER2@c_evid_wrong\endcsname{71.4}
\expandafter\def\csname an@GER2@n_correct\endcsname{338}
\expandafter\def\csname an@GER2@n_ts\endcsname{119}
\expandafter\def\csname an@GER2@c_tfail_CE\endcsname{84.6}
\expandafter\def\csname an@GER2@c_tonly_CE\endcsname{7.7}
\expandafter\def\csname an@GER2@c_tsfail_CE\endcsname{6.2}
\expandafter\def\csname an@GER2@c_tsd_CE\endcsname{1.5}
\expandafter\def\csname an@GER2@n_correct_CE\endcsname{65}
\expandafter\def\csname an@GER2@c_tfail_CF\endcsname{74.7}
\expandafter\def\csname an@GER2@c_tonly_CF\endcsname{14.5}
\expandafter\def\csname an@GER2@c_tsfail_CF\endcsname{4.8}
\expandafter\def\csname an@GER2@c_tsd_CF\endcsname{6.0}
\expandafter\def\csname an@GER2@n_correct_CF\endcsname{83}
\expandafter\def\csname an@GER2@c_tfail_H\endcsname{--}
\expandafter\def\csname an@GER2@c_tonly_H\endcsname{--}
\expandafter\def\csname an@GER2@c_tsfail_H\endcsname{--}
\expandafter\def\csname an@GER2@c_tsd_H\endcsname{--}
\expandafter\def\csname an@GER2@n_correct_H\endcsname{9}
\expandafter\def\csname an@GER2@c_tfail_A\endcsname{68.8}
\expandafter\def\csname an@GER2@c_tonly_A\endcsname{21.9}
\expandafter\def\csname an@GER2@c_tsfail_A\endcsname{6.2}
\expandafter\def\csname an@GER2@c_tsd_A\endcsname{3.1}
\expandafter\def\csname an@GER2@n_correct_A\endcsname{64}
\expandafter\def\csname an@GER2@c_tfail_P\endcsname{79.2}
\expandafter\def\csname an@GER2@c_tonly_P\endcsname{16.7}
\expandafter\def\csname an@GER2@c_tsfail_P\endcsname{2.8}
\expandafter\def\csname an@GER2@c_tsd_P\endcsname{1.4}
\expandafter\def\csname an@GER2@n_correct_P\endcsname{72}
\expandafter\def\csname an@GER2@c_tfail_G\endcsname{62.2}
\expandafter\def\csname an@GER2@c_tonly_G\endcsname{13.3}
\expandafter\def\csname an@GER2@c_tsfail_G\endcsname{17.8}
\expandafter\def\csname an@GER2@c_tsd_G\endcsname{6.7}
\expandafter\def\csname an@GER2@n_correct_G\endcsname{45}
\expandafter\def\csname an@G31P@hand_loc\endcsname{16.9}
\expandafter\def\csname an@G31P@tool_loc\endcsname{25.0}
\expandafter\def\csname an@G31P@target_loc\endcsname{18.1}
\expandafter\def\csname an@G31P@hand_role\endcsname{95.3}
\expandafter\def\csname an@G31P@tool_role\endcsname{92.7}
\expandafter\def\csname an@G31P@target_role\endcsname{86.3}
\expandafter\def\csname an@G31P@hand_joint\endcsname{16.1}
\expandafter\def\csname an@G31P@tool_joint\endcsname{23.2}
\expandafter\def\csname an@G31P@target_joint\endcsname{15.6}
\expandafter\def\csname an@G31P@tt_conf\endcsname{9.5}
\expandafter\def\csname an@G31P@n_hand\endcsname{664}
\expandafter\def\csname an@G31P@n_tool\endcsname{151}
\expandafter\def\csname an@G31P@n_target\endcsname{586}
\expandafter\def\csname an@G31P@n_obj\endcsname{737}
\expandafter\def\csname an@G31P@conc_found\endcsname{213}
\expandafter\def\csname an@G31P@conc_multi\endcsname{8.0}
\expandafter\def\csname an@G31P@conc_multi_same_tok\endcsname{0.5}
\expandafter\def\csname an@G31P@conc_sep\endcsname{3.8}
\expandafter\def\csname an@G31P@conc_miss_loc\endcsname{64.6}
\expandafter\def\csname an@G31P@conc_merged\endcsname{30.8}
\expandafter\def\csname an@G31P@conc_wrong_role\endcsname{4.5}
\expandafter\def\csname an@G31P@conc_other\endcsname{0.0}
\expandafter\def\csname an@G31P@l3_conc_share\endcsname{21.4}
\expandafter\def\csname an@G31P@n_notrec\endcsname{308}
\expandafter\def\csname an@G31P@n_conc_sep\endcsname{8}
\expandafter\def\csname an@G31P@dep_miss\endcsname{79.6}
\expandafter\def\csname an@G31P@dep_none\endcsname{5.2}
\expandafter\def\csname an@G31P@dep_type\endcsname{0.3}
\expandafter\def\csname an@G31P@dep_ok\endcsname{15.0}
\expandafter\def\csname an@G31P@std_loss_dep\endcsname{0.5}
\expandafter\def\csname an@G31P@std_loss_nodep\endcsname{0.9}
\expandafter\def\csname an@G31P@d_vacuous\endcsname{--}
\expandafter\def\csname an@G31P@n_std\endcsname{27}
\expandafter\def\csname an@G31P@n_std_vacuous\endcsname{21}
\expandafter\def\csname an@G31P@d_rate_vacuous\endcsname{91.4}
\expandafter\def\csname an@G31P@n_D\endcsname{606}
\expandafter\def\csname an@G31P@c_tfail\endcsname{60.1}
\expandafter\def\csname an@G31P@c_tonly\endcsname{24.7}
\expandafter\def\csname an@G31P@c_tsfail\endcsname{7.1}
\expandafter\def\csname an@G31P@c_tsd\endcsname{8.0}
\expandafter\def\csname an@G31P@c_evid_wrong\endcsname{75.9}
\expandafter\def\csname an@G31P@n_correct\endcsname{336}
\expandafter\def\csname an@G31P@n_ts\endcsname{212}
\expandafter\def\csname an@G31P@c_tfail_CE\endcsname{51.9}
\expandafter\def\csname an@G31P@c_tonly_CE\endcsname{26.0}
\expandafter\def\csname an@G31P@c_tsfail_CE\endcsname{7.8}
\expandafter\def\csname an@G31P@c_tsd_CE\endcsname{14.3}
\expandafter\def\csname an@G31P@n_correct_CE\endcsname{77}
\expandafter\def\csname an@G31P@c_tfail_CF\endcsname{64.7}
\expandafter\def\csname an@G31P@c_tonly_CF\endcsname{21.6}
\expandafter\def\csname an@G31P@c_tsfail_CF\endcsname{7.8}
\expandafter\def\csname an@G31P@c_tsd_CF\endcsname{5.9}
\expandafter\def\csname an@G31P@n_correct_CF\endcsname{51}
\expandafter\def\csname an@G31P@c_tfail_H\endcsname{--}
\expandafter\def\csname an@G31P@c_tonly_H\endcsname{--}
\expandafter\def\csname an@G31P@c_tsfail_H\endcsname{--}
\expandafter\def\csname an@G31P@c_tsd_H\endcsname{--}
\expandafter\def\csname an@G31P@n_correct_H\endcsname{8}
\expandafter\def\csname an@G31P@c_tfail_A\endcsname{60.5}
\expandafter\def\csname an@G31P@c_tonly_A\endcsname{28.4}
\expandafter\def\csname an@G31P@c_tsfail_A\endcsname{6.2}
\expandafter\def\csname an@G31P@c_tsd_A\endcsname{4.9}
\expandafter\def\csname an@G31P@n_correct_A\endcsname{81}
\expandafter\def\csname an@G31P@c_tfail_P\endcsname{66.2}
\expandafter\def\csname an@G31P@c_tonly_P\endcsname{22.1}
\expandafter\def\csname an@G31P@c_tsfail_P\endcsname{7.4}
\expandafter\def\csname an@G31P@c_tsd_P\endcsname{4.4}
\expandafter\def\csname an@G31P@n_correct_P\endcsname{68}
\expandafter\def\csname an@G31P@c_tfail_G\endcsname{58.8}
\expandafter\def\csname an@G31P@c_tonly_G\endcsname{21.6}
\expandafter\def\csname an@G31P@c_tsfail_G\endcsname{7.8}
\expandafter\def\csname an@G31P@c_tsd_G\endcsname{11.8}
\expandafter\def\csname an@G31P@n_correct_G\endcsname{51}
\expandafter\def\csname an@GPT@hand_loc\endcsname{30.5}
\expandafter\def\csname an@GPT@tool_loc\endcsname{42.5}
\expandafter\def\csname an@GPT@target_loc\endcsname{30.5}
\expandafter\def\csname an@GPT@hand_role\endcsname{97.9}
\expandafter\def\csname an@GPT@tool_role\endcsname{87.5}
\expandafter\def\csname an@GPT@target_role\endcsname{91.0}
\expandafter\def\csname an@GPT@hand_joint\endcsname{29.8}
\expandafter\def\csname an@GPT@tool_joint\endcsname{37.1}
\expandafter\def\csname an@GPT@target_joint\endcsname{27.7}
\expandafter\def\csname an@GPT@tt_conf\endcsname{8.2}
\expandafter\def\csname an@GPT@n_hand\endcsname{1,195}
\expandafter\def\csname an@GPT@n_tool\endcsname{256}
\expandafter\def\csname an@GPT@n_target\endcsname{988}
\expandafter\def\csname an@GPT@n_obj\endcsname{1,244}
\expandafter\def\csname an@GPT@conc_found\endcsname{261}
\expandafter\def\csname an@GPT@conc_multi\endcsname{19.2}
\expandafter\def\csname an@GPT@conc_multi_same_tok\endcsname{2.3}
\expandafter\def\csname an@GPT@conc_sep\endcsname{13.0}
\expandafter\def\csname an@GPT@conc_miss_loc\endcsname{56.0}
\expandafter\def\csname an@GPT@conc_merged\endcsname{39.6}
\expandafter\def\csname an@GPT@conc_wrong_role\endcsname{4.4}
\expandafter\def\csname an@GPT@conc_other\endcsname{0.0}
\expandafter\def\csname an@GPT@l3_conc_share\endcsname{22.1}
\expandafter\def\csname an@GPT@n_notrec\endcsname{293}
\expandafter\def\csname an@GPT@n_conc_sep\endcsname{34}
\expandafter\def\csname an@GPT@dep_miss\endcsname{65.4}
\expandafter\def\csname an@GPT@dep_none\endcsname{11.2}
\expandafter\def\csname an@GPT@dep_type\endcsname{1.6}
\expandafter\def\csname an@GPT@dep_ok\endcsname{21.8}
\expandafter\def\csname an@GPT@std_loss_dep\endcsname{0.8}
\expandafter\def\csname an@GPT@std_loss_nodep\endcsname{3.3}
\expandafter\def\csname an@GPT@d_vacuous\endcsname{62.2}
\expandafter\def\csname an@GPT@n_std\endcsname{45}
\expandafter\def\csname an@GPT@n_std_vacuous\endcsname{28}
\expandafter\def\csname an@GPT@d_rate_vacuous\endcsname{81.9}
\expandafter\def\csname an@GPT@n_D\endcsname{414}
\expandafter\def\csname an@GPT@c_tfail\endcsname{63.8}
\expandafter\def\csname an@GPT@c_tonly\endcsname{15.5}
\expandafter\def\csname an@GPT@c_tsfail\endcsname{12.4}
\expandafter\def\csname an@GPT@c_tsd\endcsname{8.3}
\expandafter\def\csname an@GPT@c_evid_wrong\endcsname{56.6}
\expandafter\def\csname an@GPT@n_correct\endcsname{541}
\expandafter\def\csname an@GPT@n_ts\endcsname{258}
\expandafter\def\csname an@GPT@c_tfail_CE\endcsname{61.5}
\expandafter\def\csname an@GPT@c_tonly_CE\endcsname{12.0}
\expandafter\def\csname an@GPT@c_tsfail_CE\endcsname{13.7}
\expandafter\def\csname an@GPT@c_tsd_CE\endcsname{12.8}
\expandafter\def\csname an@GPT@n_correct_CE\endcsname{117}
\expandafter\def\csname an@GPT@c_tfail_CF\endcsname{66.1}
\expandafter\def\csname an@GPT@c_tonly_CF\endcsname{19.3}
\expandafter\def\csname an@GPT@c_tsfail_CF\endcsname{7.3}
\expandafter\def\csname an@GPT@c_tsd_CF\endcsname{7.3}
\expandafter\def\csname an@GPT@n_correct_CF\endcsname{109}
\expandafter\def\csname an@GPT@c_tfail_H\endcsname{--}
\expandafter\def\csname an@GPT@c_tonly_H\endcsname{--}
\expandafter\def\csname an@GPT@c_tsfail_H\endcsname{--}
\expandafter\def\csname an@GPT@c_tsd_H\endcsname{--}
\expandafter\def\csname an@GPT@n_correct_H\endcsname{21}
\expandafter\def\csname an@GPT@c_tfail_A\endcsname{61.8}
\expandafter\def\csname an@GPT@c_tonly_A\endcsname{16.4}
\expandafter\def\csname an@GPT@c_tsfail_A\endcsname{14.5}
\expandafter\def\csname an@GPT@c_tsd_A\endcsname{7.3}
\expandafter\def\csname an@GPT@n_correct_A\endcsname{110}
\expandafter\def\csname an@GPT@c_tfail_P\endcsname{68.0}
\expandafter\def\csname an@GPT@c_tonly_P\endcsname{15.5}
\expandafter\def\csname an@GPT@c_tsfail_P\endcsname{10.7}
\expandafter\def\csname an@GPT@c_tsd_P\endcsname{5.8}
\expandafter\def\csname an@GPT@n_correct_P\endcsname{103}
\expandafter\def\csname an@GPT@c_tfail_G\endcsname{60.5}
\expandafter\def\csname an@GPT@c_tonly_G\endcsname{11.1}
\expandafter\def\csname an@GPT@c_tsfail_G\endcsname{18.5}
\expandafter\def\csname an@GPT@c_tsd_G\endcsname{9.9}
\expandafter\def\csname an@GPT@n_correct_G\endcsname{81}
\expandafter\def\csname an@GPTlow@hand_loc\endcsname{29.2}
\expandafter\def\csname an@GPTlow@tool_loc\endcsname{37.6}
\expandafter\def\csname an@GPTlow@target_loc\endcsname{29.6}
\expandafter\def\csname an@GPTlow@hand_role\endcsname{97.7}
\expandafter\def\csname an@GPTlow@tool_role\endcsname{90.7}
\expandafter\def\csname an@GPTlow@target_role\endcsname{91.3}
\expandafter\def\csname an@GPTlow@hand_joint\endcsname{28.5}
\expandafter\def\csname an@GPTlow@tool_joint\endcsname{34.2}
\expandafter\def\csname an@GPTlow@target_joint\endcsname{27.0}
\expandafter\def\csname an@GPTlow@tt_conf\endcsname{6.8}
\expandafter\def\csname an@GPTlow@n_hand\endcsname{1,145}
\expandafter\def\csname an@GPTlow@n_tool\endcsname{227}
\expandafter\def\csname an@GPTlow@n_target\endcsname{961}
\expandafter\def\csname an@GPTlow@n_obj\endcsname{1,188}
\expandafter\def\csname an@GPTlow@conc_found\endcsname{242}
\expandafter\def\csname an@GPTlow@conc_multi\endcsname{22.7}
\expandafter\def\csname an@GPTlow@conc_multi_same_tok\endcsname{3.7}
\expandafter\def\csname an@GPTlow@conc_sep\endcsname{12.0}
\expandafter\def\csname an@GPTlow@conc_miss_loc\endcsname{55.4}
\expandafter\def\csname an@GPTlow@conc_merged\endcsname{41.1}
\expandafter\def\csname an@GPTlow@conc_wrong_role\endcsname{3.6}
\expandafter\def\csname an@GPTlow@conc_other\endcsname{0.0}
\expandafter\def\csname an@GPTlow@l3_conc_share\endcsname{20.8}
\expandafter\def\csname an@GPTlow@n_notrec\endcsname{280}
\expandafter\def\csname an@GPTlow@n_conc_sep\endcsname{29}
\expandafter\def\csname an@GPTlow@dep_miss\endcsname{68.9}
\expandafter\def\csname an@GPTlow@dep_none\endcsname{8.4}
\expandafter\def\csname an@GPTlow@dep_type\endcsname{1.1}
\expandafter\def\csname an@GPTlow@dep_ok\endcsname{21.5}
\expandafter\def\csname an@GPTlow@std_loss_dep\endcsname{1.2}
\expandafter\def\csname an@GPTlow@std_loss_nodep\endcsname{3.7}
\expandafter\def\csname an@GPTlow@d_vacuous\endcsname{59.1}
\expandafter\def\csname an@GPTlow@n_std\endcsname{44}
\expandafter\def\csname an@GPTlow@n_std_vacuous\endcsname{26}
\expandafter\def\csname an@GPTlow@d_rate_vacuous\endcsname{81.8}
\expandafter\def\csname an@GPTlow@n_D\endcsname{401}
\expandafter\def\csname an@GPTlow@c_tfail\endcsname{59.0}
\expandafter\def\csname an@GPTlow@c_tonly\endcsname{17.2}
\expandafter\def\csname an@GPTlow@c_tsfail\endcsname{15.5}
\expandafter\def\csname an@GPTlow@c_tsd\endcsname{8.4}
\expandafter\def\csname an@GPTlow@c_evid_wrong\endcsname{56.6}
\expandafter\def\csname an@GPTlow@n_correct\endcsname{524}
\expandafter\def\csname an@GPTlow@n_ts\endcsname{288}
\expandafter\def\csname an@GPTlow@c_tfail_CE\endcsname{55.8}
\expandafter\def\csname an@GPTlow@c_tonly_CE\endcsname{13.3}
\expandafter\def\csname an@GPTlow@c_tsfail_CE\endcsname{18.6}
\expandafter\def\csname an@GPTlow@c_tsd_CE\endcsname{12.4}
\expandafter\def\csname an@GPTlow@n_correct_CE\endcsname{113}
\expandafter\def\csname an@GPTlow@c_tfail_CF\endcsname{53.8}
\expandafter\def\csname an@GPTlow@c_tonly_CF\endcsname{24.7}
\expandafter\def\csname an@GPTlow@c_tsfail_CF\endcsname{15.1}
\expandafter\def\csname an@GPTlow@c_tsd_CF\endcsname{6.5}
\expandafter\def\csname an@GPTlow@n_correct_CF\endcsname{93}
\expandafter\def\csname an@GPTlow@c_tfail_H\endcsname{--}
\expandafter\def\csname an@GPTlow@c_tonly_H\endcsname{--}
\expandafter\def\csname an@GPTlow@c_tsfail_H\endcsname{--}
\expandafter\def\csname an@GPTlow@c_tsd_H\endcsname{--}
\expandafter\def\csname an@GPTlow@n_correct_H\endcsname{24}
\expandafter\def\csname an@GPTlow@c_tfail_A\endcsname{64.8}
\expandafter\def\csname an@GPTlow@c_tonly_A\endcsname{15.2}
\expandafter\def\csname an@GPTlow@c_tsfail_A\endcsname{13.3}
\expandafter\def\csname an@GPTlow@c_tsd_A\endcsname{6.7}
\expandafter\def\csname an@GPTlow@n_correct_A\endcsname{105}
\expandafter\def\csname an@GPTlow@c_tfail_P\endcsname{63.1}
\expandafter\def\csname an@GPTlow@c_tonly_P\endcsname{17.1}
\expandafter\def\csname an@GPTlow@c_tsfail_P\endcsname{16.2}
\expandafter\def\csname an@GPTlow@c_tsd_P\endcsname{3.6}
\expandafter\def\csname an@GPTlow@n_correct_P\endcsname{111}
\expandafter\def\csname an@GPTlow@c_tfail_G\endcsname{55.1}
\expandafter\def\csname an@GPTlow@c_tonly_G\endcsname{15.4}
\expandafter\def\csname an@GPTlow@c_tsfail_G\endcsname{16.7}
\expandafter\def\csname an@GPTlow@c_tsd_G\endcsname{12.8}
\expandafter\def\csname an@GPTlow@n_correct_G\endcsname{78}
\expandafter\def\csname an@M1F@hand_loc\endcsname{15.9}
\expandafter\def\csname an@M1F@tool_loc\endcsname{22.2}
\expandafter\def\csname an@M1F@target_loc\endcsname{16.9}
\expandafter\def\csname an@M1F@hand_role\endcsname{96.5}
\expandafter\def\csname an@M1F@tool_role\endcsname{83.6}
\expandafter\def\csname an@M1F@target_role\endcsname{86.1}
\expandafter\def\csname an@M1F@hand_joint\endcsname{15.3}
\expandafter\def\csname an@M1F@tool_joint\endcsname{18.6}
\expandafter\def\csname an@M1F@target_joint\endcsname{14.5}
\expandafter\def\csname an@M1F@tt_conf\endcsname{10.6}
\expandafter\def\csname an@M1F@n_hand\endcsname{624}
\expandafter\def\csname an@M1F@n_tool\endcsname{134}
\expandafter\def\csname an@M1F@n_target\endcsname{547}
\expandafter\def\csname an@M1F@n_obj\endcsname{681}
\expandafter\def\csname an@M1F@conc_found\endcsname{173}
\expandafter\def\csname an@M1F@conc_multi\endcsname{14.5}
\expandafter\def\csname an@M1F@conc_multi_same_tok\endcsname{6.9}
\expandafter\def\csname an@M1F@conc_sep\endcsname{9.2}
\expandafter\def\csname an@M1F@conc_miss_loc\endcsname{55.8}
\expandafter\def\csname an@M1F@conc_merged\endcsname{36.3}
\expandafter\def\csname an@M1F@conc_wrong_role\endcsname{7.9}
\expandafter\def\csname an@M1F@conc_other\endcsname{0.0}
\expandafter\def\csname an@M1F@l3_conc_share\endcsname{23.2}
\expandafter\def\csname an@M1F@n_notrec\endcsname{215}
\expandafter\def\csname an@M1F@n_conc_sep\endcsname{16}
\expandafter\def\csname an@M1F@dep_miss\endcsname{91.0}
\expandafter\def\csname an@M1F@dep_none\endcsname{5.4}
\expandafter\def\csname an@M1F@dep_type\endcsname{0.0}
\expandafter\def\csname an@M1F@dep_ok\endcsname{3.5}
\expandafter\def\csname an@M1F@std_loss_dep\endcsname{0.2}
\expandafter\def\csname an@M1F@std_loss_nodep\endcsname{0.1}
\expandafter\def\csname an@M1F@d_vacuous\endcsname{--}
\expandafter\def\csname an@M1F@n_std\endcsname{29}
\expandafter\def\csname an@M1F@n_std_vacuous\endcsname{26}
\expandafter\def\csname an@M1F@d_rate_vacuous\endcsname{99.1}
\expandafter\def\csname an@M1F@n_D\endcsname{1,213}
\expandafter\def\csname an@M1F@c_tfail\endcsname{67.6}
\expandafter\def\csname an@M1F@c_tonly\endcsname{17.1}
\expandafter\def\csname an@M1F@c_tsfail\endcsname{2.3}
\expandafter\def\csname an@M1F@c_tsd\endcsname{13.1}
\expandafter\def\csname an@M1F@c_evid_wrong\endcsname{85.8}
\expandafter\def\csname an@M1F@n_correct\endcsname{222}
\expandafter\def\csname an@M1F@n_ts\endcsname{240}
\expandafter\def\csname an@M1F@c_tfail_CE\endcsname{66.1}
\expandafter\def\csname an@M1F@c_tonly_CE\endcsname{13.6}
\expandafter\def\csname an@M1F@c_tsfail_CE\endcsname{3.4}
\expandafter\def\csname an@M1F@c_tsd_CE\endcsname{16.9}
\expandafter\def\csname an@M1F@n_correct_CE\endcsname{59}
\expandafter\def\csname an@M1F@c_tfail_CF\endcsname{70.7}
\expandafter\def\csname an@M1F@c_tonly_CF\endcsname{15.9}
\expandafter\def\csname an@M1F@c_tsfail_CF\endcsname{2.4}
\expandafter\def\csname an@M1F@c_tsd_CF\endcsname{11.0}
\expandafter\def\csname an@M1F@n_correct_CF\endcsname{82}
\expandafter\def\csname an@M1F@c_tfail_H\endcsname{--}
\expandafter\def\csname an@M1F@c_tonly_H\endcsname{--}
\expandafter\def\csname an@M1F@c_tsfail_H\endcsname{--}
\expandafter\def\csname an@M1F@c_tsd_H\endcsname{--}
\expandafter\def\csname an@M1F@n_correct_H\endcsname{6}
\expandafter\def\csname an@M1F@c_tfail_A\endcsname{--}
\expandafter\def\csname an@M1F@c_tonly_A\endcsname{--}
\expandafter\def\csname an@M1F@c_tsfail_A\endcsname{--}
\expandafter\def\csname an@M1F@c_tsd_A\endcsname{--}
\expandafter\def\csname an@M1F@n_correct_A\endcsname{26}
\expandafter\def\csname an@M1F@c_tfail_P\endcsname{--}
\expandafter\def\csname an@M1F@c_tonly_P\endcsname{--}
\expandafter\def\csname an@M1F@c_tsfail_P\endcsname{--}
\expandafter\def\csname an@M1F@c_tsd_P\endcsname{--}
\expandafter\def\csname an@M1F@n_correct_P\endcsname{36}
\expandafter\def\csname an@M1F@c_tfail_G\endcsname{--}
\expandafter\def\csname an@M1F@c_tonly_G\endcsname{--}
\expandafter\def\csname an@M1F@c_tsfail_G\endcsname{--}
\expandafter\def\csname an@M1F@c_tsd_G\endcsname{--}
\expandafter\def\csname an@M1F@n_correct_G\endcsname{13}
\expandafter\def\csname an@A@hand_loc\endcsname{17.8}
\expandafter\def\csname an@A@tool_loc\endcsname{20.9}
\expandafter\def\csname an@A@target_loc\endcsname{20.5}
\expandafter\def\csname an@A@hand_role\endcsname{94.4}
\expandafter\def\csname an@A@tool_role\endcsname{84.1}
\expandafter\def\csname an@A@target_role\endcsname{89.6}
\expandafter\def\csname an@A@hand_joint\endcsname{16.8}
\expandafter\def\csname an@A@tool_joint\endcsname{17.6}
\expandafter\def\csname an@A@target_joint\endcsname{18.4}
\expandafter\def\csname an@A@tt_conf\endcsname{8.0}
\expandafter\def\csname an@A@n_hand\endcsname{698}
\expandafter\def\csname an@A@n_tool\endcsname{126}
\expandafter\def\csname an@A@n_target\endcsname{666}
\expandafter\def\csname an@A@n_obj\endcsname{792}
\expandafter\def\csname an@A@conc_found\endcsname{212}
\expandafter\def\csname an@A@conc_multi\endcsname{9.4}
\expandafter\def\csname an@A@conc_multi_same_tok\endcsname{8.5}
\expandafter\def\csname an@A@conc_sep\endcsname{4.7}
\expandafter\def\csname an@A@conc_miss_loc\endcsname{57.1}
\expandafter\def\csname an@A@conc_merged\endcsname{33.7}
\expandafter\def\csname an@A@conc_wrong_role\endcsname{9.2}
\expandafter\def\csname an@A@conc_other\endcsname{0.0}
\expandafter\def\csname an@A@l3_conc_share\endcsname{21.4}
\expandafter\def\csname an@A@n_notrec\endcsname{294}
\expandafter\def\csname an@A@n_conc_sep\endcsname{10}
\expandafter\def\csname an@A@dep_miss\endcsname{81.5}
\expandafter\def\csname an@A@dep_none\endcsname{5.7}
\expandafter\def\csname an@A@dep_type\endcsname{0.5}
\expandafter\def\csname an@A@dep_ok\endcsname{12.3}
\expandafter\def\csname an@A@std_loss_dep\endcsname{0.6}
\expandafter\def\csname an@A@std_loss_nodep\endcsname{0.4}
\expandafter\def\csname an@A@d_vacuous\endcsname{--}
\expandafter\def\csname an@A@n_std\endcsname{25}
\expandafter\def\csname an@A@n_std_vacuous\endcsname{18}
\expandafter\def\csname an@A@d_rate_vacuous\endcsname{96.0}
\expandafter\def\csname an@A@n_D\endcsname{1,117}
\expandafter\def\csname an@A@c_tfail\endcsname{54.2}
\expandafter\def\csname an@A@c_tonly\endcsname{25.4}
\expandafter\def\csname an@A@c_tsfail\endcsname{8.0}
\expandafter\def\csname an@A@c_tsd\endcsname{12.4}
\expandafter\def\csname an@A@c_evid_wrong\endcsname{87.2}
\expandafter\def\csname an@A@n_correct\endcsname{201}
\expandafter\def\csname an@A@n_ts\endcsname{321}
\expandafter\def\csname an@A@c_tfail_CE\endcsname{65.3}
\expandafter\def\csname an@A@c_tonly_CE\endcsname{16.3}
\expandafter\def\csname an@A@c_tsfail_CE\endcsname{8.2}
\expandafter\def\csname an@A@c_tsd_CE\endcsname{10.2}
\expandafter\def\csname an@A@n_correct_CE\endcsname{49}
\expandafter\def\csname an@A@c_tfail_CF\endcsname{58.8}
\expandafter\def\csname an@A@c_tonly_CF\endcsname{21.2}
\expandafter\def\csname an@A@c_tsfail_CF\endcsname{7.5}
\expandafter\def\csname an@A@c_tsd_CF\endcsname{12.5}
\expandafter\def\csname an@A@n_correct_CF\endcsname{80}
\expandafter\def\csname an@A@c_tfail_H\endcsname{--}
\expandafter\def\csname an@A@c_tonly_H\endcsname{--}
\expandafter\def\csname an@A@c_tsfail_H\endcsname{--}
\expandafter\def\csname an@A@c_tsd_H\endcsname{--}
\expandafter\def\csname an@A@n_correct_H\endcsname{0}
\expandafter\def\csname an@A@c_tfail_A\endcsname{--}
\expandafter\def\csname an@A@c_tonly_A\endcsname{--}
\expandafter\def\csname an@A@c_tsfail_A\endcsname{--}
\expandafter\def\csname an@A@c_tsd_A\endcsname{--}
\expandafter\def\csname an@A@n_correct_A\endcsname{23}
\expandafter\def\csname an@A@c_tfail_P\endcsname{47.5}
\expandafter\def\csname an@A@c_tonly_P\endcsname{35.0}
\expandafter\def\csname an@A@c_tsfail_P\endcsname{2.5}
\expandafter\def\csname an@A@c_tsd_P\endcsname{15.0}
\expandafter\def\csname an@A@n_correct_P\endcsname{40}
\expandafter\def\csname an@A@c_tfail_G\endcsname{--}
\expandafter\def\csname an@A@c_tonly_G\endcsname{--}
\expandafter\def\csname an@A@c_tsfail_G\endcsname{--}
\expandafter\def\csname an@A@c_tsd_G\endcsname{--}
\expandafter\def\csname an@A@n_correct_G\endcsname{9}
\def\HVReviewed{2,500}\def\HVPass{122}\def\HVEdited{2,055}\def\HVQAWrong{231}\def\HVNoEv{73}\def\HVCannot{19}\def\HVUnsure{746}\def\HVGold{1,639}\def\HVSecond{287}
\def\HVRemoved{323}\def\HVRemovedPct{12.9}\def\HVBoxKept{80.9}
\def\TabHVEditRows{%
Participant boxes corrected, added or removed & 19.1\% of boxes & 73.3 \\
Role tracks of a key event changed & 57.0\% of kept events & 64.7 \\
Object tracks renamed & 32.1\% of tracks & 49.7 \\
Key events removed & 30.3\% of events & 46.2 \\
Key-event boundary moved $>$0.5\,s & 33.7\% of kept events & 43.5 \\
Key events added & 14.3\% of gold events & 20.6 \\
Dependencies removed, added or retyped & 192 / 168 / 9 edges & 17.5 \\
}
\providecommand{\HVSecond}{\TODO{}}

\subsection{Task definition}
\label{sec:bench_task}

In grounded causal question answering (GCQA), given an egocentric video $V$ and a causal question $Q$, a model predicts an answer $A$ with causal evidence $(\mathcal{E},\mathcal{I},\mathcal{D})$ (\cref{fig:teaser}).

\textbf{Key events (when).} $\mathcal{E}=\{e_k\}$ are the observed manipulation events that together support the answer, each with an interval $[s_k,t_k)$ and a short description.
For counterfactual and hypothetical questions, they are the observed events the alternative rests on, and need not depend on one another.

\textbf{Interactions (what and where).} Within a key event, an interaction $\alpha\in\mathcal{I}$ groups the participants of one manipulation: one or more hands act on targets, optionally through tools.
Each participant has a bounding box and a functional role (hand, tool or target), and the interaction states which participants belong to the same manipulation, their \emph{association}.
A participant may belong to several interactions at the same moment, and a hand that only holds or stabilizes an object is not a participant.

\textbf{Dependencies (how).} $\mathcal{D}=\{(e_i,r,e_j)\}$, $r\in\{\textit{causes},\textit{enables}\}$, are directed relations from an earlier to a later key event.
$e_i$ \textit{causes} $e_j$ when it produces a state or outcome observed in $e_j$, and \textit{enables} $e_j$ when it establishes a condition that the manipulation in $e_j$ requires as performed in the video, a distinction drawn in accounts of causation~\cite{cheng1991causes,wolff2007representing}.
Temporal precedence or a shared tool or target alone establishes neither, and pairs without an annotated dependency form the negative class \textit{before}.

\textbf{Representation.} Times lie on a 0.5\,s grid and boxes are normalized to $[0,1000]$.
The annotation gives each participant's box at every annotated time of its interaction, whereas a model reports each interaction at one time with one box and role per participant, scored against the annotated boxes at that time.

\subsection{Questions and capabilities}
\label{sec:bench_filter}

The evaluation split draws on the test splits of EgoTaskQA~\cite{jia2022egotaskqa}, ProcObject-10K~\cite{guo2026procobject}, EgoTempo~\cite{plizzari2025egotempo} and EgoPlan-Bench2~\cite{qiu2024egoplanbench2}, and the training split on 11 egocentric datasets with no shared source video (\cref{tab:bench_stats}).
We distinguish six capabilities by the inference a question requires~\cite{bareinboim2022pch,gerstenberg2022counterfactual,csibra2007goals}: \emph{causal explanation} asks which earlier interaction or condition led to an observed outcome, \emph{counterfactual} questions alter an observed interaction and \emph{hypothetical} ones assume an unperformed one, \emph{anticipation} asks what happens next when the question names no interaction and no goal, \emph{planning} asks for the next interaction toward a stated goal, and \emph{goal attribution} asks for the purpose of observed interactions.
Rather than the inconsistent source task labels, we manually reclassify each question template by the causal inference it requires, following its source definition and generation rules, and check the assignments on sampled questions, with blind human labels and an independent GPT-5.4-mini classification (Appx.~A).

\subsection{Grounding annotation pipeline}
\label{sec:bench_pipeline}

We construct the evidence in six stages (\cref{fig:pipeline}a), applying stages 1--5 automatically to both splits and stage 6 to evaluation questions only (Appx.~A).
\textbf{Entities and events (stages 1--2).} To keep identities consistent across events, HOI-DETR~\cite{darkhalil2026improving} detections initialize SAM~2.1~\cite{ravi2024sam2} tracks that are then cleaned.
To represent each manipulation as one event, an event boundary detector~\cite{shou2021gebd,zheng2024efficientgebd,li2025vadtree} segments the video, an MLLM names the tracked objects, assigns their roles and captions each segment, and segments of the same manipulation are merged.
\textbf{Question-conditioned evidence (stages 3--5).} Given the question and the ground-truth answer, an MLLM selects the events that jointly support the answer.
Qwen3.6-27B decomposes each key-event caption into manipulations and links their participants, with roles, to the tracks, and each interaction's span is estimated from the box overlap between its targets and its hands or tools as a proxy for contact.
An MLLM then labels each key-event pair as \textit{causes}, \textit{enables} or \textit{before} from the captions, roles and frames.
These answer-conditioned annotations are withheld at inference.
\textbf{Human verification (stage 6).} Annotators review the video, the question--answer pair and the automatic annotations, verify and correct key events, participant identities, boxes, roles, captions and dependencies, add missing evidence, and reject invalid questions.
Since each verified caption states which hands act on which objects and with which tools, the interactions are then parsed from the verified captions, role tracks and boxes as in stage 4.
Twelve annotators reviewed \HVReviewed{} evaluation questions and rejected \HVRemoved{} (\HVRemovedPct\%, Appx.~A).

\subsection{Evidence integrity}
\label{sec:bench_integrity}

To improve evidence reliability, we apply four integrity checks, with three independent runs for each model-based check.
\textbf{(C1) Video dependence} removes questions answered correctly from the text alone in all runs.
\textbf{(C2) Evidence dependence} removes questions answered correctly in all runs with all key events masked.
\textbf{(C3) Evidence validity} is the human verification of stage 6.
\textbf{(C4) Evidence sufficiency} removes questions answered correctly in all runs on the full video but in none with only the key events visible.
C2 and C4 use the human-verified labels where they exist and the automatic labels otherwise.
From \EvAllQ{} source questions, the checks leave the \EvQ{}-question evaluation split, \EvGold{} of which have human-verified evidence (models and counts per step: Appx.~A).
These checks improve evidence reliability without establishing the necessity of each event, interaction or dependency.

\def\EvQK{3.9K}
\def\EvVideosK{2.4K}
\def\EvQ{3,901}
\def\EvGold{1,639}
\def\EvDatasets{4}
\def\EvVideos{2,444}
\def\EvDur{44.1}
\def\EvKpq{1.59}
\def\EvKdur{4.6}
\def\EvMultiK{48\%}
\def\EvIpq{1.98}
\def\EvIpk{1.25}
\def\EvPpi{2.49}
\def\EvTool{20\%}
\def\EvMultiDep{38\%}
\def\EvCauses{19}
\def\EvEnables{348}
\def\EvAllQ{6,028}
\def\EvCOneKept{5,118}
\def\EvCapQ{5,103}
\def\EvReviewed{2,485}
\def\EvVerified{1,975}
\def\EvGoldEvents{2,605}
\def\EvGoldInter{3,253}
\def\EvGoldPart{8,115}
\def\EvGoldPairs{1,180}
\def\EvTotalK{29.8K}
\def\EvGoldK{1.6K}
\def\EvRmHuman{320}
\def\EvRmNoEv{111}
\def\EvRmCtwo{743}
\def\EvRmCfour{28}
\def\EvSpKeptHuman{1,639}   %
\def\EvSpKeptAuto{2,262}   %
\def\EvSpCtwoHuman{338}
\def\EvSpCtwoAuto{405}
\def\EvSpCfourHuman{7}
\def\EvSpCfourAuto{21}
\def\EvSpNoEvHuman{8}
\def\EvSpNoEvAuto{103}
\def\EvSpHumanRej{320}
\def\EvCapHx{47.94}
\def\EvCapHw{3.18}
\def\EvCapHpct{3}
\def\EvCapBars{0/9/{29.43,18.51,3.18,12.38,28.35,8.15},1/0/{12.10,8.53,2.70,15.63,40.62,20.43}}   %

\begin{table}[t]
\centering
\small
\setlength{\tabcolsep}{2.5pt}
\begin{tabular}{@{}lrr@{}}
\toprule
& Evaluation & Training \\
\midrule
Source datasets & \EvDatasets & 11 \\
Source videos & \EvVideos & 4,624 \\
Questions & \EvQ & 25,894 \\
Gold-labeled questions & \EvGold & -- \\
\midrule
Key events (total) & \EvGoldEvents & 48,873 \\
Interactions (total) & \EvGoldInter & 64,353 \\
Participants (total) & \EvGoldPart & 163,231 \\
Questions with $\geq$2 key events & \EvMultiK & 50\% \\
Participants per interaction & \EvPpi & 2.54 \\
\midrule
Multi-event questions with a dependency & \EvMultiDep & 41\% \\
\quad \textit{causes} / \textit{enables} pairs & \EvCauses{} / \EvEnables{} & 681 / 7,787 \\
\bottomrule
\end{tabular}
\caption{\textbf{\bench statistics.} Evaluation: without the questions removed by C2 and C4 (\cref{sec:bench_integrity}); evidence statistics on its gold-labeled questions. Training: automatic labels of the questions used to train \method. Further statistics: Appx.~A. Multi-event questions with a dependency: the share of questions with $\geq$2 key events that have a \textit{causes} or \textit{enables} pair.}
\label{tab:bench_stats}
\end{table}

\subsection{\method: structured evidence grounding}
\label{sec:method}
Motivated by the failures of \cref{sec:exp_findings}, we explore \method, which predicts the structure of the causal evidence with dedicated heads on the hidden states of an MLLM, rather than writing it as text~\cite{meng2026openo3video,visioncoach2026}, and answers from it.
The MLLM is Qwen3.5-2B with one low-rank adapter for grounding and one for answering.
To recover several distinct events rather than one coarse interval, a boundary head proposes candidate segments, and a pointer selects them one at a time conditioned on the question and the events already selected.
To ground each manipulation as a unit, learned interaction queries attend to each selected event and role-specific participant slots localize its hands, tools and targets, so that locations, roles and association are supervised together.
A pairwise classifier labels each pair of selected events, and the answering adapter receives the question and the predicted evidence (event clips and descriptions, interaction close-ups with boxes and roles, dependencies) instead of the full video.
The grounding heads learn from the automatic labels, and the answering adapter from the evidence predicted for the training questions (Appx.~E).

\section{Experiments}
\label{sec:exp}

\providecommand{\AGNGold}{}\renewcommand{\AGNGold}{1,301}
\providecommand{\AGEvAuto}{}\renewcommand{\AGEvAuto}{2,933}
\providecommand{\AGEvDel}{}\renewcommand{\AGEvDel}{24.9\%}
\providecommand{\AGEvDelN}{}\renewcommand{\AGEvDelN}{729}
\providecommand{\AGEvAdd}{}\renewcommand{\AGEvAdd}{9.7\%}
\providecommand{\AGEvAddN}{}\renewcommand{\AGEvAddN}{237}
\providecommand{\AGEvBound}{}\renewcommand{\AGEvBound}{28.0\%}
\providecommand{\AGEvBoundN}{}\renewcommand{\AGEvBoundN}{617}
\providecommand{\AGRoleChg}{}\renewcommand{\AGRoleChg}{50.0\%}
\providecommand{\AGRoleChgN}{}\renewcommand{\AGRoleChgN}{1,101}
\providecommand{\AGTrackRepl}{}\renewcommand{\AGTrackRepl}{24.0\%}
\providecommand{\AGTrackReplN}{}\renewcommand{\AGTrackReplN}{529}
\providecommand{\AGTrackAddN}{}\renewcommand{\AGTrackAddN}{883}
\providecommand{\AGTrackDelN}{}\renewcommand{\AGTrackDelN}{1,721}
\providecommand{\AGNewObjN}{}\renewcommand{\AGNewObjN}{278}
\providecommand{\AGRename}{}\renewcommand{\AGRename}{25.2\%}
\providecommand{\AGBoxN}{}\renewcommand{\AGBoxN}{65,009}
\providecommand{\AGBoxEdit}{}\renewcommand{\AGBoxEdit}{14.6\%}
\providecommand{\AGBoxDel}{}\renewcommand{\AGBoxDel}{0.8\%}
\providecommand{\AGBoxAdd}{}\renewcommand{\AGBoxAdd}{6.5\%}
\providecommand{\AGRelAuto}{}\renewcommand{\AGRelAuto}{809}
\providecommand{\AGRelDel}{}\renewcommand{\AGRelDel}{125}
\providecommand{\AGRelAdd}{}\renewcommand{\AGRelAdd}{117}
\providecommand{\AGRelType}{}\renewcommand{\AGRelType}{6}
\providecommand{\AGRelChg}{}\renewcommand{\AGRelChg}{248}
\providecommand{\AGRelDropUnit}{}\renewcommand{\AGRelDropUnit}{355}
\providecommand{\AGQAnyChg}{}\renewcommand{\AGQAnyChg}{88.5\%}
\providecommand{\AGQEdited}{}\renewcommand{\AGQEdited}{90.4\%}
\providecommand{\AGReviewed}{}\renewcommand{\AGReviewed}{2,141}
\providecommand{\AGVPass}{}\renewcommand{\AGVPass}{165}
\providecommand{\AGVEdited}{}\renewcommand{\AGVEdited}{1,587}
\providecommand{\AGVQAWrong}{}\renewcommand{\AGVQAWrong}{297}
\providecommand{\AGVNoEv}{}\renewcommand{\AGVNoEv}{64}
\providecommand{\AGVCannot}{}\renewcommand{\AGVCannot}{28}
\providecommand{\AGExPool}{}\renewcommand{\AGExPool}{2,141}
\providecommand{\AGExQAWrong}{}\renewcommand{\AGExQAWrong}{292}
\providecommand{\AGExNoEv}{}\renewcommand{\AGExNoEv}{64}
\providecommand{\AGExCannot}{}\renewcommand{\AGExCannot}{28}
\providecommand{\AGExNoInter}{}\renewcommand{\AGExNoInter}{116}
\providecommand{\AGExOldGT}{}\renewcommand{\AGExOldGT}{53}
\providecommand{\AGExCOne}{}\renewcommand{\AGExCOne}{9}
\providecommand{\AGExCTwo}{}\renewcommand{\AGExCTwo}{271}
\providecommand{\AGExCFour}{}\renewcommand{\AGExCFour}{5}
\providecommand{\AGmTIoU}{}\renewcommand{\AGmTIoU}{.590}
\providecommand{\AGFThirty}{}\renewcommand{\AGFThirty}{.691}
\providecommand{\AGLone}{}\renewcommand{\AGLone}{.515}
\providecommand{\AGLtwo}{}\renewcommand{\AGLtwo}{.503}
\providecommand{\AGLthree}{}\renewcommand{\AGLthree}{.496}
\providecommand{\AGMacroFone}{}\renewcommand{\AGMacroFone}{.690}
\providecommand{\AGEdgeFone}{}\renewcommand{\AGEdgeFone}{.421}

\begin{table*}[t]
\centering
\footnotesize
\setlength{\tabcolsep}{2.5pt}
\resizebox{\textwidth}{!}{%
\begin{tabular}{@{}l c cc ccc ccc ccc@{}}
\toprule
& Answer & \multicolumn{2}{c}{RQ1: Key events} & \multicolumn{3}{c}{RQ2: Interactions} & \multicolumn{3}{c}{RQ3: Dependencies} & \multicolumn{3}{c}{RQ4: Grounded correct $\uparrow$} \\
\cmidrule(lr){2-2}\cmidrule(lr){3-4}\cmidrule(lr){5-7}\cmidrule(lr){8-10}\cmidrule(lr){11-13}
System & Acc & mTIoU & F1@.3 & \Lone & \Ltwo & \Lthree & Macro-F1 & Edge F1 & \depacc & GCR$_T$ & GCR$_{ST}$ & GCR$_{STD}$ \\
\midrule
\TabMainRowsCompact
\bottomrule
\end{tabular}}
\caption{\textbf{Main results} on the \EvGold{} gold-labeled evaluation questions, whose evidence is human-verified (all \EvQ{} evaluation questions: Appx.~C). --: not predicted by the system. Input frame rates, mTIoP, mTIoG and the spurious shares SS: Appx.~B.}
\label{tab:main}%
\end{table*}

We evaluate key-event recovery (RQ1), interaction grounding (RQ2), dependency prediction (RQ3) and grounded answering (RQ4), diagnose grounding failures, and test whether better evidence yields better answers.

\subsection{Setting and metrics}
\label{sec:exp_metrics}
\label{sec:exp_protocol}

\paragraph{Data and systems.}
We evaluate on the \EvGold{} gold-labeled questions (\EvGoldEvents{} key events, \EvGoldInter{} interactions, \EvGoldPairs{} event pairs), and on all \EvQ{} evaluation questions in Appx.~C.
Open-source and proprietary MLLMs receive the video at 4\,fps and the question and return one JSON object with the key events, their interactions, the dependencies and the answer, under a prompt with the definitions of \cref{sec:bench_task}.
Specialists run their released pipelines and are scored on what they predict: video reasoning specialists on the answer and the key events, and LLaVA-ST also on its hand and object boxes, scored as hand and target.
Causal-VLM links the events of the whole video and MECD+ links given events, those of Causal-VLM in \cref{tab:main}, and their untyped links are scored as \textit{enables}.
Inputs and reply parsing: Appx. B.

\paragraph{RQ1 \textperiodcentered{} Key events.}
Ground-truth and predicted events are matched one-to-one to maximize the total temporal IoU, and mTIoU divides the sum of the matched IoUs by the larger of the two event counts, so missed and extra events both lower it.
F1@0.3 counts matched pairs with tIoU $\geq0.3$, and only these pass boxes and dependency labels on to RQ2 and RQ3 (variants: Appx.~B).

\paragraph{RQ2 \textperiodcentered{} Interactions.}
For each predicted interaction $k$, participant boxes are compared with the ground-truth participant boxes of the matched event at its timestamp $\tau_k$, snapped to the nearest sampled frame.
Let $\mathcal P$ and $\mathcal G$ denote predicted and ground-truth participant instances.
Each participant is counted separately for every interaction it belongs to, so $|\mathcal G|$ is fixed across systems.
Participants of unmatched predicted or ground-truth events count as false positives or negatives, respectively.
Three nested conditions are checked on the same box match:
\begin{equation}
\begin{aligned}
\mathrm{TP}_1&=|\pi|,\qquad \mathrm{TP}_2=|\{(e,g)\in\pi:\hat r_e=r_g\}|,\\
\mathrm{TP}_3&=\textstyle\sum_{(k,\alpha)\in\sigma}|\hat S_k\cap S_\alpha|,\qquad
\mathrm{F1}_\ell=\dfrac{2\,\mathrm{TP}_\ell}{|\mathcal P|+|\mathcal G|}.
\end{aligned}
\end{equation}
Here $\pi$ is a one-to-one box matching at IoU $\geq0.3$ (\Lone, localization).
\Ltwo also requires the correct role.
For \Lthree, each predicted and ground-truth interaction is the set of its (track, role) pairs, $\hat S_k$ and $S_\alpha$.
These sets are matched one-to-one by $\sigma$, so a correct participant placed in the wrong interaction is lost (association).
Hence $\mathrm{TP}_3\leq\mathrm{TP}_2\leq\mathrm{TP}_1$, nested conditions on one box matching, not separate measures.

\paragraph{RQ3 \textperiodcentered{} Dependencies.}
Each ground-truth event pair takes the label predicted between its matched events (\textit{before} without an edge), and a pair with an unmatched event is an error.
Macro-F1 averages F1 over the three labels, Edge F1 scores only the \textit{causes} and \textit{enables} edges, and \depacc{} is the accuracy on pairs with both events matched.
The diagnostics also use the true and false positive rates of edges on dependent and dependency-negative pairs (TPR, FPR) and balanced accuracy (BAcc).

\paragraph{RQ4 \textperiodcentered{} Grounded answers.}
Answers are judged by GPT-OSS-120B.
For question $i$, $A_i$ marks a correct answer, and its evidence passes the thresholds $T_i=\mathbb{I}[\mathrm{mTIoU}_i\geq0.3]$ (key events), $S_i=\mathbb{I}[\Lthree_{\,i}\geq0.3]$ (interactions, credited only when reported within the span of the matched ground-truth interaction) and $D_i=\mathbb{I}[\mathrm{EdgeF1}_i\geq0.3]$ (dependencies).
A question without ground-truth dependencies has $D_i=1$ when no dependency is predicted.
With $G^T_i=T_i$, $G^{ST}_i=T_iS_i$ and $G^{STD}_i=T_iS_iD_i$,
\begin{equation}
\mathrm{GCR}_X=\frac1N\sum_iA_iG^X_i,\qquad \mathrm{SS}_X=1-\frac{\mathrm{GCR}_X}{\mathrm{Acc}}.
\end{equation}
The grounded correct rate (GCR) counts correct answers with evidence that passes the thresholds, and the spurious share (SS) is the share of correct answers whose evidence fails them (thresholds 0.5 and 0.7: Appx.~B).
Both measure agreement with the annotation, not model reasoning.

\def\TabWhereRows{%
Qwen3.5-2B & 0.3 & 0.0 & 0.0 & -- & -- & 12.3 & 3.8 \\
Qwen3.5-9B & 1.1 & 0.6 & 0.6 & 47.6 & -- & 10.6 & 6.3 \\
Qwen3.6-27B & 12.2 & 9.9 & 9.7 & 25.5 & 7.2 & \textbf{33.8} & \textbf{23.2} \\
GLM-4.6V-Flash (9B) & 10.7 & 8.0 & 7.6 & 37.3 & 7.4 & 23.7 & 14.4 \\
Gemma 4 (31B) & 7.8 & 5.8 & 5.8 & 27.6 & \textbf{3.1} & 26.2 & \underline{17.2} \\
InternVL3.5 (38B) & 0.0 & 0.0 & 0.0 & -- & -- & 10.9 & 2.5 \\
Gemini 3.8 Flash & \textbf{19.7} & \textbf{17.9} & \textbf{17.6} & 12.4 & 5.4 & \underline{27.2} & \underline{17.2} \\
\midrule
\method (2B) & 19.2 & \underline{17.5} & 17.2 & \underline{11.2} & 3.7 & 18.5 & 12.3 \\
\method (9B) & \underline{19.3} & \textbf{17.9} & \underline{17.3} & \textbf{8.4} & \underline{3.5} & 14.4 & 6.8 \\
}
\def\TabWhereNGT{7,768}\def\TabWhereNDep{367}
\providecommand{\TW}[2]{\ifcsname tw@#1@#2\endcsname\csname tw@#1@#2\endcsname\else\TODO{}\fi}
\expandafter\def\csname tw@Base2B@loc\endcsname{0.3}
\expandafter\def\csname tw@Base2B@role\endcsname{0.0}
\expandafter\def\csname tw@Base2B@grp\endcsname{0.0}
\expandafter\def\csname tw@Base2B@n_role\endcsname{23}
\expandafter\def\csname tw@Base2B@n_grp\endcsname{0}
\expandafter\def\csname tw@Base2B@found\endcsname{12.3}
\expandafter\def\csname tw@Base2B@typed\endcsname{3.8}
\expandafter\def\csname tw@Base2B@n_gt\endcsname{7,768}
\expandafter\def\csname tw@Base2B@n_dep\endcsname{367}
\expandafter\def\csname tw@Base9B@loc\endcsname{1.1}
\expandafter\def\csname tw@Base9B@role\endcsname{0.6}
\expandafter\def\csname tw@Base9B@grp\endcsname{0.6}
\expandafter\def\csname tw@Base9B@role_err\endcsname{47.6}
\expandafter\def\csname tw@Base9B@n_role\endcsname{63}
\expandafter\def\csname tw@Base9B@n_grp\endcsname{10}
\expandafter\def\csname tw@Base9B@found\endcsname{10.6}
\expandafter\def\csname tw@Base9B@typed\endcsname{6.3}
\expandafter\def\csname tw@Base9B@n_gt\endcsname{7,768}
\expandafter\def\csname tw@Base9B@n_dep\endcsname{367}
\expandafter\def\csname tw@Q36@loc\endcsname{12.2}
\expandafter\def\csname tw@Q36@role\endcsname{9.9}
\expandafter\def\csname tw@Q36@grp\endcsname{9.7}
\expandafter\def\csname tw@Q36@role_err\endcsname{25.5}
\expandafter\def\csname tw@Q36@n_role\endcsname{542}
\expandafter\def\csname tw@Q36@grp_err\endcsname{7.2}
\expandafter\def\csname tw@Q36@n_grp\endcsname{223}
\expandafter\def\csname tw@Q36@found\endcsname{33.8}
\expandafter\def\csname tw@Q36@typed\endcsname{23.2}
\expandafter\def\csname tw@Q36@n_gt\endcsname{7,768}
\expandafter\def\csname tw@Q36@n_dep\endcsname{367}
\expandafter\def\csname tw@GLM46VF@loc\endcsname{10.7}
\expandafter\def\csname tw@GLM46VF@role\endcsname{8.0}
\expandafter\def\csname tw@GLM46VF@grp\endcsname{7.6}
\expandafter\def\csname tw@GLM46VF@role_err\endcsname{37.3}
\expandafter\def\csname tw@GLM46VF@n_role\endcsname{426}
\expandafter\def\csname tw@GLM46VF@grp_err\endcsname{7.4}
\expandafter\def\csname tw@GLM46VF@n_grp\endcsname{204}
\expandafter\def\csname tw@GLM46VF@found\endcsname{23.7}
\expandafter\def\csname tw@GLM46VF@typed\endcsname{14.4}
\expandafter\def\csname tw@GLM46VF@n_gt\endcsname{7,768}
\expandafter\def\csname tw@GLM46VF@n_dep\endcsname{367}
\expandafter\def\csname tw@Gemma4@loc\endcsname{7.8}
\expandafter\def\csname tw@Gemma4@role\endcsname{5.8}
\expandafter\def\csname tw@Gemma4@grp\endcsname{5.8}
\expandafter\def\csname tw@Gemma4@role_err\endcsname{27.6}
\expandafter\def\csname tw@Gemma4@n_role\endcsname{398}
\expandafter\def\csname tw@Gemma4@grp_err\endcsname{3.1}
\expandafter\def\csname tw@Gemma4@n_grp\endcsname{131}
\expandafter\def\csname tw@Gemma4@found\endcsname{26.2}
\expandafter\def\csname tw@Gemma4@typed\endcsname{17.2}
\expandafter\def\csname tw@Gemma4@n_gt\endcsname{7,768}
\expandafter\def\csname tw@Gemma4@n_dep\endcsname{367}
\expandafter\def\csname tw@InternVL35@loc\endcsname{0.0}
\expandafter\def\csname tw@InternVL35@role\endcsname{0.0}
\expandafter\def\csname tw@InternVL35@grp\endcsname{0.0}
\expandafter\def\csname tw@InternVL35@n_role\endcsname{2}
\expandafter\def\csname tw@InternVL35@n_grp\endcsname{0}
\expandafter\def\csname tw@InternVL35@found\endcsname{10.9}
\expandafter\def\csname tw@InternVL35@typed\endcsname{2.5}
\expandafter\def\csname tw@InternVL35@n_gt\endcsname{7,768}
\expandafter\def\csname tw@InternVL35@n_dep\endcsname{367}
\expandafter\def\csname tw@G38F@loc\endcsname{19.7}
\expandafter\def\csname tw@G38F@role\endcsname{17.9}
\expandafter\def\csname tw@G38F@grp\endcsname{17.6}
\expandafter\def\csname tw@G38F@role_err\endcsname{12.4}
\expandafter\def\csname tw@G38F@n_role\endcsname{820}
\expandafter\def\csname tw@G38F@grp_err\endcsname{5.4}
\expandafter\def\csname tw@G38F@n_grp\endcsname{335}
\expandafter\def\csname tw@G38F@found\endcsname{27.2}
\expandafter\def\csname tw@G38F@typed\endcsname{17.2}
\expandafter\def\csname tw@G38F@n_gt\endcsname{7,768}
\expandafter\def\csname tw@G38F@n_dep\endcsname{367}
\expandafter\def\csname tw@A@loc\endcsname{19.2}
\expandafter\def\csname tw@A@role\endcsname{17.5}
\expandafter\def\csname tw@A@grp\endcsname{17.2}
\expandafter\def\csname tw@A@role_err\endcsname{11.2}
\expandafter\def\csname tw@A@n_role\endcsname{792}
\expandafter\def\csname tw@A@grp_err\endcsname{3.7}
\expandafter\def\csname tw@A@n_grp\endcsname{355}
\expandafter\def\csname tw@A@found\endcsname{18.5}
\expandafter\def\csname tw@A@typed\endcsname{12.3}
\expandafter\def\csname tw@A@n_gt\endcsname{7,768}
\expandafter\def\csname tw@A@n_dep\endcsname{367}
\expandafter\def\csname tw@A9@loc\endcsname{19.3}
\expandafter\def\csname tw@A9@role\endcsname{17.9}
\expandafter\def\csname tw@A9@grp\endcsname{17.3}
\expandafter\def\csname tw@A9@role_err\endcsname{8.4}
\expandafter\def\csname tw@A9@n_role\endcsname{850}
\expandafter\def\csname tw@A9@grp_err\endcsname{3.5}
\expandafter\def\csname tw@A9@n_grp\endcsname{376}
\expandafter\def\csname tw@A9@found\endcsname{14.4}
\expandafter\def\csname tw@A9@typed\endcsname{6.8}
\expandafter\def\csname tw@A9@n_gt\endcsname{7,768}
\expandafter\def\csname tw@A9@n_dep\endcsname{367}
\def\TabRoleRows{%
Qwen3.6-27B & 10.3 / 9.4 & 16.6 / 12.9 & 13.6 / 10.1 & 12.2 / 9.9 \\
Qwen3.8-27B & 14.2 / 12.7 & 22.6 / 18.7 & 18.2 / 14.7 & 16.5 / 14.0 \\
GLM-4.6V-Flash (9B) & 10.4 / 9.1 & 13.3 / 7.5 & 10.7 / 6.8 & 10.7 / 8.0 \\
\midrule
GPT-6.1 Sol & \textbf{30.5} / \textbf{29.8} & \textbf{42.5} / \textbf{37.1} & \textbf{30.5} / \textbf{27.7} & \textbf{31.4} / \textbf{29.5} \\
Gemini 3.8 Flash & \underline{18.2} / \underline{17.1} & \underline{28.9} / \underline{26.2} & 19.9 / 17.3 & \underline{19.7} / \underline{17.9} \\
\midrule
Joint SFT (2B) & 15.9 / 15.3 & 22.2 / 18.6 & 16.9 / 14.5 & 16.8 / 15.3 \\
\method{} (2B) & 17.8 / 16.8 & 20.9 / 17.6 & \underline{20.5} / \underline{18.4} & 19.2 / 17.5 \\
}
\def\TabConcRows{%
Qwen3.6-27B & 28.8 & 65.6 & 2.2 & 1.5 & 1.9 & 3.4 \\
Qwen3.8-27B & 38.4 & 51.9 & 2.0 & 3.3 & 4.4 & 5.6 \\
GLM-4.6V-Flash (9B) & 30.9 & 65.6 & 0.9 & 0.8 & 1.9 & 1.4 \\
\midrule
GPT-6.1 Sol & 38.4 & 37.8 & 4.7 & 6.7 & \textbf{12.3} & \textbf{19.3} \\
Gemini 3.8 Flash & 47.2 & 40.1 & 3.7 & 4.2 & 4.7 & 9.2 \\
\midrule
Joint SFT (2B) & 58.5 & 32.4 & 2.0 & 1.5 & \underline{5.5} & 7.0 \\
\method{} (2B) & 48.1 & 43.3 & 1.3 & 2.2 & 5.1 & \underline{9.3} \\
}
\def\TabSuppInterRows{%
Qwen3.6-27B & 25.5 & 7.2 \\
Qwen3.8-27B & 18.6 & 4.7 \\
GLM-4.6V-Flash (9B) & 37.3 & 7.4 \\
Gemma 4 (31B) & 27.6 & 3.1 \\
\midrule
Gemini 3.8 Flash & 12.4 & 5.4 \\
Gemini Robotics-ER 2 & 13.9 & 5.2 \\
Gemini 3.1 Pro & 12.3 & 5.5 \\
GPT-6.1 Sol & 9.7 & 5.0 \\
\midrule
Joint SFT (2B) & 14.4 & 2.3 \\
\method{} (2B) & 11.2 & 3.7 \\
\method{} (9B) & 8.4 & 3.5 \\
}
\def\TabRoleRowsAll{%
Qwen3.6-27B & 10.3 / 9.4 & 16.6 / 12.9 & 13.6 / 10.1 & 12.2 / 9.9 \\
Qwen3.8-27B & 14.2 / 12.7 & 22.6 / 18.7 & 18.2 / 14.7 & 16.5 / 14.0 \\
GLM-4.6V-Flash (9B) & 10.4 / 9.1 & 13.3 / 7.5 & 10.7 / 6.8 & 10.7 / 8.0 \\
Gemma 4 (31B) & 5.3 / 4.2 & 13.6 / 8.8 & 9.7 / 7.2 & 7.8 / 5.8 \\
\midrule
Gemini 3.8 Flash & \underline{18.2} / \underline{17.1} & \underline{28.9} / \underline{26.2} & 19.9 / 17.3 & \underline{19.7} / \underline{17.9} \\
Gemini Robotics-ER 2 & 9.4 / 9.0 & 15.1 / 13.9 & 10.9 / 9.3 & 10.5 / 9.5 \\
Gemini 3.1 Pro & 16.9 / 16.1 & 25.0 / 23.2 & 18.1 / 15.6 & 18.0 / 16.5 \\
GPT-6.1 Sol & \textbf{30.5} / \textbf{29.8} & \textbf{42.5} / \textbf{37.1} & \textbf{30.5} / \textbf{27.7} & \textbf{31.4} / \textbf{29.5} \\
\midrule
Joint SFT (2B) & 15.9 / 15.3 & 22.2 / 18.6 & 16.9 / 14.5 & 16.8 / 15.3 \\
\method{} (2B) & 17.8 / 16.8 & 20.9 / 17.6 & 20.5 / 18.4 & 19.2 / 17.5 \\
\method{} (9B) & 16.6 / 15.6 & 25.2 / 20.4 & \underline{21.5} / \underline{20.2} & 19.3 / \underline{17.9} \\
}
\def\TabConcRowsAll{%
Qwen3.6-27B & 28.8 & 65.6 & 2.2 & 1.5 & 1.9 & 3.4 \\
Qwen3.8-27B & 38.4 & 51.9 & 2.0 & 3.3 & 4.4 & 5.6 \\
GLM-4.6V-Flash (9B) & 30.9 & 65.6 & 0.9 & 0.8 & 1.9 & 1.4 \\
Gemma 4 (31B) & 37.9 & 60.5 & 0.4 & 0.6 & 0.6 & 1.5 \\
\midrule
Gemini 3.8 Flash & 47.2 & 40.1 & 3.7 & 4.2 & 4.7 & 9.2 \\
Gemini Robotics-ER 2 & 60.7 & 32.6 & 2.1 & 2.3 & 2.2 & 4.2 \\
Gemini 3.1 Pro & 48.6 & 39.1 & 3.9 & 3.6 & 4.7 & 8.8 \\
GPT-6.1 Sol & 38.4 & 37.8 & 4.7 & 6.7 & \textbf{12.3} & \textbf{19.3} \\
\midrule
Joint SFT (2B) & 58.5 & 32.4 & 2.0 & 1.5 & 5.5 & 7.0 \\
\method{} (2B) & 48.1 & 43.3 & 1.3 & 2.2 & 5.1 & \underline{9.3} \\
\method{} (9B) & 51.6 & 38.3 & 1.9 & 2.0 & \underline{6.3} & 7.6 \\
}
\def\TINGT{7,768}\def\TINConc{907}\def\TINNonc{2,327}\def\TINConcUnsc{0}\def\TINNoncUnsc{19}
\providecommand{\TI}[2]{\ifcsname ti@#1@#2\endcsname\csname ti@#1@#2\endcsname\else\TODO{}\fi}
\expandafter\def\csname ti@Q36@n_gt\endcsname{7,768}
\expandafter\def\csname ti@Q36@hand_loc\endcsname{10.3}
\expandafter\def\csname ti@Q36@hand_role\endcsname{9.4}
\expandafter\def\csname ti@Q36@n_hand\endcsname{3,922}
\expandafter\def\csname ti@Q36@tool_loc\endcsname{16.6}
\expandafter\def\csname ti@Q36@tool_role\endcsname{12.9}
\expandafter\def\csname ti@Q36@n_tool\endcsname{603}
\expandafter\def\csname ti@Q36@target_loc\endcsname{13.6}
\expandafter\def\csname ti@Q36@target_role\endcsname{10.1}
\expandafter\def\csname ti@Q36@n_target\endcsname{3,243}
\expandafter\def\csname ti@Q36@all_loc\endcsname{12.2}
\expandafter\def\csname ti@Q36@all_role\endcsname{9.9}
\expandafter\def\csname ti@Q36@all_assoc\endcsname{9.7}
\expandafter\def\csname ti@Q36@role_err\endcsname{25.5}
\expandafter\def\csname ti@Q36@n_role_err\endcsname{542}
\expandafter\def\csname ti@Q36@assoc_err\endcsname{7.2}
\expandafter\def\csname ti@Q36@n_assoc_err\endcsname{223}
\expandafter\def\csname ti@Q36@n_conc\endcsname{907}
\expandafter\def\csname ti@Q36@n_conc_unscorable\endcsname{0}
\expandafter\def\csname ti@Q36@conc_event_miss\endcsname{28.8}
\expandafter\def\csname ti@Q36@conc_loc_fail\endcsname{65.6}
\expandafter\def\csname ti@Q36@conc_role_fail\endcsname{2.2}
\expandafter\def\csname ti@Q36@conc_assoc_fail\endcsname{1.5}
\expandafter\def\csname ti@Q36@conc_recovered\endcsname{1.9}
\expandafter\def\csname ti@Q36@n_conc_matched\endcsname{646}
\expandafter\def\csname ti@Q36@conc_recovered_matched\endcsname{2.6}
\expandafter\def\csname ti@Q36@n_nonc\endcsname{2,327}
\expandafter\def\csname ti@Q36@n_nonc_unscorable\endcsname{19}
\expandafter\def\csname ti@Q36@nonc_event_miss\endcsname{43.7}
\expandafter\def\csname ti@Q36@nonc_loc_fail\endcsname{51.7}
\expandafter\def\csname ti@Q36@nonc_role_fail\endcsname{0.9}
\expandafter\def\csname ti@Q36@nonc_assoc_fail\endcsname{0.3}
\expandafter\def\csname ti@Q36@nonc_recovered\endcsname{3.4}
\expandafter\def\csname ti@Q36@n_nonc_matched\endcsname{1,310}
\expandafter\def\csname ti@Q36@nonc_recovered_matched\endcsname{6.0}
\expandafter\def\csname ti@Q38@n_gt\endcsname{7,768}
\expandafter\def\csname ti@Q38@hand_loc\endcsname{14.2}
\expandafter\def\csname ti@Q38@hand_role\endcsname{12.7}
\expandafter\def\csname ti@Q38@n_hand\endcsname{3,922}
\expandafter\def\csname ti@Q38@tool_loc\endcsname{22.6}
\expandafter\def\csname ti@Q38@tool_role\endcsname{18.7}
\expandafter\def\csname ti@Q38@n_tool\endcsname{603}
\expandafter\def\csname ti@Q38@target_loc\endcsname{18.2}
\expandafter\def\csname ti@Q38@target_role\endcsname{14.7}
\expandafter\def\csname ti@Q38@n_target\endcsname{3,243}
\expandafter\def\csname ti@Q38@all_loc\endcsname{16.5}
\expandafter\def\csname ti@Q38@all_role\endcsname{14.0}
\expandafter\def\csname ti@Q38@all_assoc\endcsname{13.6}
\expandafter\def\csname ti@Q38@role_err\endcsname{18.6}
\expandafter\def\csname ti@Q38@n_role_err\endcsname{726}
\expandafter\def\csname ti@Q38@assoc_err\endcsname{4.7}
\expandafter\def\csname ti@Q38@n_assoc_err\endcsname{339}
\expandafter\def\csname ti@Q38@n_conc\endcsname{907}
\expandafter\def\csname ti@Q38@n_conc_unscorable\endcsname{0}
\expandafter\def\csname ti@Q38@conc_event_miss\endcsname{38.4}
\expandafter\def\csname ti@Q38@conc_loc_fail\endcsname{51.9}
\expandafter\def\csname ti@Q38@conc_role_fail\endcsname{2.0}
\expandafter\def\csname ti@Q38@conc_assoc_fail\endcsname{3.3}
\expandafter\def\csname ti@Q38@conc_recovered\endcsname{4.4}
\expandafter\def\csname ti@Q38@n_conc_matched\endcsname{559}
\expandafter\def\csname ti@Q38@conc_recovered_matched\endcsname{7.2}
\expandafter\def\csname ti@Q38@n_nonc\endcsname{2,327}
\expandafter\def\csname ti@Q38@n_nonc_unscorable\endcsname{19}
\expandafter\def\csname ti@Q38@nonc_event_miss\endcsname{52.9}
\expandafter\def\csname ti@Q38@nonc_loc_fail\endcsname{39.7}
\expandafter\def\csname ti@Q38@nonc_role_fail\endcsname{0.9}
\expandafter\def\csname ti@Q38@nonc_assoc_fail\endcsname{0.9}
\expandafter\def\csname ti@Q38@nonc_recovered\endcsname{5.6}
\expandafter\def\csname ti@Q38@n_nonc_matched\endcsname{1,097}
\expandafter\def\csname ti@Q38@nonc_recovered_matched\endcsname{11.9}
\expandafter\def\csname ti@GLM46VF@n_gt\endcsname{7,768}
\expandafter\def\csname ti@GLM46VF@hand_loc\endcsname{10.4}
\expandafter\def\csname ti@GLM46VF@hand_role\endcsname{9.1}
\expandafter\def\csname ti@GLM46VF@n_hand\endcsname{3,922}
\expandafter\def\csname ti@GLM46VF@tool_loc\endcsname{13.3}
\expandafter\def\csname ti@GLM46VF@tool_role\endcsname{7.5}
\expandafter\def\csname ti@GLM46VF@n_tool\endcsname{603}
\expandafter\def\csname ti@GLM46VF@target_loc\endcsname{10.7}
\expandafter\def\csname ti@GLM46VF@target_role\endcsname{6.8}
\expandafter\def\csname ti@GLM46VF@n_target\endcsname{3,243}
\expandafter\def\csname ti@GLM46VF@all_loc\endcsname{10.7}
\expandafter\def\csname ti@GLM46VF@all_role\endcsname{8.0}
\expandafter\def\csname ti@GLM46VF@all_assoc\endcsname{7.6}
\expandafter\def\csname ti@GLM46VF@role_err\endcsname{37.3}
\expandafter\def\csname ti@GLM46VF@n_role_err\endcsname{426}
\expandafter\def\csname ti@GLM46VF@assoc_err\endcsname{7.4}
\expandafter\def\csname ti@GLM46VF@n_assoc_err\endcsname{204}
\expandafter\def\csname ti@GLM46VF@n_conc\endcsname{907}
\expandafter\def\csname ti@GLM46VF@n_conc_unscorable\endcsname{0}
\expandafter\def\csname ti@GLM46VF@conc_event_miss\endcsname{30.9}
\expandafter\def\csname ti@GLM46VF@conc_loc_fail\endcsname{65.6}
\expandafter\def\csname ti@GLM46VF@conc_role_fail\endcsname{0.9}
\expandafter\def\csname ti@GLM46VF@conc_assoc_fail\endcsname{0.8}
\expandafter\def\csname ti@GLM46VF@conc_recovered\endcsname{1.9}
\expandafter\def\csname ti@GLM46VF@n_conc_matched\endcsname{627}
\expandafter\def\csname ti@GLM46VF@conc_recovered_matched\endcsname{2.7}
\expandafter\def\csname ti@GLM46VF@n_nonc\endcsname{2,327}
\expandafter\def\csname ti@GLM46VF@n_nonc_unscorable\endcsname{19}
\expandafter\def\csname ti@GLM46VF@nonc_event_miss\endcsname{47.9}
\expandafter\def\csname ti@GLM46VF@nonc_loc_fail\endcsname{48.9}
\expandafter\def\csname ti@GLM46VF@nonc_role_fail\endcsname{1.3}
\expandafter\def\csname ti@GLM46VF@nonc_assoc_fail\endcsname{0.5}
\expandafter\def\csname ti@GLM46VF@nonc_recovered\endcsname{1.4}
\expandafter\def\csname ti@GLM46VF@n_nonc_matched\endcsname{1,212}
\expandafter\def\csname ti@GLM46VF@nonc_recovered_matched\endcsname{2.7}
\expandafter\def\csname ti@Gemma4@n_gt\endcsname{7,768}
\expandafter\def\csname ti@Gemma4@hand_loc\endcsname{5.3}
\expandafter\def\csname ti@Gemma4@hand_role\endcsname{4.2}
\expandafter\def\csname ti@Gemma4@n_hand\endcsname{3,922}
\expandafter\def\csname ti@Gemma4@tool_loc\endcsname{13.6}
\expandafter\def\csname ti@Gemma4@tool_role\endcsname{8.8}
\expandafter\def\csname ti@Gemma4@n_tool\endcsname{603}
\expandafter\def\csname ti@Gemma4@target_loc\endcsname{9.7}
\expandafter\def\csname ti@Gemma4@target_role\endcsname{7.2}
\expandafter\def\csname ti@Gemma4@n_target\endcsname{3,243}
\expandafter\def\csname ti@Gemma4@all_loc\endcsname{7.8}
\expandafter\def\csname ti@Gemma4@all_role\endcsname{5.8}
\expandafter\def\csname ti@Gemma4@all_assoc\endcsname{5.8}
\expandafter\def\csname ti@Gemma4@role_err\endcsname{27.6}
\expandafter\def\csname ti@Gemma4@n_role_err\endcsname{398}
\expandafter\def\csname ti@Gemma4@assoc_err\endcsname{3.1}
\expandafter\def\csname ti@Gemma4@n_assoc_err\endcsname{131}
\expandafter\def\csname ti@Gemma4@n_conc\endcsname{907}
\expandafter\def\csname ti@Gemma4@n_conc_unscorable\endcsname{0}
\expandafter\def\csname ti@Gemma4@conc_event_miss\endcsname{37.9}
\expandafter\def\csname ti@Gemma4@conc_loc_fail\endcsname{60.5}
\expandafter\def\csname ti@Gemma4@conc_role_fail\endcsname{0.4}
\expandafter\def\csname ti@Gemma4@conc_assoc_fail\endcsname{0.6}
\expandafter\def\csname ti@Gemma4@conc_recovered\endcsname{0.6}
\expandafter\def\csname ti@Gemma4@n_conc_matched\endcsname{563}
\expandafter\def\csname ti@Gemma4@conc_recovered_matched\endcsname{0.9}
\expandafter\def\csname ti@Gemma4@n_nonc\endcsname{2,327}
\expandafter\def\csname ti@Gemma4@n_nonc_unscorable\endcsname{19}
\expandafter\def\csname ti@Gemma4@nonc_event_miss\endcsname{49.8}
\expandafter\def\csname ti@Gemma4@nonc_loc_fail\endcsname{48.3}
\expandafter\def\csname ti@Gemma4@nonc_role_fail\endcsname{0.3}
\expandafter\def\csname ti@Gemma4@nonc_assoc_fail\endcsname{0.1}
\expandafter\def\csname ti@Gemma4@nonc_recovered\endcsname{1.5}
\expandafter\def\csname ti@Gemma4@n_nonc_matched\endcsname{1,167}
\expandafter\def\csname ti@Gemma4@nonc_recovered_matched\endcsname{3.1}
\expandafter\def\csname ti@G38F@n_gt\endcsname{7,768}
\expandafter\def\csname ti@G38F@hand_loc\endcsname{18.2}
\expandafter\def\csname ti@G38F@hand_role\endcsname{17.1}
\expandafter\def\csname ti@G38F@n_hand\endcsname{3,922}
\expandafter\def\csname ti@G38F@tool_loc\endcsname{28.9}
\expandafter\def\csname ti@G38F@tool_role\endcsname{26.2}
\expandafter\def\csname ti@G38F@n_tool\endcsname{603}
\expandafter\def\csname ti@G38F@target_loc\endcsname{19.9}
\expandafter\def\csname ti@G38F@target_role\endcsname{17.3}
\expandafter\def\csname ti@G38F@n_target\endcsname{3,243}
\expandafter\def\csname ti@G38F@all_loc\endcsname{19.7}
\expandafter\def\csname ti@G38F@all_role\endcsname{17.9}
\expandafter\def\csname ti@G38F@all_assoc\endcsname{17.6}
\expandafter\def\csname ti@G38F@role_err\endcsname{12.4}
\expandafter\def\csname ti@G38F@n_role_err\endcsname{820}
\expandafter\def\csname ti@G38F@assoc_err\endcsname{5.4}
\expandafter\def\csname ti@G38F@n_assoc_err\endcsname{335}
\expandafter\def\csname ti@G38F@n_conc\endcsname{907}
\expandafter\def\csname ti@G38F@n_conc_unscorable\endcsname{0}
\expandafter\def\csname ti@G38F@conc_event_miss\endcsname{47.2}
\expandafter\def\csname ti@G38F@conc_loc_fail\endcsname{40.1}
\expandafter\def\csname ti@G38F@conc_role_fail\endcsname{3.7}
\expandafter\def\csname ti@G38F@conc_assoc_fail\endcsname{4.2}
\expandafter\def\csname ti@G38F@conc_recovered\endcsname{4.7}
\expandafter\def\csname ti@G38F@n_conc_matched\endcsname{479}
\expandafter\def\csname ti@G38F@conc_recovered_matched\endcsname{9.0}
\expandafter\def\csname ti@G38F@n_nonc\endcsname{2,327}
\expandafter\def\csname ti@G38F@n_nonc_unscorable\endcsname{19}
\expandafter\def\csname ti@G38F@nonc_event_miss\endcsname{52.0}
\expandafter\def\csname ti@G38F@nonc_loc_fail\endcsname{36.9}
\expandafter\def\csname ti@G38F@nonc_role_fail\endcsname{1.4}
\expandafter\def\csname ti@G38F@nonc_assoc_fail\endcsname{0.6}
\expandafter\def\csname ti@G38F@nonc_recovered\endcsname{9.2}
\expandafter\def\csname ti@G38F@n_nonc_matched\endcsname{1,117}
\expandafter\def\csname ti@G38F@nonc_recovered_matched\endcsname{19.1}
\expandafter\def\csname ti@GER2@n_gt\endcsname{7,768}
\expandafter\def\csname ti@GER2@hand_loc\endcsname{9.4}
\expandafter\def\csname ti@GER2@hand_role\endcsname{9.0}
\expandafter\def\csname ti@GER2@n_hand\endcsname{3,922}
\expandafter\def\csname ti@GER2@tool_loc\endcsname{15.1}
\expandafter\def\csname ti@GER2@tool_role\endcsname{13.9}
\expandafter\def\csname ti@GER2@n_tool\endcsname{603}
\expandafter\def\csname ti@GER2@target_loc\endcsname{10.9}
\expandafter\def\csname ti@GER2@target_role\endcsname{9.3}
\expandafter\def\csname ti@GER2@n_target\endcsname{3,243}
\expandafter\def\csname ti@GER2@all_loc\endcsname{10.5}
\expandafter\def\csname ti@GER2@all_role\endcsname{9.5}
\expandafter\def\csname ti@GER2@all_assoc\endcsname{9.3}
\expandafter\def\csname ti@GER2@role_err\endcsname{13.9}
\expandafter\def\csname ti@GER2@n_role_err\endcsname{446}
\expandafter\def\csname ti@GER2@assoc_err\endcsname{5.2}
\expandafter\def\csname ti@GER2@n_assoc_err\endcsname{174}
\expandafter\def\csname ti@GER2@n_conc\endcsname{907}
\expandafter\def\csname ti@GER2@n_conc_unscorable\endcsname{0}
\expandafter\def\csname ti@GER2@conc_event_miss\endcsname{60.7}
\expandafter\def\csname ti@GER2@conc_loc_fail\endcsname{32.6}
\expandafter\def\csname ti@GER2@conc_role_fail\endcsname{2.1}
\expandafter\def\csname ti@GER2@conc_assoc_fail\endcsname{2.3}
\expandafter\def\csname ti@GER2@conc_recovered\endcsname{2.2}
\expandafter\def\csname ti@GER2@n_conc_matched\endcsname{356}
\expandafter\def\csname ti@GER2@conc_recovered_matched\endcsname{5.6}
\expandafter\def\csname ti@GER2@n_nonc\endcsname{2,327}
\expandafter\def\csname ti@GER2@n_nonc_unscorable\endcsname{19}
\expandafter\def\csname ti@GER2@nonc_event_miss\endcsname{66.3}
\expandafter\def\csname ti@GER2@nonc_loc_fail\endcsname{28.2}
\expandafter\def\csname ti@GER2@nonc_role_fail\endcsname{1.0}
\expandafter\def\csname ti@GER2@nonc_assoc_fail\endcsname{0.3}
\expandafter\def\csname ti@GER2@nonc_recovered\endcsname{4.2}
\expandafter\def\csname ti@GER2@n_nonc_matched\endcsname{784}
\expandafter\def\csname ti@GER2@nonc_recovered_matched\endcsname{12.4}
\expandafter\def\csname ti@G31P@n_gt\endcsname{7,768}
\expandafter\def\csname ti@G31P@hand_loc\endcsname{16.9}
\expandafter\def\csname ti@G31P@hand_role\endcsname{16.1}
\expandafter\def\csname ti@G31P@n_hand\endcsname{3,922}
\expandafter\def\csname ti@G31P@tool_loc\endcsname{25.0}
\expandafter\def\csname ti@G31P@tool_role\endcsname{23.2}
\expandafter\def\csname ti@G31P@n_tool\endcsname{603}
\expandafter\def\csname ti@G31P@target_loc\endcsname{18.1}
\expandafter\def\csname ti@G31P@target_role\endcsname{15.6}
\expandafter\def\csname ti@G31P@n_target\endcsname{3,243}
\expandafter\def\csname ti@G31P@all_loc\endcsname{18.0}
\expandafter\def\csname ti@G31P@all_role\endcsname{16.5}
\expandafter\def\csname ti@G31P@all_assoc\endcsname{16.1}
\expandafter\def\csname ti@G31P@role_err\endcsname{12.3}
\expandafter\def\csname ti@G31P@n_role_err\endcsname{737}
\expandafter\def\csname ti@G31P@assoc_err\endcsname{5.5}
\expandafter\def\csname ti@G31P@n_assoc_err\endcsname{326}
\expandafter\def\csname ti@G31P@n_conc\endcsname{907}
\expandafter\def\csname ti@G31P@n_conc_unscorable\endcsname{0}
\expandafter\def\csname ti@G31P@conc_event_miss\endcsname{48.6}
\expandafter\def\csname ti@G31P@conc_loc_fail\endcsname{39.1}
\expandafter\def\csname ti@G31P@conc_role_fail\endcsname{3.9}
\expandafter\def\csname ti@G31P@conc_assoc_fail\endcsname{3.6}
\expandafter\def\csname ti@G31P@conc_recovered\endcsname{4.7}
\expandafter\def\csname ti@G31P@n_conc_matched\endcsname{466}
\expandafter\def\csname ti@G31P@conc_recovered_matched\endcsname{9.2}
\expandafter\def\csname ti@G31P@n_nonc\endcsname{2,327}
\expandafter\def\csname ti@G31P@n_nonc_unscorable\endcsname{19}
\expandafter\def\csname ti@G31P@nonc_event_miss\endcsname{58.3}
\expandafter\def\csname ti@G31P@nonc_loc_fail\endcsname{31.2}
\expandafter\def\csname ti@G31P@nonc_role_fail\endcsname{1.2}
\expandafter\def\csname ti@G31P@nonc_assoc_fail\endcsname{0.5}
\expandafter\def\csname ti@G31P@nonc_recovered\endcsname{8.8}
\expandafter\def\csname ti@G31P@n_nonc_matched\endcsname{971}
\expandafter\def\csname ti@G31P@nonc_recovered_matched\endcsname{21.1}
\expandafter\def\csname ti@GPT@n_gt\endcsname{7,768}
\expandafter\def\csname ti@GPT@hand_loc\endcsname{30.5}
\expandafter\def\csname ti@GPT@hand_role\endcsname{29.8}
\expandafter\def\csname ti@GPT@n_hand\endcsname{3,922}
\expandafter\def\csname ti@GPT@tool_loc\endcsname{42.5}
\expandafter\def\csname ti@GPT@tool_role\endcsname{37.1}
\expandafter\def\csname ti@GPT@n_tool\endcsname{603}
\expandafter\def\csname ti@GPT@target_loc\endcsname{30.5}
\expandafter\def\csname ti@GPT@target_role\endcsname{27.7}
\expandafter\def\csname ti@GPT@n_target\endcsname{3,243}
\expandafter\def\csname ti@GPT@all_loc\endcsname{31.4}
\expandafter\def\csname ti@GPT@all_role\endcsname{29.5}
\expandafter\def\csname ti@GPT@all_assoc\endcsname{28.5}
\expandafter\def\csname ti@GPT@role_err\endcsname{9.7}
\expandafter\def\csname ti@GPT@n_role_err\endcsname{1,244}
\expandafter\def\csname ti@GPT@assoc_err\endcsname{5.0}
\expandafter\def\csname ti@GPT@n_assoc_err\endcsname{575}
\expandafter\def\csname ti@GPT@n_conc\endcsname{907}
\expandafter\def\csname ti@GPT@n_conc_unscorable\endcsname{0}
\expandafter\def\csname ti@GPT@conc_event_miss\endcsname{38.4}
\expandafter\def\csname ti@GPT@conc_loc_fail\endcsname{37.8}
\expandafter\def\csname ti@GPT@conc_role_fail\endcsname{4.7}
\expandafter\def\csname ti@GPT@conc_assoc_fail\endcsname{6.7}
\expandafter\def\csname ti@GPT@conc_recovered\endcsname{12.3}
\expandafter\def\csname ti@GPT@n_conc_matched\endcsname{559}
\expandafter\def\csname ti@GPT@conc_recovered_matched\endcsname{20.0}
\expandafter\def\csname ti@GPT@n_nonc\endcsname{2,327}
\expandafter\def\csname ti@GPT@n_nonc_unscorable\endcsname{19}
\expandafter\def\csname ti@GPT@nonc_event_miss\endcsname{42.9}
\expandafter\def\csname ti@GPT@nonc_loc_fail\endcsname{34.3}
\expandafter\def\csname ti@GPT@nonc_role_fail\endcsname{1.9}
\expandafter\def\csname ti@GPT@nonc_assoc_fail\endcsname{1.5}
\expandafter\def\csname ti@GPT@nonc_recovered\endcsname{19.3}
\expandafter\def\csname ti@GPT@n_nonc_matched\endcsname{1,328}
\expandafter\def\csname ti@GPT@nonc_recovered_matched\endcsname{33.8}
\expandafter\def\csname ti@M1F@n_gt\endcsname{7,768}
\expandafter\def\csname ti@M1F@hand_loc\endcsname{15.9}
\expandafter\def\csname ti@M1F@hand_role\endcsname{15.3}
\expandafter\def\csname ti@M1F@n_hand\endcsname{3,922}
\expandafter\def\csname ti@M1F@tool_loc\endcsname{22.2}
\expandafter\def\csname ti@M1F@tool_role\endcsname{18.6}
\expandafter\def\csname ti@M1F@n_tool\endcsname{603}
\expandafter\def\csname ti@M1F@target_loc\endcsname{16.9}
\expandafter\def\csname ti@M1F@target_role\endcsname{14.5}
\expandafter\def\csname ti@M1F@n_target\endcsname{3,243}
\expandafter\def\csname ti@M1F@all_loc\endcsname{16.8}
\expandafter\def\csname ti@M1F@all_role\endcsname{15.3}
\expandafter\def\csname ti@M1F@all_assoc\endcsname{14.9}
\expandafter\def\csname ti@M1F@role_err\endcsname{14.4}
\expandafter\def\csname ti@M1F@n_role_err\endcsname{681}
\expandafter\def\csname ti@M1F@assoc_err\endcsname{2.3}
\expandafter\def\csname ti@M1F@n_assoc_err\endcsname{309}
\expandafter\def\csname ti@M1F@n_conc\endcsname{907}
\expandafter\def\csname ti@M1F@n_conc_unscorable\endcsname{0}
\expandafter\def\csname ti@M1F@conc_event_miss\endcsname{58.5}
\expandafter\def\csname ti@M1F@conc_loc_fail\endcsname{32.4}
\expandafter\def\csname ti@M1F@conc_role_fail\endcsname{2.0}
\expandafter\def\csname ti@M1F@conc_assoc_fail\endcsname{1.5}
\expandafter\def\csname ti@M1F@conc_recovered\endcsname{5.5}
\expandafter\def\csname ti@M1F@n_conc_matched\endcsname{376}
\expandafter\def\csname ti@M1F@conc_recovered_matched\endcsname{13.3}
\expandafter\def\csname ti@M1F@n_nonc\endcsname{2,327}
\expandafter\def\csname ti@M1F@n_nonc_unscorable\endcsname{19}
\expandafter\def\csname ti@M1F@nonc_event_miss\endcsname{66.3}
\expandafter\def\csname ti@M1F@nonc_loc_fail\endcsname{24.8}
\expandafter\def\csname ti@M1F@nonc_role_fail\endcsname{1.2}
\expandafter\def\csname ti@M1F@nonc_assoc_fail\endcsname{0.6}
\expandafter\def\csname ti@M1F@nonc_recovered\endcsname{7.0}
\expandafter\def\csname ti@M1F@n_nonc_matched\endcsname{785}
\expandafter\def\csname ti@M1F@nonc_recovered_matched\endcsname{20.9}
\expandafter\def\csname ti@A@n_gt\endcsname{7,768}
\expandafter\def\csname ti@A@hand_loc\endcsname{17.8}
\expandafter\def\csname ti@A@hand_role\endcsname{16.8}
\expandafter\def\csname ti@A@n_hand\endcsname{3,922}
\expandafter\def\csname ti@A@tool_loc\endcsname{20.9}
\expandafter\def\csname ti@A@tool_role\endcsname{17.6}
\expandafter\def\csname ti@A@n_tool\endcsname{603}
\expandafter\def\csname ti@A@target_loc\endcsname{20.5}
\expandafter\def\csname ti@A@target_role\endcsname{18.4}
\expandafter\def\csname ti@A@n_target\endcsname{3,243}
\expandafter\def\csname ti@A@all_loc\endcsname{19.2}
\expandafter\def\csname ti@A@all_role\endcsname{17.5}
\expandafter\def\csname ti@A@all_assoc\endcsname{17.2}
\expandafter\def\csname ti@A@role_err\endcsname{11.2}
\expandafter\def\csname ti@A@n_role_err\endcsname{792}
\expandafter\def\csname ti@A@assoc_err\endcsname{3.7}
\expandafter\def\csname ti@A@n_assoc_err\endcsname{355}
\expandafter\def\csname ti@A@n_conc\endcsname{907}
\expandafter\def\csname ti@A@n_conc_unscorable\endcsname{0}
\expandafter\def\csname ti@A@conc_event_miss\endcsname{48.1}
\expandafter\def\csname ti@A@conc_loc_fail\endcsname{43.3}
\expandafter\def\csname ti@A@conc_role_fail\endcsname{1.3}
\expandafter\def\csname ti@A@conc_assoc_fail\endcsname{2.2}
\expandafter\def\csname ti@A@conc_recovered\endcsname{5.1}
\expandafter\def\csname ti@A@n_conc_matched\endcsname{471}
\expandafter\def\csname ti@A@conc_recovered_matched\endcsname{9.8}
\expandafter\def\csname ti@A@n_nonc\endcsname{2,327}
\expandafter\def\csname ti@A@n_nonc_unscorable\endcsname{19}
\expandafter\def\csname ti@A@nonc_event_miss\endcsname{56.8}
\expandafter\def\csname ti@A@nonc_loc_fail\endcsname{32.2}
\expandafter\def\csname ti@A@nonc_role_fail\endcsname{1.0}
\expandafter\def\csname ti@A@nonc_assoc_fail\endcsname{0.7}
\expandafter\def\csname ti@A@nonc_recovered\endcsname{9.3}
\expandafter\def\csname ti@A@n_nonc_matched\endcsname{1,006}
\expandafter\def\csname ti@A@nonc_recovered_matched\endcsname{21.6}
\expandafter\def\csname ti@A9@n_gt\endcsname{7,768}
\expandafter\def\csname ti@A9@hand_loc\endcsname{16.6}
\expandafter\def\csname ti@A9@hand_role\endcsname{15.6}
\expandafter\def\csname ti@A9@n_hand\endcsname{3,922}
\expandafter\def\csname ti@A9@tool_loc\endcsname{25.2}
\expandafter\def\csname ti@A9@tool_role\endcsname{20.4}
\expandafter\def\csname ti@A9@n_tool\endcsname{603}
\expandafter\def\csname ti@A9@target_loc\endcsname{21.5}
\expandafter\def\csname ti@A9@target_role\endcsname{20.2}
\expandafter\def\csname ti@A9@n_target\endcsname{3,243}
\expandafter\def\csname ti@A9@all_loc\endcsname{19.3}
\expandafter\def\csname ti@A9@all_role\endcsname{17.9}
\expandafter\def\csname ti@A9@all_assoc\endcsname{17.3}
\expandafter\def\csname ti@A9@role_err\endcsname{8.4}
\expandafter\def\csname ti@A9@n_role_err\endcsname{850}
\expandafter\def\csname ti@A9@assoc_err\endcsname{3.5}
\expandafter\def\csname ti@A9@n_assoc_err\endcsname{376}
\expandafter\def\csname ti@A9@n_conc\endcsname{907}
\expandafter\def\csname ti@A9@n_conc_unscorable\endcsname{0}
\expandafter\def\csname ti@A9@conc_event_miss\endcsname{51.6}
\expandafter\def\csname ti@A9@conc_loc_fail\endcsname{38.3}
\expandafter\def\csname ti@A9@conc_role_fail\endcsname{1.9}
\expandafter\def\csname ti@A9@conc_assoc_fail\endcsname{2.0}
\expandafter\def\csname ti@A9@conc_recovered\endcsname{6.3}
\expandafter\def\csname ti@A9@n_conc_matched\endcsname{439}
\expandafter\def\csname ti@A9@conc_recovered_matched\endcsname{13.0}
\expandafter\def\csname ti@A9@n_nonc\endcsname{2,327}
\expandafter\def\csname ti@A9@n_nonc_unscorable\endcsname{19}
\expandafter\def\csname ti@A9@nonc_event_miss\endcsname{59.9}
\expandafter\def\csname ti@A9@nonc_loc_fail\endcsname{29.8}
\expandafter\def\csname ti@A9@nonc_role_fail\endcsname{1.2}
\expandafter\def\csname ti@A9@nonc_assoc_fail\endcsname{1.4}
\expandafter\def\csname ti@A9@nonc_recovered\endcsname{7.6}
\expandafter\def\csname ti@A9@n_nonc_matched\endcsname{932}
\expandafter\def\csname ti@A9@nonc_recovered_matched\endcsname{19.1}

\providecommand{\TI}[2]{\TODO{}}

\subsection{Benchmark results}
\label{sec:exp_analysis}
\label{sec:exp_results}
\label{sec:exp_main}
\providecommand{\TODO}[1]{\textcolor{red}{TODO#1}}
\providecommand{\OR}[3]{\ifcsname or@#1@#2@#3\endcsname\csname or@#1@#2@#3\endcsname\else\TODO{}\fi}
\def\ORNQ{1,639}
\def\ORNRole{8,115}
\def\ORNGrp{842}
\def\ORGrpBase{72.8}
\def\ORNPertLink{0}
\expandafter\def\csname or@GLM46VF@E@L3\endcsname{.149}
\expandafter\def\csname or@GLM46VF@E@MacroF1\endcsname{.336}
\expandafter\def\csname or@GLM46VF@E@acc\endcsname{.254}
\expandafter\def\csname or@GLM46VF@E@ci\endcsname{[+1.9,+5.4]}
\expandafter\def\csname or@GLM46VF@E@dacc\endcsname{+3.6}
\expandafter\def\csname or@GLM46VF@E@down\endcsname{80}
\expandafter\def\csname or@GLM46VF@E@up\endcsname{139}
\expandafter\def\csname or@GLM46VF@EB@acc\endcsname{.243}
\expandafter\def\csname or@GLM46VF@EB@ci\endcsname{[+0.8,+4.3]}
\expandafter\def\csname or@GLM46VF@EB@dacc\endcsname{+2.5}
\expandafter\def\csname or@GLM46VF@EB@down\endcsname{88}
\expandafter\def\csname or@GLM46VF@EB@up\endcsname{129}
\expandafter\def\csname or@GLM46VF@EBR@acc\endcsname{.238}
\expandafter\def\csname or@GLM46VF@EBR@ci\endcsname{[+0.1,+3.8]}
\expandafter\def\csname or@GLM46VF@EBR@dacc\endcsname{+2.0}
\expandafter\def\csname or@GLM46VF@EBR@down\endcsname{103}
\expandafter\def\csname or@GLM46VF@EBR@up\endcsname{135}
\expandafter\def\csname or@GLM46VF@ED@acc\endcsname{.256}
\expandafter\def\csname or@GLM46VF@ED@ci\endcsname{[+2.1,+5.6]}
\expandafter\def\csname or@GLM46VF@ED@dacc\endcsname{+3.8}
\expandafter\def\csname or@GLM46VF@ED@down\endcsname{81}
\expandafter\def\csname or@GLM46VF@ED@up\endcsname{143}
\expandafter\def\csname or@GLM46VF@EI@MacroF1\endcsname{.346}
\expandafter\def\csname or@GLM46VF@EI@acc\endcsname{.250}
\expandafter\def\csname or@GLM46VF@EI@ci\endcsname{[+1.3,+4.9]}
\expandafter\def\csname or@GLM46VF@EI@dacc\endcsname{+3.1}
\expandafter\def\csname or@GLM46VF@EI@down\endcsname{93}
\expandafter\def\csname or@GLM46VF@EI@up\endcsname{144}
\expandafter\def\csname or@GLM46VF@EID@acc\endcsname{.246}
\expandafter\def\csname or@GLM46VF@EID@ci\endcsname{[+1.0,+4.6]}
\expandafter\def\csname or@GLM46VF@EID@dacc\endcsname{+2.7}
\expandafter\def\csname or@GLM46VF@EID@down\endcsname{97}
\expandafter\def\csname or@GLM46VF@EID@up\endcsname{142}
\expandafter\def\csname or@GLM46VF@Eshift@acc\endcsname{.200}
\expandafter\def\csname or@GLM46VF@Eshift@ci\endcsname{[-3.6,-0.1]}
\expandafter\def\csname or@GLM46VF@Eshift@dacc\endcsname{-1.9}
\expandafter\def\csname or@GLM46VF@Eshift@down\endcsname{125}
\expandafter\def\csname or@GLM46VF@Eshift@up\endcsname{94}
\expandafter\def\csname or@GLM46VF@TextEID@acc\endcsname{.101}
\expandafter\def\csname or@GLM46VF@TextEID@ci\endcsname{[-13.9,-9.5]}
\expandafter\def\csname or@GLM46VF@TextEID@dacc\endcsname{-11.7}
\expandafter\def\csname or@GLM46VF@TextEID@down\endcsname{269}
\expandafter\def\csname or@GLM46VF@TextEID@up\endcsname{77}
\expandafter\def\csname or@GLM46VF@V@L3\endcsname{.044}
\expandafter\def\csname or@GLM46VF@V@MacroF1\endcsname{.156}
\expandafter\def\csname or@GLM46VF@V@acc\endcsname{.218}
\expandafter\def\csname or@GLM46VF@V@mTIoU\endcsname{.204}
\expandafter\def\csname or@GLM46VF@grpprobe@exact\endcsname{11.6}
\expandafter\def\csname or@GLM46VF@grpprobe@pair\endcsname{60.1}
\expandafter\def\csname or@GLM46VF@grpprobeText@exact\endcsname{10.6}
\expandafter\def\csname or@GLM46VF@grpprobeText@pair\endcsname{59.1}
\expandafter\def\csname or@GLM46VF@roleprobe@acc\endcsname{53.4}
\expandafter\def\csname or@GLM46VF@roleprobe@accobj\endcsname{49.1}
\expandafter\def\csname or@GLM46VF@roleprobe@ttconf\endcsname{19.5}
\expandafter\def\csname or@GLM46VF@roleprobeText@acc\endcsname{42.9}
\expandafter\def\csname or@GLM46VF@roleprobeText@accobj\endcsname{43.7}
\expandafter\def\csname or@GLM46VF@roleprobeText@ttconf\endcsname{17.2}
\expandafter\def\csname or@Q36@E@L3\endcsname{.150}
\expandafter\def\csname or@Q36@E@MacroF1\endcsname{.328}
\expandafter\def\csname or@Q36@E@acc\endcsname{.248}
\expandafter\def\csname or@Q36@E@ci\endcsname{[+0.7,+4.6]}
\expandafter\def\csname or@Q36@E@dacc\endcsname{+2.7}
\expandafter\def\csname or@Q36@E@down\endcsname{104}
\expandafter\def\csname or@Q36@E@up\endcsname{148}
\expandafter\def\csname or@Q36@EB@acc\endcsname{.245}
\expandafter\def\csname or@Q36@EB@ci\endcsname{[+0.4,+4.2]}
\expandafter\def\csname or@Q36@EB@dacc\endcsname{+2.3}
\expandafter\def\csname or@Q36@EB@down\endcsname{108}
\expandafter\def\csname or@Q36@EB@up\endcsname{146}
\expandafter\def\csname or@Q36@EBR@acc\endcsname{.242}
\expandafter\def\csname or@Q36@EBR@ci\endcsname{[+0.1,+4.0]}
\expandafter\def\csname or@Q36@EBR@dacc\endcsname{+2.0}
\expandafter\def\csname or@Q36@EBR@down\endcsname{116}
\expandafter\def\csname or@Q36@EBR@up\endcsname{149}
\expandafter\def\csname or@Q36@ED@acc\endcsname{.245}
\expandafter\def\csname or@Q36@ED@ci\endcsname{[+0.5,+4.3]}
\expandafter\def\csname or@Q36@ED@dacc\endcsname{+2.4}
\expandafter\def\csname or@Q36@ED@down\endcsname{106}
\expandafter\def\csname or@Q36@ED@up\endcsname{145}
\expandafter\def\csname or@Q36@EI@MacroF1\endcsname{.350}
\expandafter\def\csname or@Q36@EI@acc\endcsname{.249}
\expandafter\def\csname or@Q36@EI@ci\endcsname{[+0.8,+4.6]}
\expandafter\def\csname or@Q36@EI@dacc\endcsname{+2.7}
\expandafter\def\csname or@Q36@EI@down\endcsname{103}
\expandafter\def\csname or@Q36@EI@up\endcsname{148}
\expandafter\def\csname or@Q36@EID@acc\endcsname{.244}
\expandafter\def\csname or@Q36@EID@ci\endcsname{[+0.2,+4.2]}
\expandafter\def\csname or@Q36@EID@dacc\endcsname{+2.3}
\expandafter\def\csname or@Q36@EID@down\endcsname{118}
\expandafter\def\csname or@Q36@EID@up\endcsname{155}
\expandafter\def\csname or@Q36@Eshift@acc\endcsname{.168}
\expandafter\def\csname or@Q36@Eshift@ci\endcsname{[-7.3,-3.4]}
\expandafter\def\csname or@Q36@Eshift@dacc\endcsname{-5.4}
\expandafter\def\csname or@Q36@Eshift@down\endcsname{180}
\expandafter\def\csname or@Q36@Eshift@up\endcsname{92}
\expandafter\def\csname or@Q36@TextEID@acc\endcsname{.103}
\expandafter\def\csname or@Q36@TextEID@ci\endcsname{[-14.1,-9.6]}
\expandafter\def\csname or@Q36@TextEID@dacc\endcsname{-11.8}
\expandafter\def\csname or@Q36@TextEID@down\endcsname{286}
\expandafter\def\csname or@Q36@TextEID@up\endcsname{92}
\expandafter\def\csname or@Q36@V@L3\endcsname{.059}
\expandafter\def\csname or@Q36@V@MacroF1\endcsname{.217}
\expandafter\def\csname or@Q36@V@acc\endcsname{.221}
\expandafter\def\csname or@Q36@V@mTIoU\endcsname{.198}
\expandafter\def\csname or@Q36@grpprobe@exact\endcsname{8.6}
\expandafter\def\csname or@Q36@grpprobe@pair\endcsname{70.7}
\expandafter\def\csname or@Q36@grpprobeText@exact\endcsname{8.9}
\expandafter\def\csname or@Q36@grpprobeText@pair\endcsname{71.8}
\expandafter\def\csname or@Q36@roleprobe@acc\endcsname{78.3}
\expandafter\def\csname or@Q36@roleprobe@accobj\endcsname{64.2}
\expandafter\def\csname or@Q36@roleprobe@ttconf\endcsname{29.0}
\expandafter\def\csname or@Q36@roleprobeText@acc\endcsname{44.2}
\expandafter\def\csname or@Q36@roleprobeText@accobj\endcsname{35.5}
\expandafter\def\csname or@Q36@roleprobeText@ttconf\endcsname{25.5}
\expandafter\def\csname or@Q36@roleNA@acc\endcsname{73.1}
\expandafter\def\csname or@Q36@roleNA@accobj\endcsname{53.3}
\expandafter\def\csname or@Q36@roleNA@ttconf\endcsname{36.1}
\expandafter\def\csname or@Q36@roleNAText@acc\endcsname{43.5}
\expandafter\def\csname or@Q36@roleNAText@accobj\endcsname{23.4}
\expandafter\def\csname or@Q36@roleNAText@ttconf\endcsname{32.5}
\expandafter\def\csname or@GLM46VF@roleNA@acc\endcsname{57.6}
\expandafter\def\csname or@GLM46VF@roleNA@accobj\endcsname{45.9}
\expandafter\def\csname or@GLM46VF@roleNA@ttconf\endcsname{22.8}
\expandafter\def\csname or@GLM46VF@roleNAText@acc\endcsname{41.1}
\expandafter\def\csname or@GLM46VF@roleNAText@accobj\endcsname{43.2}
\expandafter\def\csname or@GLM46VF@roleNAText@ttconf\endcsname{19.0}
\def\ORQNAll{1,639}\def\ORQNDep{302}
\providecommand{\ORQ}[3]{\ifcsname orq@#1@#2@#3\endcsname\csname orq@#1@#2@#3\endcsname\else\TODO{}\fi}
\expandafter\def\csname orq@Q36@VE@n\endcsname{1,639}
\expandafter\def\csname orq@Q36@VE@a\endcsname{22.1}
\expandafter\def\csname orq@Q36@VE@b\endcsname{24.8}
\expandafter\def\csname orq@Q36@VE@d\endcsname{+2.7}
\expandafter\def\csname orq@Q36@VE@lo\endcsname{+0.7}
\expandafter\def\csname orq@Q36@VE@hi\endcsname{+4.5}
\expandafter\def\csname orq@Q36@EEI@n\endcsname{1,639}
\expandafter\def\csname orq@Q36@EEI@a\endcsname{24.8}
\expandafter\def\csname orq@Q36@EEI@b\endcsname{24.3}
\expandafter\def\csname orq@Q36@EEI@d\endcsname{$-$0.5}
\expandafter\def\csname orq@Q36@EEI@lo\endcsname{$-$2.2}
\expandafter\def\csname orq@Q36@EEI@hi\endcsname{+1.2}
\expandafter\def\csname orq@Q36@EIEID@n\endcsname{302}
\expandafter\def\csname orq@Q36@EIEID@a\endcsname{32.5}
\expandafter\def\csname orq@Q36@EIEID@b\endcsname{32.5}
\expandafter\def\csname orq@Q36@EIEID@d\endcsname{+0.0}
\expandafter\def\csname orq@Q36@EIEID@lo\endcsname{$-$3.6}
\expandafter\def\csname orq@Q36@EIEID@hi\endcsname{+3.6}
\expandafter\def\csname orq@Q36@VEID@n\endcsname{302}
\expandafter\def\csname orq@Q36@VEID@a\endcsname{29.5}
\expandafter\def\csname orq@Q36@VEID@b\endcsname{32.5}
\expandafter\def\csname orq@Q36@VEID@d\endcsname{+3.0}
\expandafter\def\csname orq@Q36@VEID@lo\endcsname{$-$2.3}
\expandafter\def\csname orq@Q36@VEID@hi\endcsname{+8.3}
\expandafter\def\csname orq@Q36@Link@n\endcsname{302}
\expandafter\def\csname orq@Q36@Link@a\endcsname{32.5}
\expandafter\def\csname orq@Q36@Link@b\endcsname{35.1}
\expandafter\def\csname orq@Q36@Link@d\endcsname{+2.6}
\expandafter\def\csname orq@Q36@Link@lo\endcsname{$-$0.3}
\expandafter\def\csname orq@Q36@Link@hi\endcsname{+5.6}
\expandafter\def\csname orq@Q36@Role@n\endcsname{1,639}
\expandafter\def\csname orq@Q36@Role@a\endcsname{24.3}
\expandafter\def\csname orq@Q36@Role@b\endcsname{25.0}
\expandafter\def\csname orq@Q36@Role@d\endcsname{+0.7}
\expandafter\def\csname orq@Q36@Role@lo\endcsname{$-$0.5}
\expandafter\def\csname orq@Q36@Role@hi\endcsname{+1.9}
\expandafter\def\csname orq@Q36@Text@n\endcsname{1,639}
\expandafter\def\csname orq@Q36@Text@a\endcsname{22.1}
\expandafter\def\csname orq@Q36@Text@b\endcsname{10.6}
\expandafter\def\csname orq@Q36@Text@d\endcsname{$-$11.5}
\expandafter\def\csname orq@Q36@Text@lo\endcsname{$-$13.8}
\expandafter\def\csname orq@Q36@Text@hi\endcsname{$-$9.3}
\expandafter\def\csname orq@Q36@VE_dep@n\endcsname{302}
\expandafter\def\csname orq@Q36@VE_dep@a\endcsname{29.5}
\expandafter\def\csname orq@Q36@VE_dep@b\endcsname{35.4}
\expandafter\def\csname orq@Q36@VE_dep@d\endcsname{+6.0}
\expandafter\def\csname orq@Q36@VE_dep@lo\endcsname{+1.0}
\expandafter\def\csname orq@Q36@VE_dep@hi\endcsname{+10.9}
\expandafter\def\csname orq@Q36@NoDep@n\endcsname{302}
\expandafter\def\csname orq@Q36@NoDep@a\endcsname{32.5}
\expandafter\def\csname orq@Q36@NoDep@b\endcsname{31.1}
\expandafter\def\csname orq@Q36@NoDep@d\endcsname{$-$1.3}
\expandafter\def\csname orq@Q36@NoDep@lo\endcsname{$-$5.3}
\expandafter\def\csname orq@Q36@NoDep@hi\endcsname{+2.6}
\expandafter\def\csname orq@GLM46VF@VE@n\endcsname{1,639}
\expandafter\def\csname orq@GLM46VF@VE@a\endcsname{21.9}
\expandafter\def\csname orq@GLM46VF@VE@b\endcsname{25.4}
\expandafter\def\csname orq@GLM46VF@VE@d\endcsname{+3.5}
\expandafter\def\csname orq@GLM46VF@VE@lo\endcsname{+1.8}
\expandafter\def\csname orq@GLM46VF@VE@hi\endcsname{+5.2}
\expandafter\def\csname orq@GLM46VF@EEI@n\endcsname{1,639}
\expandafter\def\csname orq@GLM46VF@EEI@a\endcsname{25.4}
\expandafter\def\csname orq@GLM46VF@EEI@b\endcsname{24.6}
\expandafter\def\csname orq@GLM46VF@EEI@d\endcsname{$-$0.9}
\expandafter\def\csname orq@GLM46VF@EEI@lo\endcsname{$-$2.3}
\expandafter\def\csname orq@GLM46VF@EEI@hi\endcsname{+0.5}
\expandafter\def\csname orq@GLM46VF@EIEID@n\endcsname{302}
\expandafter\def\csname orq@GLM46VF@EIEID@a\endcsname{32.1}
\expandafter\def\csname orq@GLM46VF@EIEID@b\endcsname{32.5}
\expandafter\def\csname orq@GLM46VF@EIEID@d\endcsname{+0.3}
\expandafter\def\csname orq@GLM46VF@EIEID@lo\endcsname{$-$2.6}
\expandafter\def\csname orq@GLM46VF@EIEID@hi\endcsname{+3.3}
\expandafter\def\csname orq@GLM46VF@VEID@n\endcsname{302}
\expandafter\def\csname orq@GLM46VF@VEID@a\endcsname{27.5}
\expandafter\def\csname orq@GLM46VF@VEID@b\endcsname{32.5}
\expandafter\def\csname orq@GLM46VF@VEID@d\endcsname{+5.0}
\expandafter\def\csname orq@GLM46VF@VEID@lo\endcsname{+1.0}
\expandafter\def\csname orq@GLM46VF@VEID@hi\endcsname{+8.9}
\expandafter\def\csname orq@GLM46VF@Link@n\endcsname{302}
\expandafter\def\csname orq@GLM46VF@Link@a\endcsname{32.5}
\expandafter\def\csname orq@GLM46VF@Link@b\endcsname{32.1}
\expandafter\def\csname orq@GLM46VF@Link@d\endcsname{$-$0.3}
\expandafter\def\csname orq@GLM46VF@Link@lo\endcsname{$-$2.3}
\expandafter\def\csname orq@GLM46VF@Link@hi\endcsname{+1.7}
\expandafter\def\csname orq@GLM46VF@Role@n\endcsname{1,639}
\expandafter\def\csname orq@GLM46VF@Role@a\endcsname{24.6}
\expandafter\def\csname orq@GLM46VF@Role@b\endcsname{24.3}
\expandafter\def\csname orq@GLM46VF@Role@d\endcsname{$-$0.2}
\expandafter\def\csname orq@GLM46VF@Role@lo\endcsname{$-$1.3}
\expandafter\def\csname orq@GLM46VF@Role@hi\endcsname{+0.9}
\expandafter\def\csname orq@GLM46VF@Text@n\endcsname{1,639}
\expandafter\def\csname orq@GLM46VF@Text@a\endcsname{21.9}
\expandafter\def\csname orq@GLM46VF@Text@b\endcsname{11.0}
\expandafter\def\csname orq@GLM46VF@Text@d\endcsname{$-$10.9}
\expandafter\def\csname orq@GLM46VF@Text@lo\endcsname{$-$13.1}
\expandafter\def\csname orq@GLM46VF@Text@hi\endcsname{$-$8.7}
\expandafter\def\csname orq@GLM46VF@VE_dep@n\endcsname{302}
\expandafter\def\csname orq@GLM46VF@VE_dep@a\endcsname{27.5}
\expandafter\def\csname orq@GLM46VF@VE_dep@b\endcsname{34.8}
\expandafter\def\csname orq@GLM46VF@VE_dep@d\endcsname{+7.3}
\expandafter\def\csname orq@GLM46VF@VE_dep@lo\endcsname{+3.0}
\expandafter\def\csname orq@GLM46VF@VE_dep@hi\endcsname{+11.6}
\expandafter\def\csname orq@GLM46VF@NoDep@n\endcsname{302}
\expandafter\def\csname orq@GLM46VF@NoDep@a\endcsname{32.5}
\expandafter\def\csname orq@GLM46VF@NoDep@b\endcsname{33.8}
\expandafter\def\csname orq@GLM46VF@NoDep@d\endcsname{+1.3}
\expandafter\def\csname orq@GLM46VF@NoDep@lo\endcsname{$-$1.7}
\expandafter\def\csname orq@GLM46VF@NoDep@hi\endcsname{+4.3}
\expandafter\def\csname pd@Q36@roleNA\endcsname{$-$11.0}
\expandafter\def\csname pd@Q36@roleText\endcsname{$-$28.7}
\expandafter\def\csname pd@Q36@grpText\endcsname{+0.4}
\expandafter\def\csname pd@GLM46VF@roleNA\endcsname{$-$3.2}
\expandafter\def\csname pd@GLM46VF@roleText\endcsname{$-$5.3}
\expandafter\def\csname pd@GLM46VF@grpText\endcsname{$-$1.1}
\def\ORNObj{4,014}
\expandafter\def\csname or@Q36@grpDrawn@exact\endcsname{10.5}
\expandafter\def\csname pd@Q36@grpDrawn\endcsname{+1.9}
\expandafter\def\csname or@GLM46VF@grpDrawn@exact\endcsname{7.8}
\expandafter\def\csname pd@GLM46VF@grpDrawn\endcsname{$-$3.8}
\def\ORGrpRule{10.5}\def\ORGrpRulePair{62.0}
\def\ORNRoleFrames{2,856}
\providecommand{\PD}[2]{\ifcsname pd@#1@#2\endcsname\csname pd@#1@#2\endcsname\else\TODO{}\fi}
\def\TabEventRecRows{%
VideoMind & 29.9 / 29.9 & 18.0 / 0.0 & 13.3 / 0.0 & 1.0 & 10 \\
Open-o3 Video & 21.8 / 21.8 & 16.7 / 0.0 & 9.9 / 0.0 & 0.9 & 11 \\
VisionCoach & 26.2 / 26.2 & 17.9 / 0.0 & 9.7 / 0.0 & 0.9 & 10 \\
VideoChat-R1.5 & 28.6 / 28.6 & 20.6 / 2.0 & 17.9 / 1.4 & 1.3 & 13 \\
LLaVA-ST (7B) & 17.7 / 17.7 & 13.5 / 0.0 & 9.7 / 0.0 & 1.0 & 5 \\
GroundVQA & 21.8 / 21.8 & 15.9 / 0.0 & 11.2 / 0.0 & 1.0 & 8 \\
EgoThinker & 8.5 / 8.5 & 6.7 / 0.3 & 5.1 / 0.0 & 1.2 & 4 \\
Causal-VLM & 31.9 / 31.9 & 24.9 / 8.3 & 21.1 / 2.0 & 2.5 & 17 \\
Cosmos-Reason2 (8B) & 31.9 / 31.9 & 22.3 / 4.5 & 16.6 / 0.7 & 1.4 & 14 \\
\midrule
Qwen3.6-27B & 67.4 / 67.4 & 55.4 / 32.1 & 48.4 / 13.5 & 3.8 & 9 \\
Qwen3.8-27B & 58.3 / 58.3 & 47.5 / 20.8 & 36.0 / 6.8 & 3.0 & 10 \\
GLM-4.6V-Flash (9B) & 64.9 / 64.9 & 52.5 / 26.0 & 39.8 / 8.1 & 3.5 & 8 \\
Gemma 4 (31B) & 58.8 / 58.8 & 50.5 / 24.4 & 43.2 / 9.5 & 3.9 & 6 \\
GPT-6.1 Sol & 65.5 / 65.5 & 56.9 / 32.6 & 49.9 / 18.2 & 3.1 & 4 \\
Gemini 3.8 Flash & 56.6 / 56.6 & 48.3 / 22.1 & 40.4 / 8.1 & 2.8 & 4 \\
Gemini Robotics-ER 2 & 39.4 / 39.4 & 33.6 / 11.7 & 30.5 / 1.4 & 2.3 & 4 \\
Gemini 3.1 Pro & 51.6 / 51.6 & 41.2 / 17.1 & 33.5 / 3.4 & 2.2 & 6 \\
\midrule
Joint SFT & 43.0 / 43.0 & 33.2 / 6.7 & 27.4 / 2.0 & 2.8 & 3 \\
\method{} (2B) & 54.0 / 54.0 & 41.8 / 15.3 & 32.2 / 0.0 & 1.6 & 6 \\
}
\def\ERNOne{852}\def\ERNTwo{639}\def\ERNThree{148}\def\ERGTDur{4.6}
\providecommand{\ER}[2]{\ifcsname er@#1@#2\endcsname\csname er@#1@#2\endcsname\else\TODO{}\fi}
\expandafter\def\csname er@VideoMind@recall_1\endcsname{29.9}
\expandafter\def\csname er@VideoMind@all_recall_1\endcsname{29.9}
\expandafter\def\csname er@VideoMind@n_pred_1\endcsname{1.0}
\expandafter\def\csname er@VideoMind@recall_2\endcsname{18.0}
\expandafter\def\csname er@VideoMind@all_recall_2\endcsname{0.0}
\expandafter\def\csname er@VideoMind@n_pred_2\endcsname{1.0}
\expandafter\def\csname er@VideoMind@recall_3p\endcsname{13.3}
\expandafter\def\csname er@VideoMind@all_recall_3p\endcsname{0.0}
\expandafter\def\csname er@VideoMind@n_pred_3p\endcsname{1.0}
\expandafter\def\csname er@VideoMind@recall_all\endcsname{21.0}
\expandafter\def\csname er@VideoMind@all_recall_all\endcsname{15.6}
\expandafter\def\csname er@VideoMind@n_pred_all\endcsname{1.0}
\expandafter\def\csname er@VideoMind@pred_dur\endcsname{10.3}
\expandafter\def\csname er@OpenO3Video@recall_1\endcsname{21.8}
\expandafter\def\csname er@OpenO3Video@all_recall_1\endcsname{21.8}
\expandafter\def\csname er@OpenO3Video@n_pred_1\endcsname{0.9}
\expandafter\def\csname er@OpenO3Video@recall_2\endcsname{16.7}
\expandafter\def\csname er@OpenO3Video@all_recall_2\endcsname{0.0}
\expandafter\def\csname er@OpenO3Video@n_pred_2\endcsname{0.9}
\expandafter\def\csname er@OpenO3Video@recall_3p\endcsname{9.9}
\expandafter\def\csname er@OpenO3Video@all_recall_3p\endcsname{0.0}
\expandafter\def\csname er@OpenO3Video@n_pred_3p\endcsname{0.9}
\expandafter\def\csname er@OpenO3Video@recall_all\endcsname{17.1}
\expandafter\def\csname er@OpenO3Video@all_recall_all\endcsname{11.3}
\expandafter\def\csname er@OpenO3Video@n_pred_all\endcsname{0.9}
\expandafter\def\csname er@OpenO3Video@pred_dur\endcsname{10.8}
\expandafter\def\csname er@VisionCoach@recall_1\endcsname{26.2}
\expandafter\def\csname er@VisionCoach@all_recall_1\endcsname{26.2}
\expandafter\def\csname er@VisionCoach@n_pred_1\endcsname{0.9}
\expandafter\def\csname er@VisionCoach@recall_2\endcsname{17.9}
\expandafter\def\csname er@VisionCoach@all_recall_2\endcsname{0.0}
\expandafter\def\csname er@VisionCoach@n_pred_2\endcsname{0.9}
\expandafter\def\csname er@VisionCoach@recall_3p\endcsname{9.7}
\expandafter\def\csname er@VisionCoach@all_recall_3p\endcsname{0.0}
\expandafter\def\csname er@VisionCoach@n_pred_3p\endcsname{0.9}
\expandafter\def\csname er@VisionCoach@recall_all\endcsname{19.1}
\expandafter\def\csname er@VisionCoach@all_recall_all\endcsname{13.6}
\expandafter\def\csname er@VisionCoach@n_pred_all\endcsname{0.9}
\expandafter\def\csname er@VisionCoach@pred_dur\endcsname{10.4}
\expandafter\def\csname er@VideoChatR15@recall_1\endcsname{28.6}
\expandafter\def\csname er@VideoChatR15@all_recall_1\endcsname{28.6}
\expandafter\def\csname er@VideoChatR15@n_pred_1\endcsname{1.2}
\expandafter\def\csname er@VideoChatR15@recall_2\endcsname{20.6}
\expandafter\def\csname er@VideoChatR15@all_recall_2\endcsname{2.0}
\expandafter\def\csname er@VideoChatR15@n_pred_2\endcsname{1.3}
\expandafter\def\csname er@VideoChatR15@recall_3p\endcsname{17.9}
\expandafter\def\csname er@VideoChatR15@all_recall_3p\endcsname{1.4}
\expandafter\def\csname er@VideoChatR15@n_pred_3p\endcsname{1.7}
\expandafter\def\csname er@VideoChatR15@recall_all\endcsname{22.7}
\expandafter\def\csname er@VideoChatR15@all_recall_all\endcsname{15.8}
\expandafter\def\csname er@VideoChatR15@n_pred_all\endcsname{1.3}
\expandafter\def\csname er@VideoChatR15@pred_dur\endcsname{12.7}
\expandafter\def\csname er@LLaVAST@recall_1\endcsname{17.7}
\expandafter\def\csname er@LLaVAST@all_recall_1\endcsname{17.7}
\expandafter\def\csname er@LLaVAST@n_pred_1\endcsname{1.0}
\expandafter\def\csname er@LLaVAST@recall_2\endcsname{13.5}
\expandafter\def\csname er@LLaVAST@all_recall_2\endcsname{0.0}
\expandafter\def\csname er@LLaVAST@n_pred_2\endcsname{1.0}
\expandafter\def\csname er@LLaVAST@recall_3p\endcsname{9.7}
\expandafter\def\csname er@LLaVAST@all_recall_3p\endcsname{0.0}
\expandafter\def\csname er@LLaVAST@n_pred_3p\endcsname{1.0}
\expandafter\def\csname er@LLaVAST@recall_all\endcsname{14.2}
\expandafter\def\csname er@LLaVAST@all_recall_all\endcsname{9.2}
\expandafter\def\csname er@LLaVAST@n_pred_all\endcsname{1.0}
\expandafter\def\csname er@LLaVAST@pred_dur\endcsname{4.8}
\expandafter\def\csname er@GroundVQA@recall_1\endcsname{21.8}
\expandafter\def\csname er@GroundVQA@all_recall_1\endcsname{21.8}
\expandafter\def\csname er@GroundVQA@n_pred_1\endcsname{1.0}
\expandafter\def\csname er@GroundVQA@recall_2\endcsname{15.9}
\expandafter\def\csname er@GroundVQA@all_recall_2\endcsname{0.0}
\expandafter\def\csname er@GroundVQA@n_pred_2\endcsname{1.0}
\expandafter\def\csname er@GroundVQA@recall_3p\endcsname{11.2}
\expandafter\def\csname er@GroundVQA@all_recall_3p\endcsname{0.0}
\expandafter\def\csname er@GroundVQA@n_pred_3p\endcsname{1.0}
\expandafter\def\csname er@GroundVQA@recall_all\endcsname{17.0}
\expandafter\def\csname er@GroundVQA@all_recall_all\endcsname{11.3}
\expandafter\def\csname er@GroundVQA@n_pred_all\endcsname{1.0}
\expandafter\def\csname er@GroundVQA@pred_dur\endcsname{7.7}
\expandafter\def\csname er@EgoThinker@recall_1\endcsname{8.5}
\expandafter\def\csname er@EgoThinker@all_recall_1\endcsname{8.5}
\expandafter\def\csname er@EgoThinker@n_pred_1\endcsname{1.2}
\expandafter\def\csname er@EgoThinker@recall_2\endcsname{6.7}
\expandafter\def\csname er@EgoThinker@all_recall_2\endcsname{0.3}
\expandafter\def\csname er@EgoThinker@n_pred_2\endcsname{1.1}
\expandafter\def\csname er@EgoThinker@recall_3p\endcsname{5.1}
\expandafter\def\csname er@EgoThinker@all_recall_3p\endcsname{0.0}
\expandafter\def\csname er@EgoThinker@n_pred_3p\endcsname{1.0}
\expandafter\def\csname er@EgoThinker@recall_all\endcsname{6.9}
\expandafter\def\csname er@EgoThinker@all_recall_all\endcsname{4.5}
\expandafter\def\csname er@EgoThinker@n_pred_all\endcsname{1.2}
\expandafter\def\csname er@EgoThinker@pred_dur\endcsname{4.1}
\expandafter\def\csname er@CausalVLM@recall_1\endcsname{31.9}
\expandafter\def\csname er@CausalVLM@all_recall_1\endcsname{31.9}
\expandafter\def\csname er@CausalVLM@n_pred_1\endcsname{2.4}
\expandafter\def\csname er@CausalVLM@recall_2\endcsname{24.9}
\expandafter\def\csname er@CausalVLM@all_recall_2\endcsname{8.3}
\expandafter\def\csname er@CausalVLM@n_pred_2\endcsname{2.5}
\expandafter\def\csname er@CausalVLM@recall_3p\endcsname{21.1}
\expandafter\def\csname er@CausalVLM@all_recall_3p\endcsname{2.0}
\expandafter\def\csname er@CausalVLM@n_pred_3p\endcsname{2.9}
\expandafter\def\csname er@CausalVLM@recall_all\endcsname{26.5}
\expandafter\def\csname er@CausalVLM@all_recall_all\endcsname{20.0}
\expandafter\def\csname er@CausalVLM@n_pred_all\endcsname{2.5}
\expandafter\def\csname er@CausalVLM@pred_dur\endcsname{16.8}
\expandafter\def\csname er@CosmosR2@recall_1\endcsname{31.9}
\expandafter\def\csname er@CosmosR2@all_recall_1\endcsname{31.9}
\expandafter\def\csname er@CosmosR2@n_pred_1\endcsname{1.3}
\expandafter\def\csname er@CosmosR2@recall_2\endcsname{22.3}
\expandafter\def\csname er@CosmosR2@all_recall_2\endcsname{4.5}
\expandafter\def\csname er@CosmosR2@n_pred_2\endcsname{1.5}
\expandafter\def\csname er@CosmosR2@recall_3p\endcsname{16.6}
\expandafter\def\csname er@CosmosR2@all_recall_3p\endcsname{0.7}
\expandafter\def\csname er@CosmosR2@n_pred_3p\endcsname{1.7}
\expandafter\def\csname er@CosmosR2@recall_all\endcsname{24.4}
\expandafter\def\csname er@CosmosR2@all_recall_all\endcsname{18.4}
\expandafter\def\csname er@CosmosR2@n_pred_all\endcsname{1.4}
\expandafter\def\csname er@CosmosR2@pred_dur\endcsname{14.4}
\expandafter\def\csname er@Q36@recall_1\endcsname{67.4}
\expandafter\def\csname er@Q36@all_recall_1\endcsname{67.4}
\expandafter\def\csname er@Q36@n_pred_1\endcsname{3.8}
\expandafter\def\csname er@Q36@recall_2\endcsname{55.4}
\expandafter\def\csname er@Q36@all_recall_2\endcsname{32.1}
\expandafter\def\csname er@Q36@n_pred_2\endcsname{3.8}
\expandafter\def\csname er@Q36@recall_3p\endcsname{48.4}
\expandafter\def\csname er@Q36@all_recall_3p\endcsname{13.5}
\expandafter\def\csname er@Q36@n_pred_3p\endcsname{4.2}
\expandafter\def\csname er@Q36@recall_all\endcsname{58.0}
\expandafter\def\csname er@Q36@all_recall_all\endcsname{48.7}
\expandafter\def\csname er@Q36@n_pred_all\endcsname{3.8}
\expandafter\def\csname er@Q36@pred_dur\endcsname{8.5}
\expandafter\def\csname er@Q38@recall_1\endcsname{58.3}
\expandafter\def\csname er@Q38@all_recall_1\endcsname{58.3}
\expandafter\def\csname er@Q38@n_pred_1\endcsname{2.9}
\expandafter\def\csname er@Q38@recall_2\endcsname{47.5}
\expandafter\def\csname er@Q38@all_recall_2\endcsname{20.8}
\expandafter\def\csname er@Q38@n_pred_2\endcsname{3.1}
\expandafter\def\csname er@Q38@recall_3p\endcsname{36.0}
\expandafter\def\csname er@Q38@all_recall_3p\endcsname{6.8}
\expandafter\def\csname er@Q38@n_pred_3p\endcsname{3.1}
\expandafter\def\csname er@Q38@recall_all\endcsname{48.9}
\expandafter\def\csname er@Q38@all_recall_all\endcsname{39.0}
\expandafter\def\csname er@Q38@n_pred_all\endcsname{3.0}
\expandafter\def\csname er@Q38@pred_dur\endcsname{10.0}
\expandafter\def\csname er@GLM46VF@recall_1\endcsname{64.9}
\expandafter\def\csname er@GLM46VF@all_recall_1\endcsname{64.9}
\expandafter\def\csname er@GLM46VF@n_pred_1\endcsname{3.3}
\expandafter\def\csname er@GLM46VF@recall_2\endcsname{52.5}
\expandafter\def\csname er@GLM46VF@all_recall_2\endcsname{26.0}
\expandafter\def\csname er@GLM46VF@n_pred_2\endcsname{3.5}
\expandafter\def\csname er@GLM46VF@recall_3p\endcsname{39.8}
\expandafter\def\csname er@GLM46VF@all_recall_3p\endcsname{8.1}
\expandafter\def\csname er@GLM46VF@n_pred_3p\endcsname{4.0}
\expandafter\def\csname er@GLM46VF@recall_all\endcsname{54.2}
\expandafter\def\csname er@GLM46VF@all_recall_all\endcsname{44.6}
\expandafter\def\csname er@GLM46VF@n_pred_all\endcsname{3.5}
\expandafter\def\csname er@GLM46VF@pred_dur\endcsname{7.7}
\expandafter\def\csname er@Gemma4@recall_1\endcsname{58.8}
\expandafter\def\csname er@Gemma4@all_recall_1\endcsname{58.8}
\expandafter\def\csname er@Gemma4@n_pred_1\endcsname{3.6}
\expandafter\def\csname er@Gemma4@recall_2\endcsname{50.5}
\expandafter\def\csname er@Gemma4@all_recall_2\endcsname{24.4}
\expandafter\def\csname er@Gemma4@n_pred_2\endcsname{4.3}
\expandafter\def\csname er@Gemma4@recall_3p\endcsname{43.2}
\expandafter\def\csname er@Gemma4@all_recall_3p\endcsname{9.5}
\expandafter\def\csname er@Gemma4@n_pred_3p\endcsname{3.9}
\expandafter\def\csname er@Gemma4@recall_all\endcsname{51.9}
\expandafter\def\csname er@Gemma4@all_recall_all\endcsname{40.9}
\expandafter\def\csname er@Gemma4@n_pred_all\endcsname{3.9}
\expandafter\def\csname er@Gemma4@pred_dur\endcsname{6.2}
\expandafter\def\csname er@GPT@recall_1\endcsname{65.5}
\expandafter\def\csname er@GPT@all_recall_1\endcsname{65.5}
\expandafter\def\csname er@GPT@n_pred_1\endcsname{2.9}
\expandafter\def\csname er@GPT@recall_2\endcsname{56.9}
\expandafter\def\csname er@GPT@all_recall_2\endcsname{32.6}
\expandafter\def\csname er@GPT@n_pred_2\endcsname{3.2}
\expandafter\def\csname er@GPT@recall_3p\endcsname{49.9}
\expandafter\def\csname er@GPT@all_recall_3p\endcsname{18.2}
\expandafter\def\csname er@GPT@n_pred_3p\endcsname{3.4}
\expandafter\def\csname er@GPT@recall_all\endcsname{58.4}
\expandafter\def\csname er@GPT@all_recall_all\endcsname{48.4}
\expandafter\def\csname er@GPT@n_pred_all\endcsname{3.1}
\expandafter\def\csname er@GPT@pred_dur\endcsname{3.7}
\expandafter\def\csname er@G38F@recall_1\endcsname{56.6}
\expandafter\def\csname er@G38F@all_recall_1\endcsname{56.6}
\expandafter\def\csname er@G38F@n_pred_1\endcsname{2.6}
\expandafter\def\csname er@G38F@recall_2\endcsname{48.3}
\expandafter\def\csname er@G38F@all_recall_2\endcsname{22.1}
\expandafter\def\csname er@G38F@n_pred_2\endcsname{2.9}
\expandafter\def\csname er@G38F@recall_3p\endcsname{40.4}
\expandafter\def\csname er@G38F@all_recall_3p\endcsname{8.1}
\expandafter\def\csname er@G38F@n_pred_3p\endcsname{3.1}
\expandafter\def\csname er@G38F@recall_all\endcsname{49.6}
\expandafter\def\csname er@G38F@all_recall_all\endcsname{38.7}
\expandafter\def\csname er@G38F@n_pred_all\endcsname{2.8}
\expandafter\def\csname er@G38F@pred_dur\endcsname{4.3}
\expandafter\def\csname er@GER2@recall_1\endcsname{39.4}
\expandafter\def\csname er@GER2@all_recall_1\endcsname{39.4}
\expandafter\def\csname er@GER2@n_pred_1\endcsname{2.2}
\expandafter\def\csname er@GER2@recall_2\endcsname{33.6}
\expandafter\def\csname er@GER2@all_recall_2\endcsname{11.7}
\expandafter\def\csname er@GER2@n_pred_2\endcsname{2.4}
\expandafter\def\csname er@GER2@recall_3p\endcsname{30.5}
\expandafter\def\csname er@GER2@all_recall_3p\endcsname{1.4}
\expandafter\def\csname er@GER2@n_pred_3p\endcsname{2.6}
\expandafter\def\csname er@GER2@recall_all\endcsname{35.0}
\expandafter\def\csname er@GER2@all_recall_all\endcsname{25.2}
\expandafter\def\csname er@GER2@n_pred_all\endcsname{2.3}
\expandafter\def\csname er@GER2@pred_dur\endcsname{3.9}
\expandafter\def\csname er@G31P@recall_1\endcsname{51.6}
\expandafter\def\csname er@G31P@all_recall_1\endcsname{51.6}
\expandafter\def\csname er@G31P@n_pred_1\endcsname{2.1}
\expandafter\def\csname er@G31P@recall_2\endcsname{41.2}
\expandafter\def\csname er@G31P@all_recall_2\endcsname{17.1}
\expandafter\def\csname er@G31P@n_pred_2\endcsname{2.4}
\expandafter\def\csname er@G31P@recall_3p\endcsname{33.5}
\expandafter\def\csname er@G31P@all_recall_3p\endcsname{3.4}
\expandafter\def\csname er@G31P@n_pred_3p\endcsname{2.5}
\expandafter\def\csname er@G31P@recall_all\endcsname{43.2}
\expandafter\def\csname er@G31P@all_recall_all\endcsname{33.8}
\expandafter\def\csname er@G31P@n_pred_all\endcsname{2.2}
\expandafter\def\csname er@G31P@pred_dur\endcsname{6.1}
\expandafter\def\csname er@M1F@recall_1\endcsname{43.0}
\expandafter\def\csname er@M1F@all_recall_1\endcsname{43.0}
\expandafter\def\csname er@M1F@n_pred_1\endcsname{2.2}
\expandafter\def\csname er@M1F@recall_2\endcsname{33.2}
\expandafter\def\csname er@M1F@all_recall_2\endcsname{6.7}
\expandafter\def\csname er@M1F@n_pred_2\endcsname{3.4}
\expandafter\def\csname er@M1F@recall_3p\endcsname{27.4}
\expandafter\def\csname er@M1F@all_recall_3p\endcsname{2.0}
\expandafter\def\csname er@M1F@n_pred_3p\endcsname{3.7}
\expandafter\def\csname er@M1F@recall_all\endcsname{35.3}
\expandafter\def\csname er@M1F@all_recall_all\endcsname{25.1}
\expandafter\def\csname er@M1F@n_pred_all\endcsname{2.8}
\expandafter\def\csname er@M1F@pred_dur\endcsname{3.0}
\expandafter\def\csname er@A@recall_1\endcsname{54.0}
\expandafter\def\csname er@A@all_recall_1\endcsname{54.0}
\expandafter\def\csname er@A@n_pred_1\endcsname{1.5}
\expandafter\def\csname er@A@recall_2\endcsname{41.8}
\expandafter\def\csname er@A@all_recall_2\endcsname{15.3}
\expandafter\def\csname er@A@n_pred_2\endcsname{1.6}
\expandafter\def\csname er@A@recall_3p\endcsname{32.2}
\expandafter\def\csname er@A@all_recall_3p\endcsname{0.0}
\expandafter\def\csname er@A@n_pred_3p\endcsname{1.7}
\expandafter\def\csname er@A@recall_all\endcsname{44.0}
\expandafter\def\csname er@A@all_recall_all\endcsname{34.0}
\expandafter\def\csname er@A@n_pred_all\endcsname{1.6}
\expandafter\def\csname er@A@pred_dur\endcsname{6.2}
\def\TabCorrectSplitRows{%
Qwen3.6-27B & 363 & 76.6 & 18.7 & 2.5 & 2.2 & 89 & 66.3 & 25.8 & 3.4 & 4.5 \\
Qwen3.8-27B & 343 & 72.6 & 19.8 & 4.7 & 2.9 & 91 & 58.2 & 29.7 & 8.8 & 3.3 \\
GLM-4.6V-Flash & 358 & 75.4 & 22.9 & 1.1 & 0.6 & 81 & 63.0 & 35.8 & 0.0 & 1.2 \\
GPT-6.1 Sol & 541 & 63.8 & 15.5 & 12.4 & 8.3 & 117 & 43.6 & 30.8 & 11.1 & 14.5 \\
Gemini 3.8 Flash & 430 & 67.7 & 18.8 & 6.7 & 6.7 & 111 & 53.2 & 28.8 & 8.1 & 9.9 \\
Joint SFT & 222 & 67.6 & 17.1 & 2.3 & 13.1 & 56 & 73.2 & 16.1 & 5.4 & 5.4 \\
\method{} (2B) & 201 & 54.2 & 25.4 & 8.0 & 12.4 & 56 & 44.6 & 25.0 & 17.9 & 12.5 \\
}
\def\ETNDep{302}

\begin{figure}[t]
\centering
\includegraphics[width=\linewidth]{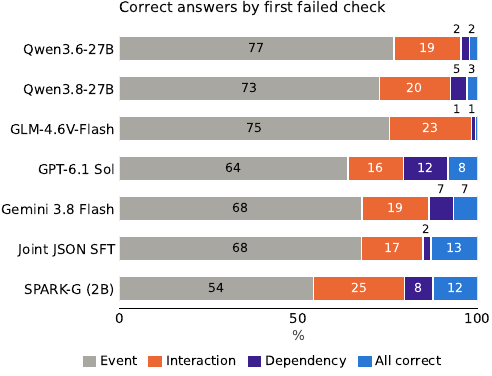}
\caption{\textbf{Correct answers by the first failed grounding check} (\%): key events (mTIoU $<0.3$), interactions (\Lthree{} $<0.3$), dependencies (Edge F1 $<0.3$), or all correct. Dependent pairs: Appx.~C.}
\label{fig:errors}
\end{figure}

\providecommand{\AN}[2]{\TODO{}}\providecommand{\OR}[3]{\TODO{}}\providecommand{\ORQ}[3]{\TODO{}}\providecommand{\ER}[2]{\TODO{}}
\providecommand{\ANNInter}{\TODO{}}\providecommand{\ANConcInter}{\TODO{}}\providecommand{\ANConcShare}{\TODO{}}\providecommand{\ANConcEvents}{\TODO{}}
\providecommand{\ANNDepPairs}{\TODO{}}\providecommand{\ANNoDepQ}{\TODO{}}
\providecommand{\ORNQ}{\TODO{}}\providecommand{\ORNRole}{\TODO{}}\providecommand{\ORNGrp}{\TODO{}}\providecommand{\ORGrpBase}{\TODO{}}\providecommand{\ORQNDep}{\TODO{}}
\Cref{tab:main} reports representative systems (all 26: Appx.~B).
Key events are recovered only partly (RQ1): the best mTIoU among systems not trained on our data is \TM{GPT}{mTIoU} (GPT-6.1 Sol).
Few participants are recovered with the correct location, role and association (RQ2), with an \Lthree{} of at most \TM{GPT}{L3}.
Dependencies are often wrong even between matched events (RQ3), with \depacc{} between \TM{Base9B}{DepAcc} (Qwen3.5-9B) and \TM{InternVL35}{DepAcc} (InternVL3.5-38B) for the open-source MLLMs.
Accuracy hides these failures (RQ4): GPT-6.1 Sol answers \TM{GPT}{Acc} correctly, but its GCR falls to \TM{GPT}{GCR_T}, \TM{GPT}{GCR_ST} and \TM{GPT}{GCR_STD} as events, interactions and dependencies are required, and for every system most correct answers already fail the event check (\cref{fig:errors}).
Pooled over the question--system predictions of \FApoolN{} systems, \FApoolRA{} of the \FApoolLoc{} correct answers whose events and participants are localized (\FApoolRAPct\%) fail on roles or association, errors that localization alone does not reveal (Appx.~C).
As \ANNoDepQ\% of the gold-labeled questions have no annotated dependency, which predicting none passes, GCR$_{STD}$ on the 302 questions with one is at most .056 (Appx.~C).
Pooled over \FApoolN{} systems, hypothetical questions have the lowest accuracy (.116) and GCR$_T$ (.031) and counterfactual questions the highest (.270, .084), a deficit that remains after controlling for key-event count and dependency presence (Appx.~C).

\subsection{Diagnosing failures in causal evidence grounding}
\label{sec:exp_findings}
\label{sec:exp_ablation}

We diagnose grounding failures through the model outputs and controlled experiments that give the human-verified evidence of earlier stages (\cref{tab:probes}).

\begin{table}[t]
\centering
\footnotesize
\setlength{\tabcolsep}{3pt}
\begin{tabular}{@{}llcccc@{}}
\toprule
\multicolumn{6}{@{}l}{\textit{Edge rate by dependency (\%)}} \\
System & Events & Cov. & TPR$\uparrow$ & FPR$\downarrow$ & BAcc \\
\midrule
Causal-VLM & Pred. & 7 & 94 & 94 & 50$^\circ$ \\
MECD+$^\dagger$ & GT & -- & 78 & 76 & 51$^\circ$ \\
\midrule
\multicolumn{6}{@{}l}{\textit{Edge rate on dependency-negative pairs (FPR, \%)}} \\
System & Events & \multicolumn{2}{l}{Pairs split by} & yes & no \\
\midrule
MECD+$^\dagger$ & GT & \multicolumn{2}{l}{Adjacent} & 85 & 66 \\
Qwen3.6-27B$^\ddagger$ & GT & \multicolumn{2}{l}{Shared tool/target} & 70 & 54 \\
GLM-4.6V-Flash$^\ddagger$ & GT & \multicolumn{2}{l}{Shared tool/target} & 94 & 73 \\
Cosmos-Reason2$^\ddagger$ & GT & \multicolumn{2}{l}{Shared tool/target} & 81 & 58 \\
\bottomrule
\end{tabular}
\caption{\textbf{Diagnosing dependency recovery.} Events: Pred.: the system's predicted events matched to the ground-truth events (tIoU $\geq$0.3); GT: the ground-truth events are given. Cov.: share of the 1,180 ground-truth event pairs whose two events are matched, on which the rates are computed. TPR, FPR: rate of predicted edges on dependent (\textit{causes} or \textit{enables}) and on dependency-negative pairs. BAcc (\cref{sec:exp_metrics}): 50 is chance; $^\circ$: 95\% CI includes 50. The rates are computed on different pair subsets, so BAcc does not rank the systems. Bottom: FPR on dependency-negative pairs that are adjacent or not, or share a tool or target or not; raw rates, the regressions controlling for the other pair features are in Appx.~C. $^\dagger$Questions with $\geq$3 key events. $^\ddagger$Given the ground-truth events as captions, with the video.}
\label{tab:dep_diag}
\end{table}

\providecommand{\TODO}[1]{\textcolor{red}{TODO#1}}
\providecommand{\ST}[3]{\ifcsname st@#1@#2@#3\endcsname\csname st@#1@#2@#3\endcsname\else\textcolor{red}{?}\fi}
\def\STNConcMom{211}
\def\STNConcInt{437}
\def\STNInt{3,083}
\def\STRule{.382}
\def\STRuleExact{.194}
\expandafter\def\csname st@GLM46VF@E@EdgeF1\endcsname{.410}
\expandafter\def\csname st@GLM46VF@E@L1\endcsname{.213}
\expandafter\def\csname st@GLM46VF@E@L2\endcsname{.155}
\expandafter\def\csname st@GLM46VF@E@L3\endcsname{.149}
\expandafter\def\csname st@GLM46VF@E@MacroF1\endcsname{.336}
\expandafter\def\csname st@GLM46VF@E@acc\endcsname{25.4}
\expandafter\def\csname st@GLM46VF@E@assoc\endcsname{.195}
\expandafter\def\csname st@GLM46VF@E@dacc\endcsname{+3.6$^\ast$}
\expandafter\def\csname st@GLM46VF@E@daccabs\endcsname{3.6}
\expandafter\def\csname st@GLM46VF@Eshift@daccabs\endcsname{5.5}
\expandafter\def\csname st@Q36@E@daccabs\endcsname{2.7}
\expandafter\def\csname st@Q36@Eshift@daccabs\endcsname{8.1}
\expandafter\def\csname st@GLM46VF@E@depBAcc\endcsname{.581}
\expandafter\def\csname st@GLM46VF@E@depF1\endcsname{.458}
\expandafter\def\csname st@GLM46VF@E@evcov\endcsname{98}
\expandafter\def\csname st@GLM46VF@E@intF1\endcsname{.596}
\expandafter\def\csname st@GLM46VF@E@intmulti\endcsname{.303}
\expandafter\def\csname st@GLM46VF@E@role\endcsname{.594}
\expandafter\def\csname st@GLM46VF@E@roleF1\endcsname{.742}
\expandafter\def\csname st@GLM46VF@E@sig\endcsname{$^\ast$}
\expandafter\def\csname st@GLM46VF@EB@EdgeF1\endcsname{.319}
\expandafter\def\csname st@GLM46VF@EB@L1\endcsname{.699}
\expandafter\def\csname st@GLM46VF@EB@L2\endcsname{.593}
\expandafter\def\csname st@GLM46VF@EB@L3\endcsname{.580}
\expandafter\def\csname st@GLM46VF@EB@L3conc\endcsname{.254}
\expandafter\def\csname st@GLM46VF@EB@MacroF1\endcsname{.318}
\expandafter\def\csname st@GLM46VF@EB@acc\endcsname{24.3}
\expandafter\def\csname st@GLM46VF@EB@assoc\endcsname{.037}
\expandafter\def\csname st@GLM46VF@EB@dacc\endcsname{$-$1.1}
\expandafter\def\csname st@GLM46VF@EB@depBAcc\endcsname{.575}
\expandafter\def\csname st@GLM46VF@EB@depF1\endcsname{.448}
\expandafter\def\csname st@GLM46VF@EB@evcov\endcsname{94}
\expandafter\def\csname st@GLM46VF@EB@intF1\endcsname{.608}
\expandafter\def\csname st@GLM46VF@EB@intmulti\endcsname{.208}
\expandafter\def\csname st@GLM46VF@EB@role\endcsname{.722}
\expandafter\def\csname st@GLM46VF@EB@roleF1\endcsname{.860}
\expandafter\def\csname st@GLM46VF@EB@sig\endcsname{{}}
\expandafter\def\csname st@GLM46VF@EBR@EdgeF1\endcsname{.358}
\expandafter\def\csname st@GLM46VF@EBR@L1\endcsname{.736}
\expandafter\def\csname st@GLM46VF@EBR@L2\endcsname{.727}
\expandafter\def\csname st@GLM46VF@EBR@L3\endcsname{.709}
\expandafter\def\csname st@GLM46VF@EBR@L3conc\endcsname{.228}
\expandafter\def\csname st@GLM46VF@EBR@MacroF1\endcsname{.327}
\expandafter\def\csname st@GLM46VF@EBR@acc\endcsname{23.8}
\expandafter\def\csname st@GLM46VF@EBR@assoc\endcsname{.016}
\expandafter\def\csname st@GLM46VF@EBR@dacc\endcsname{$-$0.5}
\expandafter\def\csname st@GLM46VF@EBR@depBAcc\endcsname{.595}
\expandafter\def\csname st@GLM46VF@EBR@depF1\endcsname{.467}
\expandafter\def\csname st@GLM46VF@EBR@evcov\endcsname{96}
\expandafter\def\csname st@GLM46VF@EBR@intF1\endcsname{.872}
\expandafter\def\csname st@GLM46VF@EBR@intmulti\endcsname{.233}
\expandafter\def\csname st@GLM46VF@EBR@role\endcsname{.975}
\expandafter\def\csname st@GLM46VF@EBR@roleF1\endcsname{.987}
\expandafter\def\csname st@GLM46VF@EBR@sig\endcsname{{}}
\expandafter\def\csname st@GLM46VF@ED@EdgeF1\endcsname{.534}
\expandafter\def\csname st@GLM46VF@ED@L1\endcsname{.215}
\expandafter\def\csname st@GLM46VF@ED@L2\endcsname{.157}
\expandafter\def\csname st@GLM46VF@ED@L3\endcsname{.151}
\expandafter\def\csname st@GLM46VF@ED@MacroF1\endcsname{.548}
\expandafter\def\csname st@GLM46VF@ED@acc\endcsname{25.6}
\expandafter\def\csname st@GLM46VF@EI@EdgeF1\endcsname{.316}
\expandafter\def\csname st@GLM46VF@EI@L1\endcsname{.842}
\expandafter\def\csname st@GLM46VF@EI@L2\endcsname{.838}
\expandafter\def\csname st@GLM46VF@EI@L3\endcsname{.836}
\expandafter\def\csname st@GLM46VF@EI@L3conc\endcsname{.839}
\expandafter\def\csname st@GLM46VF@EI@MacroF1\endcsname{.346}
\expandafter\def\csname st@GLM46VF@EI@acc\endcsname{25.0}
\expandafter\def\csname st@GLM46VF@EI@assoc\endcsname{.776}
\expandafter\def\csname st@GLM46VF@EI@dacc\endcsname{+1.2}
\expandafter\def\csname st@GLM46VF@EI@depBAcc\endcsname{.569}
\expandafter\def\csname st@GLM46VF@EI@depF1\endcsname{.426}
\expandafter\def\csname st@GLM46VF@EI@evcov\endcsname{96}
\expandafter\def\csname st@GLM46VF@EI@intF1\endcsname{.974}
\expandafter\def\csname st@GLM46VF@EI@intmulti\endcsname{.885}
\expandafter\def\csname st@GLM46VF@EI@role\endcsname{.994}
\expandafter\def\csname st@GLM46VF@EI@roleF1\endcsname{.996}
\expandafter\def\csname st@GLM46VF@EI@sig\endcsname{{}}
\expandafter\def\csname st@GLM46VF@EID@EdgeF1\endcsname{.674}
\expandafter\def\csname st@GLM46VF@EID@L1\endcsname{.848}
\expandafter\def\csname st@GLM46VF@EID@L2\endcsname{.843}
\expandafter\def\csname st@GLM46VF@EID@L3\endcsname{.841}
\expandafter\def\csname st@GLM46VF@EID@MacroF1\endcsname{.935}
\expandafter\def\csname st@GLM46VF@EID@acc\endcsname{24.6}
\expandafter\def\csname st@GLM46VF@EID@dacc\endcsname{$-$0.4}
\expandafter\def\csname st@GLM46VF@EID@sig\endcsname{{}}
\expandafter\def\csname st@GLM46VF@Eshift@EdgeF1\endcsname{.006}
\expandafter\def\csname st@GLM46VF@Eshift@L1\endcsname{.027}
\expandafter\def\csname st@GLM46VF@Eshift@L2\endcsname{.020}
\expandafter\def\csname st@GLM46VF@Eshift@L3\endcsname{.018}
\expandafter\def\csname st@GLM46VF@Eshift@MacroF1\endcsname{.038}
\expandafter\def\csname st@GLM46VF@Eshift@acc\endcsname{20.0}
\expandafter\def\csname st@GLM46VF@Eshift@assoc\endcsname{.222}
\expandafter\def\csname st@GLM46VF@Eshift@dacc\endcsname{$-$5.5$^\ast$}
\expandafter\def\csname st@GLM46VF@Eshift@depBAcc\endcsname{.468}
\expandafter\def\csname st@GLM46VF@Eshift@depF1\endcsname{.444}
\expandafter\def\csname st@GLM46VF@Eshift@evcov\endcsname{9}
\expandafter\def\csname st@GLM46VF@Eshift@intF1\endcsname{.524}
\expandafter\def\csname st@GLM46VF@Eshift@intmulti\endcsname{.216}
\expandafter\def\csname st@GLM46VF@Eshift@role\endcsname{.634}
\expandafter\def\csname st@GLM46VF@Eshift@roleF1\endcsname{.755}
\expandafter\def\csname st@GLM46VF@Eshift@sig\endcsname{$^\ast$}
\expandafter\def\csname st@GLM46VF@TextEID@EdgeF1\endcsname{.945}
\expandafter\def\csname st@GLM46VF@TextEID@L1\endcsname{.976}
\expandafter\def\csname st@GLM46VF@TextEID@L2\endcsname{.972}
\expandafter\def\csname st@GLM46VF@TextEID@L3\endcsname{.971}
\expandafter\def\csname st@GLM46VF@TextEID@MacroF1\endcsname{.850}
\expandafter\def\csname st@GLM46VF@TextEID@acc\endcsname{10.1}
\expandafter\def\csname st@GLM46VF@TextEID@dacc\endcsname{$-$14.5$^\ast$}
\expandafter\def\csname st@GLM46VF@TextEID@sig\endcsname{$^\ast$}
\expandafter\def\csname st@GLM46VF@V@EdgeF1\endcsname{.030}
\expandafter\def\csname st@GLM46VF@V@L1\endcsname{.062}
\expandafter\def\csname st@GLM46VF@V@L2\endcsname{.046}
\expandafter\def\csname st@GLM46VF@V@L3\endcsname{.044}
\expandafter\def\csname st@GLM46VF@V@MacroF1\endcsname{.156}
\expandafter\def\csname st@GLM46VF@V@acc\endcsname{21.8}
\expandafter\def\csname st@GLM46VF@V@assoc\endcsname{.217}
\expandafter\def\csname st@GLM46VF@V@depBAcc\endcsname{.589}
\expandafter\def\csname st@GLM46VF@V@depF1\endcsname{.463}
\expandafter\def\csname st@GLM46VF@V@evcov\endcsname{54}
\expandafter\def\csname st@GLM46VF@V@intF1\endcsname{.600}
\expandafter\def\csname st@GLM46VF@V@intmulti\endcsname{.444}
\expandafter\def\csname st@GLM46VF@V@role\endcsname{.642}
\expandafter\def\csname st@GLM46VF@V@roleF1\endcsname{.763}
\expandafter\def\csname st@Q36@E@EdgeF1\endcsname{.345}
\expandafter\def\csname st@Q36@E@L1\endcsname{.199}
\expandafter\def\csname st@Q36@E@L2\endcsname{.153}
\expandafter\def\csname st@Q36@E@L3\endcsname{.150}
\expandafter\def\csname st@Q36@E@MacroF1\endcsname{.328}
\expandafter\def\csname st@Q36@E@acc\endcsname{24.8}
\expandafter\def\csname st@Q36@E@assoc\endcsname{.289}
\expandafter\def\csname st@Q36@E@dacc\endcsname{+2.7$^\ast$}
\expandafter\def\csname st@Q36@E@depBAcc\endcsname{.584}
\expandafter\def\csname st@Q36@E@depF1\endcsname{.462}
\expandafter\def\csname st@Q36@E@evcov\endcsname{94}
\expandafter\def\csname st@Q36@E@intF1\endcsname{.674}
\expandafter\def\csname st@Q36@E@intmulti\endcsname{.400}
\expandafter\def\csname st@Q36@E@role\endcsname{.716}
\expandafter\def\csname st@Q36@E@roleF1\endcsname{.789}
\expandafter\def\csname st@Q36@E@sig\endcsname{$^\ast$}
\expandafter\def\csname st@Q36@EB@EdgeF1\endcsname{.311}
\expandafter\def\csname st@Q36@EB@L1\endcsname{.726}
\expandafter\def\csname st@Q36@EB@L2\endcsname{.663}
\expandafter\def\csname st@Q36@EB@L3\endcsname{.653}
\expandafter\def\csname st@Q36@EB@L3conc\endcsname{.311}
\expandafter\def\csname st@Q36@EB@MacroF1\endcsname{.337}
\expandafter\def\csname st@Q36@EB@acc\endcsname{24.5}
\expandafter\def\csname st@Q36@EB@assoc\endcsname{.059}
\expandafter\def\csname st@Q36@EB@dacc\endcsname{$-$0.4}
\expandafter\def\csname st@Q36@EB@depBAcc\endcsname{.591}
\expandafter\def\csname st@Q36@EB@depF1\endcsname{.473}
\expandafter\def\csname st@Q36@EB@evcov\endcsname{93}
\expandafter\def\csname st@Q36@EB@intF1\endcsname{.732}
\expandafter\def\csname st@Q36@EB@intmulti\endcsname{.276}
\expandafter\def\csname st@Q36@EB@role\endcsname{.835}
\expandafter\def\csname st@Q36@EB@roleF1\endcsname{.916}
\expandafter\def\csname st@Q36@EB@sig\endcsname{{}}
\expandafter\def\csname st@Q36@EBR@EdgeF1\endcsname{.315}
\expandafter\def\csname st@Q36@EBR@L1\endcsname{.780}
\expandafter\def\csname st@Q36@EBR@L2\endcsname{.768}
\expandafter\def\csname st@Q36@EBR@L3\endcsname{.752}
\expandafter\def\csname st@Q36@EBR@L3conc\endcsname{.284}
\expandafter\def\csname st@Q36@EBR@MacroF1\endcsname{.342}
\expandafter\def\csname st@Q36@EBR@acc\endcsname{24.2}
\expandafter\def\csname st@Q36@EBR@assoc\endcsname{.041}
\expandafter\def\csname st@Q36@EBR@dacc\endcsname{$-$0.3}
\expandafter\def\csname st@Q36@EBR@depBAcc\endcsname{.596}
\expandafter\def\csname st@Q36@EBR@depF1\endcsname{.471}
\expandafter\def\csname st@Q36@EBR@evcov\endcsname{94}
\expandafter\def\csname st@Q36@EBR@intF1\endcsname{.878}
\expandafter\def\csname st@Q36@EBR@intmulti\endcsname{.276}
\expandafter\def\csname st@Q36@EBR@role\endcsname{.976}
\expandafter\def\csname st@Q36@EBR@roleF1\endcsname{.987}
\expandafter\def\csname st@Q36@EBR@sig\endcsname{{}}
\expandafter\def\csname st@Q36@ED@EdgeF1\endcsname{.428}
\expandafter\def\csname st@Q36@ED@L1\endcsname{.202}
\expandafter\def\csname st@Q36@ED@L2\endcsname{.156}
\expandafter\def\csname st@Q36@ED@L3\endcsname{.153}
\expandafter\def\csname st@Q36@ED@MacroF1\endcsname{.508}
\expandafter\def\csname st@Q36@ED@acc\endcsname{24.5}
\expandafter\def\csname st@Q36@EI@EdgeF1\endcsname{.289}
\expandafter\def\csname st@Q36@EI@L1\endcsname{.800}
\expandafter\def\csname st@Q36@EI@L2\endcsname{.793}
\expandafter\def\csname st@Q36@EI@L3\endcsname{.793}
\expandafter\def\csname st@Q36@EI@L3conc\endcsname{.897}
\expandafter\def\csname st@Q36@EI@MacroF1\endcsname{.350}
\expandafter\def\csname st@Q36@EI@acc\endcsname{24.9}
\expandafter\def\csname st@Q36@EI@assoc\endcsname{.833}
\expandafter\def\csname st@Q36@EI@dacc\endcsname{+0.7}
\expandafter\def\csname st@Q36@EI@depBAcc\endcsname{.596}
\expandafter\def\csname st@Q36@EI@depF1\endcsname{.476}
\expandafter\def\csname st@Q36@EI@evcov\endcsname{92}
\expandafter\def\csname st@Q36@EI@intF1\endcsname{.977}
\expandafter\def\csname st@Q36@EI@intmulti\endcsname{.939}
\expandafter\def\csname st@Q36@EI@role\endcsname{.989}
\expandafter\def\csname st@Q36@EI@roleF1\endcsname{.993}
\expandafter\def\csname st@Q36@EI@sig\endcsname{{}}
\expandafter\def\csname st@Q36@EID@EdgeF1\endcsname{.464}
\expandafter\def\csname st@Q36@EID@L1\endcsname{.811}
\expandafter\def\csname st@Q36@EID@L2\endcsname{.804}
\expandafter\def\csname st@Q36@EID@L3\endcsname{.804}
\expandafter\def\csname st@Q36@EID@MacroF1\endcsname{.849}
\expandafter\def\csname st@Q36@EID@acc\endcsname{24.4}
\expandafter\def\csname st@Q36@EID@dacc\endcsname{$-$0.5}
\expandafter\def\csname st@Q36@EID@sig\endcsname{{}}
\expandafter\def\csname st@Q36@Eshift@EdgeF1\endcsname{.009}
\expandafter\def\csname st@Q36@Eshift@L1\endcsname{.041}
\expandafter\def\csname st@Q36@Eshift@L2\endcsname{.032}
\expandafter\def\csname st@Q36@Eshift@L3\endcsname{.031}
\expandafter\def\csname st@Q36@Eshift@MacroF1\endcsname{.054}
\expandafter\def\csname st@Q36@Eshift@acc\endcsname{16.8}
\expandafter\def\csname st@Q36@Eshift@assoc\endcsname{.182}
\expandafter\def\csname st@Q36@Eshift@dacc\endcsname{$-$8.1$^\ast$}
\expandafter\def\csname st@Q36@Eshift@depBAcc\endcsname{.559}
\expandafter\def\csname st@Q36@Eshift@depF1\endcsname{.449}
\expandafter\def\csname st@Q36@Eshift@evcov\endcsname{21}
\expandafter\def\csname st@Q36@Eshift@intF1\endcsname{.634}
\expandafter\def\csname st@Q36@Eshift@intmulti\endcsname{.412}
\expandafter\def\csname st@Q36@Eshift@role\endcsname{.718}
\expandafter\def\csname st@Q36@Eshift@roleF1\endcsname{.790}
\expandafter\def\csname st@Q36@Eshift@sig\endcsname{$^\ast$}
\expandafter\def\csname st@Q36@TextEID@EdgeF1\endcsname{.739}
\expandafter\def\csname st@Q36@TextEID@L1\endcsname{.970}
\expandafter\def\csname st@Q36@TextEID@L2\endcsname{.962}
\expandafter\def\csname st@Q36@TextEID@L3\endcsname{.962}
\expandafter\def\csname st@Q36@TextEID@MacroF1\endcsname{.754}
\expandafter\def\csname st@Q36@TextEID@acc\endcsname{10.3}
\expandafter\def\csname st@Q36@TextEID@dacc\endcsname{$-$14.1$^\ast$}
\expandafter\def\csname st@Q36@TextEID@sig\endcsname{$^\ast$}
\expandafter\def\csname st@Q36@V@EdgeF1\endcsname{.040}
\expandafter\def\csname st@Q36@V@L1\endcsname{.075}
\expandafter\def\csname st@Q36@V@L2\endcsname{.061}
\expandafter\def\csname st@Q36@V@L3\endcsname{.059}
\expandafter\def\csname st@Q36@V@MacroF1\endcsname{.217}
\expandafter\def\csname st@Q36@V@acc\endcsname{22.1}
\expandafter\def\csname st@Q36@V@assoc\endcsname{.281}
\expandafter\def\csname st@Q36@V@depBAcc\endcsname{.612}
\expandafter\def\csname st@Q36@V@depF1\endcsname{.531}
\expandafter\def\csname st@Q36@V@evcov\endcsname{58}
\expandafter\def\csname st@Q36@V@intF1\endcsname{.703}
\expandafter\def\csname st@Q36@V@intmulti\endcsname{.400}
\expandafter\def\csname st@Q36@V@role\endcsname{.761}
\expandafter\def\csname st@Q36@V@roleF1\endcsname{.829}
\providecommand{\ST}[3]{\textcolor{red}{?}}\providecommand{\STRule}{\TODO{}}\providecommand{\STNConcMom}{\TODO{}}\providecommand{\STNConcInt}{\TODO{}}\providecommand{\STRuleExact}{\TODO{}}
\begin{table}[t]
\centering
\scriptsize
\setlength{\tabcolsep}{1.8pt}
\begin{tabular}{@{}lcccc@{}}
\toprule
Input & Acc & \Ltwo & \Lthree & Edge F1 \\
\midrule
$V$ & \ST{Q36}{V}{acc} / \ST{GLM46VF}{V}{acc} & \ST{Q36}{V}{L2} / \ST{GLM46VF}{V}{L2} & \ST{Q36}{V}{L3} / \ST{GLM46VF}{V}{L3} & \ST{Q36}{V}{EdgeF1} / \ST{GLM46VF}{V}{EdgeF1} \\
$V{+}\mathcal{E}$ & \ST{Q36}{E}{acc}\ST{Q36}{E}{sig} / \ST{GLM46VF}{E}{acc}\ST{GLM46VF}{E}{sig} & \ST{Q36}{E}{L2} / \ST{GLM46VF}{E}{L2} & \ST{Q36}{E}{L3} / \ST{GLM46VF}{E}{L3} & \ST{Q36}{E}{depF1} / \ST{GLM46VF}{E}{depF1} \\
$V{+}\mathcal{E}{+}B$ & \ST{Q36}{EB}{acc}\ST{Q36}{EB}{sig} / \ST{GLM46VF}{EB}{acc}\ST{GLM46VF}{EB}{sig} & \ST{Q36}{EB}{roleF1} / \ST{GLM46VF}{EB}{roleF1} & \ST{Q36}{EB}{intF1} / \ST{GLM46VF}{EB}{intF1} & \ST{Q36}{EB}{depF1} / \ST{GLM46VF}{EB}{depF1} \\
$V{+}\mathcal{E}{+}B{+}R$ & \ST{Q36}{EBR}{acc}\ST{Q36}{EBR}{sig} / \ST{GLM46VF}{EBR}{acc}\ST{GLM46VF}{EBR}{sig} & -- & \ST{Q36}{EBR}{intF1} / \ST{GLM46VF}{EBR}{intF1} & \ST{Q36}{EBR}{depF1} / \ST{GLM46VF}{EBR}{depF1} \\
$V{+}\mathcal{E}{+}\mathcal{I}$ & \ST{Q36}{EI}{acc}\ST{Q36}{EI}{sig} / \ST{GLM46VF}{EI}{acc}\ST{GLM46VF}{EI}{sig} & -- & -- & \ST{Q36}{EI}{depF1} / \ST{GLM46VF}{EI}{depF1} \\
$V{+}\mathcal{E}{+}\mathcal{I}{+}\mathcal{D}$ & \ST{Q36}{EID}{acc}\ST{Q36}{EID}{sig} / \ST{GLM46VF}{EID}{acc}\ST{GLM46VF}{EID}{sig} & -- & -- & -- \\
\cmidrule{1-5}
$V{+}\tilde{\mathcal{E}}$ & \ST{Q36}{Eshift}{acc}\ST{Q36}{Eshift}{sig} / \ST{GLM46VF}{Eshift}{acc}\ST{GLM46VF}{Eshift}{sig} & \ST{Q36}{Eshift}{L2} / \ST{GLM46VF}{Eshift}{L2} & \ST{Q36}{Eshift}{L3} / \ST{GLM46VF}{Eshift}{L3} & \ST{Q36}{Eshift}{EdgeF1} / \ST{GLM46VF}{Eshift}{EdgeF1} \\
$\mathcal{E}{+}\mathcal{I}{+}\mathcal{D}$ & \ST{Q36}{TextEID}{acc}\ST{Q36}{TextEID}{sig} / \ST{GLM46VF}{TextEID}{acc}\ST{GLM46VF}{TextEID}{sig} & -- & -- & -- \\
\bottomrule
\end{tabular}
\caption{\textbf{Human-verified evidence given step by step} (Qwen3.6-27B / GLM-4.6V-Flash), as text with the video $V$: key events $\mathcal{E}$, boxes $B$, roles $R$, interactions $\mathcal{I}$ (adding the association) and dependencies $\mathcal{D}$. Controls: misplaced events $\tilde{\mathcal{E}}$ and no video. Only what is not given is scored. The rows without given events and the \Ltwo{} and \Lthree{} of $V{+}\mathcal{E}$ are scored as in \cref{tab:main}, and the other cells only on the given elements a reply keeps, so their denominators differ (coverage: Appx.~D). --: given. $^\ast$95\% CI of the accuracy change against the row above (for the controls: $V{+}\mathcal{E}$, $V{+}\mathcal{E}{+}\mathcal{I}{+}\mathcal{D}$) excludes 0. Full table: Appx.~D.}
\label{tab:probes}
\end{table}

\paragraph{Finding 1: Models fail to recover complete multi-event evidence.}

\textit{(i) Recovery degrades with the number of key events.}
Qwen3.6-27B localizes \ER{Q36}{recall_1}\% of the key events of single-event questions but \ER{Q36}{recall_3p}\% for questions with $\geq$3 key events, recovering all of them for only \ER{Q36}{all_recall_3p}\%  (Appx.~C).

\textit{(ii) Single-interval outputs explain only part of the loss.}
Several video reasoning specialists return one interval per question, which can recover at most half of the key events of a two-event question.
Yet VideoMind, the best of them, finds only \ER{VideoMind}{recall_1}\% of the events of single-event questions (one interval suffices), against \ER{Q36}{recall_1}\% for Qwen3.6-27B.

\paragraph{Finding 2: Correct object localization does not imply functional role understanding.}

\textit{(i) Role errors persist after localization.}
Among the tools and targets they localize correctly, Qwen3.6-27B and GLM-4.6V-Flash misassign \TI{Q36}{role_err}\% and \TI{GLM46VF}{role_err}\% of the roles (Appx.~C), errors that a box-only evaluation would overlook.

\textit{(ii) Role errors persist when localization is given.}
Given the verified key events and participant boxes but not their roles ($V{+}\mathcal{E}{+}B$), the two models assign the correct role to \ST{Q36}{EB}{role} and \ST{GLM46VF}{EB}{role} of the tools and targets their replies keep, and most errors swap tool and target.

\paragraph{Finding 3: Correctly identified participants are not necessarily associated with the right interactions.}

\textit{(i) Association errors remain when boxes and roles are correct.}
Pooled over eleven systems, 76 of the 286 interactions reported at moments with multiple ground-truth interactions and with every participant localized in the correct role (27\%) associate the participants wrongly, mostly by merging participants of different manipulations or omitting a participant shared with another interaction (Appx.~C).

\textit{(ii) Association errors persist given verified participants and roles.}
In $V{+}\mathcal{E}{+}B{+}R$ (\cref{tab:probes}), the input lists the verified participants with their roles by annotated moment, so only the association within a moment is left to the model.
\Lthree{} is \ST{Q36}{EBR}{intF1} and \ST{GLM46VF}{EBR}{intF1} over all moments, most of which hold one interaction, but \ST{Q36}{EBR}{intmulti} and \ST{GLM46VF}{EBR}{intmulti} at the \STNConcMom{} moments with multiple interactions, below a rule joining each object to the hand with the largest box overlap (\STRule).

\paragraph{Finding 4: Coarse events and correlation biases hinder dependency recovery.}

\textit{(i) Missing and coarse events limit dependency recovery.}
Most dependent pairs are lost because an event of the pair is not matched (\AN{Q36}{dep_miss}\% for Qwen3.6-27B, Appx.~C).
Causal-VLM predicts events of 17.5\,s on average, against 4.5\,s for the key events of dependent pairs, and places both events of 53\% of these pairs in one interval.
Given the verified key events, the Edge F1 of Qwen3.6-27B and GLM-4.6V-Flash on the event pairs their replies keep is \ST{Q36}{E}{depF1} and \ST{GLM46VF}{E}{depF1} (\FDqVEdgeKept{} and \FDgVEdgeKept{} on the fewer pairs they localize without them), and the boxes, roles and interactions do not raise it consistently (\cref{tab:probes}).

\textit{(ii) Dependency predictions do not separate dependent from dependency-negative pairs.}
On the 7\% of pairs whose two events it localizes, Causal-VLM predicts edges for 94\% of the dependent and 94\% of the dependency-negative pairs, and MECD+, given the ground-truth events, reaches a BAcc of 51\% (\cref{tab:dep_diag}).
MECD+ predicts edges for 85\% of the adjacent and 66\% of the non-adjacent dependency-negative pairs, an association that stays significant when annotated dependency and other features are controlled.

\textit{(iii) Shared tools or targets are associated with false-positive dependencies.}
Given the ground-truth events as captions and the video, GLM-4.6V-Flash, Qwen3.6-27B and Cosmos-Reason2 predict more edges for dependency-negative pairs that share a tool or target (\cref{tab:dep_diag}), significantly so for GLM-4.6V-Flash and Cosmos-Reason2 after controlling for adjacency, gap and caption similarity (Appx.~C).
These diagnostics show which pair features accompany predicted edges, not how models reach them.

\subsection{Effect of verified evidence on answering}
\label{sec:exp_utilization}
Giving the verified evidence as text alongside the video (\cref{tab:probes}, without captions or object names), verified key events significantly raise the accuracy by \ST{Q36}{E}{daccabs} and \ST{GLM46VF}{E}{daccabs} points, whereas intervals of the same number and length at other times lower it by \ST{Q36}{Eshift}{daccabs} and \ST{GLM46VF}{Eshift}{daccabs} points against the verified ones, highlighting the importance of accurate temporal grounding.
We observe no significant accuracy gain from adding interaction or dependency annotations in the tested textual representation (subsets: Appx.~D).
Whether models can better exploit these annotations through other representations or reasoning strategies remains open.

\subsection{Results of \method}
\label{sec:exp_learning}
Against the joint SFT baseline, which generates the evidence and answer in a single output with the same backbone and the same \bench training split, \method raises the key-event F1@0.3 from \TM{M1F}{F1_30} to \TM{A}{F1_30}, \Lthree{} from \TM{M1F}{L3} to \TM{A}{L3} and Edge F1 from \TM{M1F}{EdgeF1} to \TM{A}{EdgeF1} (\cref{tab:main}). Its key-event F1@0.3 and Edge F1 are also the highest among all evaluated models.
These gains do not translate into higher answer accuracy (\TM{A}{Acc} against \TM{M1F}{Acc}) or GCR$_{STD}$.

\section{Conclusion}
\label{sec:conclusion}

We introduced \bench, which evaluates causal answers on egocentric videos together with the key events, role-aware hand--object interactions and typed event dependencies that support them.
Current MLLMs and causal video reasoning models recover multi-event evidence incompletely, confuse functional roles and participant associations, and link adjacent events or events sharing a tool or target, so most of their correct answers are not supported by correct evidence.
\method, which predicts this evidence explicitly before answering, improves its recovery, and \bench offers a testbed for models whose causal answers rest on correct grounding.

\paragraph{Limitations.}
Functional state changes are not fully covered.
The few \textit{causes} pairs (\EvCauses{}, against \EvEnables{} \textit{enables} pairs) limit the evaluation of dependency types.
Predicting the annotated evidence correctly does not show that a model used it to answer and the controlled experiments test one representation.
\method propagates event localization errors, and its separate grounding stage adds computation over joint generation (Appx.~E).

\section*{LLM Usage Statement}

\paragraph{Use in the research.}
Generative models are part of the data pipeline and the evaluation protocol,
as described in \cref{sec:bench_pipeline,sec:bench_integrity,sec:exp_protocol}.
\emph{Annotation (\cref{sec:bench_pipeline}).} GPT-5.4-mini (evaluation videos) and
Qwen3.6-27B (training videos) name the tracked objects, assign their roles
and caption event segments; GPT-5.4-mini (evaluation videos) and Qwen3.6-27B (training videos) select the
key events given the question and the answer; Qwen3.6-27B splits key-event
captions into interactions; and GPT-5.4-mini (evaluation videos) and Qwen3.6-27B (training videos) label
the dependency of each key-event pair.
Labels of the training split are used as generated. The key events, captions, role tracks, boxes and dependencies of the
\EvGold{} gold-labeled questions were checked and corrected by human annotators (stage~6), and their
interactions are re-derived from them by Qwen3.6-27B.
\emph{Evidence integrity (\cref{sec:bench_integrity}).} GPT-5.4-mini answers the
questions from the text alone (C1), and InternVL3.5-38B answers them
with the key events blacked out (C2) or with only the key events visible (C4), and GPT-OSS-120B judges these answers.
An independent GPT-5.4-mini pass checks the capability
labels.
\emph{Evaluation (\cref{sec:exp_protocol}).} GPT-OSS-120B judges the correctness of
open-ended answers.
The MLLMs in Tab.~3 are evaluated as subjects of the study, and SPARK-G is
built on Qwen3.5-2B. These models are objects of the research and did not
produce the content of the paper.

\paragraph{Use in writing and code.}
ChatGPT (GPT-6 Astra Max) and Claude (Claude Opus 5.5, through Claude Code) assisted in revising
the wording and structure of text written by the authors, and Claude Code assisted in checking bibliographic entries.
Claude Code (Claude Opus 5.5) was used as a coding assistant: parts of the evaluation, analysis and figure scripts
were generated with it following the authors' specifications, and the authors reviewed, modified and tested this code.
No text, figure or citation was included without author revision, and all
references were verified by the authors.

The authors reviewed all AI-assisted content and take full responsibility
for the paper, including all claims, results, code and references.
{
    \small
    \bibliographystyle{ieeenat_fullname}
    \bibliography{supp}
}

\clearpage
\def\FAgptCorrect{541}
\def\FAgptT{196}
\def\FAgptLoc{138}
\def\FAgptRole{127}
\def\FAgptAssoc{119}
\def\FAgptS{112}
\def\FAgptAll{45}
\def\FAgptAllPctCorrect{8.3}
\def\FAgptAllPctN{2.7}
\def\FApoolN{12}
\def\FApoolLoc{587}
\def\FApoolRA{96}
\def\FApoolRAPct{16}
\def\FApoolRole{74}
\def\FApoolAssoc{22}
\def\FCnDep{302}
\def\FCqD{$-$2.3}
\def\FCqLo{$-$6.0}
\def\FCqHi{+1.3}
\def\FCgD{$-$0.3}
\def\FCgLo{$-$3.3}
\def\FCgHi{+2.6}
\def\FDqVEdgeKept{.531}
\def\FDqVPairsKept{335}
\def\FDqVDepPairsKept{124}
\def\FDgVEdgeKept{.463}
\def\FDgVPairsKept{253}
\def\FDgVDepPairsKept{87}
\def\FDPairsAll{1,180}
\def\FDDepPairsAll{367}

\def\HVReviewed{2,500}\def\HVPass{122}\def\HVEdited{2,055}\def\HVQAWrong{231}\def\HVNoEv{73}\def\HVCannot{19}\def\HVUnsure{746}\def\HVGold{1,639}\def\HVSecond{287}
\def\HVRemoved{323}\def\HVRemovedPct{12.9}\def\HVBoxKept{80.9}
\def\TabHVEditRows{%
Participant boxes corrected, added or removed & 19.1\% of boxes & 73.3 \\
Role tracks of a key event changed & 57.0\% of kept events & 64.7 \\
Object tracks renamed & 32.1\% of tracks & 49.7 \\
Key events removed & 30.3\% of events & 46.2 \\
Key-event boundary moved $>$0.5\,s & 33.7\% of kept events & 43.5 \\
Key events added & 14.3\% of gold events & 20.6 \\
Dependencies removed, added or retyped & 192 / 168 / 9 edges & 17.5 \\
}

\expandafter\def\csname or@Q36@roleNA@acc\endcsname{73.1}
\expandafter\def\csname or@Q36@roleNA@accobj\endcsname{53.3}
\expandafter\def\csname or@Q36@roleNA@ttconf\endcsname{36.1}
\expandafter\def\csname or@Q36@roleNAText@acc\endcsname{43.5}
\expandafter\def\csname or@Q36@roleNAText@accobj\endcsname{23.4}
\expandafter\def\csname or@Q36@roleNAText@ttconf\endcsname{32.5}
\expandafter\def\csname or@GLM46VF@roleNA@acc\endcsname{57.6}
\expandafter\def\csname or@GLM46VF@roleNA@accobj\endcsname{45.9}
\expandafter\def\csname or@GLM46VF@roleNA@ttconf\endcsname{22.8}
\expandafter\def\csname or@GLM46VF@roleNAText@acc\endcsname{41.1}
\expandafter\def\csname or@GLM46VF@roleNAText@accobj\endcsname{43.2}
\expandafter\def\csname or@GLM46VF@roleNAText@ttconf\endcsname{19.0}

\def\TabCorrectSplitRows{%
Qwen3.6-27B & 363 & 76.6 & 18.7 & 2.5 & 2.2 & 89 & 66.3 & 25.8 & 3.4 & 4.5 \\
Qwen3.8-27B & 343 & 72.6 & 19.8 & 4.7 & 2.9 & 91 & 58.2 & 29.7 & 8.8 & 3.3 \\
GLM-4.6V-Flash & 358 & 75.4 & 22.9 & 1.1 & 0.6 & 81 & 63.0 & 35.8 & 0.0 & 1.2 \\
GPT-6.1 Sol & 541 & 63.8 & 15.5 & 12.4 & 8.3 & 117 & 43.6 & 30.8 & 11.1 & 14.5 \\
Gemini 3.8 Flash & 430 & 67.7 & 18.8 & 6.7 & 6.7 & 111 & 53.2 & 28.8 & 8.1 & 9.9 \\
Joint SFT & 222 & 67.6 & 17.1 & 2.3 & 13.1 & 56 & 73.2 & 16.1 & 5.4 & 5.4 \\
\method{} (2B) & 201 & 54.2 & 25.4 & 8.0 & 12.4 & 56 & 44.6 & 25.0 & 17.9 & 12.5 \\
}
\def\ETNDep{302}

\providecommand{\TODO}[1]{\textcolor{red}{TODO#1}}
\providecommand{\ORD}[3]{\ifcsname ord@#1@#2@#3\endcsname\csname ord@#1@#2@#3\endcsname\else\TODO{}\fi}
\expandafter\def\csname ord@GLM46VF@E@L3\endcsname{.149}
\expandafter\def\csname ord@GLM46VF@E@MacroF1\endcsname{.336}
\expandafter\def\csname ord@GLM46VF@E@acc\endcsname{.254}
\expandafter\def\csname ord@GLM46VF@E@ci\endcsname{[+1.9,+5.4]}
\expandafter\def\csname ord@GLM46VF@E@dacc\endcsname{+3.6}
\expandafter\def\csname ord@GLM46VF@E@down\endcsname{80}
\expandafter\def\csname ord@GLM46VF@E@up\endcsname{139}
\expandafter\def\csname ord@GLM46VF@EB@acc\endcsname{.243}
\expandafter\def\csname ord@GLM46VF@EB@ci\endcsname{[+0.8,+4.3]}
\expandafter\def\csname ord@GLM46VF@EB@dacc\endcsname{+2.5}
\expandafter\def\csname ord@GLM46VF@EB@down\endcsname{88}
\expandafter\def\csname ord@GLM46VF@EB@up\endcsname{129}
\expandafter\def\csname ord@GLM46VF@EBR@acc\endcsname{.238}
\expandafter\def\csname ord@GLM46VF@EBR@ci\endcsname{[+0.1,+3.8]}
\expandafter\def\csname ord@GLM46VF@EBR@dacc\endcsname{+2.0}
\expandafter\def\csname ord@GLM46VF@EBR@down\endcsname{103}
\expandafter\def\csname ord@GLM46VF@EBR@up\endcsname{135}
\expandafter\def\csname ord@GLM46VF@ED@acc\endcsname{.256}
\expandafter\def\csname ord@GLM46VF@ED@ci\endcsname{[+2.1,+5.6]}
\expandafter\def\csname ord@GLM46VF@ED@dacc\endcsname{+3.8}
\expandafter\def\csname ord@GLM46VF@ED@down\endcsname{81}
\expandafter\def\csname ord@GLM46VF@ED@up\endcsname{143}
\expandafter\def\csname ord@GLM46VF@EI@MacroF1\endcsname{.347}
\expandafter\def\csname ord@GLM46VF@EI@acc\endcsname{.250}
\expandafter\def\csname ord@GLM46VF@EI@ci\endcsname{[+1.3,+4.9]}
\expandafter\def\csname ord@GLM46VF@EI@dacc\endcsname{+3.1}
\expandafter\def\csname ord@GLM46VF@EI@down\endcsname{93}
\expandafter\def\csname ord@GLM46VF@EI@up\endcsname{144}
\expandafter\def\csname ord@GLM46VF@EID@acc\endcsname{.246}
\expandafter\def\csname ord@GLM46VF@EID@ci\endcsname{[+1.0,+4.6]}
\expandafter\def\csname ord@GLM46VF@EID@dacc\endcsname{+2.7}
\expandafter\def\csname ord@GLM46VF@EID@down\endcsname{97}
\expandafter\def\csname ord@GLM46VF@EID@up\endcsname{142}
\expandafter\def\csname ord@GLM46VF@Eshift@acc\endcsname{.200}
\expandafter\def\csname ord@GLM46VF@Eshift@ci\endcsname{[-3.6,-0.1]}
\expandafter\def\csname ord@GLM46VF@Eshift@dacc\endcsname{-1.9}
\expandafter\def\csname ord@GLM46VF@Eshift@down\endcsname{125}
\expandafter\def\csname ord@GLM46VF@Eshift@up\endcsname{94}
\expandafter\def\csname ord@GLM46VF@TextEID@acc\endcsname{.101}
\expandafter\def\csname ord@GLM46VF@TextEID@ci\endcsname{[-13.9,-9.5]}
\expandafter\def\csname ord@GLM46VF@TextEID@dacc\endcsname{-11.7}
\expandafter\def\csname ord@GLM46VF@TextEID@down\endcsname{269}
\expandafter\def\csname ord@GLM46VF@TextEID@up\endcsname{77}
\expandafter\def\csname ord@GLM46VF@V@L3\endcsname{.044}
\expandafter\def\csname ord@GLM46VF@V@MacroF1\endcsname{.157}
\expandafter\def\csname ord@GLM46VF@V@acc\endcsname{.218}
\expandafter\def\csname ord@GLM46VF@V@mTIoU\endcsname{.204}
\expandafter\def\csname ord@GLM46VF@grpprobe@exact\endcsname{11.6}
\expandafter\def\csname ord@GLM46VF@grpprobe@pair\endcsname{60.1}
\expandafter\def\csname ord@GLM46VF@grpprobeText@exact\endcsname{10.6}
\expandafter\def\csname ord@GLM46VF@grpprobeText@pair\endcsname{59.1}
\expandafter\def\csname ord@GLM46VF@roleprobe@acc\endcsname{53.4}
\expandafter\def\csname ord@GLM46VF@roleprobe@accobj\endcsname{49.1}
\expandafter\def\csname ord@GLM46VF@roleprobe@ttconf\endcsname{19.6}
\expandafter\def\csname ord@GLM46VF@roleprobeText@acc\endcsname{42.9}
\expandafter\def\csname ord@GLM46VF@roleprobeText@accobj\endcsname{43.7}
\expandafter\def\csname ord@GLM46VF@roleprobeText@ttconf\endcsname{17.2}
\expandafter\def\csname ord@Q36@E@L3\endcsname{.150}
\expandafter\def\csname ord@Q36@E@MacroF1\endcsname{.328}
\expandafter\def\csname ord@Q36@E@acc\endcsname{.248}
\expandafter\def\csname ord@Q36@E@ci\endcsname{[+0.7,+4.6]}
\expandafter\def\csname ord@Q36@E@dacc\endcsname{+2.7}
\expandafter\def\csname ord@Q36@E@down\endcsname{104}
\expandafter\def\csname ord@Q36@E@up\endcsname{148}
\expandafter\def\csname ord@Q36@EB@acc\endcsname{.245}
\expandafter\def\csname ord@Q36@EB@ci\endcsname{[+0.4,+4.2]}
\expandafter\def\csname ord@Q36@EB@dacc\endcsname{+2.3}
\expandafter\def\csname ord@Q36@EB@down\endcsname{108}
\expandafter\def\csname ord@Q36@EB@up\endcsname{146}
\expandafter\def\csname ord@Q36@EBR@acc\endcsname{.242}
\expandafter\def\csname ord@Q36@EBR@ci\endcsname{[+0.1,+4.0]}
\expandafter\def\csname ord@Q36@EBR@dacc\endcsname{+2.0}
\expandafter\def\csname ord@Q36@EBR@down\endcsname{116}
\expandafter\def\csname ord@Q36@EBR@up\endcsname{149}
\expandafter\def\csname ord@Q36@ED@acc\endcsname{.245}
\expandafter\def\csname ord@Q36@ED@ci\endcsname{[+0.5,+4.3]}
\expandafter\def\csname ord@Q36@ED@dacc\endcsname{+2.4}
\expandafter\def\csname ord@Q36@ED@down\endcsname{106}
\expandafter\def\csname ord@Q36@ED@up\endcsname{145}
\expandafter\def\csname ord@Q36@EI@MacroF1\endcsname{.350}
\expandafter\def\csname ord@Q36@EI@acc\endcsname{.249}
\expandafter\def\csname ord@Q36@EI@ci\endcsname{[+0.8,+4.6]}
\expandafter\def\csname ord@Q36@EI@dacc\endcsname{+2.7}
\expandafter\def\csname ord@Q36@EI@down\endcsname{103}
\expandafter\def\csname ord@Q36@EI@up\endcsname{148}
\expandafter\def\csname ord@Q36@EID@acc\endcsname{.244}
\expandafter\def\csname ord@Q36@EID@ci\endcsname{[+0.2,+4.2]}
\expandafter\def\csname ord@Q36@EID@dacc\endcsname{+2.3}
\expandafter\def\csname ord@Q36@EID@down\endcsname{118}
\expandafter\def\csname ord@Q36@EID@up\endcsname{155}
\expandafter\def\csname ord@Q36@Eshift@acc\endcsname{.168}
\expandafter\def\csname ord@Q36@Eshift@ci\endcsname{[-7.3,-3.4]}
\expandafter\def\csname ord@Q36@Eshift@dacc\endcsname{-5.4}
\expandafter\def\csname ord@Q36@Eshift@down\endcsname{180}
\expandafter\def\csname ord@Q36@Eshift@up\endcsname{92}
\expandafter\def\csname ord@Q36@TextEID@acc\endcsname{.103}
\expandafter\def\csname ord@Q36@TextEID@ci\endcsname{[-14.1,-9.6]}
\expandafter\def\csname ord@Q36@TextEID@dacc\endcsname{-11.8}
\expandafter\def\csname ord@Q36@TextEID@down\endcsname{286}
\expandafter\def\csname ord@Q36@TextEID@up\endcsname{92}
\expandafter\def\csname ord@Q36@V@L3\endcsname{.059}
\expandafter\def\csname ord@Q36@V@MacroF1\endcsname{.215}
\expandafter\def\csname ord@Q36@V@acc\endcsname{.221}
\expandafter\def\csname ord@Q36@V@mTIoU\endcsname{.198}
\expandafter\def\csname ord@Q36@grpprobe@exact\endcsname{8.6}
\expandafter\def\csname ord@Q36@grpprobe@pair\endcsname{70.7}
\expandafter\def\csname ord@Q36@grpprobeText@exact\endcsname{8.9}
\expandafter\def\csname ord@Q36@grpprobeText@pair\endcsname{71.8}
\expandafter\def\csname ord@Q36@roleprobe@acc\endcsname{78.3}
\expandafter\def\csname ord@Q36@roleprobe@accobj\endcsname{64.2}
\expandafter\def\csname ord@Q36@roleprobe@ttconf\endcsname{29.0}
\expandafter\def\csname ord@Q36@roleprobeText@acc\endcsname{44.2}
\expandafter\def\csname ord@Q36@roleprobeText@accobj\endcsname{35.5}
\expandafter\def\csname ord@Q36@roleprobeText@ttconf\endcsname{25.5}

\renewcommand{\topfraction}{0.9}\renewcommand{\dbltopfraction}{0.9}\renewcommand{\bottomfraction}{0.5}
\renewcommand{\textfraction}{0.07}\renewcommand{\floatpagefraction}{0.85}\renewcommand{\dblfloatpagefraction}{0.85}
\setcounter{topnumber}{3}\setcounter{dbltopnumber}{3}\setcounter{totalnumber}{5}
\maketitlesupplementary
\begin{table}[!t]
\centering
\footnotesize
\setlength{\tabcolsep}{3pt}
\renewcommand{\arraystretch}{1.08}
\begin{tabular}{@{}>{\raggedright\arraybackslash}p{0.17\linewidth}>{\raggedright\arraybackslash}p{0.40\linewidth}>{\raggedright\arraybackslash}p{0.37\linewidth}@{}}
\toprule
Main paper & Claim or detail & Supplementary \\
\midrule
\cref{tab:bench_compare} & Comparison with related benchmarks & \cref{tab:supp_bench_compare} \\
\cref{sec:bench_filter} & Capability reclassification and its check, training sources & \cref{sec:supp_capability}, \cref{tab:supp_capmap,tab:supp_train_sources} \\
\cref{fig:pipeline}, \cref{sec:bench_pipeline} & Six-stage grounding annotation pipeline & \cref{sec:supp_bench}, \cref{fig:postproc} \\
\cref{sec:bench_pipeline} & Human verification (stage 6) and accuracy of the automatic labels & \cref{sec:supp_hv}, \cref{tab:supp_hv_guide,tab:supp_hv_edits,tab:supp_hv_agree}, \cref{fig:supp_hv_example} \\
\cref{sec:bench_pipeline} & Reliability of the interaction associations and agreement between annotators & \cref{sec:supp_reliability}, \cref{tab:supp_rel_iaa} \\
\cref{tab:bench_stats} & Statistics on all evaluation questions & \cref{tab:supp_bench_stats} \\
\cref{sec:bench_integrity} & Integrity checks C1--C4, models and counts per step & \cref{sec:supp_integrity}, \cref{tab:split} \\
\cref{sec:exp_protocol} & Model inputs, reply parsing, answer judging and its agreement with human judgments & \cref{sec:supp_eval,sec:supp_prompt} \\
\cref{sec:exp_protocol}, \cref{tab:main} & Grounding thresholds and metric variants & \cref{sec:supp_thr}, \cref{tab:supp_thr} \\
\cref{tab:main} & All 26 systems, mTIoP, mTIoG and spurious shares & \cref{tab:supp_main_full} \\
\cref{sec:exp_main}, \cref{fig:errors} & Role and association failures of correct answers, results by capability, questions with a dependency, all evaluation questions & \cref{sec:supp_further}, \cref{fig:supp_incremental}, \cref{tab:supp_capability,tab:supp_capability2,tab:supp_deppos,tab:supp_fulltest} \\
\cref{sec:exp_findings}, Finding~1 & Incomplete recovery of multi-event evidence & \cref{sec:supp_events} \\
\cref{sec:exp_findings}, Finding~2 & Role errors after correct localization & \cref{sec:supp_roles} \\
\cref{sec:exp_findings}, Finding~3 & Association errors with correct participants and roles & \cref{sec:supp_inter} \\
\cref{sec:exp_findings}, Finding~4, \cref{tab:dep_diag} & Coarse events and correlation biases in dependency recovery & \cref{sec:supp_dep}, \cref{tab:supp_dep_diag,tab:supp_e2} \\
\cref{sec:exp_utilization}, \cref{tab:probes} & Effect of verified evidence on answering, full table and coverage & \cref{app:oracle}, \cref{tab:supp_steps_full} \\
\cref{sec:method,sec:exp_learning} & \method{} implementation, results, ablations and cost & \cref{app:method} \\
-- & Dataset documentation, release and examples by capability & \cref{sec:supp_datasheet}, \cref{fig:supp_cap_examples1,fig:supp_cap_examples2,fig:supp_cap_examples3} \\
\bottomrule
\end{tabular}
\par\vspace{2pt}
{\footnotesize\raggedright Unless stated otherwise, results are on the \EvGold{} gold-labeled evaluation questions, and 95\% confidence intervals are bootstrap intervals that resample questions. Tables and figures name any other resampling unit, such as source videos.\par}
\caption*{\textbf{Reading guide.} Claims of the main paper and the sections, tables and figures of this supplementary material that support them.}
\end{table}

\appendix
\let\arxivsection\section
\renewcommand{\section}{\FloatBarrier\arxivsection}
\raggedbottom
\counterwithin{table}{section}
\counterwithin{figure}{section}
\counterwithin{equation}{section}
\def\FAgptCorrect{541}
\def\FAgptT{196}
\def\FAgptLoc{138}
\def\FAgptRole{127}
\def\FAgptAssoc{119}
\def\FAgptS{112}
\def\FAgptAll{45}
\def\FAgptAllPctCorrect{8.3}
\def\FAgptAllPctN{2.7}
\def\FApoolN{12}
\def\FApoolLoc{587}
\def\FApoolRA{96}
\def\FApoolRAPct{16}
\def\FApoolRole{74}
\def\FApoolAssoc{22}
\def\FCnDep{302}
\def\FCqD{$-$2.3}
\def\FCqLo{$-$6.0}
\def\FCqHi{+1.3}
\def\FCgD{$-$0.3}
\def\FCgLo{$-$3.3}
\def\FCgHi{+2.6}
\def\FDqVEdgeKept{.531}
\def\FDqVPairsKept{335}
\def\FDqVDepPairsKept{124}
\def\FDgVEdgeKept{.463}
\def\FDgVPairsKept{253}
\def\FDgVDepPairsKept{87}
\def\FDPairsAll{1,180}
\def\FDDepPairsAll{367}

\section{\bench{} construction}
\label{app:bench}
\begin{table}[t]
\centering
\footnotesize
\setlength{\tabcolsep}{3pt}
\resizebox{\linewidth}{!}{%
\begin{tabular}{@{}l rr c l@{}}
\toprule
Benchmark & \#Videos & \#QAs & Anno. & QA format \\
\midrule
Causal-VidQA~\cite{li2022causalvidqa} & 26.9K & 107.6K & M & MCQ \\
NExT-GQA~\cite{xiao2024nextgqa} & 1.6K & 8.9K & M & MCQ \\
CaST-Bench~\cite{zhang2026castbench} & 1,015 & 2,066 & A\,\&\,M & MCQ, OE \\
V-STaR~\cite{cheng2025vstar} & 2,094 & 2,094\,$\times$\,2 & A\,\&\,M & OE \\
\midrule
EgoTaskQA~\cite{jia2022egotaskqa} & 2K & 40K & A\,\&\,M & OE, Y/N \\
HourVideo~\cite{chandrasegaran2024hourvideo} & 500 & 13.0K & A\,\&\,M & MCQ \\
HD-EPIC~\cite{perrett2025hdepic} & 156 & 26.6K & A\,\&\,M & MCQ \\
EgoPlan-Bench2~\cite{qiu2024egoplanbench2} & 1.1K & 1.3K & A\,\&\,M & MCQ \\
CausalVQA~\cite{foss2025causalvqa} & 779$^{a}$ & 1.6K & A\,\&\,M & MCQ \\
MultiHop-EgoQA~\cite{gelm2025multihop} & 360 & 1,080 & A\,\&\,M & OE \\
ProcObject-10K~\cite{guo2026procobject} & 1.8K & 10.5K (1,050 test) & A\,\&\,M & OE \\
Minerva-Ego~\cite{nagrani2026minervaego} & 156 & 1.2K & M & MCQ + trace \\
EgoCoT-Bench~\cite{dai2026egocotbench} & 351 & 3.2K & A\,\&\,M & MCQ \\
\midrule
\bench (ours) & 7.1K & \EvTotalK & A\,\&\,M & OE, Y/N \\   %
\bottomrule
\end{tabular}}
\caption{\textbf{Size, annotation and QA format of the benchmarks} of \cref{tab:bench_compare}. Anno.: automatic (A) / manual (M). MCQ: multiple choice, OE: open-ended, Y/N: yes/no. $^{a}$Segments of 298 videos. Counts from the original papers.}
\label{tab:supp_bench_compare}
\end{table}

This appendix details the construction of \bench{} in the order of \cref{sec:bench_filter,sec:bench_pipeline,sec:bench_integrity}: capability reclassification and the sources of the training split (\cref{sec:supp_capability}), the detailed version of the grounding annotation pipeline of \cref{fig:pipeline}, stage by stage (\cref{sec:supp_bench}), human verification (\cref{sec:supp_hv}), the integrity checks with the resulting splits (\cref{sec:supp_integrity}), and the reliability of the annotations (\cref{sec:supp_reliability}).
\Cref{tab:supp_bench_compare} gives the size, annotation and QA format of the benchmarks compared in \cref{tab:bench_compare}. \Cref{tab:supp_bench_stats} gives the statistics of \cref{tab:bench_stats} on all evaluation questions.

\begin{table}[t]
\centering
\small
\setlength{\tabcolsep}{2.5pt}
\begin{tabular}{@{}lr@{}}
\toprule
& Evaluation (all) \\
\midrule
Source datasets & \EvDatasets \\
Source videos & \EvVideos \\
Questions & \EvQ \\
Gold-labeled questions & \EvGold \\
\midrule
Key events (total) & 6,743 \\
Interactions (total) & 9,077 \\
Participants (total) & 22,889 \\
Questions with $\geq$2 key events & 55\% \\
Participants per interaction & 2.52 \\
\midrule
Multi-event questions with a dependency & 50\% \\
\quad \textit{causes} / \textit{enables} pairs & 80 / 1,343 \\
\bottomrule
\end{tabular}
\caption{\textbf{\bench statistics on all evaluation questions}, with the rows of \cref{tab:bench_stats}, whose evidence statistics cover the gold-labeled questions. Evidence: human-verified labels where they exist and automatic labels (stages 1--5) elsewhere. Multi-event questions with a dependency: the share of questions with $\geq$2 key events that have a \textit{causes} or \textit{enables} pair.}
\label{tab:supp_bench_stats}
\end{table}

\subsection{Capability reclassification}
\label{sec:supp_capability}
The evaluation split draws on four source datasets (\cref{tab:supp_capmap}). We map each question template of EgoTaskQA and each question category of the other three datasets to one of the six causal capabilities of \cref{sec:bench_filter}, by the information the question provides and the inference it requires. The mapping follows the definitions and generation programs of the source questions rather than their category names, and categories outside the six capabilities, such as the descriptive questions of EgoTaskQA, are not used.
For example, a question about the next action is planning if it states a goal (EgoPlan-Bench2) and anticipation otherwise (EgoTaskQA, EgoTempo~\cite{plizzari2025egotempo}). A question that alters an action performed in the video is counterfactual (EgoTaskQA, ProcObject-10K), while one that assumes a future action is hypothetical, even where the source types it as predictive and counterfactual (EgoTaskQA).

Each template or category maps to one capability, but two EgoTaskQA types contain templates of different capabilities: among the kept questions, 485 of its 1,893 explanatory questions are goal attribution, and 186 of its 257 questions typed predictive and counterfactual are hypothetical.
For 15 questions that pass C1, all from one EgoTaskQA template, the generating program does not fix whether the negated action occurs in the clip, so their capability (counterfactual or hypothetical) is not confirmed and they are not used. Three of them were also rejected in human verification; \cref{tab:split} counts these three under human verification and the other 12 in a separate row. Finally, EgoPlan-Bench2 is multiple-choice in its source; \bench{} asks its questions without the options and uses the text of the correct option as the ground-truth answer.

\begin{table}[h]
\centering
\footnotesize
\setlength{\tabcolsep}{3pt}
\resizebox{\linewidth}{!}{%
\begin{tabular}{@{}llcrr@{}}
\toprule
Source & Question type & Capability & Questions & Kept \\
\midrule
EgoTaskQA & explanatory (world, multi-agent scope) & CE & 1,759 & 1,408 \\
 & explanatory (intent scope) & G & 498 & 485 \\
 & counterfactual & CF & 1,098 & 887 \\
 & predictive and counterfactual (action negated) & CF & 86 & 71 \\
 & predictive and counterfactual (action assumed) & H & 207 & 186 \\
 & predictive & A & 589 & 559 \\
EgoPlan-Bench2 & all & P & 1,321 & 1,164 \\
EgoTempo & future action prediction & A & 50 & 45 \\
ProcObject-10K & mistake recognition & CE & 210 & 196 \\
 & counterfactual reasoning & CF & 210 & 102 \\
\midrule
Total & & & 6,028 & 5,103 \\
\bottomrule
\end{tabular}}
\caption{\textbf{Assignment of the source question types to the capabilities.} Questions: questions of each listed type in the source test splits. Kept: questions kept by the question-only filter C1 whose capability is confirmed at the question level. CE: causal explanation, CF: counterfactual, H: hypothetical, A: anticipation, P: planning, G: goal attribution.}
\label{tab:supp_capmap}
\end{table}

\Cref{fig:capability} shows the resulting capability composition of the evaluation split, its gold-labeled questions and the training split.

\definecolor{capCE}{HTML}{6F9FDB}\definecolor{capCF}{HTML}{9EC6F3}\definecolor{capH}{HTML}{C5CBD2}
\definecolor{capA}{HTML}{44BDB8}\definecolor{capP}{HTML}{83CAC8}\definecolor{capG}{HTML}{DFDDDA}\definecolor{capTx}{HTML}{1D3B4A}
\input{sec/stats_eval_auto}
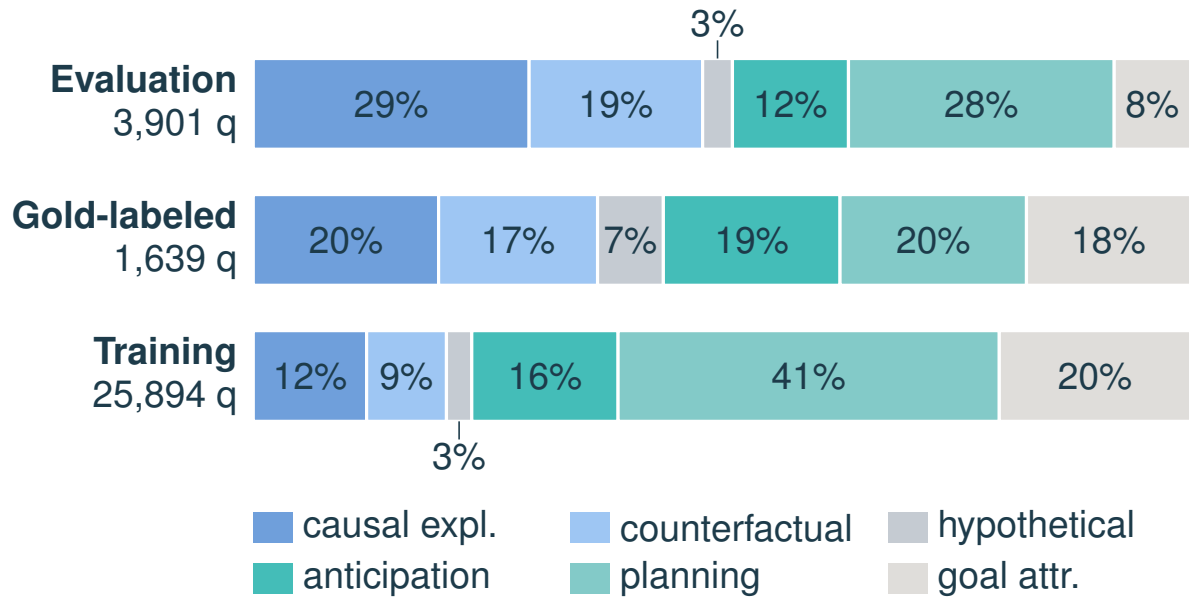
\begin{figure}[t]
\centering
\begin{tikzpicture}[x=1mm,y=1mm,font=\sffamily\scriptsize,text=capTx]
\def\W{62}
\def\CapBarsFig{0/18/{29.43,18.51,3.18,12.38,28.35,8.15},1/9/{19.77,16.96,7.02,18.85,19.89,17.51},2/0/{12.09,8.52,2.70,15.61,40.67,20.41}}
\foreach \row/\y/\vals in \CapBarsFig {
  \pgfmathsetmacro{\acc}{0}
  \foreach \v [count=\i] in \vals {
    \ifcase\i\or\def\c{capCE}\or\def\c{capCF}\or\def\c{capH}\or\def\c{capA}\or\def\c{capP}\or\def\c{capG}\fi
    \pgfmathsetmacro{\xa}{19+\acc*\W/100}\pgfmathsetmacro{\xb}{19+(\acc+\v)*\W/100}
    \draw[fill=\c,draw=white,line width=.6pt] (\xa,\y) rectangle (\xb,\y+6);
    \pgfmathparse{\v>6?1:0}\ifnum\pgfmathresult=1 \pgfmathtruncatemacro{\vr}{round(\v)}\node[inner sep=0] at ({(\xa+\xb)/2},\y+3) {\vr\%};\fi
    \pgfmathparse{\acc+\v}\global\let\acc\pgfmathresult
  }
}
\node[anchor=east,align=right,inner sep=0] at (18,21) {\textbf{Evaluation}\\\EvQ{} q};
\node[anchor=east,align=right,inner sep=0] at (18,12) {\textbf{Gold-labeled}\\\EvGold{} q};
\node[anchor=east,align=right,inner sep=0] at (18,3) {\textbf{Training}\\25,894 q};
\ifdim\EvCapHw pt<6pt \draw[capTx,line width=.3pt] ({19+\EvCapHx*\W/100+\EvCapHw*\W/200},24) -- ++(0,1.2) node[above,inner sep=.5pt] {\EvCapHpct\%};\fi
\draw[capTx,line width=.3pt] ({19+20.61*\W/100+2.70*\W/200},0) -- ++(0,-1.2) node[below,inner sep=.5pt] {3\%};
\foreach \c/\n [count=\i from 0] in {capCE/causal expl.,capCF/counterfactual,capH/hypothetical,capA/anticipation,capP/planning,capG/goal attr.} {
  \pgfmathsetmacro{\lx}{19+mod(\i,3)*21}\pgfmathsetmacro{\ly}{-7-floor(\i/3)*3.4}
  \draw[fill=\c,draw=none] (\lx,\ly-1) rectangle ++(2.6,2); \node[anchor=west,inner sep=0] at (\lx+3.2,\ly) {\n};
}
\end{tikzpicture}
\caption{\textbf{Capability composition} of the evaluation split (after C1, without the questions removed by C2 and C4), its gold-labeled questions with human-verified evidence and the training split (share of questions).}
\label{fig:capability}
\end{figure}

\paragraph{Check of the assignment.}
We checked the assignment on a stratified sample of 150 questions of the evaluation split, 15 per row of \cref{tab:supp_capmap}, so the agreement rates below are unweighted over the rows.
Two annotators labeled each question with one of the six capabilities or \emph{none}, from the video clip, the question and the answer, given only the definitions of the six capabilities. The source dataset, the source question type and the assignment were hidden.
Independently, GPT-5.4-mini labeled the same questions from the same definitions, frames sampled from the clip, the question and the answer (three requests at temperature 0, majority label).
The assignment agrees with 269 of the 300 human labels (89.7\%, Cohen's $\kappa=0.87$), and the two annotators agree with each other on 139 of the 150 questions (92.7\%, $\kappa=0.91$).
It agrees with GPT-5.4-mini on 129 of the 150 questions (86.0\%, $\kappa=0.83$).

\paragraph{Training split.}
The training split draws on 11 egocentric datasets (\cref{tab:supp_train_sources}).
Six provide questions: EgoPlan-IT, the instruction-tuning set of EgoPlan-Bench~\cite{chen2023egoplanbench}, EgoTools~\cite{tian2026egotools}, EgoSAT~\cite{lei2026egosat}, CFG-Bench~\cite{liu2026cfgbench}, and the training splits of EgoTaskQA~\cite{jia2022egotaskqa} and ProcObject-10K~\cite{guo2026procobject}.
As in the evaluation split, only the question types that fall under the six capabilities are used.
The other five provide procedural annotations but no questions: HoloAssist~\cite{wang2023holoassist}, Ego-Exo4D~\cite{grauman2024egoexo4d}, EgoPER~\cite{lee2024egoper}, Ego4D Goal-Step~\cite{song2023goalstep} and CaptainCook4D~\cite{peddi2024captaincook4d}.
Their questions and answers are built from the annotations with templates; an LLM assists in this step for all sources but HoloAssist (see the LLM Usage Statement).
For question types that do not fix a capability (in EgoTools, CFG-Bench and HoloAssist), Qwen3.6-27B judges each question on four criteria from its text, and fixed rules derive the capability; the 1,613 questions without a confirmed capability are removed (\cref{tab:split}).
EgoTaskQA and ProcObject-10K also supply the evaluation split, from their test splits, and training questions that share a source video with data set aside for evaluation are removed (\cref{tab:split}).

\begin{table}[h]
\centering
\scriptsize
\setlength{\tabcolsep}{3pt}
\begin{tabular}{@{}lllr@{}}
\toprule
Source & Questions & Capabilities & Count \\
\midrule
EgoPlan-IT~\cite{chen2023egoplanbench} & source & P & 8,672 \\
EgoTools~\cite{tian2026egotools} & source & G, P, CE & 6,240 \\
EgoSAT~\cite{lei2026egosat} & source & A & 4,043 \\
HoloAssist~\cite{wang2023holoassist} & built & CE, CF, H & 2,807 \\
ProcObject-10K~\cite{guo2026procobject} & source & CF, CE & 2,064 \\
CFG-Bench~\cite{liu2026cfgbench} & source & G, CE, P, CF & 1,113 \\
Ego-Exo4D~\cite{grauman2024egoexo4d} & built & G, H & 570 \\
EgoPER~\cite{lee2024egoper} & built & CF, H & 296 \\
Ego4D Goal-Step~\cite{song2023goalstep} & built & H & 43 \\
EgoTaskQA~\cite{jia2022egotaskqa} & source & CF, CE, G & 31 \\
CaptainCook4D~\cite{peddi2024captaincook4d} & built & H & 15 \\
\midrule
Total & & & 25,894 \\
\bottomrule
\end{tabular}
\caption{\textbf{Sources of the training split.} Questions: questions of the source dataset (source) or questions built from its annotations (built). Capabilities are listed by decreasing number of questions, with the abbreviations of \cref{tab:supp_capmap}.}
\label{tab:supp_train_sources}
\end{table}

\subsection{Grounding annotation pipeline}
\label{sec:supp_bench}
\input{fig/fig_postproc}
\paragraph{Overview.}
This section is the detailed version of \cref{fig:pipeline}: it follows the six stages of \cref{sec:bench_pipeline} in order and gives the settings that the main paper omits, together with the automatic post-processing that ends stages 1 and 2 (\cref{fig:postproc}).   %
The MLLM of stages 2, 3 and 5 is GPT-5.4-mini for the evaluation questions and Qwen3.6-27B for the training questions.
The numbers below are computed on the clips of the \EvQ{} evaluation questions, and the training clips are processed in the same way.

\paragraph{Stage 1: entities.}
The clips are re-encoded at 15\,fps.
HOI-DETR~\cite{darkhalil2026improving} runs on frames sampled at 3\,fps, and its detections with a score of at least 0.4 are matched to the SAM~2.1~\cite{ravi2024sam2} tracks of their type (hand or object) by Hungarian matching on box IoU, with a minimum IoU of 0.4 for hands and 0.5 for objects.
An unmatched detection starts a new track only if more than 70\% of its box lies outside the masks of the existing tracks of its type, and a matched track whose mask confidence drops while its IoU with the detection is below 0.75 is re-prompted with the detection box under the same identity.

\paragraph{Track post-processing.}
The tracks can still carry three kinds of identity errors: an object detected twice is followed by two tracks, a track drifts from its object onto another one, and an object detected again after an occlusion starts a new track.
Three steps correct these errors in turn.
\emph{Duplicate removal} first finds pairs of tracks of the same type whose boxes intersect and whose masks reach an IoU of at least 0.7 on at least five frames, bridging gaps of one frame.
On these frames, only the track that agrees better with the HOI-DETR detections is kept, and the longer-lived one on a tie within 0.05.
In \cref{fig:postproc}a, O6 and O7 follow the same mug over frames 485--664 with a mean mask IoU of 0.995, so O7 is removed on these frames.
Over the evaluation clips, this step removes 19.1\% of the track boxes and 176 tracks entirely.
\emph{Identity-drift splitting} then follows the appearance of each object track with DINOv2 ViT-L/14 features~\cite{oquab2024dinov2}, averaged over its mask on every third frame and grouped into tracklets of four samples.
When two consecutive tracklets have a cosine similarity below 0.6 to the last three tracklets of the track, the track is cut at the frame of the lowest box IoU between consecutive frames, and the part after the cut receives a new identity.
In \cref{fig:postproc}b, O3 moves from the blue bottle cap at frame 115 to the coffee bag at frame 129 with a similarity of 0.57, and the cut at frame 128 gives the coffee bag its own identity, O7.
\emph{Re-detection linking} finally joins tracks of one object that were split by an occlusion or a cut.
A later object track is linked to an earlier one if the two are active together on at most six frames, their features reach a similarity of at least 0.7, and their box centers lie within 0.1 of the image diagonal after camera motion is compensated by a homography (SIFT and RANSAC with at least 20 inliers).
The two sides of one cut are never linked back.
In \cref{fig:postproc}c, the spoon is tracked as O5 until frame 486 and as O10 from frame 537, and O10 is linked to O5 with a similarity of 0.90 and a center distance of 0.05.
Over the evaluation clips, the three steps remove 176 tracks and make 12,759 cuts and 4,406 links, turning the 35,185 tracks of the tracker into 43,362 identities.

\paragraph{Stage 2: event units.}
EfficientGEBD~\cite{zheng2024efficientgebd} scores the clip at up to 10\,fps in windows of 100 frames with a stride of 50, and a boundary is placed at the center of each run of scores $\geq$0.5.
The MLLM captions each segment from up to 80 of its frames and receives every track active in it as its identifier and the trajectory of its box center.
Each track is named once, from a crop outlined by its mask, and keeps this name in every segment.
A reply is requested again, up to twice, if it leaves an active track unnamed, renames a named track or gives a name of more than ten words.

\paragraph{Segment merging.}
The boundary detector cuts the clip at changes of motion, so one manipulation can span several segments with nearly the same caption, whereas a key event should cover one manipulation.
For each pair of adjacent segments, the MLLM therefore lists the (action, object) units of the two captions.
Passive actions (\textit{hold}, \textit{steady}, \textit{support}, \textit{rest}, \textit{keep}, \textit{stand}, \textit{idle}) are ignored unless a caption has no other unit, and the two segments are merged if one set of units equals, contains or overlaps the other.
Chains of merged pairs form one event unit, whose caption is kept from the segment whose units cover the other's or is otherwise written by the MLLM to describe the actions of both in order.
In \cref{fig:postproc}d, s09 only holds the bottle that s10 pours from, so the two stay separate, whereas s10 and s11 both pour from the bottle into the cup and become one event unit from 14.2 to 19.4\,s.
Over the evaluation clips, 42.2\% of the segments are merged, and the 48,082 segments form 35,441 event units.

\paragraph{Stage 3: key events.}
The MLLM receives the clip at 4\,fps with a timestamp for every frame pair, the question, its ground-truth answer and the event units with their identifiers, intervals and captions, and returns the units that support the answer.
The ground-truth answer is provided only during annotation, to identify the observed events that support it, and is not available to the evaluated models.
The MLLM samples one reply with temperature 0.7 and top-$p$ 0.8, and a reply that cites an unknown unit is requested again, up to twice.

\paragraph{Stage 4: interactions.}
Qwen3.6-27B (temperature 0) splits the caption of each key event into interactions, each listing its hands, tools and targets by track identifier, with the roles of the caption as hints.
An interaction needs a hand and a target, and a key event left without such an interaction is not kept.
An interaction is active at a 0.5\,s time step when, in either frame of the step, one of its target boxes intersects one of its hand or tool boxes.
Box intersection is an operational approximation of the active span of an interaction, not a verification of physical contact.
The same parsing is applied to the human-verified captions and role tracks of stage~6, so key events without a complete interaction are also removed from the verified labels. 8.3\% of the key events' captions yield no interaction, most of them describing a hold.

\paragraph{Stage 5: dependencies.}
The MLLM receives the clip at 4\,fps, the question and its answer, and the key events with their intervals, descriptions and roles, and returns the \textit{causes} and \textit{enables} pairs, each with a reason.
A reply is requested again, up to twice, if it refers to an unknown event, links an event to itself, uses another label, omits a reason, repeats a pair or forms a cycle.
Questions with one key event require no dependency prediction. Event pairs without a \textit{causes} or \textit{enables} label are assigned \textit{before}, which denotes no annotated dependency.

\paragraph{Stage 6: human verification.}
The verification procedure, the annotation guideline and the edit statistics are given in \cref{sec:supp_hv}.

\subsection{Human verification}
\label{sec:supp_hv}
\begin{table*}[t]
\centering
\footnotesize
\renewcommand{\arraystretch}{1.12}
\begin{tabular}{@{}p{0.15\textwidth}p{0.81\textwidth}@{}}
\toprule
Step & Instruction \\
\midrule
1. Question and answer & Before checking the labels, decide which observations the answer needs. A cause needs the causing action and the state it changes. A counterfactual or hypothetical question needs the observed event that the question alters and the later state needed to judge the outcome. A next action needs the work done so far and the final visible state. A goal needs the actions that reveal it, and a mistake the action and the outcome that shows it is a mistake. A state or location needs the moment it is created or clearly visible. When the answer itself is not shown (counterfactual, hypothetical, next action), the evidence is the observed events it rests on. \\
2. Video & Watch the whole video (at most 1.5$\times$ speed) with the automatic key events on the timeline. \\
3. Key events & Remove events the answer does not need, including other instances of a repeated action (keep the instance the question refers to), and add missing events. An interval covers only the visible event. It starts when the hand begins to move toward the object and ends when the action is over and its result is visible. Correct boundaries that are off by 1\,s or more, leave differences within 0.5\,s, do not pad intervals and do not select the whole video. \\
4. Roles and tracks & In each key event, assign only three roles. A \emph{hand} is a hand of the camera wearer that manipulates an object or a tool. A \emph{tool} is an object that the hand manipulates to act on another object, and a \emph{target} is an object directly acted on by the hand or the tool. Background objects get no role. Keep one of two tracks of the same object, and add an untracked participant as a new object with a short English name and its boxes. \\
5. Boxes & On the 0.5\,s grid, draw the tightest box around the visible part of each participant. A hand box spans fingertips to wrist (gloves included, forearm excluded). Delete the box of a frame in which the participant is fully occluded or out of view (the interval is unchanged), never extend a box over occluded parts, and leave differences within 10\% of the object size. \\
6. Captions and names & Correct only wrong content (action, object, left or right hand, state change), not style. A caption is one or two English sentences naming each participant with its track ID, and the tracks in the caption and in the roles agree. \\
7. Dependencies & Keep a dependency only for a direct physical link. It is \emph{causes} if the earlier event directly produces the outcome observed in the later one, and \emph{enables} if the manipulation in the later event cannot be performed as shown without the condition the earlier event creates. Temporal succession, a shared object or activity, repetition, preparation or intention, a chain $A\!\to\!B\!\to\!C$ (no $A\!\to\!C$) and pre-existing conditions are not dependencies. For each pair, name the state the earlier event creates, find how the later event uses it or what it results in, and decide \emph{causes}, else \emph{enables}, else remove. Remove when unsure. \\
8. Verdict & \emph{pass} (nothing changed), \emph{edited}, \emph{no evidence} (no scene in the video supports the answer), \emph{QA wrong} (the video clearly contradicts the answer) or \emph{cannot judge} (video unreadable), with a note for no evidence and QA wrong. Disagreeing with a plausible answer is not QA wrong. \emph{Unsure} flags a doubtful review, also after more than 5 minutes on one question. \\
\bottomrule
\end{tabular}
\caption{\textbf{Human verification guideline} (stage~6).}
\label{tab:supp_hv_guide}
\end{table*}
\begin{figure*}[tb]\centering
\begin{minipage}[t]{0.565\textwidth}\centering
\vspace{0pt}\includegraphics[width=\linewidth]{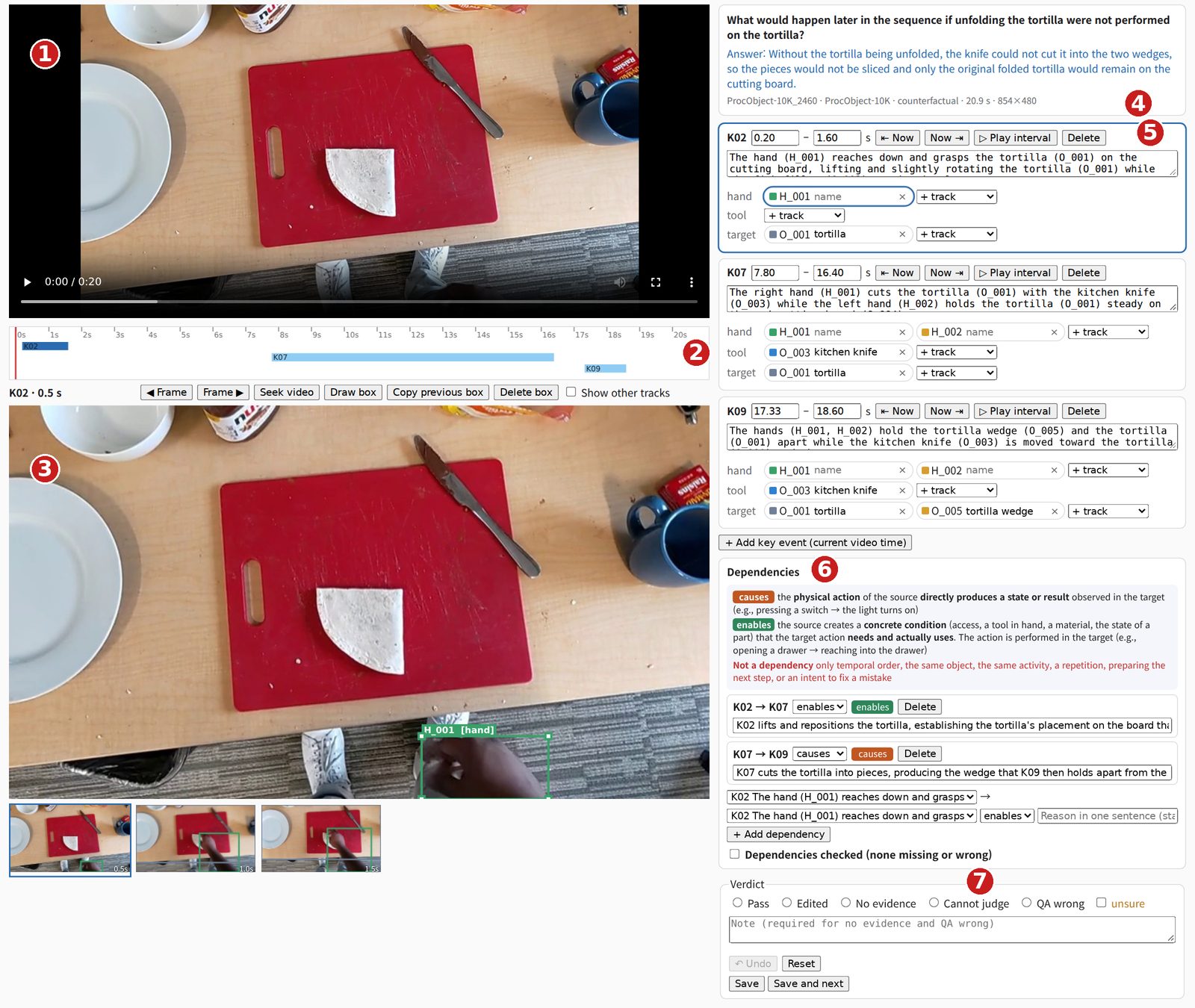}\\[1pt]{\footnotesize (a) Review tool}
\end{minipage}\hfill
\begin{minipage}[t]{0.415\textwidth}\centering
\vspace{0pt}\includegraphics[width=\linewidth]{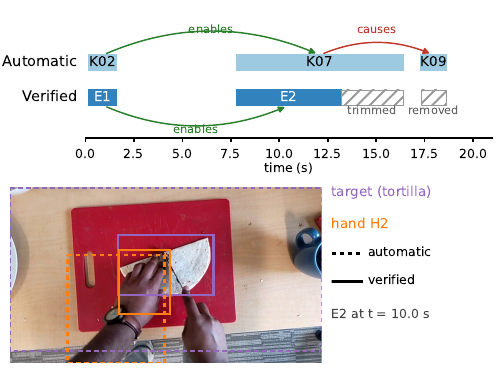}\\[1pt]{\footnotesize (b) Automatic and verified labels}\\[5pt]
\includegraphics[width=\linewidth]{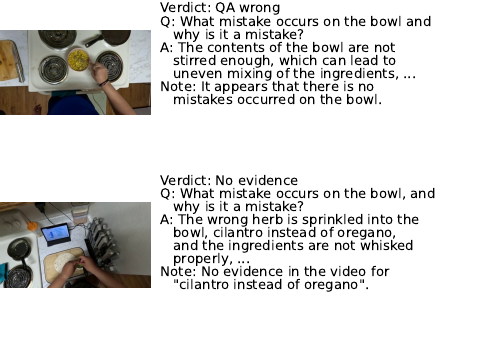}\\[1pt]{\footnotesize (c) Removed questions}
\end{minipage}
\caption{\textbf{Human verification (stage~6).} (a) The review tool on one gold-labeled question with its automatic labels before editing: \textcircled{\scriptsize 1} the video, \textcircled{\scriptsize 2} the key events on the timeline, \textcircled{\scriptsize 3} the frame at a 0.5\,s grid time with the boxes of the role tracks, which the annotator draws, copies or deletes, \textcircled{\scriptsize 4} the question and answer, \textcircled{\scriptsize 5} each key event with its interval, caption and role tracks, \textcircled{\scriptsize 6} the dependencies with the definitions of \textit{causes} and \textit{enables}, and \textcircled{\scriptsize 7} the verdict and note. (b) Automatic and verified labels of this question: the annotator removed the third key event and its dependency, shortened the second key event and corrected the boxes of its target and of the second hand (dashed: automatic, solid: verified, at 10.0\,s). (c) Two removed questions with the annotator's note.}
\label{fig:supp_hv_example}
\end{figure*}

\paragraph{Procedure and verdicts.}
In stage~6, annotators reviewed \HVReviewed{} evaluation questions in a web tool that shows the video, the question and answer, and the automatic key events with their intervals, captions, named role tracks, boxes on a 0.5\,s grid and dependencies (\cref{fig:supp_hv_example}a).
The reviewed questions were drawn at random from the questions kept by C1, balancing the six capabilities. As the questions kept by C1 are not balanced over capabilities, the gold-labeled questions differ in composition from the evaluation split (\cref{fig:capability}).
For each question the annotator gives one verdict: \emph{pass} (labels correct as shown), \emph{edited} (labels corrected), \emph{QA wrong} (question or answer incorrect), \emph{no evidence} (no scene in the video supports the answer) or \emph{cannot judge}. The last three remove the question.
Of the \HVReviewed{} reviewed questions, \HVPass{} pass, \HVEdited{} are edited, and \HVQAWrong{}, \HVNoEv{} and \HVCannot{} are removed as QA wrong, no evidence and cannot judge. Annotators marked \HVUnsure{} of their reviews as unsure. \Cref{fig:supp_hv_example} shows an edited question (b) and two removed questions (c).
Twelve trained annotators made the final reviews, with a median review time of 180\,s (details in \cref{sec:supp_datasheet}).
Each question was initially reviewed by one annotator. Questions that the first review removed or marked as unsure were eligible for a second review when the recorded reason did not clearly support the verdict, and another annotator reviewed 187 such gold-labeled questions, whose second verdict replaces the first. The written guideline (\cref{tab:supp_hv_guide}), worked examples and the unsure flag were designed to promote consistency across reviews.

\paragraph{Guideline.}
Annotators followed a written guideline with worked examples from pilot reviews, summarized in \cref{tab:supp_hv_guide} in the terms of \cref{sec:bench_task}.
It asks them to correct what is wrong, add what is missing and leave correct labels unchanged, and to annotate only what is visible.
The interactions are not verified directly: annotators verify the key events, their captions, role tracks and boxes, and the dependencies, from which the interactions are parsed (stage~4, \cref{sec:supp_bench}).

\begin{table}[t]
\centering
\footnotesize
\setlength{\tabcolsep}{3pt}
\resizebox{\linewidth}{!}{%
\begin{tabular}{@{}lrrr@{}}
\toprule
& Questions (\%) & Key events (\%) & Tracks (\%) \\
\midrule
\multicolumn{4}{@{}l}{\textbf{Removed}} \\
\quad QA wrong & 231 (9.2) & -- & -- \\
\quad No evidence & 73 (2.9) & -- & -- \\
\quad Cannot judge & 19 (0.8) & -- & -- \\
\midrule
\multicolumn{4}{@{}l}{\textbf{Edited}} \\
\quad\textit{Key events} & & & \\
\qquad Removed & 1,012 (40.5) & 1,471 (31.0) & -- \\
\qquad Boundary moved $>$0.5\,s & 936 (37.4) & 1,113 (23.5) & -- \\
\qquad Added & 441 (17.6) & 555 (11.7) & -- \\
\quad\textit{Interactions} & & & \\
\qquad Participant boxes corrected, added or removed & 1,586 (63.4) & 2,423 (51.1) & 3,917 (32.5) \\
\qquad Role tracks changed & 1,390 (55.6) & 1,862 (39.2) & 4,260 (35.4) \\
\qquad Object tracks renamed & 1,049 (42.0) & 1,518 (32.0) & 1,492 (12.4) \\
\quad\textit{Dependencies} & & & \\
\qquad Removed, added or retyped & 355 (14.2) & 780 (16.4) & -- \\
\midrule
Total & 2,500 & 4,744 & 12,045 \\
\bottomrule
\end{tabular}}
\caption{\textbf{Changes made in human verification} on the \HVReviewed{} reviewed questions. Removed: questions removed by their verdict. Edited: on the questions that are not removed, the numbers of questions, key events and role tracks with each edit. Shares are relative to the Total row: all reviewed questions, and the automatic key events and the tracks that hold a role in them, on the questions that are not removed. Added key events are counted relative to the automatic key events.}
\label{tab:supp_hv_edits}
\end{table}

\begin{table*}[t]
\centering
\footnotesize
\setlength{\tabcolsep}{4pt}
\resizebox{\textwidth}{!}{%
\begin{tabular}{@{}l cc ccc ccc@{}}
\toprule
& \multicolumn{2}{c}{Key events} & \multicolumn{3}{c}{Interactions} & \multicolumn{3}{c}{Dependencies} \\
\cmidrule(lr){2-3}\cmidrule(lr){4-6}\cmidrule(lr){7-9}
System & mTIoU & F1@.3 & \Lone & \Ltwo & \Lthree & Macro-F1 & Edge F1 & \depacc \\
\midrule
GPT-6.1 Sol & .227 / .184 & .400 / .325 & .238 / .124 & .224 / .109 & .217 / .106 & .244 / .106 & .061 / .044 & .612 / .489 \\
Gemini 3.1 Pro & .221 / .173 & .359 / .292 & .155 / .096 & .141 / .084 & .138 / .081 & .115 / .057 & .056 / .042 & .581 / .476 \\
Qwen3.6-27B & .200 / .184 & .346 / .319 & .075 / .044 & .062 / .035 & .061 / .033 & .222 / .136 & .041 / .043 & .564 / .519 \\
GLM-4.6V-Flash (9B) & .203 / .189 & .344 / .322 & .062 / .049 & .046 / .036 & .043 / .035 & .154 / .097 & .029 / .039 & .524 / .488 \\
Open-o3 Video & .139 / .121 & .217 / .209 & -- & -- & -- & -- & -- & -- \\
Cosmos-Reason2 (8B) & .167 / .138 & .260 / .225 & .008 / .008 & .005 / .004 & .005 / .004 & .024 / .011 & .014 / .005 & .457 / .375 \\
\method (2B) & .272 / .239 & .449 / .396 & .207 / .162 & .190 / .150 & .186 / .146 & .101 / .070 & .119 / .078 & .578 / .629 \\
\midrule
Kendall $\tau$ & 0.71 & 0.71 & 0.73 & 0.73 & 0.73 & 0.73 & 0.87 & 0.33 \\
\bottomrule
\end{tabular}}
\caption{\textbf{Scores against the human-verified and the automatic labels.} Each cell gives a system's score with the human-verified labels / with the automatic labels of stages 1--5 before human editing as the reference, for the same predictions and with the scorer and settings of \cref{tab:main}. Questions: the 1,494 of the 1,550 gold-labeled questions with verdict edited whose automatic labels contain a key event with a complete interaction. Kendall $\tau$: agreement of the system orders under the two references (systems with a score).}
\label{tab:supp_hv_agree}
\end{table*}

\paragraph{Changes by evidence type.}
\Cref{tab:supp_hv_edits} counts the questions, key events and role tracks that human verification changed.
Of the \HVReviewed{} reviewed questions, \HVRemovedPct\% are removed as QA wrong, no evidence or cannot judge, most of them as QA wrong.
On the other questions, key events are removed in 40.5\% of the reviewed questions, have a boundary moved by more than 0.5\,s in 37.4\% and are added in 17.6\%.
The interactions receive the most edits: participant boxes change in 63.4\% of the reviewed questions and role tracks in 55.6\%, which affects 32.5\% and 35.4\% of the tracks that hold a role in an automatic key event.
Dependencies are edited in 14.2\% of the reviewed questions.

\paragraph{Accuracy of the automatic labels.}
To measure the automatic labels in the units of the benchmark, we convert the labels of stages 1--5 of the \EvGold{} gold-labeled questions with the same parser and converter as the gold labels and score them as a prediction against the gold labels, with the scorer and thresholds of \cref{tab:main} (each automatic interaction at its time with the most visible participants).
They reach a key-event F1@.3 of .636 [.619, .655], a \Lone{}, \Ltwo{} and \Lthree{} of .442 [.424, .462], .430 [.411, .450] and .422 [.402, .442], and an Edge F1 of .325 [.287, .364] (95\% CIs, bootstrap over questions), with 3,142 key events and 896 dependencies against 2,605 and 367 in the gold labels.
A localized participant has the correct role in 97\% and the correct interaction in 98\% of the cases, so most of the loss lies in the key events and the participant boxes. For 56 questions, the automatic labels contain no key event with a complete interaction and count as empty predictions.
These values are upper bounds, since the gold labels were edited from these labels and errors that the annotators kept count as correct. They are also not comparable to the systems of \cref{tab:main}, since stage~3 selects the key events with the reference answer.
The training split is labeled by the same stages, with Qwen3.6-27B in place of GPT-5.4-mini in stages 2, 3 and 5, and is not verified, so the accuracy of its labels is not measured directly.

\paragraph{Automatic against verified labels.}
\Cref{tab:supp_hv_agree} scores the same predictions of seven systems of \cref{tab:main} twice, once against the human-verified labels and once against the automatic labels of stages 1--5 before human editing, on the 1,494 edited gold-labeled questions whose automatic labels contain a key event with a complete interaction.
System scores are generally lower against the automatic labels, particularly for interactions and dependencies (GPT-6.1 Sol: \Lthree{} from 0.217 to 0.106, Macro-F1 from 0.244 to 0.106). The exceptions are Edge F1 of Qwen3.6-27B and GLM-4.6V-Flash and \depacc{} of \method{} (2B).
The order of the systems is largely preserved: Kendall $\tau$ between the two orders is 0.71--0.87 on every metric except \depacc{} (0.33).
This comparison measures how sensitive the evaluation of systems is to the human revisions of the labels. It is not a measure of inter-annotator agreement, which \cref{sec:supp_rel_iaa} reports.

\subsection{Integrity check protocols}
\label{sec:supp_integrity}
C3 (evidence validity) is the human verification of \cref{sec:supp_hv}. The other three checks are model-based.
\paragraph{C1.} GPT-5.4-mini (reasoning effort none) answers each question from its text alone in three independent requests, and its answers are judged by GPT-5.4-mini at temperature 0 after exact-match, yes/no and refusal rules. A question answered correctly in all three is removed. %
C1 runs on the questions of the types selected in \cref{sec:supp_capability}, in both splits.
\emph{Counts} (\cref{tab:split}): evaluation \EvAllQ{} $\to$ \EvCOneKept{} questions (15.1\% removed), training 117,304 $\to$ 53,582 questions.
\paragraph{C2 and C4.} InternVL3.5-38B (thinking off) answers with temperature 0.6, top-$p$ 0.95 and seeds 0, 1 and 2, from frames at 4\,fps (at most 148, $448^2$, with timestamps), where masked frames are black. GPT-OSS-120B (temperature 0) judges the answers. C2 removes a question answered correctly in all three runs with the key-event frames masked, and C4 one answered correctly in all three runs on the full video and in none with only the key-event frames visible. 
\emph{Counts} (evaluation split, \cref{tab:split}): C2 4,672 $\to$ 3,929 questions (\EvRmCtwo{} removed, 15.9\%), C4 3,929 $\to$ \EvQ{} questions (\EvRmCfour{} removed, 0.7\%).
\paragraph{Scope.} C1 uses GPT-5.4-mini, and C2 and C4 use InternVL3.5-38B. Each model-based check makes three requests per input under the decoding settings above. These checks remove questions that show answer shortcuts under the tested models and settings. They do not establish that the kept questions require every annotated evidence component.

\paragraph{Training and evaluation splits.}\label{sec:supp_split}
\Cref{tab:split} gives the number of questions after each step, for the training and the evaluation split. Of the \EvQ{} evaluation questions, \EvGold{} are gold-labeled.
Of the 2,177 questions kept in human verification, 538 do not enter the gold-labeled set because they fail C2 or C4, have no key event with a complete interaction after verification or have no confirmed capability, which leaves the 1,639 gold-labeled questions.
\begin{table}[t]
\centering
\footnotesize
\setlength{\tabcolsep}{3pt}
\resizebox{\linewidth}{!}{%
\begin{tabular}{@{}lrr@{}}
\toprule
& Training split & Evaluation split \\
\midrule
Source questions & 300,179 & 11,654 \\
After capability reclassification & 117,304 & \EvAllQ{} \\
After C1 & 53,582 & \EvCOneKept{} \\
\midrule
\quad removed: overlap with evaluation & 16,302 & -- \\
\quad removed: rejected in human verification & -- & \HVRemoved{} \\
\quad removed: capability not confirmed at the question level & 1,613 & 12 \\
\quad removed: no key event with a complete interaction & 9,773 & \EvRmNoEv{} \\
\quad removed: C2 (evidence dependence) & -- & \EvRmCtwo{} \\
\quad removed: C4 (evidence sufficiency) & -- & \EvRmCfour{} \\
\midrule
Split & 25,894 & \EvQ{} \\
\quad gold-labeled (human-verified labels) & -- & \EvGold{} \\
\bottomrule
\end{tabular}}
\caption{\textbf{Construction of the training and evaluation splits.} A question removed by both C2 and C4 is counted under C2. Gold-labeled: evaluation questions whose evidence is human-verified (\cref{tab:main}).}
\label{tab:split}
\end{table}

\subsection{Annotation reliability}
\label{sec:supp_reliability}
Annotators verify the key events, their captions, role tracks and boxes, and the dependencies directly, whereas the interactions and the association of their participants are parsed afterwards from the verified captions (stage~4). \Cref{sec:supp_rel_assoc} therefore tests whether people reproduce the parsed association, and \cref{sec:supp_rel_iaa} measures how far independent verifications of the directly verified labels agree.

\subsubsection{Interaction association}
\label{sec:supp_rel_assoc}
In stage~6, Qwen3.6-27B splits each verified key-event caption into interactions, each a single manipulation of a target (stage~4, \cref{sec:supp_bench}).
To test whether people reproduce these association labels, two trained annotators group the participants into interactions at 45 moments with multiple interactions: all such moments of 40 gold-labeled questions drawn at random from the 202 that contain one (\STNConcMom{} moments, \cref{sec:supp_inter_given}).
For each moment they see the video and the frame with the verified participants, roles and boxes, but not the question, the answer, the captions or the interaction labels, and they work independently.
Following the guideline, an interaction is one manipulation of a target by one or more hands, possibly through a tool, so a tool joins the hand that uses it, an object acted on by two interactions joins both, and proximity alone joins no participants.
They also mark the span of each interaction on the 0.5\,s grid inside the key event, covering the time a hand or tool acts on the target, without reaching.

\begin{table}[tb]\centering\footnotesize\setlength{\tabcolsep}{3pt}
\resizebox{\linewidth}{!}{%
\begin{tabular}{@{}l c ccc cccc@{}}\toprule
Comparison & \Lthree{} & Grouping (\%) & $\kappa$ & tIoU & Verdict (\%) & F1@.3 & \Ltwo{} & Edge F1 \\\midrule
\multicolumn{9}{@{}l}{\textit{Interaction association}} \\
\quad Annotators vs.\ labels & .567 & 40.0 & .46 & .581 & -- & -- & -- & -- \\
\quad Annotator vs.\ annotator & .682 & 60.0 & .58 & .687 & -- & -- & -- & -- \\
\quad Overlap rule vs.\ labels & .400 & 26.7 & .29 & -- & -- & -- & -- & -- \\\midrule
\multicolumn{9}{@{}l}{\textit{Verification}} \\
\quad Among three annotators & -- & -- & -- & -- & 90.3 & .639 & .505 & .267 \\
\bottomrule\end{tabular}}
\caption{\textbf{Annotation reliability.} Interaction association on 45 moments (\cref{sec:supp_rel_assoc}): the two annotators against the labels (mean) and against each other. Grouping: moments with the same set of interactions. $\kappa$: whether two participants share an interaction (Cohen's $\kappa$). tIoU: span overlap of interactions with the same participants. Verification on 62 gold-labeled questions (\cref{sec:supp_rel_iaa}), mean over the three pairs of annotators. Verdict: questions with the same verdict (pass, edited, QA wrong, no evidence or cannot judge, \cref{sec:supp_hv}) from all three. Overlap rule: one interaction per hand, each tool or target added to the hand whose box overlaps it most. Other columns: one label set scored against the other as in \cref{tab:main}, with participants matched by box IoU $\geq$0.3 over the shared 0.5\,s grid times. The verification row has no \Lthree{}, since annotators verify the captions from which the interactions are parsed, not the interactions.}
\label{tab:supp_rel_assoc}\label{tab:supp_rel_iaa}
\end{table}

As a reference without learning, the overlap rule of \cref{sec:supp_inter_given} starts one interaction per hand and adds each tool or target to the hand whose box overlaps it most at that moment.
Annotators who group the participants directly reproduce the association labels with an Int-F1 of .567 on average (95\% CI [.457, .674], bootstrap over the 45 moments). This is .167 [.066, .267] above this rule (.400 [.274, .529]) but below their agreement with each other (.682, \cref{tab:supp_rel_assoc}), with a 95\% CI of [.006, .232] for the difference.
The parsed labels thus follow how people group the participants more closely than box overlap does, though less closely than a second annotator.
Exact grouping agreement is lower (40.0\% with the labels, 60.0\% between the annotators), because a moment counts only when all its interactions match.
The association errors of Finding~3 persist when an annotator's association replaces the labels as the reference. At these 45 moments, Qwen3.6-27B and GLM-4.6V-Flash in $V{+}\mathcal{E}{+}B{+}R$ (\cref{sec:supp_inter_given}) then reach an Int-F1 of only .177 and .187, .390 [.244, .531] and .380 [.235, .519] below the .567 at which the annotators reproduce the labels.

\subsubsection{Agreement between annotators}
\label{sec:supp_rel_iaa}
In stage~6 each question was reviewed by one annotator, and different annotators reviewed different questions (\cref{sec:supp_hv}).
To measure how far their labels agree, two further trained annotators each verify the same 62 gold-labeled questions: 42 with a moment of multiple interactions that are not in \cref{sec:supp_rel_assoc}, and 20 with an annotated dependency.
They start from the automatic labels of stages 1--5 in the tool and guideline of stage~6 (\cref{tab:supp_hv_guide}), without the gold labels, earlier reviews or each other's labels, as the stage-6 annotator did.
As in stage~6, each annotator gives one verdict per question, pass (labels correct as shown), edited (labels corrected) or one of three verdicts that remove the question (QA wrong, no evidence and cannot judge, \cref{sec:supp_hv}), and corrects the labels of the questions they keep.
With the review from which the gold labels are built, this gives three independent verifications of each question, compared pairwise (\cref{tab:supp_rel_iaa}).

The three annotators give the same verdict on 90.3\% of the questions (95\% CI [82.3, 96.8], bootstrap over the 62 questions), and among the participants that two of them localize in the same key event, they assign the same role in 97.2\% of the cases ([95.7, 98.5]).
The roles of the participants are therefore verified consistently.
Key-event F1@.3 (.639 [.579, .698]) and \Ltwo{} (.505 [.449, .561]) are lower mainly because annotators differ in which key events and participants they add or remove, which also removes or adds the participants of those events.
Dependencies agree least (Edge F1 .267 [.176, .357]), mostly because many dependencies involve a key event that only one annotator keeps (194 of the 320 dependencies, summed over the three pairs of annotators). %
On the 120 event pairs that both annotators keep, they make the same dependency decision for 70.8\% of the pairs. For 26.7\% of the pairs they differ in whether the pair is dependent, and for 2.5\% in the type of the dependency.

These agreements describe how consistently the labels are verified, with the question and answer in view, not how well people perform the task.
Pooled over the 62 questions and averaged over the three pairs of annotators, they do not correspond to the per-question checks $T_i$, $S_i$ and $D_i$ of the grounded correct rate (\cref{sec:exp_metrics}), so an Edge F1 below 0.3 does not mean that the dependencies of one annotator fail $D_i$ against those of another.
The shared automatic starting labels likely raise the agreement, as they raise that of the automatic labels with the gold labels (\cref{sec:supp_hv}), and the selection of questions with multiple interactions or a dependency likely lowers it.

\section{Evaluation protocol}
\label{app:protocol}

\subsection{Model inputs and evaluation details}
\label{sec:supp_eval}
\begin{table*}[t]
\centering
\footnotesize
\setlength{\tabcolsep}{2.5pt}
\resizebox{\textwidth}{!}{%
\begin{tabular}{@{}l c c cccc ccc ccc ccc ccc@{}}
\toprule
& & Answer & \multicolumn{4}{c}{RQ1: Key events (when)} & \multicolumn{3}{c}{RQ2: Interactions (what/where)} & \multicolumn{3}{c}{RQ3: Dependencies (how)} & \multicolumn{6}{c}{RQ4: Grounded answers} \\
\cmidrule(lr){3-3}\cmidrule(lr){4-7}\cmidrule(lr){8-10}\cmidrule(lr){11-13}\cmidrule(lr){14-19}
& & & & & & & & & & & & & \multicolumn{3}{c}{Grounded correct $\uparrow$} & \multicolumn{3}{c}{Spurious share $\downarrow$} \\
\cmidrule(lr){14-16}\cmidrule(lr){17-19}
System & FPS & Acc & mTIoU & mTIoP & mTIoG & F1@.3 & \Lone & \Ltwo & \Lthree & Macro-F1 & Edge F1 & \depacc & GCR$_T$ & GCR$_{ST}$ & GCR$_{STD}$ & SS$_T$ & SS$_{ST}$ & SS$_{STD}$ \\
\midrule
\TabMainRows
\bottomrule
\end{tabular}}
\caption{\textbf{Main results with all columns} (\cref{tab:main}) on the \EvGold{} gold-labeled evaluation questions, whose evidence is human-verified (all \EvQ{} evaluation questions: \cref{sec:supp_fulltest}). FPS: input frames per second (fr.: frames in total, $^\ast$lowered to 1.6--3.9 for the 60 longest videos to fit the model's input limit); --: not predicted by the system.}
\label{tab:supp_main_full}%
\end{table*}

\providecolor{pDk}{HTML}{306692}\providecolor{pTx}{HTML}{1B2A41}\providecolor{pBg}{HTML}{DBE0EB}
\providecolor{tHand}{HTML}{E67E22}\providecolor{tTool}{HTML}{C2185B}\providecolor{tTarg}{HTML}{2E8B57}
\providecolor{cWhen}{HTML}{2F6DB3}\providecolor{cWhat}{HTML}{7A3FA0}\providecolor{cHow}{HTML}{1E7B6A}\providecolor{cAns}{HTML}{A8551B}
\providecommand{\fscr}{\fontsize{7.7}{9.2}\selectfont}
\begin{figure*}[t]
\centering
\begin{tikzpicture}[x=1mm,y=1mm,font=\sffamily\fscr,text=pTx,
  ipanel/.style={draw=pDk,fill=white,rounded corners=2.5pt,line width=1pt},
  tag/.style={anchor=west,fill=#1,text=white,rounded corners=1pt,inner xsep=1.6pt,inner ysep=.8pt,font=\sffamily\fscr\bfseries},
  jsn/.style={anchor=west,inner sep=0pt,font=\sffamily\fontsize{6.5}{7.6}\selectfont,text=pTx}]
\draw[ipanel] (0,0) rectangle (127.6,51.0);
\node[jsn] at (20.40,47.90) {\{\textcolor{cWhen}{\textbf{"evidence"}}:\ [};
\node[jsn] at (20.40,44.65) {\ \ \{\textcolor{pTx!72}{"id":}\ \textbf{"E1"},\ \textcolor{pTx!72}{"start":}\ \textbf{0.0},\ \textcolor{pTx!72}{"end":}\ \textbf{1.5},\ \textcolor{pTx!72}{"description":}\ "The\ hand\ lifts\ and\ unfolds\ the\ tortilla."\},};
\node[jsn] at (20.40,41.40) {\ \ \{\textcolor{pTx!72}{"id":}\ \textbf{"E2"},\ \textcolor{pTx!72}{"start":}\ \textbf{8.0},\ \textcolor{pTx!72}{"end":}\ \textbf{13.0},\ \textcolor{pTx!72}{"description":}\ "The\ right\ hand\ cuts\ it\ with\ the\ knife."\}],};
\node[jsn] at (20.40,38.15) {\ \textcolor{cWhat}{\textbf{"grounding"}}:\ [};
\node[jsn] at (20.40,34.90) {\ \ \{\textcolor{pTx!72}{"event\_id":}\ \textbf{"E1"},\ \textcolor{pTx!72}{"interactions":}\ [\{\textcolor{pTx!72}{"id":}\ \textbf{"I1"},\ \textcolor{pTx!72}{"t":}\ \textbf{1.0},\ \textcolor{pTx!72}{"boxes":}\ [};
\node[jsn] at (20.40,31.65) {\ \ \ \ \{\textcolor{pTx!72}{"role":}\ \textcolor{tHand}{\textbf{"hand"}},\ \textcolor{pTx!72}{"track":}\ \textbf{"H1"},\ \textcolor{pTx!72}{"box":}\ [\textbf{533},\ \textbf{423},\ \textbf{681},\ \textbf{706}]\},};
\node[jsn] at (20.40,28.40) {\ \ \ \ \{\textcolor{pTx!72}{"role":}\ \textcolor{tTarg}{\textbf{"target"}},\ \textcolor{pTx!72}{"track":}\ \textbf{"O1"},\ \textcolor{pTx!72}{"box":}\ [\textbf{451},\ \textbf{403},\ \textbf{595},\ \textbf{662}]\}]\}]\},};
\node[jsn] at (20.40,25.15) {\ \ \{\textcolor{pTx!72}{"event\_id":}\ \textbf{"E2"},\ \textcolor{pTx!72}{"interactions":}\ [\{\textcolor{pTx!72}{"id":}\ \textbf{"I1"},\ \textcolor{pTx!72}{"t":}\ \textbf{10.5},\ \textcolor{pTx!72}{"boxes":}\ [};
\node[jsn] at (20.40,21.90) {\ \ \ \ \{\textcolor{pTx!72}{"role":}\ \textcolor{tHand}{\textbf{"hand"}},\ \textcolor{pTx!72}{"track":}\ \textbf{"H1"},\ \textcolor{pTx!72}{"box":}\ [\textbf{522},\ \textbf{568},\ \textbf{660},\ \textbf{1000}]\},};
\node[jsn] at (20.40,18.65) {\ \ \ \ \{\textcolor{pTx!72}{"role":}\ \textcolor{tTool}{\textbf{"tool"}},\ \textcolor{pTx!72}{"track":}\ \textbf{"O2"},\ \textcolor{pTx!72}{"box":}\ [\textbf{484},\ \textbf{350},\ \textbf{539},\ \textbf{579}]\},};
\node[jsn] at (20.40,15.40) {\ \ \ \ \{\textcolor{pTx!72}{"role":}\ \textcolor{tTarg}{\textbf{"target"}},\ \textcolor{pTx!72}{"track":}\ \textbf{"O1"},\ \textcolor{pTx!72}{"box":}\ [\textbf{347},\ \textbf{270},\ \textbf{652},\ \textbf{610}]\}]\}]\}],};
\node[jsn] at (20.40,12.15) {\ \textcolor{cHow}{\textbf{"relations"}}:\ [\{\textcolor{pTx!72}{"source":}\ \textbf{"E1"},\ \textcolor{pTx!72}{"target":}\ \textbf{"E2"},\ \textcolor{pTx!72}{"type":}\ "enables"\}],};
\node[jsn] at (20.40,8.90) {\ \textcolor{cAns}{\textbf{"answer"}}:\ "Without\ the\ tortilla\ being\ unfolded,\ the\ knife\ could\ not\ cut\ it\ into\ the\ two\ wedges,};
\node[jsn] at (20.40,5.65) {\ \ \ so\ the\ pieces\ would\ not\ be\ sliced\ and\ only\ the\ original\ folded\ tortilla};
\node[jsn] at (20.40,2.40) {\ \ \ would\ remain\ on\ the\ cutting\ board."\}};
\draw[cWhen,line width=1.3pt] (19.20,49.25) -- (19.20,40.05);
\node[tag=cWhen] at (1.20,47.90) {When};
\draw[cWhat,line width=1.3pt] (19.20,39.50) -- (19.20,14.05);
\node[tag=cWhat] at (1.20,38.15) {What \& where};
\draw[cHow,line width=1.3pt] (19.20,13.50) -- (19.20,10.80);
\node[tag=cHow] at (1.20,12.15) {How};
\draw[cAns,line width=1.3pt] (19.20,10.25) -- (19.20,1.05);
\node[tag=cAns] at (1.20,8.90) {Answer};
\end{tikzpicture}
\caption{The full JSON output of \cref{fig:teaser}, with boxes as $[x_{\min},y_{\min},x_{\max},y_{\max}]$ normalized to 0--1000.}
\label{fig:teaser_json}
\end{figure*}
\def\TabThrRows{%
Qwen3.5-2B & .027 & .001 & .001 & .77 & .99 & .013 & .001 & .001 & .89 & .99 & .004 & .000 & .000 & .96 & 1.00 \\   %
Qwen3.5-9B & .052 & .002 & .002 & .67 & .99 & .019 & .001 & .001 & .88 & 1.00 & .012 & .000 & .000 & .92 & 1.00 \\   %
Qwen3.6-27B & .052 & .010 & .005 & .77 & .95 & .016 & .003 & .002 & .93 & .99 & .007 & .001 & .001 & .97 & .99 \\   %
Qwen3.8-27B & .057 & .016 & .006 & .73 & .92 & .012 & .006 & .005 & .94 & .97 & .008 & .001 & .001 & .96 & 1.00 \\   %
Qwen3.5-397B-A17B & .049 & .002 & .002 & .75 & .99 & .016 & .001 & .001 & .92 & 1.00 & .008 & .000 & .000 & .96 & 1.00 \\   %
GLM-4.6V-Flash (9B)~\cite{hong2025glm45v} & .054 & .004 & .001 & .75 & .98 & .012 & .000 & .000 & .95 & 1.00 & .005 & .000 & .000 & .97 & 1.00 \\   %
Gemma 4 (31B)~\cite{gemma2026gemma4} & .054 & .003 & .003 & .71 & .98 & .015 & .001 & .001 & .92 & 1.00 & .005 & .000 & .000 & .97 & 1.00 \\   %
InternVL3.5 (38B)~\cite{wang2025internvl35} & .023 & .000 & .000 & .86 & 1.00 & .005 & .000 & .000 & .97 & 1.00 & .003 & .000 & .000 & .98 & 1.00 \\   %
InternVL3.5 (241B-A28B)~\cite{wang2025internvl35} & .014 & .000 & .000 & .91 & 1.00 & .002 & .000 & .000 & .99 & 1.00 & .001 & .000 & .000 & .99 & 1.00 \\   %
EgoThinker~\cite{egothinker2025} & .008 & -- & -- & .97 & -- & .001 & -- & -- & 1.00 & -- & .000 & -- & -- & 1.00 & -- \\   %
Open-o3 Video~\cite{meng2026openo3video} & .027 & -- & -- & .83 & -- & .012 & -- & -- & .93 & -- & .007 & -- & -- & .95 & -- \\   %
VisionCoach~\cite{visioncoach2026} & .032 & -- & -- & .81 & -- & .014 & -- & -- & .92 & -- & .005 & -- & -- & .97 & -- \\   %
VideoChat-R1.5~\cite{videochatr15_2025} & .020 & -- & -- & .84 & -- & .007 & -- & -- & .94 & -- & .004 & -- & -- & .97 & -- \\   %
VideoMind~\cite{videomind2025} & .027 & -- & -- & .81 & -- & .012 & -- & -- & .91 & -- & .006 & -- & -- & .96 & -- \\   %
LLaVA-ST (7B)~\cite{li2025llavast} & .013 & .002 & -- & .85 & .98 & .004 & .000 & -- & .96 & 1.00 & .001 & .000 & -- & .99 & 1.00 \\   %
GroundVQA~\cite{di2024groundvqa} & .003 & -- & -- & .91 & -- & .001 & -- & -- & .98 & -- & .000 & -- & -- & 1.00 & -- \\   %
Cosmos-Reason2 (8B)~\cite{nvidia2025cosmosreason2} & .027 & .000 & .000 & .81 & 1.00 & .012 & .000 & .000 & .92 & 1.00 & .005 & .000 & .000 & .96 & 1.00 \\   %
Gemini 3.8 Flash & .085 & .035 & .018 & .68 & .87 & .024 & .007 & .006 & .91 & .97 & .010 & .001 & .001 & .96 & 1.00 \\   %
Gemini Robotics-ER 2 & .052 & .021 & .007 & .75 & .90 & .015 & .003 & .003 & .93 & .99 & .005 & .001 & .001 & .98 & .99 \\   %
Gemini 3.1 Pro & .082 & .031 & .016 & .60 & .85 & .030 & .009 & .008 & .85 & .96 & .014 & .002 & .002 & .93 & .99 \\   %
GPT-6.1 Sol & .120 & .068 & .027 & .64 & .79 & .031 & .017 & .012 & .91 & .95 & .008 & .005 & .004 & .98 & .98 \\   %
Joint SFT (2B) & .044 & .021 & .018 & .68 & .85 & .018 & .008 & .008 & .87 & .94 & .011 & .003 & .003 & .92 & .98 \\   %
\method (2B) & .056 & .025 & .015 & .54 & .80 & .023 & .007 & .005 & .81 & .94 & .010 & .001 & .000 & .92 & 1.00 \\   %
\method (9B) & .084 & .033 & .018 & .54 & .82 & .036 & .012 & .010 & .80 & .93 & .017 & .002 & .002 & .91 & .99 \\   %
}
\def\TabThrN{1639}

\begin{table*}[t]
\centering
\footnotesize
\setlength{\tabcolsep}{2.4pt}
\resizebox{\textwidth}{!}{%
\begin{tabular}{@{}l ccccc ccccc ccccc@{}}
\toprule
& \multicolumn{5}{c}{Threshold 0.3 (\cref{tab:main})} & \multicolumn{5}{c}{Threshold 0.5} & \multicolumn{5}{c}{Threshold 0.7} \\
\cmidrule(lr){2-6}\cmidrule(lr){7-11}\cmidrule(lr){12-16}
System & GCR$_T$ & GCR$_{ST}$ & GCR$_{STD}$ & SS$_T$ & SS$_{ST}$ & GCR$_T$ & GCR$_{ST}$ & GCR$_{STD}$ & SS$_T$ & SS$_{ST}$ & GCR$_T$ & GCR$_{ST}$ & GCR$_{STD}$ & SS$_T$ & SS$_{ST}$ \\
\midrule
\TabThrRows
\bottomrule
\end{tabular}}
\caption{\textbf{Grounded correct rate and spurious share at stricter thresholds} (\TabThrN{} gold-labeled questions). The thresholds of $T_i$ (mTIoU), $S_i$ (\Lthree) and $D_i$ (Edge F1) are set to 0.3, 0.5 or 0.7; matchings inside the scores are unchanged. --: not predicted by the system.}
\label{tab:supp_thr}
\end{table*}
\paragraph{Inputs.}
The open-source MLLMs, including the Qwen3.5 models~\cite{qwen2026qwen35}, receive frames sampled at 4\,fps (two frames per 0.5\,s timestamp), subject to model-specific frame and token limits (FPS column of \cref{tab:supp_main_full}), run with thinking off where their chat template has it, and are told that a hand that only holds an object is not a participant.
Qwen3.5-397B-A17B runs with GPTQ 4-bit weights and InternVL3.5 (241B-A28B) with FP8 weights.   %
GPT-6.1 Sol receives frames at 4\,fps, each preceded by its timestamp and resized with the pixel budget of the Qwen3.5 video processor, through the OpenAI Batch API at the default reasoning effort (medium).
Gemini 3.8 Flash, Gemini Robotics-ER 2 and Gemini 3.1 Pro receive the video natively through the Gemini Batch API at the lowest thinking level of each model. The video is sampled at 4\,fps, except for Gemini Robotics-ER 2 on the 60 longest videos, which exceed its input limit and are therefore sampled at 1.6--3.9\,fps.
Cosmos-Reason2 runs with its official system prompt, reasoning prompt, 4\,fps sampling and decoding parameters.
Its user prompt contains the task description and JSON format given to the other MLLMs.

\paragraph{Video reasoning specialists.}
\Cref{tab:main} scores each specialist on the elements that its released pipeline predicts, and -- marks the others.
EgoThinker, VideoChat-R1.5, VideoMind and GroundVQA return an answer and temporal intervals.
Open-o3 Video and VisionCoach also return object boxes, which carry no participant role.
These six systems are therefore scored on the answer and the key events, and their interaction, dependency, GCR$_{ST}$, GCR$_{STD}$, SS$_{ST}$ and SS$_{STD}$ cells are empty.

\paragraph{Causal reasoning models.}\label{sec:supp_causal}
Causal-VLM and MECD+ do not select events for a question, answer no question and predict no boxes, so their answer, interaction and grounded-answer cells are empty.
Their links are untyped, so every predicted link is scored as \textit{enables} and every event pair without a predicted link as \textit{before}.
Causal-VLM describes the events of the whole video and links them.
MECD+ predicts the causal links between given events.
We run the authors' code with our reproduction of its best checkpoint. This checkpoint reaches 65.3 on the original benchmark, below the reported 71.3, so its results on \bench{} are those of our reproduction rather than an upper bound for MECD+. We give it the events of Causal-VLM, with the auxiliary event texts generated by Qwen3.6-27B and at most 11 events per video.
Its key-event scores therefore equal those of Causal-VLM, and \cref{tab:main} reports only its dependencies.
Its link score is saved for every pair. Given the ground-truth events instead, it is run on the questions with at least three key events, as its model requires (\cref{sec:supp_dep}).

\paragraph{LLaVA-ST.}
LLaVA-ST has no prompt that returns our evidence format, so it is run as a fixed composition of five of its native single-task prompts with the released inference settings (100 uniform frames): open-ended video QA for the answer, temporal grounding of the question for one key event, temporal referring for its description, and spatio-temporal grounding of the object and of the hand that the camera wearer manipulates within that interval.
Their outputs are converted deterministically into one key event with one interaction, whose two boxes are scored as target and hand.
LLaVA-ST never predicts a tool or a dependency, so its dependency, GCR$_{STD}$ and SS$_{STD}$ cells are empty.

\paragraph{Parsing.}
The same rule applies to every MLLM reply (\cref{fig:teaser_json} shows a complete reply).
A reply that is not valid JSON is cut back to its longest prefix that ends with a complete element, and the open brackets are closed, so only complete events, interactions and dependencies are kept.
Elements that do not follow the format are dropped before scoring and count neither as correct nor as false-positive predictions: an event without a numeric start and end, an interaction without a numeric time, a participant whose role is not hand, tool or target or whose box does not have four values, and a dependency with a label other than \textit{causes} or \textit{enables}.
A dependency with a valid label that refers to an unknown or unmatched event counts as a false positive.
A reply without an answer counts as wrong.
On the gold-labeled questions, no JSON element can be recovered from 383 replies of InternVL3.5 (38B), 734 of InternVL3.5 (241B-A28B) and 40 of Cosmos-Reason2, and from no reply of the other systems. These replies count as empty outputs.
Format failures do not explain the low grounding scores of the two InternVL3.5 models: on the questions whose replies parse, their \Lone{} stays at .000 although they recover key events (F1@.3 .200 and .105), against .072 and .073 for Qwen3.6-27B on the same questions.
Boxes are read in the convention that the prompt specifies, $[x_{\min},y_{\min},x_{\max},y_{\max}]$ normalized to 0--1000.
As a check of this convention, each axis of the boxes of InternVL3.5 and Qwen3.5-397B-A17B is also multiplied by 1 or by $1000/s$ for $s\in\{448,512,640,720,896,1080,1280,1920\}$, as if written in pixels.
The scaling whose boxes overlap the ground truth most often among these 81, chosen on these questions, keeps their \Lone{} at or below .035, against .075 for Qwen3.6-27B.
InternVL3.5 (38B), run in BF16, has the same \Lone{} of .000 as the 241B model run in FP8.
\depacc{} is computed on the pairs with both events matched: 105 and 138 pairs for the two InternVL3.5 models and 115 to 335 for the other open-source MLLMs.
Each predicted interaction is read with the participants written in it, at the input timestamp $\tau$ nearest to its reported time.
Each predicted participant takes the track of the ground-truth participant matched to its box by $\pi$ (\cref{sec:exp_protocol}), and a participant whose box matches none takes no track and counts only as a false positive.
Track identifiers in a reply are therefore not used, and LLaVA-ST and \method{}, which write none, are scored like the other systems.

\paragraph{Answer judging.}
An LLM judge, GPT-OSS-120B (low reasoning effort, temperature 0), judges each answer from the question, the ground-truth answer and the predicted answer, without the video or the evidence.
An empty answer is judged wrong, and when both the reference and the prediction are a bare yes or no, they are compared directly.
The prompt of the LLM judge asks whether the prediction conveys the requested answer: it accepts paraphrases and more specific answers that keep the required content, and rejects answers that change the entity, role, action, state, polarity or temporal and hypothetical conditions asked for, contradictions and refusals.
The same LLM judge and prompt score every system.

\paragraph{Agreement with human judgments.}
To test whether the verdicts of the LLM judge agree with human judgment, two annotators mark 120 predicted answers as correct or incorrect.
The predicted answers are drawn from those that the LLM judge scores for the 19 answering systems of \cref{tab:main} on the \EvGold{} gold-labeled questions: 60 that the LLM judge marks correct and 60 that it marks incorrect, with 10 per capability in each group and no question used twice.
Like the LLM judge, the annotators see only the question, the ground-truth answer and the predicted answer and apply the criteria of its prompt, without seeing the system or the verdict of the LLM judge.
The LLM judge agrees with the two annotators on 88.3\% and 91.7\% of the predicted answers (Cohen's $\kappa$ .77 [.65, .87] and .83 [.73, .92]), close to the agreement between the annotators (91.7\%, $\kappa$ .83 [.73, .92]). The intervals are 95\% bootstrap intervals that resample predicted answers within each verdict and capability group.
The disagreements lie mostly in one direction. Of the 60 predicted answers that the LLM judge marks correct, the annotators mark 57 and 59 correct, but of the 60 that it marks incorrect, they mark only 49 and 51 incorrect.
Of the 6 predicted answers that the LLM judge marks incorrect and both annotators mark correct, 4 agree with the ground-truth answer except that they omit one of its details, such as the tool used, or add an action.
The LLM judge is therefore more conservative than the annotators: it rarely marks correct a predicted answer that they mark incorrect, but marks incorrect some predicted answers that they mark correct.

\paragraph{Temporal variants.}
mTIoP and mTIoG replace the union in mTIoU by the predicted and the ground-truth interval and average over the $N$ predicted and the $M$ ground-truth events, so mTIoP penalizes irrelevant evidence and mTIoG missed evidence.

\paragraph{Statistics.}
Confidence intervals are 95\% bootstrap intervals that resample questions, unless a table or figure states another resampling unit (for example, source videos in \cref{tab:answer}).

\subsection{System prompt of the MLLMs}
\label{sec:supp_prompt}
The general-purpose MLLMs of \cref{tab:main} and the joint SFT baseline (Joint SFT (2B) in \cref{tab:main}, labeled Joint JSON SFT in \cref{fig:errors}) receive the system prompt below, followed by the video and the user message ``Question: \textless question\textgreater''. Cosmos-Reason2 receives the same text in its user prompt (\cref{sec:supp_eval}).
Models that receive frames instead of a video read ``the timestamp of the frame'' for ``the timestamp of the temporal patch'', and the Gemini models receive the boxes without the coordinate order, which the scorer converts.
The output format is the one shown in \cref{fig:teaser_json}.

\nolinenumbers
\begin{lstlisting}
You are an egocentric video question-answering assistant.

Given a video and a question, identify and temporally ground the key events needed to answer the question, ground the hand-object interactions in each key event, give the dependencies between the key events, and answer the question.

# Definitions
- Hand: a hand of the person wearing the camera.
- Interaction: a manipulation that changes the state of one or more objects, that is, their position, their shape, or their relation to other objects (for example lifting and moving, cutting, pouring, attaching, wiping). Opening or closing a container and turning a device on or off also count as interactions.
- Participants: the hands, tools and targets that take part in one interaction. Each participant has one role:
  - hand: a hand that performs the manipulation;
  - tool: an object that a hand uses to change the state of the target; when pouring from a held container, the container is the tool and the receiving object is the target;
  - target: an object whose state the interaction changes; when food is moved into a pan, the food is the target and the pan is not.
- Hold: a hand in contact with an object whose state does not change, including the moment of grasping it. A hold is not an interaction, and a hand that only holds an object is not a participant.

# Tasks
1. Analyze the question to determine what visual evidence is needed, such as actions, interactions, prerequisites, or state changes.
2. Examine the video to locate relevant events. Select all events needed to answer the question and identify their temporal boundaries based on visual evidence.
3. In each key event, find each distinct interaction and all of its participants. For each interaction, choose one moment at which the manipulation is observed and its participants are best visible, and locate every participant visible in that frame.
4. Give the dependencies between the key events.
5. Answer the question based on the evidence.

# Rules
- Ground only observed events, including for hypothetical, counterfactual, and anticipation questions.
- Keep each interval focused on the relevant event while preserving the context needed to interpret it.
- Describe each event concisely using observable actions or state changes.
- List events chronologically with sequential IDs (E1, E2, ...). Use timestamps in seconds.
- An interaction has exactly one set of participants. Start a new interaction whenever its hands, tools or targets change, including when an object that was a target becomes a tool (picking up a kettle and then pouring from it into a cup are two interactions). Consecutive movements by the same participants, such as lifting and carrying a bowl, form one interaction.
- Hands that act together belong to one interaction: two hands lifting one pot are one interaction with two hands. Objects are targets of the same interaction only when they are operated on together. The place where an object is put down is not a target.
- Do not write holds as interactions, and do not include a hand that only holds an object. Grasping with a movement (lifting, carrying, moving) is an interaction.
- In "grounding", keep the event IDs and order of "evidence". Number the interactions of each event I1, I2, ... in time order and write each distinct interaction once. Write an event with no interaction with an empty "interactions" list.
- Write each interaction's moment in "t" as the timestamp of the temporal patch in which it is observed, between the event's start and end, in seconds with one decimal.
- In "boxes", write one entry for every participant of the interaction that is visible at that moment (every hand, every tool, every target), each with its role and a box [x_min, y_min, x_max, y_max] of integers normalized to 0-1000 of the frame width and height from the top-left, as the participant appears at that moment. Omit the tool when no tool is used, and do not guess a box for a participant that is not visible at that moment. A participant appears once per interaction.
- Give each entity an ID that stays the same within an event, even if its role changes: H1, H2, ... for hands and O1, O2, ... for objects (a kettle that is first a target and then a tool keeps its ID).
- In "relations", list the causal dependencies between the key events you selected, from an earlier event (source) to a later event (target): "causes" when the source event physically produces a state or result that the target event shows, "enables" when the source event establishes a concrete condition, such as access, a tool in hand or an available material, that a separate interaction in the target event requires and uses. Temporal order or shared objects alone are neither; omit such pairs. Write an empty list if there is no dependency.
- Output only the JSON object.

# Output Format
Return one JSON object with exactly these fields:
{
  "evidence": [{"id": "E1", "start": <seconds>, "end": <seconds>, "description": "..."}, ...],
  "grounding": [{"event_id": "E1", "interactions": [{"id": "I1", "t": <seconds>,
                 "boxes": [{"role": "hand", "track": "H1", "box": [x_min, y_min, x_max, y_max]},
                           {"role": "hand", "track": "H2", "box": [x_min, y_min, x_max, y_max]},
                           {"role": "tool", "track": "O1", "box": [x_min, y_min, x_max, y_max]},
                           {"role": "target", "track": "O2", "box": [x_min, y_min, x_max, y_max]}]}]}, ...],
  "relations": [{"source": "E1", "target": "E2", "type": "enables"}, ...],
  "answer": "..."
}
Use the same event IDs in "evidence", "grounding" and "relations"; give every evidence event a grounding entry (empty
"interactions" if none); "boxes" lists every visible participant of the interaction, with as many hands, tools and targets as
take part, and no tool when none is used; boxes are integers normalized to 0-1000; "answer" is the direct, concise answer to the question.
\end{lstlisting}
\linenumbers

\subsection{Grounding thresholds}
\label{sec:supp_thr}
\Cref{tab:supp_thr} recomputes the grounded correct rate and the spurious share of \cref{tab:main} with the thresholds of $T_i$, $S_i$ and $D_i$ raised from 0.3 to 0.5 and 0.7, on the same \TabThrN{} gold-labeled questions and predictions.
The matchings inside each score (events at temporal IoU $\geq$0.3, boxes at IoU $\geq$0.3) are unchanged, so only the bar that a question's mTIoU, \Lthree{} and Edge F1 must pass is raised.
Increasing the grounding thresholds reduces grounded correct rates across systems, and with interactions also required almost no correct answer passes at 0.5 or 0.7.
\method (2B) keeps a lower event-level spurious share than joint SFT at 0.3 and 0.5 (.54 and .81 against .68 and .87), and with interactions also required the two are close at the stricter thresholds (GCR$_{ST}$ .007 and .008 at 0.5).

\section{Benchmark results and failure diagnostics}
\label{app:results}
This appendix gives the evidence behind \cref{sec:exp_main,sec:exp_findings}, with \cref{sec:supp_events,sec:supp_roles,sec:supp_inter,sec:supp_dep} following Findings~1--4 in order.
Its analyses use the predictions of \cref{tab:main}.

\subsection{Benchmark results}
\label{sec:supp_further}
\begin{figure}[tb]\centering
\includegraphics[width=\linewidth]{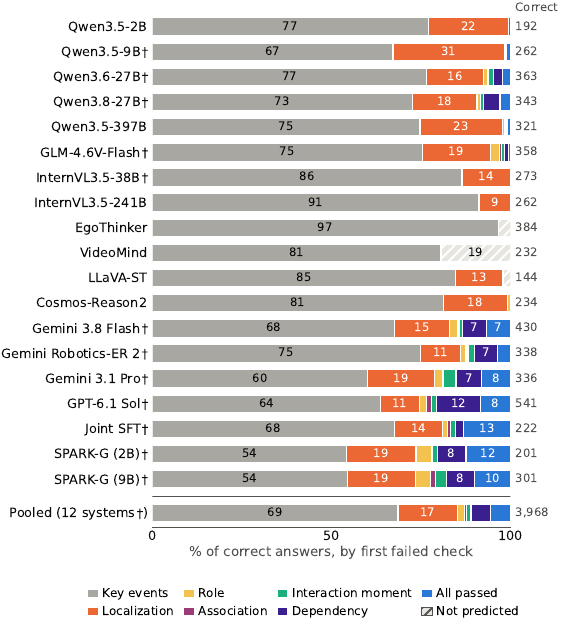}
\caption{\textbf{Correct answers by first failed grounding check} (\%, \EvGold{} gold-labeled questions, outputs of \cref{tab:main}; right: number of correct answers). The checks are applied in order: key events (mTIoU$_i\geq0.3$), participant localization, roles and association (per-question F1 of the nested \Lone{}, \Ltwo{} and \Lthree{} counts $\geq0.3$, box matching of \cref{sec:exp_metrics}), the interaction moment (each counted interaction is reported within the span of the matched ground-truth interaction, the criterion of GCR$_{ST}$) and dependencies (the criterion of GCR$_{STD}$). All passed equals GCR$_{STD}$/Acc. Not predicted: interactions (EgoThinker, VideoMind) or dependencies (LLaVA-ST) are not predicted by the system. Shares below 6\% are not labeled. The figure shows the 19 systems of \cref{tab:main} that give an answer. Pooled: the twelve systems marked $\dagger$, the pool of \cref{sec:exp_main} (GPT-6.1 Sol, Gemini 3.8 Flash, Gemini Robotics-ER 2, Gemini 3.1 Pro, Qwen3.5-9B, Qwen3.6-27B, Qwen3.8-27B, GLM-4.6V-Flash, InternVL3.5 (38B), joint SFT and \method{} (2B, 9B)).}
\label{fig:supp_incremental}
\end{figure}

\paragraph{Grounding checks one at a time.}
\Cref{fig:supp_incremental} applies the grounding checks of \cref{fig:errors} to the correct answers one at a time and splits the interaction check into participant localization, roles, association and the interaction moment.
Each bar gives the share of the correct answers of a system that first fail each check, so the key-event, dependency and all-passed shares are those of \cref{fig:errors}. For example, \AN{GPT}{c_tfail}\% of the correct answers of GPT-6.1 Sol fail on the key events and \AN{GPT}{c_tsd}\% pass all checks.
Pooled over the twelve systems marked $\dagger$ in the figure, which include the three systems trained on our training split (joint SFT and \method{} (2B, 9B)), \FApoolRA{} of the \FApoolLoc{} correct answers whose events and participants are localized fail on roles (\FApoolRole) or association (\FApoolAssoc).
The share is the same for two defined pools: the 16 systems of \cref{tab:main} that predict every evidence type and an answer (100 of 597, 16.8\%), and these systems without the three trained on our training split (69 of 422, 16.4\%).

\paragraph{Results by causal capability.}
\Cref{tab:supp_capability,tab:supp_capability2} split the results of \cref{tab:main} by capability for the twelve pooled systems of \cref{fig:supp_incremental} (324 causal explanation, 278 counterfactual, 115 hypothetical, 309 anticipation, 326 planning and 287 goal attribution questions).
As the capabilities differ in structure (counterfactual questions have 1.88 key events per question against 1.47--1.67 and twice the share of dependency-positive questions, 32\% against 15--17\%), each observed score is compared with its expected value under 5,000 permutations of the capability labels within strata of key-event count (1, 2, 3, $\geq$4) and dependency positivity.
The expected values vary little across the capabilities (\eg, accuracy .198--.214), and $\dagger$ in \cref{tab:supp_capability} marks the scores that depart from them (Holm-adjusted across the six capabilities within each metric).

Answering differs most.
Counterfactual questions are answered best and hypothetical questions worst (accuracy .270 and .116, expected .214 and .202), and GCR$_T$ follows the same order (.084 and .031).
The hypothetical accuracy and GCR$_T$ and the counterfactual accuracy remain significant after an additional Bonferroni correction across the six metrics (adjusted $p\leq.029$).
Over the 16 systems of \cref{tab:main} that predict every evidence type and an answer, with and without the three trained on our training split, hypothetical questions also have the lowest accuracy (.117 and .133) and GCR$_T$ (.027 and .029, Holm-adjusted $p\leq.041$), and counterfactual questions the highest accuracy (.259 and .248).
A counterfactual question supposes that an action shown in the video had gone differently, whereas a hypothetical question supposes an action that the video does not show, which suggests that models answer better when the supposition concerns an observed action.

Evidence recovery differs much less (mTIoU .193--.229 and \Lthree{} .082--.115, against accuracy .116--.270), and the best capability differs by component: planning for key events (mTIoU .229), goal attribution for interactions (\Lthree{} .115) and counterfactual for dependencies (Edge F1 .059, as its many dependency-positive questions predict), while anticipation exceeds its expected Edge F1 (.053 against .038).
In every capability, the interactions and dependencies remain the weakest evidence.

\paragraph{Questions with a dependency.}
For a question without an annotated dependency, the dependency check of GCR$_{STD}$ passes whenever no dependency is predicted, which applies to \ANNoDepQ\% of the gold-labeled questions.
\Cref{sec:exp_main} therefore also reports GCR$_{STD}$ on the 302 dependency-positive questions, those of the 787 with $\geq$2 key events that have an annotated \textit{causes} or \textit{enables} pair (485 are dependency-negative, \cref{tab:supp_deppos}).
There, GCR$_{STD}$ is at most .056 (GPT-6.1 Sol, 17 of 302 questions).

These questions are not harder: pooled over the twelve systems, their accuracy is .269 against .195, their GCR$_T$ .108 against .060 and their mTIoU .241 against .206, and every system answers them more accurately.

The two dependency types differ (\cref{tab:supp_deppos}, right).
Of the 19 annotated \textit{causes} pairs, twelve of the 18 systems of \cref{tab:main} that predict dependencies recover none and the other six at most two (recall $\leq$11\%), and at most 6\% of their predicted \textit{causes} edges are correct, whereas every system recovers some of the 348 \textit{enables} pairs (recall 1--23\%, precision up to 35\%).
With only 19 \textit{causes} pairs, this difference rests on few cases.

\begin{table*}[tb]\centering\scriptsize\setlength{\tabcolsep}{3pt}
\begin{tabular}{@{}l c cc ccc ccc ccc@{}}\toprule
& Answer & \multicolumn{2}{c}{RQ1: Key events} & \multicolumn{3}{c}{RQ2: Interactions} & \multicolumn{3}{c}{RQ3: Dependencies} & \multicolumn{3}{c}{RQ4: Grounded correct $\uparrow$} \\
\cmidrule(lr){2-2}\cmidrule(lr){3-4}\cmidrule(lr){5-7}\cmidrule(lr){8-10}\cmidrule(lr){11-13}
Capability (questions) & Acc & mTIoU & F1@.3 & \Lone & \Ltwo & \Lthree & Macro-F1 & Edge F1 & \depacc & GCR$_T$ & GCR$_{ST}$ & GCR$_{STD}$ \\\midrule
\multicolumn{13}{@{}l}{\textit{Pooled over the twelve systems: observed / expected under stratified label permutation}} \\
\quad Causal explanation (324) & .236/.199$^\dagger$ & .193/.206 & .284/.320$^\dagger$ &  &  & .089/.100 &  & .030/.036 &  & .076/.061$^\dagger$ &  &  \\
\quad Counterfactual (278) & .270/.214$^\dagger$ & .204/.213 & .335/.339 &  &  & .109/.104 &  & .059/.063 &  & .084/.071 &  &  \\
\quad Hypothetical (115) & .116/.202$^\dagger$ & .196/.209 & .321/.327 &  &  & .082/.100 &  & .038/.036 &  & .031/.063$^\dagger$ &  &  \\
\quad Anticipation (309) & .185/.200 & .206/.207 & .332/.322 &  &  & .090/.100 &  & .053/.038$^\dagger$ &  & .059/.063 &  &  \\
\quad Planning (326) & .182/.199 & .229/.207$^\dagger$ & .336/.322 &  &  & .107/.099 &  & .030/.035 &  & .054/.061 &  &  \\
\quad Goal attribution (287) & .172/.198 & .211/.206 & .342/.320 &  &  & .115/.100$^\dagger$ &  & .032/.035 &  & .057/.061 &  &  \\
\midrule
\multicolumn{13}{@{}l}{\textit{Qwen3.5-9B}} \\
\quad Causal explanation (324) & .185 & .211 & .205 & .007 & .005 & .005 & .062 & .011 & .50 & .068 & .003 & .003 \\
\quad Counterfactual (278) & .216 & .197 & .243 & .006 & .004 & .004 & .075 & .025 & .55 & .058 & .004 & .004 \\
\quad Hypothetical (115) & .113 & .185 & .265 & .009 & .002 & .002 & .020 & .000 & .67 & .043 & .000 & .000 \\
\quad Anticipation (309) & .104 & .228 & .256 & .011 & .005 & .005 & .132 & .040 & .59 & .042 & .003 & .003 \\
\quad Planning (326) & .156 & .192 & .199 & .008 & .005 & .005 & .056 & .008 & .50 & .037 & .000 & .000 \\
\quad Goal attribution (287) & .160 & .238 & .269 & .010 & .006 & .006 & .044 & .011 & .35 & .063 & .000 & .000 \\
\midrule
\multicolumn{13}{@{}l}{\textit{Qwen3.6-27B}} \\
\quad Causal explanation (324) & .287 & .212 & .369 & .090 & .076 & .075 & .161 & .042 & .51 & .093 & .025 & .009 \\
\quad Counterfactual (278) & .259 & .214 & .381 & .089 & .075 & .075 & .255 & .059 & .54 & .058 & .007 & .007 \\
\quad Hypothetical (115) & .148 & .187 & .338 & .053 & .040 & .040 & .216 & .043 & .70 & .017 & .000 & .000 \\
\quad Anticipation (309) & .178 & .169 & .305 & .055 & .042 & .041 & .208 & .048 & .64 & .026 & .003 & .003 \\
\quad Planning (326) & .190 & .176 & .281 & .048 & .037 & .036 & .256 & .020 & .58 & .028 & .003 & .000 \\
\quad Goal attribution (287) & .223 & .228 & .404 & .110 & .090 & .087 & .138 & .032 & .49 & .070 & .017 & .007 \\
\midrule
\multicolumn{13}{@{}l}{\textit{Qwen3.8-27B}} \\
\quad Causal explanation (324) & .259 & .195 & .328 & .093 & .079 & .079 & .092 & .029 & .45 & .086 & .018 & .003 \\
\quad Counterfactual (278) & .320 & .219 & .357 & .121 & .100 & .097 & .121 & .062 & .50 & .097 & .032 & .022 \\
\quad Hypothetical (115) & .070 & .190 & .345 & .080 & .072 & .070 & .166 & .044 & .71 & .026 & .000 & .000 \\
\quad Anticipation (309) & .175 & .181 & .305 & .085 & .074 & .071 & .104 & .037 & .53 & .032 & .010 & .006 \\
\quad Planning (326) & .193 & .218 & .339 & .119 & .102 & .098 & .126 & .035 & .54 & .037 & .012 & .003 \\
\quad Goal attribution (287) & .157 & .204 & .368 & .129 & .107 & .105 & .090 & .029 & .41 & .049 & .014 & .000 \\
\midrule
\multicolumn{13}{@{}l}{\textit{GLM-4.6V-Flash}} \\
\quad Causal explanation (324) & .188 & .168 & .284 & .049 & .035 & .033 & .084 & .005 & .44 & .031 & .000 & .000 \\
\quad Counterfactual (278) & .259 & .201 & .343 & .070 & .055 & .052 & .176 & .051 & .64 & .068 & .007 & .000 \\
\quad Hypothetical (115) & .226 & .225 & .380 & .054 & .037 & .035 & .131 & .026 & .46 & .061 & .009 & .000 \\
\quad Anticipation (309) & .210 & .204 & .351 & .048 & .032 & .030 & .108 & .043 & .53 & .052 & .003 & .003 \\
\quad Planning (326) & .132 & .232 & .347 & .069 & .053 & .050 & .186 & .016 & .54 & .052 & .003 & .003 \\
\quad Goal attribution (287) & .317 & .206 & .381 & .079 & .060 & .058 & .151 & .040 & .51 & .066 & .004 & .000 \\
\midrule
\multicolumn{13}{@{}l}{\textit{InternVL3.5-38B}} \\
\quad Causal explanation (324) & .225 & .082 & .146 & .000 & .000 & .000 & .045 & .009 & .58 & .025 & .000 & .000 \\
\quad Counterfactual (278) & .104 & .065 & .166 & .000 & .000 & .000 & .053 & .018 & .93 & .011 & .000 & .000 \\
\quad Hypothetical (115) & .217 & .142 & .265 & .000 & .000 & .000 & .085 & .027 & .70 & .035 & .000 & .000 \\
\quad Anticipation (309) & .181 & .116 & .200 & .000 & .000 & .000 & .138 & .015 & .65 & .029 & .000 & .000 \\
\quad Planning (326) & .117 & .128 & .186 & .001 & .001 & .001 & .057 & .000 & .52 & .025 & .000 & .000 \\
\quad Goal attribution (287) & .181 & .076 & .172 & .000 & .000 & .000 & .040 & .000 & .88 & .017 & .000 & .000 \\
\bottomrule\end{tabular}
\caption{\textbf{Results by causal capability} on the \EvGold{} gold-labeled questions, aggregated over the questions of each capability as in \cref{tab:main} (continued in \cref{tab:supp_capability2}). Top: pooled over the twelve systems of \cref{fig:supp_incremental}, observed / expected value under the stratified permutation of the capability labels (\cref{sec:supp_further}), for the six tested metrics. $^\dagger$Holm-adjusted $p<.05$ over the six capabilities.}
\label{tab:supp_capability}\end{table*}

\begin{table*}[tb]\centering\scriptsize\setlength{\tabcolsep}{3pt}
\begin{tabular}{@{}l c cc ccc ccc ccc@{}}\toprule
& Answer & \multicolumn{2}{c}{RQ1: Key events} & \multicolumn{3}{c}{RQ2: Interactions} & \multicolumn{3}{c}{RQ3: Dependencies} & \multicolumn{3}{c}{RQ4: Grounded correct $\uparrow$} \\
\cmidrule(lr){2-2}\cmidrule(lr){3-4}\cmidrule(lr){5-7}\cmidrule(lr){8-10}\cmidrule(lr){11-13}
Capability (questions) & Acc & mTIoU & F1@.3 & \Lone & \Ltwo & \Lthree & Macro-F1 & Edge F1 & \depacc & GCR$_T$ & GCR$_{ST}$ & GCR$_{STD}$ \\\midrule
\multicolumn{13}{@{}l}{\textit{Gemini 3.8 Flash}} \\
\quad Causal explanation (324) & .312 & .204 & .321 & .141 & .133 & .132 & .132 & .037 & .50 & .102 & .049 & .015 \\
\quad Counterfactual (278) & .299 & .223 & .387 & .175 & .159 & .154 & .148 & .086 & .64 & .112 & .043 & .032 \\
\quad Hypothetical (115) & .148 & .196 & .335 & .142 & .122 & .121 & .063 & .015 & .45 & .035 & .000 & .000 \\
\quad Anticipation (309) & .246 & .203 & .370 & .144 & .124 & .121 & .141 & .058 & .48 & .074 & .023 & .010 \\
\quad Planning (326) & .258 & .224 & .375 & .169 & .154 & .153 & .246 & .045 & .57 & .064 & .028 & .012 \\
\quad Goal attribution (287) & .240 & .219 & .356 & .171 & .159 & .155 & .100 & .043 & .67 & .094 & .049 & .028 \\
\midrule
\multicolumn{13}{@{}l}{\textit{Gemini Robotics-ER 2}} \\
\quad Causal explanation (324) & .201 & .141 & .240 & .074 & .065 & .064 & .070 & .031 & .47 & .031 & .015 & .003 \\
\quad Counterfactual (278) & .299 & .179 & .314 & .095 & .089 & .088 & .124 & .074 & .66 & .075 & .032 & .018 \\
\quad Hypothetical (115) & .078 & .146 & .247 & .076 & .069 & .067 & .010 & .000 & .50 & .017 & .000 & .000 \\
\quad Anticipation (309) & .207 & .165 & .285 & .098 & .091 & .089 & .093 & .049 & .48 & .065 & .019 & .006 \\
\quad Planning (326) & .221 & .178 & .297 & .083 & .071 & .070 & .144 & .023 & .48 & .046 & .009 & .003 \\
\quad Goal attribution (287) & .157 & .169 & .291 & .128 & .115 & .113 & .054 & .025 & .50 & .059 & .038 & .011 \\
\midrule
\multicolumn{13}{@{}l}{\textit{Gemini 3.1 Pro}} \\
\quad Causal explanation (324) & .238 & .214 & .347 & .153 & .144 & .142 & .112 & .051 & .59 & .114 & .052 & .034 \\
\quad Counterfactual (278) & .183 & .215 & .362 & .164 & .145 & .143 & .105 & .055 & .62 & .065 & .025 & .011 \\
\quad Hypothetical (115) & .070 & .185 & .316 & .122 & .115 & .109 & .121 & .064 & .70 & .026 & .000 & .000 \\
\quad Anticipation (309) & .262 & .217 & .374 & .143 & .131 & .127 & .141 & .081 & .63 & .104 & .029 & .013 \\
\quad Planning (326) & .209 & .233 & .349 & .141 & .127 & .123 & .146 & .050 & .62 & .071 & .025 & .009 \\
\quad Goal attribution (287) & .178 & .232 & .378 & .179 & .167 & .164 & .090 & .043 & .48 & .073 & .035 & .021 \\
\midrule
\multicolumn{13}{@{}l}{\textit{GPT-6.1 Sol}} \\
\quad Causal explanation (324) & .361 & .218 & .368 & .221 & .211 & .203 & .253 & .045 & .54 & .139 & .096 & .046 \\
\quad Counterfactual (278) & .392 & .227 & .396 & .237 & .221 & .213 & .295 & .066 & .65 & .133 & .058 & .029 \\
\quad Hypothetical (115) & .183 & .206 & .366 & .183 & .172 & .169 & .186 & .065 & .57 & .061 & .009 & .000 \\
\quad Anticipation (309) & .356 & .233 & .439 & .245 & .231 & .221 & .211 & .079 & .62 & .136 & .078 & .026 \\
\quad Planning (326) & .316 & .231 & .419 & .246 & .231 & .225 & .224 & .062 & .62 & .101 & .052 & .018 \\
\quad Goal attribution (287) & .282 & .224 & .383 & .245 & .230 & .222 & .149 & .032 & .55 & .112 & .080 & .028 \\
\midrule
\multicolumn{13}{@{}l}{\textit{Joint SFT (2B)}} \\
\quad Causal explanation (324) & .182 & .194 & .171 & .107 & .097 & .096 & .064 & .024 & .79 & .062 & .037 & .031 \\
\quad Counterfactual (278) & .295 & .189 & .235 & .147 & .138 & .136 & .086 & .033 & .55 & .086 & .040 & .032 \\
\quad Hypothetical (115) & .052 & .169 & .188 & .121 & .105 & .101 & .000 & .000 & .00 & .009 & .009 & .009 \\
\quad Anticipation (309) & .084 & .194 & .264 & .126 & .112 & .109 & .070 & .056 & .88 & .023 & .006 & .003 \\
\quad Planning (326) & .110 & .302 & .462 & .228 & .205 & .200 & .034 & .031 & .75 & .052 & .021 & .021 \\
\quad Goal attribution (287) & .045 & .212 & .281 & .152 & .141 & .137 & .041 & .015 & .60 & .011 & .004 & .004 \\
\midrule
\multicolumn{13}{@{}l}{\textit{\method{} (2B)}} \\
\quad Causal explanation (324) & .151 & .223 & .372 & .182 & .168 & .162 & .073 & .092 & .53 & .052 & .028 & .015 \\
\quad Counterfactual (278) & .288 & .266 & .446 & .212 & .194 & .193 & .113 & .153 & .56 & .119 & .058 & .036 \\
\quad Hypothetical (115) & .000 & .255 & .415 & .174 & .157 & .155 & .117 & .200 & .71 & .000 & .000 & .000 \\
\quad Anticipation (309) & .074 & .292 & .483 & .203 & .181 & .176 & .165 & .122 & .53 & .052 & .016 & .006 \\
\quad Planning (326) & .123 & .325 & .506 & .243 & .218 & .218 & .080 & .046 & .73 & .064 & .021 & .018 \\
\quad Goal attribution (287) & .031 & .251 & .414 & .193 & .185 & .180 & .097 & .089 & .71 & .017 & .014 & .007 \\
\midrule
\multicolumn{13}{@{}l}{\textit{\method{} (9B)}} \\
\quad Causal explanation (324) & .244 & .255 & .399 & .189 & .176 & .167 & .065 & .083 & .56 & .108 & .052 & .031 \\
\quad Counterfactual (278) & .331 & .257 & .432 & .197 & .184 & .175 & .070 & .096 & .54 & .122 & .032 & .011 \\
\quad Hypothetical (115) & .087 & .260 & .423 & .184 & .167 & .157 & .019 & .000 & .33 & .043 & .009 & .009 \\
\quad Anticipation (309) & .139 & .268 & .437 & .186 & .171 & .165 & .146 & .131 & .53 & .078 & .026 & .013 \\
\quad Planning (326) & .160 & .305 & .483 & .234 & .212 & .210 & .084 & .070 & .62 & .071 & .034 & .015 \\
\quad Goal attribution (287) & .087 & .272 & .407 & .196 & .186 & .181 & .035 & .022 & .71 & .056 & .028 & .024 \\
\bottomrule\end{tabular}
\caption{\textbf{Results by causal capability} (continued from \cref{tab:supp_capability}).}
\label{tab:supp_capability2}\end{table*}

\begin{table*}[tb]\centering\footnotesize\setlength{\tabcolsep}{3pt}
\resizebox{\textwidth}{!}{%
\begin{tabular}{@{}l ccccc cccc cc cc@{}}\toprule
 & \multicolumn{5}{c}{Dependency-positive (302)} & \multicolumn{4}{c}{Dependency-negative (485)} & \multicolumn{2}{c}{Precision (\%)} & \multicolumn{2}{c}{Recall (\%)} \\
\cmidrule(lr){2-6}\cmidrule(lr){7-10}\cmidrule(lr){11-12}\cmidrule(l){13-14}
System & Acc & GCR$_T$ & GCR$_{STD}$ & mTIoU & F1@.3 & Acc & GCR$_T$ & mTIoU & F1@.3 & \textit{causes} & \textit{enables} & \textit{causes} (19) & \textit{enables} (348) \\\midrule
Pooled (twelve systems) & .269 & .108 & .016 & .241 & .390 & .195 & .060 & .206 & .320 & -- & -- & -- & -- \\
\midrule
Qwen3.5-2B & .159 & .033 & .000 & .138 & .133 & .103 & .012 & .140 & .097 & -- & 3 & 0 & 4 \\
Qwen3.5-9B & .232 & .056 & .000 & .206 & .282 & .140 & .039 & .197 & .223 & 0 & 6 & 0 & 7 \\
Qwen3.6-27B & .295 & .099 & .013 & .259 & .434 & .202 & .064 & .222 & .366 & 3 & 12 & 11 & 23 \\
Qwen3.8-27B & .301 & .126 & .010 & .252 & .407 & .196 & .047 & .220 & .350 & 0 & 12 & 0 & 16 \\
Qwen3.5-397B & .248 & .070 & .000 & .247 & .356 & .194 & .058 & .226 & .332 & 2 & 8 & 11 & 15 \\
GLM-4.6V-Flash & .268 & .099 & .003 & .241 & .409 & .237 & .050 & .208 & .351 & 6 & 9 & 5 & 15 \\
InternVL3.5-38B & .225 & .023 & .000 & .105 & .207 & .134 & .021 & .101 & .177 & 0 & 3 & 0 & 3 \\
InternVL3.5-241B & .212 & .023 & .000 & .079 & .117 & .138 & .017 & .085 & .110 & 0 & 4 & 0 & 5 \\
Cosmos-Reason2 & .225 & .040 & .000 & .150 & .281 & .132 & .017 & .121 & .222 & 0 & 5 & 0 & 1 \\
Causal-VLM & -- & -- & -- & .144 & .242 & -- & -- & .129 & .209 & -- & 3 & 0 & 7 \\
MECD+ & -- & -- & -- & .144 & .242 & -- & -- & .129 & .209 & -- & 4 & 0 & 5 \\
Gemini 3.8 Flash & .367 & .172 & .036 & .273 & .447 & .239 & .078 & .226 & .364 & 4 & 16 & 5 & 18 \\
Gemini Robotics-ER 2 & .285 & .106 & .013 & .227 & .374 & .216 & .043 & .171 & .275 & 2 & 13 & 5 & 12 \\
Gemini 3.1 Pro & .242 & .113 & .020 & .255 & .428 & .204 & .078 & .205 & .337 & 0 & 17 & 0 & 16 \\
GPT-6.1 Sol & .387 & .218 & .056 & .279 & .477 & .322 & .142 & .256 & .430 & 5 & 15 & 11 & 22 \\
Joint SFT & .185 & .050 & .010 & .202 & .286 & .134 & .029 & .170 & .224 & 0 & 10 & 0 & 4 \\
\method{} (2B) & .185 & .103 & .023 & .310 & .513 & .117 & .054 & .256 & .411 & 0 & 35 & 0 & 13 \\
\method{} (9B) & .252 & .133 & .010 & .279 & .475 & .192 & .074 & .235 & .401 & 0 & 27 & 0 & 7 \\
\bottomrule\end{tabular}}
\caption{\textbf{Multi-event questions with and without an annotated dependency.} Of the 787 gold-labeled questions with $\geq$2 key events, 302 are dependency-positive ($\geq$1 annotated \textit{causes} or \textit{enables} pair) and 485 dependency-negative. Metrics as in \cref{tab:main}. GCR$_{STD}$ is reported on the dependency-positive questions only, where its dependency check requires a correct edge. Precision and recall of the dependent pairs per type (19 \textit{causes} and 348 \textit{enables} pairs), with the directed evaluator of \cref{tab:main} (--: no edge of the type predicted). Pooled: the twelve systems of \cref{fig:supp_incremental}. MECD+ links the events of Causal-VLM, hence their identical event scores.}
\label{tab:supp_deppos}\end{table*}

\def\TabFullRows{%
\multicolumn{19}{@{}l}{\textit{Open-source MLLMs}} \\
Qwen3.5-2B & 4 & .112 & .143 & .218 & .458 & .112 & .003 & .001 & .001 & .067 & .016 & .54 & .021 & .001 & .000 & .82 & 1.00 & 1.00 \\
Qwen3.5-9B & 4 & .158 & .195 & .295 & .520 & .233 & .007 & .004 & .004 & .059 & .023 & .52 & .043 & .002 & .001 & .73 & .99 & .99 \\
Qwen3.6-27B & 4 & .216 & .189 & .239 & \textbf{.667} & .319 & .058 & .045 & .044 & \textbf{.160} & .040 & \underline{.57} & .050 & .007 & .003 & .77 & .97 & .99 \\
Qwen3.8-27B & 4 & .220 & .198 & .249 & \underline{.632} & .337 & .093 & .079 & .077 & .106 & .045 & .52 & .061 & .013 & .006 & .72 & .94 & .97 \\
Qwen3.5-397B-A17B & 4 & .187 & .192 & .255 & .607 & .289 & .017 & .012 & .011 & .145 & .032 & .54 & .047 & .001 & .001 & .75 & .99 & .99 \\
GLM-4.6V-Flash (9B)~\cite{hong2025glm45v} & 4 & .208 & .196 & .287 & .568 & .337 & .055 & .040 & .038 & .129 & .036 & .56 & .050 & .004 & .002 & .76 & .98 & .99 \\
Gemma 4 (31B)~\cite{gemma2026gemma4} & 4 & .169 & .176 & .277 & .493 & .289 & .046 & .033 & .033 & .137 & .035 & .56 & .041 & .003 & .002 & .76 & .98 & .99 \\
InternVL3.5 (38B)~\cite{wang2025internvl35} & $\leq$112 fr. & .163 & .100 & .216 & .307 & .175 & .001 & .000 & .000 & .050 & .013 & \textbf{.64} & .023 & .000 & .000 & .86 & 1.00 & 1.00 \\
InternVL3.5 (241B-A28B)~\cite{wang2025internvl35} & $\leq$112 fr. & .177 & .076 & .169 & .317 & .110 & .001 & .000 & .000 & .071 & .017 & \underline{.57} & .017 & .000 & .000 & .90 & 1.00 & 1.00 \\
\midrule
\multicolumn{19}{@{}l}{\textit{Video reasoning specialists}} \\
EgoThinker~\cite{egothinker2025} & 1 & .228 & .056 & .157 & .116 & .092 & -- & -- & -- & -- & -- & -- & .011 & -- & -- & .95 & -- & -- \\
Open-o3 Video~\cite{meng2026openo3video} & 16 fr. & .161 & .142 & .390 & .283 & .233 & -- & -- & -- & -- & -- & -- & .028 & -- & -- & .82 & -- & -- \\
VisionCoach~\cite{visioncoach2026} & 16 fr. & .159 & .151 & .391 & .292 & .252 & -- & -- & -- & -- & -- & -- & .032 & -- & -- & .80 & -- & -- \\
VideoChat-R1.5~\cite{videochatr15_2025} & 2 & .119 & .148 & .248 & .389 & .249 & -- & -- & -- & -- & -- & -- & .024 & -- & -- & .80 & -- & -- \\
VideoMind~\cite{videomind2025} & 1 / 2 & .141 & .161 & .294 & .318 & .265 & -- & -- & -- & -- & -- & -- & .028 & -- & -- & .80 & -- & -- \\
LLaVA-ST (7B)~\cite{li2025llavast} & 100 fr. & .101 & .104 & .233 & .173 & .174 & .019 & .013 & .013 & -- & -- & -- & .014 & .001 & -- & .86 & .99 & -- \\
GroundVQA~\cite{di2024groundvqa} & 1.87 & .039 & .121 & .248 & .220 & .206 & -- & -- & -- & -- & -- & -- & .004 & -- & -- & .91 & -- & -- \\
\midrule
\multicolumn{19}{@{}l}{\textit{Causal reasoning / grounding specialists}} \\
Cosmos-Reason2 (8B)~\cite{nvidia2025cosmosreason2} & 4 & .151 & .153 & .282 & .412 & .249 & .008 & .004 & .004 & .020 & .010 & .53 & .031 & .000 & .000 & .79 & 1.00 & 1.00 \\
Causal-VLM~\cite{nadeem2026causalvlm} & 64 fr. & -- & .126 & .184 & .629 & .223 & -- & -- & -- & .042 & .015 & .40 & -- & -- & -- & -- & -- & -- \\
MECD+~\cite{chen2025mecdplus} & 2 & -- & -- & -- & -- & -- & -- & -- & -- & .041 & .015 & .46 & -- & -- & -- & -- & -- & -- \\
\midrule
\multicolumn{19}{@{}l}{\textit{Proprietary models}} \\
Gemini 3.8 Flash & 4 & \underline{.279} & .191 & .342 & .393 & .324 & .124 & .112 & .109 & .107 & .049 & \underline{.57} & .072 & .026 & .012 & .74 & .91 & .96 \\
Gemini Robotics-ER 2 & 4$^\ast$ & .211 & .151 & .295 & .297 & .262 & .073 & .064 & .062 & .075 & .031 & .52 & .049 & .015 & .007 & .77 & .93 & .97 \\
Gemini 3.1 Pro & 4 & .218 & .196 & .345 & .394 & .326 & .121 & .107 & .105 & .079 & .048 & \underline{.57} & .067 & .023 & .014 & .69 & .89 & .94 \\
GPT-6.1 Sol & 4 & \textbf{.324} & .206 & .341 & .458 & .371 & .178 & .162 & .157 & \underline{.148} & .051 & \underline{.57} & \textbf{.092} & \textbf{.045} & \textbf{.021} & .72 & .86 & .94 \\
\midrule
\multicolumn{19}{@{}l}{\textit{Fine-tuned on our training data}} \\
Joint SFT (2B) & 4 & .135 & .206 & \underline{.424} & .316 & .273 & .131 & .118 & .115 & .036 & .022 & \underline{.57} & .041 & .018 & .015 & .69 & .87 & \underline{.89} \\
\method (2B) & 4 / $\leq$8 & .125 & \textbf{.258} & .419 & .427 & \textbf{.428} & \textbf{.186} & \textbf{.166} & \textbf{.163} & .063 & \textbf{.077} & .56 & .057 & .024 & .016 & \textbf{.55} & \textbf{.81} & \textbf{.87} \\
\method (9B) & 4 / $\leq$8 & .178 & \underline{.254} & \textbf{.440} & .413 & \underline{.411} & \underline{.180} & \underline{.164} & \underline{.158} & .044 & \underline{.053} & .53 & \underline{.076} & \underline{.030} & \underline{.019} & \underline{.57} & \underline{.83} & \underline{.89} \\
}

\begin{table*}[tb]
\centering
\footnotesize
\setlength{\tabcolsep}{2.5pt}
\resizebox{\textwidth}{!}{%
\begin{tabular}{@{}l c c cccc ccc ccc ccc ccc@{}}
\toprule
& & Answer & \multicolumn{4}{c}{RQ1: Key events (when)} & \multicolumn{3}{c}{RQ2: Interactions (what/where)} & \multicolumn{3}{c}{RQ3: Dependencies (how)} & \multicolumn{6}{c}{RQ4: Grounded answers} \\
\cmidrule(lr){3-3}\cmidrule(lr){4-7}\cmidrule(lr){8-10}\cmidrule(lr){11-13}\cmidrule(lr){14-19}
& & & & & & & & & & & & & \multicolumn{3}{c}{Grounded correct $\uparrow$} & \multicolumn{3}{c}{Spurious share $\downarrow$} \\
\cmidrule(lr){14-16}\cmidrule(lr){17-19}
System & FPS & Acc & mTIoU & mTIoP & mTIoG & F1@.3 & \Lone & \Ltwo & \Lthree & Macro-F1 & Edge F1 & \depacc & GCR$_T$ & GCR$_{ST}$ & GCR$_{STD}$ & SS$_T$ & SS$_{ST}$ & SS$_{STD}$ \\
\midrule
\TabFullRows
\bottomrule
\end{tabular}}
\caption{\textbf{Main results on all \EvQ{} evaluation questions} (columns as in \cref{tab:supp_main_full}). Evidence is scored against the human-verified labels on the \EvGold{} gold-labeled questions and against the automatic labels on the other \EvSpKeptAuto{} questions. Bold and underline mark the best and second-best results among the systems shown with results. --: not predicted by the system.}
\label{tab:supp_fulltest}
\end{table*}

\paragraph{All evaluation questions.}
\label{sec:supp_fulltest}
\Cref{tab:supp_fulltest} reports the results on all \EvQ{} evaluation questions.
Evidence is scored against the human-verified labels on the \EvGold{} gold-labeled questions and against the automatic labels of stages 1--5 on the other \EvSpKeptAuto{} questions (\cref{sec:supp_split}), and answers are judged as in \cref{tab:main}.

GPT-6.1 Sol answers most accurately (.324) and has the highest GCR$_T$, GCR$_{ST}$ and GCR$_{STD}$ (.092, .045 and .021), so its correct answers are also the most often grounded.
\method{} (2B) localizes the evidence best, with the highest mTIoU (.258), F1@.3 (.428), \Lthree{} (.163) and Edge F1 (.077), followed by \method{} (9B), but answers less accurately (.125 and .178).
Qwen3.6-27B has the highest dependency Macro-F1 (.160).
\method{} thus has the highest mTIoU, F1@.3, \Lthree{} and Edge F1 on these questions whereas GPT-6.1 Sol is best at giving correct answers supported by the evidence.

Compared with the gold-labeled questions (\cref{tab:main}), 88\% of the values of the systems other than the proprietary models change by at most .02.
The proprietary models change more, mainly in the interaction and dependency columns (GPT-6.1 Sol: \Lthree{} .157 against .213, Macro-F1 .148 against .235), while their accuracy changes by at most .017.
The order of the systems is largely kept: over all systems, Kendall $\tau$ between the two orders is .87--.96 on every column except Edge F1 (.81) and \depacc{} (.65).

\raggedbottom  %
\subsection{Finding 1: recovery of multi-event evidence}
\label{sec:supp_events}
Finding~1 makes two claims: (1) key-event recovery degrades with the number of key events, and (2) single-interval outputs explain only part of this loss.
\Cref{sec:supp_events_n,sec:supp_events_out} give the evidence for claims (1) and (2).

\subsubsection{Recovery degrades with the number of key events}
\label{sec:supp_events_n}
The key-event recall of every system decreases as the number of key events required by a question increases.
This decrease is not an artifact of the shorter events and longer videos of these questions: it persists when event duration, video duration and source dataset are held fixed.

\paragraph{Recall by number of key events.}
Recall falls with the number of key events for every system (\cref{tab:supp_eventrec}).
For Qwen3.6-27B, it drops from \ER{Q36}{recall_1}\% on single-event questions to \ER{Q36}{recall_3p}\% on questions with $\geq$3 key events.
Recovering all key events of a question becomes rare: on questions with $\geq$3 key events, no system does so for more than 18.2\% of the questions (GPT-6.1 Sol), and Qwen3.6-27B does so for \ER{Q36}{all_recall_3p}\%.

\begin{table}[!t]\centering\footnotesize\setlength{\tabcolsep}{2.5pt}
\begin{tabular}{@{}lccccc@{}}\toprule
& \multicolumn{3}{c}{Key events per question} & & \\\cmidrule(lr){2-4}
System & 1 (\ERNOne) & 2 (\ERNTwo) & $\geq$3 (\ERNThree) & \#Pred. & Dur. \\\midrule
\TabEventRecRows
\bottomrule\end{tabular}
\caption{\textbf{Recovery of multi-event evidence} (outputs of \cref{tab:main}). Per number of key events of the question (questions in parentheses): recall of the key events (one-to-one matching at tIoU $\geq$0.3) / share of questions whose key events are all recovered (\%). \#Pred.: mean number of predicted events per question. Dur.: mean duration (s) of the predicted events (key events: \ERGTDur\,s).}
\label{tab:supp_eventrec}
\end{table}

\begin{table}[t]\centering\footnotesize\setlength{\tabcolsep}{2.2pt}
\resizebox{\linewidth}{!}{%
\begin{tabular}{@{}l rr rr@{}}\toprule
& \multicolumn{2}{c}{Raw difference} & \multicolumn{2}{c}{Adjusted difference [95\% CI]} \\\cmidrule(lr){2-3}\cmidrule(l){4-5}
System & 2 & $\geq$3 & 2 & $\geq$3 \\\midrule
VideoMind & $-$11.9 & $-$16.7 & $-$6.2 {\scriptsize[$-$9.2, $-$3.2]} & $-$7.1 {\scriptsize[$-$10.7, $-$3.2]} \\
Causal-VLM & $-$7.0 & $-$10.9 & $-$0.4 {\scriptsize[$-$3.7, 2.6]} & 1.8 {\scriptsize[$-$3.1, 6.5]} \\
\midrule
Qwen3.6-27B & $-$12.0 & $-$18.9 & $-$5.1 {\scriptsize[$-$9.0, $-$1.5]} & $-$5.0 {\scriptsize[$-$9.9, $-$0.3]} \\
Qwen3.8-27B & $-$10.8 & $-$22.3 & $-$3.5 {\scriptsize[$-$7.1, 0.2]} & $-$7.6 {\scriptsize[$-$12.5, $-$2.7]} \\
GLM-4.6V-Flash & $-$12.4 & $-$25.1 & $-$6.7 {\scriptsize[$-$10.6, $-$2.9]} & $-$13.9 {\scriptsize[$-$19.0, $-$8.6]} \\
Gemma 4 (31B) & $-$8.3 & $-$15.6 & $-$3.0 {\scriptsize[$-$7.1, 0.9]} & $-$5.3 {\scriptsize[$-$10.8, 0.1]} \\
\midrule
GPT-6.1 Sol & $-$8.6 & $-$15.6 & $-$6.2 {\scriptsize[$-$10.5, $-$2.1]} & $-$9.6 {\scriptsize[$-$15.5, $-$3.9]} \\
Gemini 3.8 Flash & $-$8.3 & $-$16.2 & $-$5.4 {\scriptsize[$-$9.6, $-$1.4]} & $-$10.1 {\scriptsize[$-$15.6, $-$4.4]} \\
Gemini Robotics-ER 2 & $-$5.8 & $-$8.9 & $-$2.3 {\scriptsize[$-$6.1, 1.7]} & $-$2.0 {\scriptsize[$-$7.1, 3.5]} \\
Gemini 3.1 Pro & $-$10.5 & $-$18.2 & $-$6.0 {\scriptsize[$-$10.3, $-$1.8]} & $-$9.0 {\scriptsize[$-$14.2, $-$3.9]} \\
\midrule
Joint SFT (2B) & $-$9.8 & $-$15.6 & $-$7.9 {\scriptsize[$-$12.1, $-$4.1]} & $-$12.9 {\scriptsize[$-$18.0, $-$7.8]} \\
\method{} (2B) & $-$12.2 & $-$21.8 & $-$7.6 {\scriptsize[$-$11.6, $-$3.6]} & $-$12.8 {\scriptsize[$-$17.8, $-$7.6]} \\
\bottomrule\end{tabular}}
\caption{\textbf{Key-event recall by number of key events, adjusted for duration} (points against single-event questions, 2,605 key events of the \EvGold{} gold-labeled questions). Raw: difference of the recall of \cref{tab:supp_eventrec}. Adjusted: average change in the probability that a key event is matched, from a per-system logistic regression of the match on the number of key events of the question (1, 2, $\geq$3), log key-event duration, log video duration and the source dataset. 95\% CIs from 2,000 bootstrap resamples of questions. Per-system associations, not corrected for multiple comparisons.}
\label{tab:supp_eventctrl}
\end{table}

\paragraph{Event and video duration.}
The decrease persists when the shorter key events and longer videos of questions with more key events are held fixed.
Questions with more key events have shorter key events (mean 5.4, 4.4 and 4.0\,s for 1, 2 and $\geq$3 key events) and longer videos (41.9, 45.0 and 54.0\,s), and short events are harder to match at tIoU $\geq$0.3, so the decrease could reflect duration rather than the number of key events.
Within each tertile of key-event duration (cut at 2.4 and 4.9\,s), however, the recall of Qwen3.6-27B still falls with the number of key events (short events: 39, 26 and 24\%, middle: 68, 60 and 55\%, long: 86, 80 and 77\%).
Adjusting for event duration, video duration and source dataset (\cref{tab:supp_eventctrl}), the difference between 1 and $\geq$3 key events is $-5.0$ points [$-9.9$, $-0.3$] for Qwen3.6-27B, against a raw difference of $-18.9$ points.
For 10 of the other 11 systems the adjusted difference at $\geq$3 key events also stays negative, and its 95\% CI excludes zero for 8 of them.

\subsubsection{Single-interval outputs explain only part of the loss}
\label{sec:supp_events_out}
Returning a single interval limits recall on multi-event questions, but it explains only part of the loss.
The systems that return one interval recover far fewer key events than this limit allows, and recall also declines for a system whose number of outputs sets no limit.

\paragraph{Video reasoning specialists.}
The specialists that return one interval are limited mainly by where they place it, not by returning only one.
With one-to-one matching, the highest recall that a system's number of predicted events allows is $\sum\min(n_{\text{pred}},n_{\text{GT}})/\sum n_{\text{GT}}$, so one interval recovers at most half of the key events of a two-event question.
The video reasoning specialists (VideoMind, Open-o3 Video, VisionCoach, VideoChat-R1.5, LLaVA-ST, GroundVQA and EgoThinker) return 0.9--1.3 intervals per question (\cref{tab:supp_eventrec}), which allows a recall of 40--53\% on the 787 multi-event questions, but they recover only 6--20\%.
They also miss most events of single-event questions, where one interval suffices: VideoMind, the best of them, finds \ER{VideoMind}{recall_1}\% of these events, against \ER{Q36}{recall_1}\% for Qwen3.6-27B.

\paragraph{Systems with enough outputs.}
The decline of \cref{sec:supp_events_n} also appears when the number of outputs sets no limit.
Qwen3.6-27B predicts 3.8 events per question, which allows a recall of 97\% on the multi-event questions, and recovers 54\%.
On the questions where it predicts at least as many events as there are key events, its recall still falls with the number of key events (67.4, 56.0 and 49.5\% for 1, 2 and $\geq$3 key events).

\subsection{Finding 2: role confusion after correct localization}
\label{sec:supp_roles}
Finding~2 makes two claims: (1) role errors persist after localization, and (2) role errors persist when localization is given.
\Cref{sec:supp_roles_out,sec:supp_roles_given} give the evidence for claims (1) and (2).
Each analysis conditions on correctly localized participants, so that role errors are separated from localization failures.

\subsubsection{Role errors persist after localization}
\label{sec:supp_roles_out}
Every system that outputs participants gives wrong roles to some of the tools and targets it localizes correctly, errors that scoring the boxes alone (e.g., \Lone{}) would not reveal.

\paragraph{Role errors in the outputs.}
Correct localization does not ensure a correct role.
Among the ground-truth tools and targets that a system localizes at box IoU $\geq$0.3, the share given another role ranges from 8.4\% (\method{} (9B)) to 37.3\% (GLM-4.6V-Flash) over the systems with \Lone{} $\geq$0.05 (\cref{tab:supp_where_all}).
Qwen3.6-27B gives another role to 25.5\% of 542 such objects and GLM-4.6V-Flash to 37.3\% of 426.

\begin{table}[tbp]
\centering\footnotesize\setlength{\tabcolsep}{2.1pt}
\resizebox{\linewidth}{!}{%
\begin{tabular}{@{}lcccc cc@{}}\toprule
& \multicolumn{4}{c}{Participant recovery} & Role & Assoc. \\\cmidrule(lr){2-5}
System & Hand & Tool & Target & All & error & error \\\midrule
Qwen3.6-27B & 10.3 / 9.4 & 16.6 / 12.9 & 13.6 / 10.1 & 12.2 / 9.9 & 25.5 & 7.2 \\
Qwen3.8-27B & 14.2 / 12.7 & 22.6 / 18.7 & 18.2 / 14.7 & 16.5 / 14.0 & 18.6 & 4.7 \\
GLM-4.6V-Flash (9B) & 10.4 / 9.1 & 13.3 / 7.5 & 10.7 / 6.8 & 10.7 / 8.0 & 37.3 & 7.4 \\
Gemma 4 (31B) & 5.3 / 4.2 & 13.6 / 8.8 & 9.7 / 7.2 & 7.8 / 5.8 & 27.6 & 3.1 \\
\midrule
Gemini 3.8 Flash & \underline{18.2} / \underline{17.1} & \underline{28.9} / \underline{26.2} & 19.9 / 17.3 & \underline{19.7} / \underline{17.9} & 12.4 & 5.4 \\
Gemini Robotics-ER 2 & 9.4 / 9.0 & 15.1 / 13.9 & 10.9 / 9.3 & 10.5 / 9.5 & 13.9 & 5.2 \\
Gemini 3.1 Pro & 16.9 / 16.1 & 25.0 / 23.2 & 18.1 / 15.6 & 18.0 / 16.5 & 12.3 & 5.5 \\
GPT-6.1 Sol & \textbf{30.5} / \textbf{29.8} & \textbf{42.5} / \textbf{37.1} & \textbf{30.5} / \textbf{27.7} & \textbf{31.4} / \textbf{29.5} & 9.7 & 5.0 \\
\midrule
Joint SFT (2B) & 15.9 / 15.3 & 22.2 / 18.6 & 16.9 / 14.5 & 16.8 / 15.3 & 14.4 & 2.3 \\
\method{} (2B) & 17.8 / 16.8 & 20.9 / 17.6 & 20.5 / 18.4 & 19.2 / 17.5 & 11.2 & 3.7 \\
\method{} (9B) & 16.6 / 15.6 & 25.2 / 20.4 & \underline{21.5} / \underline{20.2} & 19.3 / \underline{17.9} & 8.4 & 3.5 \\
\bottomrule\end{tabular}}
\caption{\textbf{Participant recovery by role and conditional error rates} (\%) for all systems with \Lone{} $\geq$0.05 in \cref{tab:main}. The other systems localize almost no participants. Each cell gives $\mathrm{TP}_1^r/|\mathcal G_r|$ / $\mathrm{TP}_2^r/|\mathcal G_r|$ for the ground-truth participants of role $r$: the share localized (box IoU $\geq$0.3) and the share localized with role $r$. Role error: localized ground-truth tools and targets given another role. Association error: participants with the correct box and role, at timestamps where $\geq$2 ground-truth interactions are visible, placed in another interaction (unit: participant, see \cref{sec:supp_inter}).}
\label{tab:supp_where_all}\label{tab:supp_inter}
\end{table}

\subsubsection{Role errors persist when localization is given}
\label{sec:supp_roles_given}
Given the verified boxes, the models still assign wrong roles to many tools and targets, and almost all of these errors swap tool and target.
The errors remain when the frame of the interaction is also given, so they arise in the role decision itself rather than in finding the right moment.

\paragraph{Roles given the boxes.}
Given the verified boxes, both models still assign many tools and targets the wrong role, almost always by swapping the two.
In $V{+}\mathcal{E}{+}B$ (\cref{app:oracle}), the models receive the verified key events and participant boxes with their track IDs but not their roles, and we score the roles of the tools and targets that appear in their predicted interactions (3,056 for Qwen3.6-27B and 3,040 for GLM-4.6V-Flash). With hands included, their predicted interactions contain 89\% of the 6,840 ground-truth participants (\cref{sec:supp_oracle}).
Qwen3.6-27B assigns the correct role to .835 of them (95\% CI [.819, .851]) and GLM-4.6V-Flash to .722 ([.701, .740]), with CIs from 2,000 bootstrap resamples of source recordings.
Nearly all of their errors swap tool and target (502 of 503 and 841 of 845), mostly by calling a target a tool.
These accuracies are lower than the .86 that a rule labeling every object a target would reach, since targets outnumber tools by about six to one.
The models do distinguish the two roles, with a balanced accuracy of .862 and .745 against .500 for that rule, but still confuse them often.

\providecolor{aHand}{HTML}{D7301F}\providecolor{aTool}{HTML}{E08214}\providecolor{aTarg}{HTML}{2166AC}
\begin{figure*}[t]\centering\footnotesize
\begin{minipage}[t]{0.485\textwidth}\centering
\includegraphics[width=\linewidth]{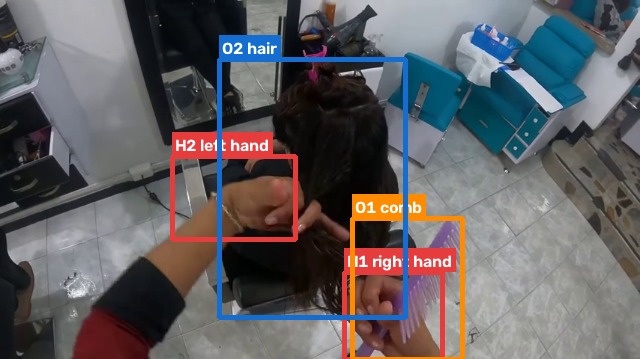}\\[3pt]
\begin{minipage}{0.97\linewidth}\raggedright
\textbf{(a) Merged} (Qwen3.8-27B): the left hand (H2) separates a section of the hair (O2), and the right hand (H1) combs it with the comb (O1).\\[2pt]
{\ttfamily\fontsize{6.2}{7.3}\selectfont \textbf{Ground truth}: "interactions": [\\\ \ \{"id": "I1", "boxes": [\\\ \ \ \ \{"role": \textcolor{aHand}{"hand"}, "track": "H2"\},\\\ \ \ \ \{"role": \textcolor{aTarg}{"target"}, "track": "O2"\}]\},\\\ \ \{"id": "I2", "boxes": [\\\ \ \ \ \{"role": \textcolor{aHand}{"hand"}, "track": "H1"\},\\\ \ \ \ \{"role": \textcolor{aTool}{"tool"}, "track": "O1"\},\\\ \ \ \ \{"role": \textcolor{aTarg}{"target"}, "track": "O2"\}]\}]}\\[2pt]
{\ttfamily\fontsize{6.2}{7.3}\selectfont \textbf{Model}: "interactions": [\\\ \ \{"id": "I1", "boxes": [\\\ \ \ \ \{"role": \textcolor{aHand}{"hand"}, "track": "H2"\},\\\ \ \ \ \{"role": \textcolor{aHand}{"hand"}, "track": "H1"\},\\\ \ \ \ \{"role": \textcolor{aTool}{"tool"}, "track": "O1"\},\\\ \ \ \ \{"role": \textcolor{aTarg}{"target"}, "track": "O2"\}]\}]}
\end{minipage}
\end{minipage}\hfill
\begin{minipage}[t]{0.485\textwidth}\centering
\includegraphics[width=\linewidth]{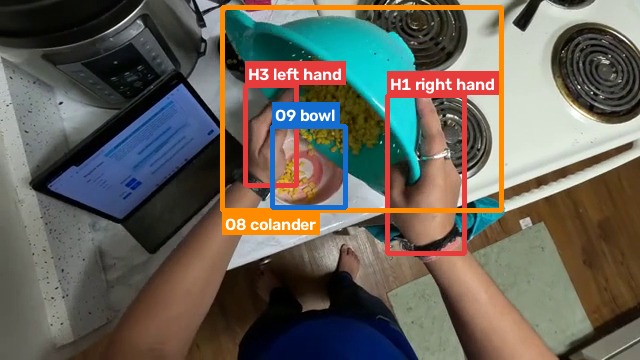}\\[3pt]
\begin{minipage}{0.97\linewidth}\raggedright
\textbf{(b) Omitted shared participant} (GPT-6.1 Sol): the right hand (H1) lifts the colander (O8), and both hands pour the corn from it into the bowl (O9).\\[2pt]
{\ttfamily\fontsize{6.2}{7.3}\selectfont \textbf{Ground truth}: "interactions": [\\\ \ \{"id": "I1", "boxes": [\\\ \ \ \ \{"role": \textcolor{aHand}{"hand"}, "track": "H1"\},\\\ \ \ \ \{"role": \textcolor{aTarg}{"target"}, "track": "O8"\}]\},\\\ \ \{"id": "I2", "boxes": [\\\ \ \ \ \{"role": \textcolor{aHand}{"hand"}, "track": "H1"\},\\\ \ \ \ \{"role": \textcolor{aHand}{"hand"}, "track": "H3"\},\\\ \ \ \ \{"role": \textcolor{aTool}{"tool"}, "track": "O8"\},\\\ \ \ \ \{"role": \textcolor{aTarg}{"target"}, "track": "O9"\}]\}]}\\[2pt]
{\ttfamily\fontsize{6.2}{7.3}\selectfont \textbf{Model}: "interactions": [\\\ \ \{"id": "I1", "boxes": [\\\ \ \ \ \{"role": \textcolor{aHand}{"hand"}, "track": "H3"\},\\\ \ \ \ \{"role": \textcolor{aTarg}{"target"}, "track": "O9"\}]\}]}
\end{minipage}
\end{minipage}
\caption{\textbf{Association errors at moments with two ground-truth interactions}, from the outputs of \cref{tab:main} counted as wrong in \cref{fig:supp_assoc_groups}, written in the output format of \cref{fig:teaser_json} without box coordinates. Boxes in the frames: the verified participants with their track identifiers (hand red, tool orange, target blue), and every box of the model's interaction matches one of them with the same role. (a) The model joins the participants of both interactions in one. (b) The model omits H1 and O8, which also belong to the other interaction.}
\label{fig:supp_assoc_ex}
\end{figure*}

\paragraph{Role probe.}
Role errors remain even when the moment of the interaction is fixed, so they do not come only from searching the video.
\Cref{tab:supp_probes} gives one frame per ground-truth interaction, the annotated time at which most of its participants have a box (\ORNRoleFrames{} distinct frames), with the verified boxes (\ORNRole{} participants), with or without the interactions, instead of the video and the events.
With the frame and the interactions, Qwen3.6-27B gives the correct role to \OR{Q36}{roleprobe}{accobj}\% of the tools and targets and GLM-4.6V-Flash to \OR{GLM46VF}{roleprobe}{accobj}\%, and \OR{Q36}{roleprobe}{ttconf}\% and \OR{GLM46VF}{roleprobe}{ttconf}\% receive the other object role.
Without the frame, their accuracy on tools and targets is \OR{Q36}{roleprobeText}{accobj}\% and \OR{GLM46VF}{roleprobeText}{accobj}\%, so the models use the frame for their role decisions, but it does not remove the confusion.

\begin{table}[!h]\centering\footnotesize\setlength{\tabcolsep}{2.5pt}
\resizebox{\linewidth}{!}{%
\begin{tabular}{@{}l ccc ccc@{}}\toprule
& \multicolumn{3}{c}{Qwen3.6-27B} & \multicolumn{3}{c}{GLM-4.6V-Flash} \\\cmidrule(lr){2-4}\cmidrule(l){5-7}
Given & All & Tools, targets & Swap & All & Tools, targets & Swap \\\midrule
boxes, interactions, frame & \OR{Q36}{roleprobe}{acc} & \OR{Q36}{roleprobe}{accobj} & \OR{Q36}{roleprobe}{ttconf} & \OR{GLM46VF}{roleprobe}{acc} & \OR{GLM46VF}{roleprobe}{accobj} & \OR{GLM46VF}{roleprobe}{ttconf} \\
boxes, frame & \OR{Q36}{roleNA}{acc} & \OR{Q36}{roleNA}{accobj} & \OR{Q36}{roleNA}{ttconf} & \OR{GLM46VF}{roleNA}{acc} & \OR{GLM46VF}{roleNA}{accobj} & \OR{GLM46VF}{roleNA}{ttconf} \\
boxes, interactions & \OR{Q36}{roleprobeText}{acc} & \OR{Q36}{roleprobeText}{accobj} & \OR{Q36}{roleprobeText}{ttconf} & \OR{GLM46VF}{roleprobeText}{acc} & \OR{GLM46VF}{roleprobeText}{accobj} & \OR{GLM46VF}{roleprobeText}{ttconf} \\
boxes & \OR{Q36}{roleNAText}{acc} & \OR{Q36}{roleNAText}{accobj} & \OR{Q36}{roleNAText}{ttconf} & \OR{GLM46VF}{roleNAText}{acc} & \OR{GLM46VF}{roleNAText}{accobj} & \OR{GLM46VF}{roleNAText}{ttconf} \\
\bottomrule\end{tabular}}
\caption{\textbf{Role probe} (\%, \ORNRole{} participants, \ORNObj{} tools and targets). All: all participants. Swap: tools and targets given the other object role.}
\label{tab:supp_probes}
\end{table}

\subsection{Finding 3: association errors with correct participants and roles}
\label{sec:supp_inter}
Finding~3 makes two claims: (1) association errors remain when boxes and roles are correct, and (2) association errors persist given verified participants and roles.
\Cref{sec:supp_inter_out,sec:supp_inter_given} give the evidence for claims (1) and (2).
Each analysis evaluates participants whose boxes and roles are correct or provided, so that association errors are separated from localization and role errors.

\subsubsection{Association errors remain when boxes and roles are correct}
\label{sec:supp_inter_out}
About a quarter of the interactions that systems report with correct boxes and roles at moments with several interactions associate the participants wrongly, mostly by merging participants of different manipulations or omitting a participant shared with another interaction.

\begin{figure}[t]\centering
\includegraphics[width=\linewidth]{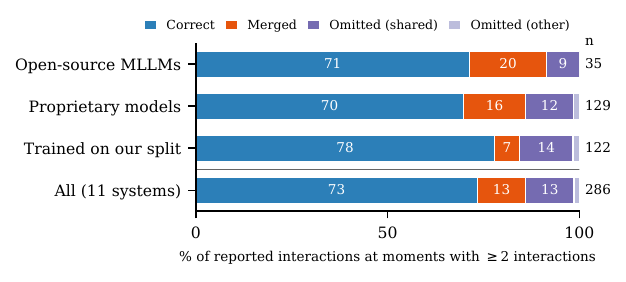}
\caption{\textbf{Association at the reported moments} by system group (\%, unit: reported interaction at a moment with $\geq$2 ground-truth interactions whose boxes all match ground-truth participants with their roles, $n$ on the right, systems of \cref{tab:supp_where_all}). Merged: participants of two or more ground-truth interactions. Omitted: a strict subset of one ground-truth interaction, shared when the missing participant also takes part in another ground-truth interaction at that moment.}
\label{fig:supp_assoc_groups}
\end{figure}

\paragraph{Association errors in the reported interactions.}
Pooled over eleven systems, 76 of the 286 interactions reported with correct boxes and roles at moments with several ground-truth interactions are wrong (26.6\%).
We pool the outputs of the eleven systems with \Lone{} $\geq$0.05 (\cref{tab:supp_where_all}) and keep the reported interactions of matched events in which every box matches, one-to-one at IoU $\geq$0.3, a ground-truth participant that is present at the reported moment and has the same role.
Their association is correct when the matched participants are exactly those of one ground-truth interaction at that moment.
Among these 286 interactions, the 95\% CI of the error rate is [20.1, 33.7] (10,000 bootstrap resamples of the 140 questions), and of the 76 wrong ones, 36 merge participants of different interactions and 40 omit a participant, in 36 cases one shared with another interaction (examples in \cref{fig:supp_assoc_ex}, breakdown by system group in \cref{fig:supp_assoc_groups}).
The rate is similar across system groups (28.6\% for the open-source MLLMs, 30.2\% for the proprietary models and 22.1\% for the systems trained on our training split) and stays within 25.8--28.2\% when any one system is left out, although GPT-6.1 Sol contributes the most interactions (76).
At the participant level, which counts every participant with a correct box and role at all timestamps with $\geq$2 visible ground-truth interactions, 2.3--7.4\% are placed in another interaction (Assoc.\ error in \cref{tab:supp_where_all}).

\subsubsection{Association errors persist given verified participants and roles}
\label{sec:supp_inter_given}
Even when the verified participants and roles are given, the models group them poorly.
Their grouping is worse than a simple rule that attaches each object to the hand whose box overlaps it most, and in single frames it is no more accurate than putting all participants into one interaction.

\paragraph{Given the verified participants and roles.}
Given only the grouping to decide, both models score below a simple rule.
In $V{+}\mathcal{E}{+}B{+}R$, the models receive the video, the verified key events and the verified participant boxes with their roles at each annotated moment, so only the grouping within a moment is left to them.
As a reference, we use a rule without learning: each hand starts its own interaction, and each tool or target joins the hand whose box overlaps it most.
Finding~3 (ii) reports an \Lthree{} of \ST{Q36}{EBR}{intmulti} (Qwen3.6-27B) and \ST{GLM46VF}{EBR}{intmulti} (GLM-4.6V-Flash) at the moments with multiple interactions, below the \STRule{} that this rule reaches on all \STNConcMom{} such moments (202 questions).
As in \cref{tab:probes}, only the given participants that appear in the predicted interactions are scored, so a moment counts when at least two of them appear (195 and 191 of the \STNConcMom{} moments), and the predicted interactions are compared with the ground-truth interactions over these participants.
Applied to the same participants, the rule still scores higher (.395 and .397).
The models thus group the participants worse than attaching each object to its most overlapping hand, so the failure lies in the grouping itself even when all other evidence is given.

\paragraph{Association probe.}
The models also fail to group the participants when they are shown the exact moment of the interactions.
In $V{+}\mathcal{E}{+}B{+}R$, the models receive the participants of each annotated moment but must locate that moment in the whole video to see them, so a wrong grouping could also come from failing to find the right frame.
The probe removes this step: it gives one frame at each of \ORNGrp{} moments with at least two ground-truth interactions (the middle frame of each run of consecutive annotated times with the same interactions), together with the verified boxes and roles, and asks the model to group the boxes into interactions.
The models decide whether two participants belong to the same interaction correctly for \OR{Q36}{grpprobe}{pair}\% (Qwen3.6-27B) and \OR{GLM46VF}{grpprobe}{pair}\% (GLM-4.6V-Flash) of the pairs, below the \ORGrpBase\% of putting all participants into one interaction.
Grouping the participants thus remains difficult even when the moment is given, so the errors of $V{+}\mathcal{E}{+}B{+}R$ are not explained by locating the moment alone.

\FloatBarrier  %
\subsection{Finding 4: dependency recovery}
\label{sec:supp_dep}
Finding~4 makes three claims: (1) missing and coarse events limit dependency recovery, (2) dependency predictions do not separate dependent from dependency-negative pairs, and (3) shared tools or targets are associated with false-positive dependencies.
\Cref{sec:supp_dep_events,sec:supp_dep_disc,sec:supp_dep_shared} give the evidence for claims (1)--(3).

\begin{table*}[tb]
\centering
\footnotesize
\setlength{\tabcolsep}{2.4pt}
\begin{minipage}[t]{0.31\textwidth}\centering
\textbf{(a)} Event granularity\\[2pt]
\begin{tabular}{@{}lcccc@{}}
\toprule
Model & Merg. & Sep. & Miss. & Avg dur. \\
 & (\%) & (\%) & (\%) & (s) \\
\midrule
Causal-VLM & 53 & 34 & 12 & 17.5 \\
Cosmos-Reason2 & 41 & 12 & 47 & 14.2 \\
Qwen3.6-27B & 36 & 55 & 9 & 9.0 \\
\method{} (2B) & 18 & 25 & 57 & 6.1 \\
Ground truth & & & & 4.5 \\
\bottomrule
\end{tabular}
\end{minipage}\hfill
\begin{minipage}[t]{0.33\textwidth}\centering
\textbf{(b)} Edge rate by dependency\\[2pt]
\begin{tabular}{@{}lrccc@{}}
\toprule
Model & Pair cov. & TPR$\uparrow$ & FPR$\downarrow$ & BAcc \\
 & (\%) & (\%) & (\%) & (\%) \\
\midrule
MECD+$^\dagger$ & -- & 78 & 76 & 51$^\circ$ \\
Causal-VLM & 7 & 94 & 94 & 50$^\circ$ \\
Cosmos-Reason2 & 4 & 28 & 40 & 44$^\circ$ \\
Qwen3.6-27B & 28 & 73 & 50 & 61 \\
GLM-4.6V-Flash & 21 & 68 & 50 & 59 \\
Joint SFT & 7 & 39 & 21 & 59$^\circ$ \\
\method{} (2B) & 12 & 69 & 46 & 62 \\
\bottomrule
\end{tabular}
\end{minipage}\hfill
\begin{minipage}[t]{0.34\textwidth}\centering
\textbf{(c)} Edge rate on dependency-negative pairs\\[2pt]
\setlength{\tabcolsep}{2pt}
\begin{tabular}{@{}lcccc@{}}
\toprule
Model & \multicolumn{2}{c}{Adjacent (\%)} & \multicolumn{2}{c}{Shared obj. (\%)} \\
 & yes & no & yes & no \\
\midrule
MECD+$^\dagger$ & 85 & 66 & 78 & 75 \\   %
Cosmos-Reason2$^\ddagger$ & 71 & 63 & 81 & 58 \\
Qwen3.6-27B$^\ddagger$ & 64 & 54 & 70 & 54 \\
GLM-4.6V-Flash$^\ddagger$ & 82 & 86 & 94 & 73 \\
\bottomrule
\end{tabular}
\end{minipage}

\caption{\textbf{Diagnosing dependency recovery.} (a)~Each event of the 367 ground-truth dependent pairs is assigned the predicted interval that overlaps it longest, so two events can share one interval, unlike the one-to-one matching of \cref{tab:main}. Merg./Sep.: both events in the same / in different intervals. Miss.: an event overlaps no interval. Each row sums to 100\%. Avg dur.: mean duration of the predicted events of the questions with $\geq$2 key events (ground truth: events of the dependent pairs). (b)~Pair cov.: share of all 1,180 ground-truth event pairs (dependent and dependency-negative) whose two events the system localizes (tIoU $\geq$0.3), on which the rates are computed. TPR, FPR: rate of predicted edges on dependent (\textit{causes} or \textit{enables}) and on dependency-negative pairs. BAcc (\cref{sec:exp_metrics}): 50 is chance, and $^\circ$ marks a 95\% CI that includes 50. The pairs differ across systems, so BAcc does not rank them. (c)~Rate of predicted edges on dependency-negative pairs, computed separately within each group: adjacent or not, and sharing a tool or target or not. Raw rates, the effects controlled for the other pair features are in \cref{sec:supp_dep}. $^\dagger$Given the ground-truth events (questions with $\geq$3 key events). $^\ddagger$Given the ground-truth events as captions, with the video. Cosmos-Reason2 returns a parsed label for 88.5\% of these pairs, and its rates are computed on those pairs.}
\label{tab:supp_dep_diag}\label{tab:supp_e1}
\end{table*}

\begin{figure}[tbp]
\centering
\includegraphics[width=\linewidth]{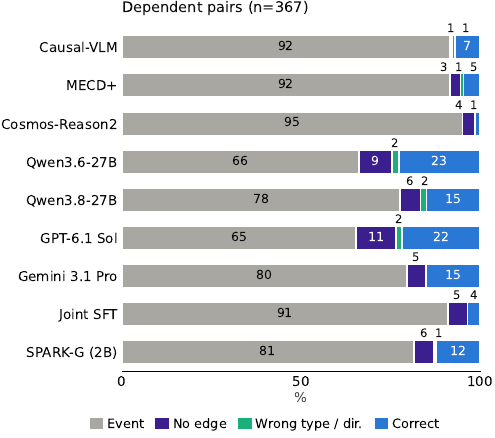}
\caption{\textbf{Ground-truth dependent pairs by the first failed check} (\%, the 367 dependent pairs, directed evaluator of \cref{tab:main}): an event not matched (tIoU $<0.3$), no edge, wrong type or direction, or correct.}
\label{fig:supp_errors_deps}
\end{figure}

\paragraph{Pairs and scoring.}
The diagnostics use the 1,180 ordered pairs of key events of the 787 gold-labeled questions with at least two key events (19 \textit{causes}, 348 \textit{enables} and 813 \textit{before}).
A pair is dependent if it is labeled \textit{causes} or \textit{enables}, two events are adjacent if no other key event lies between them, and two events share an object if a tool or target track appears in both (hands are not counted).
A system that predicts the events from the video is scored on the pairs whose two events it matches (tIoU $\geq$0.3, one-to-one, as in \cref{tab:main}), and MECD+ is given the ground-truth events of the 148 questions with at least three key events (541 pairs).
The pair-level rates (TPR, FPR, BAcc) count an edge in either direction and of either type.

\subsubsection{Missing and coarse events limit dependency recovery}
\label{sec:supp_dep_events}
Most dependent pairs are lost before any edge is judged, because an event of the pair is not recovered, and coarse predicted events often cover both events of a pair at once.

\paragraph{Event recovery.}
For every system, most dependent pairs fail because an event of the pair is not matched.
\Cref{fig:supp_errors_deps} splits the 367 dependent pairs by the first failed check, and for Qwen3.6-27B, 243 of them (66.2\%) fail at the events.

\paragraph{Event granularity.}
Coarse predicted events place both events of a dependent pair in one interval.
Causal-VLM predicts events of 17.5\,s on average, against 4.5\,s for the key events of dependent pairs, and places both events of 53\% of the dependent pairs in one interval (\cref{tab:supp_dep_diag}a).
This panel assigns each event to the predicted interval that overlaps it longest, which needs only overlap, so an interval that covers an event loosely counts here but not in the one-to-one matching of \cref{fig:supp_errors_deps}.
A merged pair means that one predicted interval covers both events, not that the model treats the two manipulations as one.

\begin{table}[tbp]\centering\footnotesize\setlength{\tabcolsep}{2.4pt}
\resizebox{\linewidth}{!}{%
\begin{tabular}{@{}l l c ccc@{}}\toprule
& Input & Dep.\ pairs scored & TPR & FPR & BAcc [95\% CI] \\\midrule
Qwen3.6-27B & video ($V$) & 124 (33.8\%) & .726 & .502 & .612 {\scriptsize[.560, .663]} \\
 & + verified key events ($V{+}\mathcal{E}$) & 361 (98.4\%) & .911 & .744 & .584 {\scriptsize[.562, .605]} \\
 & \multicolumn{2}{l}{both, same pairs (122 dep., 210 dep.-neg.)} & & & .611 $\rightarrow$ .577, $-$.034 {\scriptsize[$-$.094, .026]} \\
\midrule
GLM-4.6V-Flash & video ($V$) & 87 (23.7\%) & .678 & .500 & .589 {\scriptsize[.527, .650]} \\
 & + verified key events ($V{+}\mathcal{E}$) & 365 (99.5\%) & .800 & .637 & .581 {\scriptsize[.554, .607]} \\
 & \multicolumn{2}{l}{both, same pairs (87 dep., 165 dep.-neg.)} & & & .588 $\rightarrow$ .578, $-$.010 {\scriptsize[$-$.085, .063]} \\
\bottomrule\end{tabular}}
\caption{\textbf{Dependency discrimination from the video only and with the verified key events}, one scorer for both rows: of the 1,180 pairs, those whose two events appear in the model's predicted dependencies (video only: events matched at tIoU $\geq$0.3). Dep.\ pairs scored: of the 367 dependent pairs. TPR / FPR: rate of predicted edges (either direction) on the dependent / dependency-negative pairs scored. Same pairs: pairs scored in both conditions, BAcc from the video only $\rightarrow$ with the verified key events and the paired difference. CIs: 2,000 bootstrap resamples of questions.}
\label{tab:supp_depgiven}
\end{table}

\begin{table*}[tb]\centering\scriptsize\setlength{\tabcolsep}{2.5pt}
\begin{minipage}[t]{0.24\textwidth}\centering
{\footnotesize\textbf{(a)} MECD+$^\dagger$}\\[2pt]
\scalebox{0.86}{%
\begin{tabular}{@{}lc@{}}\toprule
Feature (standardized) & Coefficient \\\midrule
Dependent (GT) & $-$.10 {\scriptsize[$-$.30, .15]} \\
Adjacent & .50 {\scriptsize[.30, .75]} \\
log(1+gap) & $-$.12 {\scriptsize[$-$.40, .21]} \\
Shared target & $-$.02 {\scriptsize[$-$.25, .20]} \\
Shared tool, not target & $-$.11 {\scriptsize[$-$.28, .08]} \\
Caption similarity & .23 {\scriptsize[$-$.02, .51]} \\
\bottomrule\end{tabular}}
\end{minipage}\hfill
\begin{minipage}[t]{0.75\textwidth}\centering
{\footnotesize\textbf{(b)} MLLMs given the ground-truth events (A: captions, B: + the video)}\\[2pt]
\scalebox{0.86}{%
\begin{tabular}{@{}lcccccc@{}}\toprule
Feature (standardized) & Qwen A & Qwen B & GLM A & GLM B & Cosmos A & Cosmos B \\\midrule
Dependent (GT) & .69 {\scriptsize[.53, .86]} & .69 {\scriptsize[.53, .91]} & .53 {\scriptsize[.35, .73]} & .82 {\scriptsize[.56, 1.22]} & .53 {\scriptsize[.39, .68]} & .51 {\scriptsize[.35, .69]} \\
Adjacent & $-$.07 {\scriptsize[$-$.19, .08]} & -.00 {\scriptsize[$-$.15, .12]} & $-$.36 {\scriptsize[$-$.59, $-$.18]} & $-$.33 {\scriptsize[$-$.61, $-$.14]} & $-$.03 {\scriptsize[$-$.15, .10]} & .10 {\scriptsize[$-$.06, .24]} \\
log(1+gap) & $-$.45 {\scriptsize[$-$.63, $-$.29]} & $-$.44 {\scriptsize[$-$.62, $-$.28]} & $-$.55 {\scriptsize[$-$.78, $-$.35]} & $-$.63 {\scriptsize[$-$.91, $-$.41]} & $-$.35 {\scriptsize[$-$.52, $-$.22]} & $-$.24 {\scriptsize[$-$.39, $-$.06]} \\
Shared object & .16 {\scriptsize[.00, .30]} & .10 {\scriptsize[$-$.07, .25]} & .50 {\scriptsize[.27, .72]} & .54 {\scriptsize[.29, .84]} & .27 {\scriptsize[.13, .44]} & .38 {\scriptsize[.23, .55]} \\
Caption similarity & .27 {\scriptsize[.13, .44]} & .23 {\scriptsize[.06, .41]} & 1.07 {\scriptsize[.83, 1.37]} & .54 {\scriptsize[.25, .83]} & .37 {\scriptsize[.19, .54]} & .39 {\scriptsize[.20, .59]} \\
\bottomrule\end{tabular}}
\end{minipage}
\caption{\textbf{Correlates of predicted edges}: logistic regression of the predicted edge on standardized pair features (coefficients, 95\% CI from 300 bootstrap resamples of questions). (a)~MECD+ given the ground-truth events ($^\dagger$questions with $\geq$3 key events), used in \cref{sec:supp_dep_disc}. Its adjacency coefficient stays significant after Holm's correction over the six features (adjusted $p<0.002$, 10,000 resamples). (b)~Qwen3.6-27B, GLM-4.6V-Flash and Cosmos-Reason2 given the ground-truth events as captions (A) or as captions with the video (B), on all pairs whose labels parse, used in \cref{sec:supp_dep_shared}. The shared-object coefficients of GLM-4.6V-Flash and Cosmos-Reason2 stay significant after Holm's correction over the 30 coefficients (adjusted $p\leq0.018$). The other CIs are unadjusted for multiple comparisons.}
\label{tab:supp_e2}\label{tab:supp_e2gt}\end{table*}

\paragraph{Given the verified key events.}
Supplying the verified key events removes this bottleneck but not the dependency errors.
\Cref{tab:supp_depgiven} compares the dependencies that Qwen3.6-27B and GLM-4.6V-Flash predict from the video only ($V$), where they must first predict the events themselves, with those they predict from the video and the verified key events ($V{+}\mathcal{E}$, \cref{tab:probes}).
From the video only, just 33.8\% and 23.7\% of the dependent pairs have both events in the output and can be scored, whereas with the verified key events 98.4\% and 99.5\% can, so the missing events are indeed what limits recovery from the video only.
The discrimination between dependent and dependency-negative pairs, however, remains weak and does not improve with the verified key events: on the pairs scored in both conditions, the BAcc changes by $-$.034 [$-$.094, .026] and $-$.010 [$-$.085, .063], and it stays near .58--.61 against the .50 of chance, because the models also predict edges for most dependency-negative pairs (FPR .74 and .64 with the verified key events).
Once the events are given, the remaining errors therefore lie in distinguishing dependent from dependency-negative pairs (\cref{sec:supp_dep_disc}).

\subsubsection{Dependency predictions do not separate dependent from dependency-negative pairs}
\label{sec:supp_dep_disc}
Dependency predictions do not separate dependent from dependency-negative pairs, and the edges of MECD+ are associated with temporal adjacency but not significantly with the annotated dependency.

\paragraph{Edge rates by dependency.}
Edge predictions separate dependent from dependency-negative pairs at most partially (\cref{tab:supp_dep_diag}b, each system on the pairs it localizes).
The causal models do not separate them: on the 7\% of the pairs whose two events it localizes, Causal-VLM predicts edges for 94\% of both, and MECD+ reaches a BAcc of 51\% given the ground-truth events.
Qwen3.6-27B, GLM-4.6V-Flash and \method{} (2B) predict edges for dependent pairs 18 to 23 points more often, but still for about half of the dependency-negative pairs, so their BAcc stays at 59--62\%.
Even \method{} (2B), trained on our training split and given the ground-truth events, still misjudges whether a dependency exists for many of the 1,180 pairs (\cref{sec:supp_ablation}).

\paragraph{Temporal adjacency in MECD+.}
MECD+ links adjacent events whether or not they are dependent.
Among the adjacent pairs, it predicts edges for 85.3\% of the dependent and 84.8\% of the dependency-negative ones, and on dependency-negative pairs for 84.8\% of the adjacent but 65.9\% of the non-adjacent ones (\cref{tab:supp_dep_diag}c).
Holding the annotated dependency, the gap between the events, shared objects and caption similarity fixed in a logistic regression (\cref{tab:supp_e2}a), adjacency raises its edge probability on dependency-negative pairs by 17.9 points [9.8, 27.6], significant after Holm's correction over the six features, whereas the coefficient of the annotated dependency has a CI that includes zero.

\subsubsection{Shared tools or targets are associated with false-positive dependencies}
\label{sec:supp_dep_shared}
Given the ground-truth events, the MLLMs predict more edges for dependency-negative pairs that share a tool or target, and this association remains for GLM-4.6V-Flash and Cosmos-Reason2 after controlling for the other pair features.

\paragraph{Edge rates by shared object.}
Dependency-negative pairs that share a tool or target receive edges more often than those that share none.
Here, Qwen3.6-27B, GLM-4.6V-Flash and Cosmos-Reason2 receive the ground-truth key events of a question (identifier, interval and caption) and label every pair \textit{none}, \textit{causes} or \textit{enables}.
Given the ground-truth events as captions and the video, Qwen3.6-27B predicts edges for 70\% of the dependency-negative pairs that share an object against 54\% of those that do not, GLM-4.6V-Flash for 94\% against 73\% and Cosmos-Reason2 for 81\% against 58\% (\cref{tab:supp_dep_diag}c), although the prompt states that shared objects alone are no dependency.

\paragraph{Controlling for other pair features.}
The association remains for GLM-4.6V-Flash and Cosmos-Reason2 when the annotated dependency, adjacency, gap and caption similarity are held fixed.
In the regressions given the ground-truth events (\cref{tab:supp_e2}b), shared tools or targets are positively associated with the predicted edges of these two models in both input conditions, and these four coefficients stay significant after Holm's correction over all 30 coefficients of the six MLLM regressions (adjusted $p\leq0.018$, 10,000 question resamples).
For Qwen3.6-27B the coefficient is small and does not survive the correction (unadjusted $p=0.04$ with captions and $0.25$ with the video).
These diagnostics show which pair features accompany predicted edges, not how models reach them.

\section{Effect of verified evidence on answering}
\label{app:oracle}
\label{sec:supp_oracle}
These controlled experiments (\cref{sec:exp_utilization}) give selected components of the human-verified evidence to two MLLMs instead of having them recover it from the video.
Verified key events significantly raise the accuracy, and the same intervals placed at other times lower it, so the answers depend on where the supplied evidence lies in time.
Adding the participants, roles, interactions or dependencies yields no clear further accuracy gain under the tested representation. Given the boxes, the models recover most roles, and given the roles as well, most interactions, whereas the dependency scores stay at the level reached with the key events.

\subsection{Conditions and scoring}
\Cref{tab:supp_steps_full} extends \cref{tab:probes} with the \Lone{} and dependency Macro-F1 of \cref{tab:main}.
All conditions use the system prompt, output format and scoring of \cref{tab:main} and differ only in the human-verified evidence given as text before the question.
This evidence consists of the key events as intervals ($\mathcal{E}$), the participant boxes with their track IDs ($B$) and roles ($R$), for each event its interactions with the time of the interaction and the role, track ID and box of each participant ($\mathcal{I}$), and the \textit{causes} and \textit{enables} pairs of key events, with every unlisted pair marked as having no dependency ($\mathcal{D}$).
Events are identified only by their time and objects only by their ID and box, so the evidence contains no caption or object name that could reveal the answer.
$\mathcal{E}{+}\mathcal{I}{+}\mathcal{D}$ gives this text without the video, and $V$ gives the video without evidence, as in \cref{tab:main}.

Each condition is scored only on the evidence that it does not provide as input.
The rows $V$ and $V{+}\tilde{\mathcal{E}}$ and the \Lone{}, \Ltwo{} and \Lthree{} of $V{+}\mathcal{E}$ are scored as in \cref{tab:main}, over all ground-truth elements.
The other cells count only the given events, participants and event pairs that the model output retains, so a given element the output omits is not counted as an error.
The outputs retain 92--98\% of the given key events and 92--98\% of the event pairs, and 89--95\% of the given participants, so these scores cover nearly all of the given evidence.
Accuracy is over all \ORQNAll{} questions, with 95\% CIs of its changes from a paired bootstrap over questions (10,000 resamples). Changes are computed before rounding, so they can differ by 0.1 point from the differences of the rounded accuracies in \cref{tab:supp_steps_full}.
The role and association scores of these conditions support Findings~2 and~3 and are analyzed in \cref{sec:supp_roles,sec:supp_inter}.

\begin{table}[tp]
\centering
\footnotesize
\setlength{\tabcolsep}{2.2pt}
\resizebox{\linewidth}{!}{%
\begin{tabular}{@{}lcccccc@{}}
\toprule
Input & Acc & \Lone & \Ltwo & \Lthree & Macro-F1 & Edge F1 \\
\midrule
\multicolumn{7}{@{}l}{\textit{Qwen3.6-27B}} \\
$V$ & \ST{Q36}{V}{acc} & \ST{Q36}{V}{L1} & \ST{Q36}{V}{L2} & \ST{Q36}{V}{L3} & \ORD{Q36}{V}{MacroF1} & \ST{Q36}{V}{EdgeF1} \\
$V{+}\mathcal{E}$ & \ST{Q36}{E}{acc}\ST{Q36}{E}{sig} & \ST{Q36}{E}{L1} & \ST{Q36}{E}{L2} & \ST{Q36}{E}{L3} & .330 & \ST{Q36}{E}{depF1} \\
$V{+}\mathcal{E}{+}B$ & \ST{Q36}{EB}{acc}\ST{Q36}{EB}{sig} & -- & \ST{Q36}{EB}{roleF1} & \ST{Q36}{EB}{intF1} & .339 & \ST{Q36}{EB}{depF1} \\
$V{+}\mathcal{E}{+}B{+}R$ & \ST{Q36}{EBR}{acc}\ST{Q36}{EBR}{sig} & -- & -- & \ST{Q36}{EBR}{intF1} & .344 & \ST{Q36}{EBR}{depF1} \\
$V{+}\mathcal{E}{+}\mathcal{I}$ & \ST{Q36}{EI}{acc}\ST{Q36}{EI}{sig} & -- & -- & -- & .352 & \ST{Q36}{EI}{depF1} \\
$V{+}\mathcal{E}{+}\mathcal{I}{+}\mathcal{D}$ & \ST{Q36}{EID}{acc}\ST{Q36}{EID}{sig} & -- & -- & -- & -- & -- \\
\cmidrule{1-7}
$V{+}\tilde{\mathcal{E}}$ & \ST{Q36}{Eshift}{acc}\ST{Q36}{Eshift}{sig} & \ST{Q36}{Eshift}{L1} & \ST{Q36}{Eshift}{L2} & \ST{Q36}{Eshift}{L3} & \ST{Q36}{Eshift}{MacroF1} & \ST{Q36}{Eshift}{EdgeF1} \\
$\mathcal{E}{+}\mathcal{I}{+}\mathcal{D}$ & \ST{Q36}{TextEID}{acc}\ST{Q36}{TextEID}{sig} & -- & -- & -- & -- & -- \\
\midrule
\multicolumn{7}{@{}l}{\textit{GLM-4.6V-Flash}} \\
$V$ & \ST{GLM46VF}{V}{acc} & \ST{GLM46VF}{V}{L1} & \ST{GLM46VF}{V}{L2} & \ST{GLM46VF}{V}{L3} & \ORD{GLM46VF}{V}{MacroF1} & \ST{GLM46VF}{V}{EdgeF1} \\
$V{+}\mathcal{E}$ & \ST{GLM46VF}{E}{acc}\ST{GLM46VF}{E}{sig} & \ST{GLM46VF}{E}{L1} & \ST{GLM46VF}{E}{L2} & \ST{GLM46VF}{E}{L3} & .338 & \ST{GLM46VF}{E}{depF1} \\
$V{+}\mathcal{E}{+}B$ & \ST{GLM46VF}{EB}{acc}\ST{GLM46VF}{EB}{sig} & -- & \ST{GLM46VF}{EB}{roleF1} & \ST{GLM46VF}{EB}{intF1} & .331 & \ST{GLM46VF}{EB}{depF1} \\
$V{+}\mathcal{E}{+}B{+}R$ & \ST{GLM46VF}{EBR}{acc}\ST{GLM46VF}{EBR}{sig} & -- & -- & \ST{GLM46VF}{EBR}{intF1} & .334 & \ST{GLM46VF}{EBR}{depF1} \\
$V{+}\mathcal{E}{+}\mathcal{I}$ & \ST{GLM46VF}{EI}{acc}\ST{GLM46VF}{EI}{sig} & -- & -- & -- & .354 & \ST{GLM46VF}{EI}{depF1} \\
$V{+}\mathcal{E}{+}\mathcal{I}{+}\mathcal{D}$ & \ST{GLM46VF}{EID}{acc}\ST{GLM46VF}{EID}{sig} & -- & -- & -- & -- & -- \\
\cmidrule{1-7}
$V{+}\tilde{\mathcal{E}}$ & \ST{GLM46VF}{Eshift}{acc}\ST{GLM46VF}{Eshift}{sig} & \ST{GLM46VF}{Eshift}{L1} & \ST{GLM46VF}{Eshift}{L2} & \ST{GLM46VF}{Eshift}{L3} & \ST{GLM46VF}{Eshift}{MacroF1} & \ST{GLM46VF}{Eshift}{EdgeF1} \\
$\mathcal{E}{+}\mathcal{I}{+}\mathcal{D}$ & \ST{GLM46VF}{TextEID}{acc}\ST{GLM46VF}{TextEID}{sig} & -- & -- & -- & -- & -- \\
\bottomrule
\end{tabular}}
\caption{\textbf{Step-wise human-verified evidence} (\cref{tab:probes}) with the \Lone{} and Macro-F1 of \cref{tab:main}. Controls below the thin rule: $\tilde{\mathcal{E}}$ moves each key event to another time of the same length (tIoU $<0.1$ with every key event), and $\mathcal{E}{+}\mathcal{I}{+}\mathcal{D}$ omits the video. $V$, $V{+}\tilde{\mathcal{E}}$ and the \Lone{}, \Ltwo{} and \Lthree{} of $V{+}\mathcal{E}$: scored as in \cref{tab:main}. Other cells: only the given elements that the output retains. --: given. Acc in \%. $^\ast$95\% CI of the accuracy change against the row above (for the controls: $V{+}\mathcal{E}$, $V{+}\mathcal{E}{+}\mathcal{I}{+}\mathcal{D}$) excludes 0.}
\label{tab:supp_steps_full}
\end{table}

\subsection{Results}
\label{sec:supp_oracle_res}
\paragraph{Answer accuracy.}
Verified key events significantly raise the accuracy, by 2.7 points for Qwen3.6-27B (95\% CI [$+$0.7, $+$4.6]) and 3.6 points for GLM-4.6V-Flash ([$+$1.9, $+$5.4]).
Intervals of the same number and length at other times lower it by 8.1 ([$-$10.1, $-$5.9]) and 5.5 ([$-$7.3, $-$3.7]) points against the verified ones.
The accuracy thus depends on where the supplied events lie in time, not only on their presence, which highlights the role of accurate temporal grounding.
Adding the boxes, roles, interactions or dependencies changes the accuracy by $-$1.1 to $+$1.2 points against the row above, with 95\% CIs that include 0.
The same holds for the dependencies on the \ORQNDep{} questions with an annotated dependency, where they apply (95\% CIs [$-$6.0, $+$1.3] and [$-$3.3, $+$2.6]).
Under the tested textual representation, these annotations thus show no clear accuracy gain beyond the verified key events.
Without the video, the full evidence as text lowers the accuracy by 14.1 and 14.5 points, so the answers still rely on visual information.

\paragraph{Evidence scores.}
Verified key events extend the evidence that the models recover from the video.
\Lone{} rises from \ST{Q36}{V}{L1} to \ST{Q36}{E}{L1} (Qwen3.6-27B) and from \ST{GLM46VF}{V}{L1} to \ST{GLM46VF}{E}{L1} (GLM-4.6V-Flash), and \Lthree{} from \ST{Q36}{V}{L3} to \ST{Q36}{E}{L3} and from \ST{GLM46VF}{V}{L3} to \ST{GLM46VF}{E}{L3}.
With the verified key events, the models judge the dependencies of 1,162 of the 1,180 ground-truth event pairs, and their Edge F1 on the pairs their replies keep is \ST{Q36}{E}{depF1} and \ST{GLM46VF}{E}{depF1}. From the video alone, they localize both events of only 335 and 253 pairs, and their Edge F1 on these fewer pairs is \ST{Q36}{V}{depF1} and \ST{GLM46VF}{V}{depF1}.   %

Given the boxes, the models assign the roles of the retained participants with a \Ltwo{} of \ST{Q36}{EB}{roleF1} and \ST{GLM46VF}{EB}{roleF1} (.835 and .722 on the tools and targets alone, \cref{sec:supp_roles_given}).
The remaining errors confuse the roles of correctly localized participants, mostly tool and target (Finding~2, \cref{sec:supp_roles}).
Given the roles as well, \Lthree{} rises from \ST{Q36}{EB}{intF1} and \ST{GLM46VF}{EB}{intF1} to \ST{Q36}{EBR}{intF1} and \ST{GLM46VF}{EBR}{intF1}.
At the \STNConcMom{} moments with multiple interactions, however, it stays at \ST{Q36}{EBR}{intmulti} and \ST{GLM46VF}{EBR}{intmulti}, below the \STRule{} of a rule based on box overlap (Finding~3, \cref{sec:supp_inter}).

Unlike the role and interaction scores, the dependency scores do not rise with the further evidence.
With the boxes, roles or interactions given, Edge F1 stays between .426 and .476 and Macro-F1 between .331 and .354, against \ST{Q36}{E}{depF1} and \ST{GLM46VF}{E}{depF1} and .330 and .338 with the key events only.
Compared with the key events only, giving the interactions changes Edge F1 by $+$.014 for Qwen3.6-27B and $-$.032 for GLM-4.6V-Flash, so their effect on dependency recovery differs between the models and shows no consistent gain (Finding~4, \cref{sec:supp_dep}).

\section{\method: details and analyses}
\label{app:method}

\subsection{\method: full description}
\label{sec:supp_method}

\input{fig/fig3_method_integrated}

\paragraph{Design motivation.}
Each component targets a grounding failure diagnosed in \cref{sec:exp_findings}.
(1)~They miss many of the cause and precondition events that a causal question requires (\cref{tab:main}, \cref{sec:supp_events}): \method learns event boundaries from the video, divides the video at these boundaries into candidate segments and selects the key events among them one at a time, each conditioned on the question and on the events selected before.
(2)~They miss participants, confuse their roles and group them into the wrong interactions (\cref{sec:supp_roles,sec:supp_inter}): extending the interaction queries of DETR-style human--object interaction detection~\cite{carion2020detr,tamura2021qpic,kim2021hotr,zhang2021cdn} from images to video, \method predicts the interactions of each key event as a set of learned queries, each consisting of role-specific slots for hands, tools and targets that predict the box of one participant in that role.
(3)~They do not separate dependent from dependency-negative event pairs (\cref{sec:supp_dep}): \method classifies every pair of its key events as \textit{causes}, \textit{enables} or no dependency from a representation of each event, built from the MLLM hidden states of its frames, its predicted description and the question, together with the time gap, the durations and the overlap of the two events.
(4)~Many of their correct answers are not supported by the evidence (\cref{tab:main}): \method answers only from the evidence that its trained grounding module predicts, namely clips of the key events, frames of each interaction cropped to its participants with their boxes and roles, and the dependencies.
\Cref{sec:supp_method_results} examines each of these failures in the results of \method.

\paragraph{Overview.}
\Cref{fig:method} shows the pipeline.
\method answers in two stages: a grounding module (Stage~1, what to ground) predicts the evidence, and an answering module (Stage~2, how to answer) answers the question from this evidence,
\begin{equation}
\begin{gathered}
\underbrace{(\hat{\mathcal E},H)=f_T(V,Q),\ \ \hat{\mathcal A}_k=f_S(H,\hat e_k),\ \ \hat{\mathcal R}=f_R(H,\hat{\mathcal E})}_{\text{what to ground}},\\
\underbrace{\hat A=f_A(Q,\hat{\mathcal E},\hat{\mathcal A},\hat{\mathcal R};V)}_{\text{how to answer}}.
\end{gathered}
\label{eq:modules}
\end{equation}
The predicted evidence has the form defined in \cref{sec:bench_task}: $\hat{\mathcal E}=\{\hat e_k\}$ are the predicted key events, each a time interval with a one-sentence description, $\hat{\mathcal A}_k$ the interactions in event $\hat e_k$, which give the box and the role of every participant and the interaction it belongs to (association), $\hat{\mathcal R}$ the dependencies between the predicted events, and $H$ the hidden states of the MLLM while computing $f_T$.
They correspond to $\mathcal E$, $\mathcal I$ and $\mathcal D$ of \cref{sec:bench_task}, and $\hat A$ is the predicted answer.

\emph{Stage~1.} The MLLM reads the video and the question once, and three heads read its hidden states $H$ (top to bottom in \cref{fig:method}).
$f_T$ retrieves the key events among candidate segments and describes them, $f_S$ predicts the interactions of each key event with the box and role of every participant, and $f_R$ labels each pair of key events as \textit{before} (no dependency), \textit{causes} or \textit{enables}.
The three heads are trained jointly.

\emph{Stage~2.} $f_A$, the same MLLM with a second adapter, answers from the predicted evidence only: the clip of each key event, the frame of each interaction cropped to its participants with their boxes and roles as text, and the dependencies.
In \cref{fig:method}, it answers from E1 (0.4--4.0\,s), E2 (7.0--10.4\,s), the interaction of each event and the dependency E1 \textit{enables} E2.

The MLLM is Qwen3.5-2B~\cite{qwen2026qwen35} (Qwen3.5-9B for the 9B variant in \cref{tab:main}, \cref{sec:supp_9b}) with a frozen vision encoder and two low-rank adapters on its language model, $\theta_G$ for grounding and $\theta_A$ for answering.
Both modules are trained on the 25,894 questions of the \bench{} training split with their automatic labels (\cref{sec:bench_pipeline}), which share no source video with the evaluation split (\cref{tab:split}).

\subsubsection{What to ground: the grounding module}
\label{sec:method_grounding}

\paragraph{Key events (when).}
\label{sec:method_events}
\emph{Input.} $f_T$ reads the video at 4\,fps as temporal patches of two frames, each preceded by its timestamp, followed by the question.
\emph{Step 1, event segmentation.} A boundary head, two transformer layers over the spatial means $\bar v_t$ of the patch hidden states, predicts a boundary probability $\beta_t$ per patch. It does not depend on the question, which comes after the video in the input.
Peaks with $\beta_t\geq\theta_b$ split the video, and runs of one or two adjacent segments form up to 48 candidates $c_j$.
\emph{Step 2, sequential key-event retrieval.} Each candidate is appended after the question as its interval and a token embedded from the video states in and around it, so its hidden state $z_j$ depends on the question.
From a retrieval token, a pointer selects one candidate per step $n$,
\begin{equation}
\ell_{n,j}=\langle W_gg_n,W_zz_j\rangle/\sqrt d,\qquad \ell_{n,\mathrm{END}}=\mathrm{MLP}(g_n),
\end{equation}
with $g_n$ the hidden state of the last token. The chosen candidate is fed back as the next token.
Chosen candidates and those with temporal IoU $\geq$0.5 to them are masked, END is allowed from the second step, and at most four events are retrieved. At inference, retrieval is greedy and stops when $\ell_{n,\mathrm{END}}+b_{\mathrm{END}}$, with a constant bias $b_{\mathrm{END}}$, exceeds every candidate logit.
\emph{Step 3, description.} The retrieved events are written in temporal order with their candidates' times, and the MLLM generates a one-sentence description for each.
$f_S$ and $f_R$ represent event $e$ by $h_e=W[\bar v_e;\bar d_e;h_Q]$, a projection of the mean hidden states over its patches ($\bar v_e$), over its description ($\bar d_e$) and over the question ($h_Q$).
\emph{Training.} $\mathcal{L}_{\mathrm{bnd}}$ is a binary cross-entropy against Gaussians ($\sigma=0.5$\,s) at the boundaries of the pipeline's event units (\cref{sec:bench_pipeline}).
Key events are assigned one-to-one to candidates at temporal IoU $\geq$0.3. Since the key events form a set, the target of each step is any assigned event not yet retrieved,
\begin{equation}
\mathcal{L}_{\mathrm{ptr},n}=-\log\textstyle\sum_e\exp\big(\sum_ju_e(j)\log p_n(j)\big),
\end{equation}
with $u_e$ the soft label of event $e$ over the candidates ($\propto$ squared IoU, for IoU $\geq$0.5) and $p_n$ the softmax over the allowed candidates and END. The remaining event scored highest is fed back next, and END is the target after the last. The loss marginalizes over the remaining assigned events, so they can be retrieved in any order.
$\mathcal{L}_T$ is the description cross-entropy. Key events without a candidate are added with their annotated intervals at a lower weight.
During training, the candidates move from perturbed event-unit boundaries to the model's own segmentation, and $\theta_b$ is set on the training questions (\cref{sec:supp_retrieval}).

\paragraph{Interactions (what and where).}
\label{sec:method_actions}
\emph{Interaction queries.} For each event, $M{=}4$ learned queries $q_k+W_eh_e+W_qh_Q$ pass a cross-attention block over the event's patches $m_t=W_v\bar v_t+\mathrm{enc}(t)$ ($\mathrm{enc}$: relative time), a self-attention layer among the queries, which lets two queries avoid the same interaction, and a feed-forward layer, giving features $a_k$.
Each $a_k$ yields an interaction probability $p_k$ and an \emph{interaction span} (\cref{sec:bench_task}), possibly in several segments, at whose maximum the interaction is read,
\begin{equation}
\hat\Sigma_k(t)=\sigma\big(\langle W_m a_k,m_t\rangle/\sqrt d+b\big),\qquad \tau_k=\arg\max_t\hat\Sigma_k(t).
\end{equation}
\emph{Participant slots.} For each role $r\in\{\text{hand},\text{tool},\text{target}\}$, $N_r$ slots with queries $q_{k,r,s}=q_r(h_Q)+e_{r,s}+W_ra_k$ attend over the projected hidden states $v'_{\tau_k,p}$ of the spatial tokens $p$ of $\tau_k$ (with positional encoding) and read out a participation probability $\pi_{k,r,s}$ and a box
\begin{equation}
\hat b_{k,r,s}=\mathrm{MLP}_r\big([\textstyle\sum_p\alpha_{k,r,s,p}v'_{\tau_k,p};\ \sum_p\alpha_{k,r,s,p}c_p;\ q_{k,r,s}]\big),
\end{equation}
with $\alpha$ the attention over patches and $c_p$ the patch centres. There are two hand, one tool and two target slots, and neither slots nor queries have a fixed identity.
\emph{Training by set matching.} For each ground-truth interaction, one token is drawn from its span and all queries are read there.
Within a role, the Hungarian algorithm matches slots to the participants visible at that token on box distance, GIoU and participation probability. Participants not visible there take a remaining slot by participation alone, without box loss, so $\pi$ models participation, not visibility.
A second Hungarian matching assigns queries to interactions on the interaction probability, the temporal IoU of the spans and the role costs, and
\begin{equation}
\mathcal{L}_S=2\mathcal{L}_{\mathrm{act}}+2\mathcal{L}_{\mathrm{time}}+\mathcal{L}_{\mathrm{box}}+4\mathcal{L}_{\mathrm{patch}}+2\mathcal{L}_{\mathrm{part}}
\end{equation}
combines losses on the interaction probabilities, the spans, the boxes of matched visible participants, the slots' attention against the participants' box coverage, and the participation probabilities.
Association needs no separate loss: a query must explain the participants of one ground-truth interaction jointly.
\emph{Inference.} Queries with $p_k\geq0.5$ become interactions, with the tokens whose span is $\geq$0.5 and the slots with $\pi_{k,r,s}\geq0.5$ as participants, each with its box at $\tau_k$.
\emph{Capacity.} The fixed capacity rarely binds on the gold-labeled questions: 4 of the \EvGold{} questions have more than four key events, 10 of the 2,605 key events have more than four interactions, and 47 of the \EvGoldInter{} interactions have more participants of a role than there are slots (32 with more than two hands).

\paragraph{Dependencies (how).}
\label{sec:method_relations}
For each pair of retrieved events $(e_i,e_j)$, $i<j$, $f_R$ predicts
\begin{equation}
p(y_{ij})=\mathrm{softmax}\big(\mathrm{MLP}([h_i;h_j;h_i\odot h_j;\delta_{ij}])\big),
\end{equation}
$y_{ij}\in\{\textit{before},\textit{causes},\textit{enables}\}$, where $\delta_{ij}$ encodes the index gap, the time gap, the two durations and their overlap.
$\mathcal{L}_R$ is a cross-entropy weighted by inverse square-root class frequencies. At inference, the pairs labeled \textit{causes} or \textit{enables} are output as dependencies.

\paragraph{Joint training.}
\label{sec:method_training}
$\theta_G$ is a low-rank adapter (rank 16) on every layer of the language model, and all heads are initialized randomly.
All parts are trained in one forward pass over the video, the question, the candidates, the teacher-forced retrieval steps and the event blocks, with
\begin{equation}
\mathcal{L}_G=\mathcal{L}_T+w_{\mathrm{bnd}}\mathcal{L}_{\mathrm{bnd}}+w_{\mathrm{ptr}}\mathcal{L}_{\mathrm{ptr}}+w_R\mathcal{L}_R+w_S\mathcal{L}_S+\mathcal{L}_{\mathrm{aux}},
\end{equation}
where $\mathcal{L}_{\mathrm{aux}}$ trains two auxiliary heads and a candidate score that are not used at inference (\cref{sec:supp_retrieval}), and each weight $w$ sets the gradient norm of its loss on $\theta_G$ relative to that of $\mathcal{L}_T$, measured early in training and then fixed.
As $\mathcal{L}_S$ and $\mathcal{L}_R$ also reach $\theta_G$, they shape the hidden states the heads read (ablations: \cref{sec:supp_ablation}).
Training runs 4 epochs of AdamW (learning rate $10^{-4}$ for $\theta_G$, $3\times10^{-4}$ for the heads, global batch size 16), and the final checkpoint is used.

\subsubsection{How to answer: evidence-constrained answering}
\label{sec:method_answer}

$f_A$ sees the video only through the predicted evidence, given as one sequence in temporal order (\cref{fig:method}).
For each predicted event, the sequence contains a header with its interval and description and a clip of its frames (48 frames over all events, at most 8\,fps, with their original timestamps).
Each interaction of the event follows as the frame at $\tau_k$ cropped to the union of its participants' boxes, with the box of each participant, relative to the crop, and its role as text.
The predicted \textit{causes} and \textit{enables} dependencies come next, each with the intervals and descriptions of its two events.
A fixed instruction explains this format, the roles and the dependency types, and the question follows the evidence.
$\theta_A$, initialized from $\theta_G$, is trained after the frozen grounding module for 3 epochs with the answer cross-entropy, on the evidence it predicts for the training questions.

\subsubsection{Implementation details}
\label{sec:supp_retrieval}
This part completes the description of $f_T$ and of the loss weights of the grounding module (\cref{sec:method_grounding}) and gives the settings of the 9B backbone.

\paragraph{Event segmentation.}
The boundary head maps the spatial mean of the hidden states of each temporal patch to 256 dimensions, adds a sinusoidal encoding of the patch time and applies two bidirectional transformer layers (4 heads). A linear layer gives the boundary logit of each patch.
A boundary is a local maximum of $\beta$ with $\beta_t\geq\theta_b$, placed at the time of its patch. The boundaries and the two ends of the video define the segments, and every run of one or two adjacent segments is a candidate.
Candidates are ranked by the mean boundary probability at their two ends (0.5 at an end of the video), then by shorter duration and earlier start, and the first 48 are kept. Candidate times are rounded to 0.1\,s.

\paragraph{Candidate tokens and retrieval pointer.}
The input embedding of the token of candidate $c_j$ is computed from the hidden states of a first forward pass over the video and the question, without gradient into the MLLM: attention pooling with two learned queries over all visual tokens of the patches inside $c_j$, the mean patch states in 1\,s windows before and after each of its two ends and inside it, the boundary probabilities at its ends, and features of its start, end and duration. An MLP maps these features to the embedding space, followed by a layer normalization with a learned scale.
The chosen candidate is fed back through a learned linear map of its embedding, initialized to the identity, plus a learned vector.
The pointer logits are computed for every candidate at every step, and masked candidates are excluded from the softmax. At inference, the END bias $b_{\mathrm{END}}=-2.25$ is added to the END logit at each step. Retrieval also stops when four events are retrieved or no candidate is allowed.

\paragraph{Selection of $b_{\mathrm{END}}$.}
$b_{\mathrm{END}}$ is the only hyperparameter value of \method{} chosen on evaluation questions. $\theta_b$ is set on the training questions, the loss weights by gradient norms, and both modules use their final checkpoints.
$b_{\mathrm{END}}$ was chosen once, with the final checkpoint and before the evaluation on the gold-labeled questions, as the value on a grid from $-6$ to $6$ in steps of 0.25 that gives the highest key-event mTIoU at the retrieval step on 615 evaluation questions that had passed C1 and human verification. Of these 615 questions, 462 are among the \EvGold{} gold-labeled questions.
On these 615 questions, mTIoU stays between .243 and .263 and F1@0.3 between .399 and .432 for $b_{\mathrm{END}}$ from $-3$ to $-1$ (\cref{tab:supp_bend}a), and the default $b_{\mathrm{END}}=0$ reaches an mTIoU of .246 and an F1@0.3 of .396, against .263 and .429 at the chosen value.
On the 1,177 gold-labeled questions outside these 615, \method{} (2B) keeps the highest key-event F1@0.3 and Edge F1 of all evaluated systems, its gains over joint SFT hold, and its accuracy stays below that of joint SFT (\cref{tab:supp_bend}b). The comparisons of \cref{sec:exp_learning} change little when the questions used to choose $b_{\mathrm{END}}$ are left out.

\begin{table}[t]
\centering\footnotesize\setlength{\tabcolsep}{3.2pt}
\textbf{(a)} Key-event retrieval by $b_{\mathrm{END}}$\\[2pt]
\begin{tabular}{@{}lccccccccc@{}}
\toprule
$b_{\mathrm{END}}$ & $-6$ & $-4$ & $-3$ & $-2.75$ & $\mathbf{-2.25}$ & $-2$ & $-1$ & $0$ & $\geq1$ \\
\midrule
mTIoU  & .182 & .195 & .243 & .253 & \textbf{.263} & .258 & .246 & .246 & .246 \\
F1@0.3 & .377 & .385 & .422 & .432 & \textbf{.429} & .426 & .399 & .396 & .395 \\
Events & 3.9 & 3.7 & 2.3 & 2.0 & 1.6 & 1.4 & 1.0 & 1.0 & 1.0 \\
\bottomrule
\end{tabular}
\\[6pt]
\textbf{(b)} Outside the 615 questions / all gold-labeled questions\\[2pt]
\begin{tabular}{@{}lcccc@{}}
\toprule
System & Acc & F1@0.3 & \Lthree{} & Edge F1 \\
\midrule
\method{} (2B) & .133 / .123 & .447 / .443 & .191 / .184 & .105 / .116 \\
\method{} (9B) & .201 / .184 & .437 / .432 & .183 / .178 & .077 / .081 \\
Joint SFT & .139 / .135 & .261 / .256 & .138 / .129 & .027 / .030 \\
GPT-6.1 Sol & .359 / .330 & .407 / .399 & .229 / .213 & .057 / .058 \\
\bottomrule
\end{tabular}
\caption{\textbf{Selection of $b_{\mathrm{END}}$.} (a)~Key-event retrieval of \method{} (2B) by $b_{\mathrm{END}}$ on the 615 questions used to choose it, scored at the retrieval step as in the selection. Events: mean number of retrieved events per question. Bold: the chosen value. (b)~Results on the 1,177 gold-labeled questions outside these 615 / on all \EvGold{} gold-labeled questions (\cref{tab:main}), scored as in \cref{tab:main}.}
\label{tab:supp_bend}
\end{table}

\paragraph{Retrieval targets.}
The one-to-one assignment of key events to candidates maximizes the number of assigned events and then their total temporal IoU, under the retrieval constraints (distinct candidates, pairwise IoU below 0.5, at most four) and with IoU $\geq0.3$ for every assigned pair.
At each step, the soft label $u_e$ of a remaining event $e$ spreads over the allowed candidates with IoU $\geq0.5$ to $e$, in proportion to the squared IoU. If the assigned candidate of $e$ has IoU below 0.5, $u_e$ is one-hot on it, and a step in which all remaining events have such labels has weight 0.5.
The event that is fed back next is chosen by a decoding pass without gradient as the remaining event with the highest $\sum_ju_e(j)\log p_n(j)$, and its assigned candidate is fed back.
END is the target after the last assigned event unless four events are chosen or no candidate is allowed. A question without an assigned event has no pointer loss.

\paragraph{Candidates during training.}
During the first 15\% of the training steps, the candidates are built, as above, from the event-unit boundaries after random perturbation (shifted, removed, split or added boundaries) and ranked by the boundary probabilities of the first pass.
Over the next 60\%, the candidates from the module's own segmentation replace them with a probability that rises linearly from 0 to 1, with a lower threshold of 0.3. In the last 25\%, only these are used, at threshold $\theta_b$.
$\theta_b$ is set once, at the start of the last phase, to the value in $\{0.05,0.10,\dots,0.90\}$ that maximizes the F1 of the predicted boundaries (matched within $\pm0.5$\,s) against the event-unit boundaries accumulated over the training questions of the middle phase (first-pass boundary probabilities of the model at that step), and the same value is used at inference.
In the event blocks, assigned key events are written with the intervals of their candidates, and key events without a candidate are added with their annotated intervals, with weight 0.3 after the first phase, so that $f_S$ and $f_R$ are trained on every key event. A dependency pair takes the lower weight of its two events.
The interaction targets of an event are restricted to the tokens inside its block interval, and an event left without an interaction there gets no interaction loss.
The description loss covers only the description tokens, since the times are copied from the candidates.

\paragraph{Loss weights.}
$\mathcal{L}_{\mathrm{aux}}=0.1\,\mathcal{L}_{\mathrm{patch}}^{\mathrm{aux}}+0.05\,\mathcal{L}_{\mathrm{temp}}^{\mathrm{aux}}+0.5\,\mathcal{L}_{\mathrm{q}}$, where the two auxiliary heads predict from the hidden states which spatial tokens of a span token are covered by the participants' boxes and which temporal patches lie in a key event, and $\mathcal{L}_{\mathrm{q}}$ regresses a score of each candidate token, an MLP on $z_j$, to the highest temporal IoU of the candidate with a key event (smooth $L_1$).
For $k\in\{\mathrm{bnd},\mathrm{ptr},R,S\}$, $w_k=\mathrm{clip}(\rho_k\lVert\nabla_{\theta_G}\mathcal{L}_T\rVert/\lVert\nabla_{\theta_G}\mathcal{L}_k\rVert,\,0.05,\,10)$ with $\rho_{\mathrm{ptr}}=1$ and $\rho_k=0.1$ otherwise. The gradient norms are averaged over a few training questions at the first step and again after 150 steps, after which the weights are fixed.
Each loss is averaged over the questions of the batch that have its labels.

\paragraph{9B backbone.}
\label{sec:supp_9b}
\Cref{tab:main} also reports \method{} with Qwen3.5-9B as the MLLM.
Both modules are trained as for the 2B backbone, with the same data, schedule, losses and END bias, except that the 9B answering module uses a global batch size of 16 instead of 4, since it is trained on four nodes instead of one.
All ablations use the 2B backbone, and the results of the 9B model by causal capability are in \cref{tab:supp_capability2}.

\subsection{Results against the design motivation}
\label{sec:supp_method_results}
We revisit the four failures that motivate \method (\cref{sec:supp_method}, design motivation) on the \EvGold{} gold-labeled questions (\cref{tab:main}) and on all \EvQ{} evaluation questions (\cref{tab:supp_fulltest}).
For each failure, we summarize what \method addresses and the problem that remains.
\Cref{fig:supp_qual} compares the outputs of Qwen3.6-27B and \method on one gold-labeled question. Qwen3.6-27B answers correctly, but only one of its five events matches a key event, none of its 21 participant boxes matches a ground-truth participant and none of its three dependencies is annotated. \method recovers both key events, three of its nine boxes match the ground truth and one of its three dependencies is the annotated one, and it answers correctly from this evidence.

\providecolor{pDk}{HTML}{306692}\providecolor{pTx}{HTML}{1B2A41}
\providecolor{cWhen}{HTML}{2F6DB3}\providecolor{cWhat}{HTML}{7A3FA0}\providecolor{cHow}{HTML}{1E7B6A}\providecolor{cAns}{HTML}{A8551B}
\providecolor{qRed}{HTML}{D62728}\providecolor{qBlue}{HTML}{1F5FBF}
\begin{figure*}[t]\centering
{\sffamily\fontsize{7}{8.4}\selectfont\raggedright \textbf{Question:} What would happen later in the sequence if putting the mug into the microwave were not performed on the mug?\quad \textbf{Ground-truth answer:} If the mug were not heated in the microwave, the tea would remain cold, affecting the ability to stir and hold a warm beverage later.\par}\vspace{2pt}
\includegraphics[width=\textwidth]{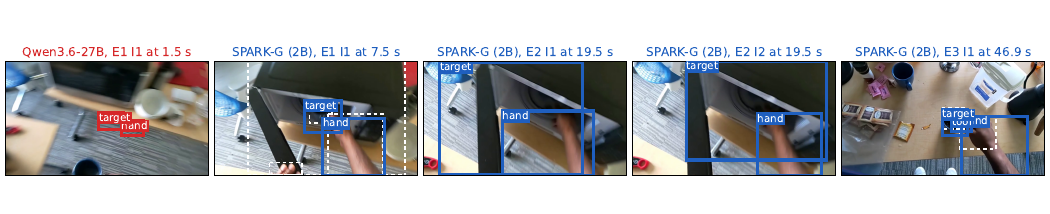}\\[2pt]
\begin{minipage}[t]{0.493\textwidth}{\sffamily\fontsize{6.6}{7.6}\selectfont\textbf{\textcolor{qRed}{Qwen3.6-27B}}}\\[1pt]
\fcolorbox{pDk}{white}{\begin{minipage}[t]{0.975\linewidth}\raggedright\sffamily\fontsize{5.5}{6.45}\selectfont\color{pTx}
\hangindent=2.2em \{\textcolor{cWhen}{\textbf{"evidence"}}: [\\
\hangindent=2.2em \quad \textcolor{qRed}{\{\textcolor{pTx!72}{"id":} \textbf{"E1"}, \textcolor{pTx!72}{"start":} \textbf{0.0}, \textcolor{pTx!72}{"end":} \textbf{3.5}, \textcolor{pTx!72}{"description":} "A person holds a blue mug with a spoon inside and moves it towards a microwave."\}},\\
\hangindent=2.2em \quad \{\textcolor{pTx!72}{"id":} \textbf{"E2"}, \textcolor{pTx!72}{"start":} \textbf{3.5}, \textcolor{pTx!72}{"end":} \textbf{10.5}, \textcolor{pTx!72}{"description":} "The person opens the microwave door, places the mug inside, and closes the door."\},\\
\hangindent=2.2em \quad \textcolor{qRed}{\{\textcolor{pTx!72}{"id":} \textbf{"E3"}, \textcolor{pTx!72}{"start":} \textbf{10.5}, \textcolor{pTx!72}{"end":} \textbf{18.0}, \textcolor{pTx!72}{"description":} "The person presses buttons on the microwave to start it, and the microwave begins to operate."\}},\\
\hangindent=2.2em \quad \textcolor{qRed}{\{\textcolor{pTx!72}{"id":} \textbf{"E4"}, \textcolor{pTx!72}{"start":} \textbf{18.0}, \textcolor{pTx!72}{"end":} \textbf{22.0}, \textcolor{pTx!72}{"description":} "The person opens the microwave door and removes the mug."\}},\\
\hangindent=2.2em \quad \textcolor{qRed}{\{\textcolor{pTx!72}{"id":} \textbf{"E5"}, \textcolor{pTx!72}{"start":} \textbf{22.0}, \textcolor{pTx!72}{"end":} \textbf{56.5}, \textcolor{pTx!72}{"description":} "The person places the mug on the table, stirs the contents with a spoon, and then lifts the mug."\}},\\
\hangindent=2.2em \textcolor{cWhat}{\textbf{"grounding"}}: [\\
\hangindent=2.2em \quad \{\textcolor{pTx!72}{"event\_id":} \textbf{"E1"}, \textcolor{pTx!72}{"interactions":} [\\
\hangindent=2.2em \qquad \{\textcolor{pTx!72}{"id":} \textbf{"I1"}, \textcolor{pTx!72}{"t":} \textbf{1.5}, \textcolor{pTx!72}{"boxes":} [\\
\hangindent=2.2em \qquad\quad \textcolor{qRed}{\{\textcolor{pTx!72}{"role":} \textbf{"hand"}, \textcolor{pTx!72}{"track":} \textbf{"H1"}, \textcolor{pTx!72}{"box":} [570, 520, 680, 650]\}},\\
\hangindent=2.2em \qquad\quad \textcolor{qRed}{\{\textcolor{pTx!72}{"role":} \textbf{"target"}, \textcolor{pTx!72}{"track":} \textbf{"O1"}, \textcolor{pTx!72}{"box":} [460, 450, 580, 600]\}}]\}]\},\\
\hangindent=2.2em \quad \{\textcolor{pTx!72}{"event\_id":} \textbf{"E2"}, \textcolor{pTx!72}{"interactions":} [\\
\hangindent=2.2em \qquad \{\textcolor{pTx!72}{"id":} \textbf{"I2"}, \textcolor{pTx!72}{"t":} \textbf{5.0}, \textcolor{pTx!72}{"boxes":} [\\
\hangindent=2.2em \qquad\quad \textcolor{qRed}{\{\textcolor{pTx!72}{"role":} \textbf{"hand"}, \textcolor{pTx!72}{"track":} \textbf{"H1"}, \textcolor{pTx!72}{"box":} [100, 200, 300, 400]\}},\\
\hangindent=2.2em \qquad\quad \textcolor{qRed}{\{\textcolor{pTx!72}{"role":} \textbf{"target"}, \textcolor{pTx!72}{"track":} \textbf{"O2"}, \textcolor{pTx!72}{"box":} [200, 100, 600, 400]\}}]\},\\
\hangindent=2.2em \qquad \{\textcolor{pTx!72}{"id":} \textbf{"I3"}, \textcolor{pTx!72}{"t":} \textbf{7.5}, \textcolor{pTx!72}{"boxes":} [\\
\hangindent=2.2em \qquad\quad \textcolor{qRed}{\{\textcolor{pTx!72}{"role":} \textbf{"hand"}, \textcolor{pTx!72}{"track":} \textbf{"H1"}, \textcolor{pTx!72}{"box":} [500, 300, 600, 450]\}},\\
\hangindent=2.2em \qquad\quad \textcolor{qRed}{\{\textcolor{pTx!72}{"role":} \textbf{"target"}, \textcolor{pTx!72}{"track":} \textbf{"O1"}, \textcolor{pTx!72}{"box":} [400, 350, 500, 450]\}}]\},\\
\hangindent=2.2em \qquad \{\textcolor{pTx!72}{"id":} \textbf{"I4"}, \textcolor{pTx!72}{"t":} \textbf{9.5}, \textcolor{pTx!72}{"boxes":} [\\
\hangindent=2.2em \qquad\quad \textcolor{qRed}{\{\textcolor{pTx!72}{"role":} \textbf{"hand"}, \textcolor{pTx!72}{"track":} \textbf{"H1"}, \textcolor{pTx!72}{"box":} [100, 200, 300, 400]\}},\\
\hangindent=2.2em \qquad\quad \textcolor{qRed}{\{\textcolor{pTx!72}{"role":} \textbf{"target"}, \textcolor{pTx!72}{"track":} \textbf{"O2"}, \textcolor{pTx!72}{"box":} [200, 100, 600, 400]\}}]\}]\},\\
\hangindent=2.2em \quad \{\textcolor{pTx!72}{"event\_id":} \textbf{"E3"}, \textcolor{pTx!72}{"interactions":} [\\
\hangindent=2.2em \qquad \{\textcolor{pTx!72}{"id":} \textbf{"I5"}, \textcolor{pTx!72}{"t":} \textbf{11.0}, \textcolor{pTx!72}{"boxes":} [\\
\hangindent=2.2em \qquad\quad \textcolor{qRed}{\{\textcolor{pTx!72}{"role":} \textbf{"hand"}, \textcolor{pTx!72}{"track":} \textbf{"H1"}, \textcolor{pTx!72}{"box":} [400, 500, 500, 600]\}},\\
\hangindent=2.2em \qquad\quad \textcolor{qRed}{\{\textcolor{pTx!72}{"role":} \textbf{"target"}, \textcolor{pTx!72}{"track":} \textbf{"O2"}, \textcolor{pTx!72}{"box":} [200, 100, 600, 400]\}}]\}]\},\\
\hangindent=2.2em \quad \{\textcolor{pTx!72}{"event\_id":} \textbf{"E4"}, \textcolor{pTx!72}{"interactions":} [\\
\hangindent=2.2em \qquad \{\textcolor{pTx!72}{"id":} \textbf{"I6"}, \textcolor{pTx!72}{"t":} \textbf{19.0}, \textcolor{pTx!72}{"boxes":} [\\
\hangindent=2.2em \qquad\quad \textcolor{qRed}{\{\textcolor{pTx!72}{"role":} \textbf{"hand"}, \textcolor{pTx!72}{"track":} \textbf{"H1"}, \textcolor{pTx!72}{"box":} [100, 200, 300, 400]\}},\\
\hangindent=2.2em \qquad\quad \textcolor{qRed}{\{\textcolor{pTx!72}{"role":} \textbf{"target"}, \textcolor{pTx!72}{"track":} \textbf{"O2"}, \textcolor{pTx!72}{"box":} [200, 100, 600, 400]\}}]\},\\
\hangindent=2.2em \qquad \{\textcolor{pTx!72}{"id":} \textbf{"I7"}, \textcolor{pTx!72}{"t":} \textbf{20.5}, \textcolor{pTx!72}{"boxes":} [\\
\hangindent=2.2em \qquad\quad \textcolor{qRed}{\{\textcolor{pTx!72}{"role":} \textbf{"hand"}, \textcolor{pTx!72}{"track":} \textbf{"H1"}, \textcolor{pTx!72}{"box":} [500, 300, 600, 450]\}},\\
\hangindent=2.2em \qquad\quad \textcolor{qRed}{\{\textcolor{pTx!72}{"role":} \textbf{"target"}, \textcolor{pTx!72}{"track":} \textbf{"O1"}, \textcolor{pTx!72}{"box":} [400, 350, 500, 450]\}}]\}]\},\\
\hangindent=2.2em \quad \{\textcolor{pTx!72}{"event\_id":} \textbf{"E5"}, \textcolor{pTx!72}{"interactions":} [\\
\hangindent=2.2em \qquad \{\textcolor{pTx!72}{"id":} \textbf{"I8"}, \textcolor{pTx!72}{"t":} \textbf{23.0}, \textcolor{pTx!72}{"boxes":} [\\
\hangindent=2.2em \qquad\quad \textcolor{qRed}{\{\textcolor{pTx!72}{"role":} \textbf{"hand"}, \textcolor{pTx!72}{"track":} \textbf{"H1"}, \textcolor{pTx!72}{"box":} [570, 520, 680, 650]\}},\\
\hangindent=2.2em \qquad\quad \textcolor{qRed}{\{\textcolor{pTx!72}{"role":} \textbf{"target"}, \textcolor{pTx!72}{"track":} \textbf{"O1"}, \textcolor{pTx!72}{"box":} [460, 450, 580, 600]\}}]\},\\
\hangindent=2.2em \qquad \{\textcolor{pTx!72}{"id":} \textbf{"I9"}, \textcolor{pTx!72}{"t":} \textbf{47.0}, \textcolor{pTx!72}{"boxes":} [\\
\hangindent=2.2em \qquad\quad \textcolor{qRed}{\{\textcolor{pTx!72}{"role":} \textbf{"hand"}, \textcolor{pTx!72}{"track":} \textbf{"H1"}, \textcolor{pTx!72}{"box":} [570, 520, 680, 650]\}},\\
\hangindent=2.2em \qquad\quad \textcolor{qRed}{\{\textcolor{pTx!72}{"role":} \textbf{"tool"}, \textcolor{pTx!72}{"track":} \textbf{"O3"}, \textcolor{pTx!72}{"box":} [550, 400, 650, 500]\}},\\
\hangindent=2.2em \qquad\quad \textcolor{qRed}{\{\textcolor{pTx!72}{"role":} \textbf{"target"}, \textcolor{pTx!72}{"track":} \textbf{"O1"}, \textcolor{pTx!72}{"box":} [460, 450, 580, 600]\}}]\},\\
\hangindent=2.2em \qquad \{\textcolor{pTx!72}{"id":} \textbf{"I10"}, \textcolor{pTx!72}{"t":} \textbf{56.0}, \textcolor{pTx!72}{"boxes":} [\\
\hangindent=2.2em \qquad\quad \textcolor{qRed}{\{\textcolor{pTx!72}{"role":} \textbf{"hand"}, \textcolor{pTx!72}{"track":} \textbf{"H1"}, \textcolor{pTx!72}{"box":} [570, 520, 680, 650]\}},\\
\hangindent=2.2em \qquad\quad \textcolor{qRed}{\{\textcolor{pTx!72}{"role":} \textbf{"target"}, \textcolor{pTx!72}{"track":} \textbf{"O1"}, \textcolor{pTx!72}{"box":} [460, 450, 580, 600]\}}]\}]\}],\\
\hangindent=2.2em \textcolor{cHow}{\textbf{"relations"}}: [\textcolor{qRed}{\{\textcolor{pTx!72}{"source":} \textbf{"E2"}, \textcolor{pTx!72}{"target":} \textbf{"E3"}, \textcolor{pTx!72}{"type":} "enables"\}}, \textcolor{qRed}{\{\textcolor{pTx!72}{"source":} \textbf{"E3"}, \textcolor{pTx!72}{"target":} \textbf{"E4"}, \textcolor{pTx!72}{"type":} "causes"\}}, \textcolor{qRed}{\{\textcolor{pTx!72}{"source":} \textbf{"E4"}, \textcolor{pTx!72}{"target":} \textbf{"E5"}, \textcolor{pTx!72}{"type":} "enables"\}}],\\
\hangindent=2.2em \textcolor{cAns}{\textbf{"answer"}}: "If the mug were not put into the microwave, the contents of the mug would not be heated, and the subsequent actions of removing the mug from the microwave, stirring, and lifting it would not occur in the same manner."\}
\end{minipage}}
\end{minipage}\hfill
\begin{minipage}[t]{0.493\textwidth}{\sffamily\fontsize{6.6}{7.6}\selectfont\textbf{\textcolor{qBlue}{SPARK-G (2B)}}}\\[1pt]
\fcolorbox{pDk}{white}{\begin{minipage}[t]{0.975\linewidth}\raggedright\sffamily\fontsize{5.5}{6.45}\selectfont\color{pTx}
\hangindent=2.2em \{\textcolor{cWhen}{\textbf{"evidence"}}: [\\
\hangindent=2.2em \quad \textcolor{qBlue}{\{\textcolor{pTx!72}{"id":} \textbf{"E1"}, \textcolor{pTx!72}{"start":} \textbf{6.4}, \textcolor{pTx!72}{"end":} \textbf{9.0}, \textcolor{pTx!72}{"description":} "Another person's hand places the blue ceramic mug into the black microwave oven."\}},\\
\hangindent=2.2em \quad \{\textcolor{pTx!72}{"id":} \textbf{"E2"}, \textcolor{pTx!72}{"start":} \textbf{19.0}, \textcolor{pTx!72}{"end":} \textbf{21.0}, \textcolor{pTx!72}{"description":} "Another person's hand opens the door of the black microwave oven, reaches inside, and removes the blue ceramic mug."\},\\
\hangindent=2.2em \quad \textcolor{qBlue}{\{\textcolor{pTx!72}{"id":} \textbf{"E3"}, \textcolor{pTx!72}{"start":} \textbf{45.0}, \textcolor{pTx!72}{"end":} \textbf{51.0}, \textcolor{pTx!72}{"description":} "Another person's hand uses the metal butter knife to stir the contents of the blue ceramic mug."\}},\\
\hangindent=2.2em \textcolor{cWhat}{\textbf{"grounding"}}: [\\
\hangindent=2.2em \quad \{\textcolor{pTx!72}{"event\_id":} \textbf{"E1"}, \textcolor{pTx!72}{"interactions":} [\\
\hangindent=2.2em \qquad \{\textcolor{pTx!72}{"id":} \textbf{"I1"}, \textcolor{pTx!72}{"t":} \textbf{7.5}, \textcolor{pTx!72}{"boxes":} [\\
\hangindent=2.2em \qquad\quad \textcolor{qBlue}{\{\textcolor{pTx!72}{"role":} \textbf{"hand"}, \textcolor{pTx!72}{"box":} [533, 500, 838, 1000]\}},\\
\hangindent=2.2em \qquad\quad \textcolor{qBlue}{\{\textcolor{pTx!72}{"role":} \textbf{"target"}, \textcolor{pTx!72}{"box":} [444, 347, 625, 625]\}}]\}]\},\\
\hangindent=2.2em \quad \{\textcolor{pTx!72}{"event\_id":} \textbf{"E2"}, \textcolor{pTx!72}{"interactions":} [\\
\hangindent=2.2em \qquad \{\textcolor{pTx!72}{"id":} \textbf{"I1"}, \textcolor{pTx!72}{"t":} \textbf{19.5}, \textcolor{pTx!72}{"boxes":} [\\
\hangindent=2.2em \qquad\quad \{\textcolor{pTx!72}{"role":} \textbf{"hand"}, \textcolor{pTx!72}{"box":} [388, 436, 836, 1000]\},\\
\hangindent=2.2em \qquad\quad \{\textcolor{pTx!72}{"role":} \textbf{"target"}, \textcolor{pTx!72}{"box":} [79, 7, 783, 990]\}]\},\\
\hangindent=2.2em \qquad \{\textcolor{pTx!72}{"id":} \textbf{"I2"}, \textcolor{pTx!72}{"t":} \textbf{19.5}, \textcolor{pTx!72}{"boxes":} [\\
\hangindent=2.2em \qquad\quad \{\textcolor{pTx!72}{"role":} \textbf{"hand"}, \textcolor{pTx!72}{"box":} [614, 461, 930, 1000]\},\\
\hangindent=2.2em \qquad\quad \{\textcolor{pTx!72}{"role":} \textbf{"target"}, \textcolor{pTx!72}{"box":} [263, 0, 955, 865]\}]\}]\},\\
\hangindent=2.2em \quad \{\textcolor{pTx!72}{"event\_id":} \textbf{"E3"}, \textcolor{pTx!72}{"interactions":} [\\
\hangindent=2.2em \qquad \{\textcolor{pTx!72}{"id":} \textbf{"I1"}, \textcolor{pTx!72}{"t":} \textbf{46.9}, \textcolor{pTx!72}{"boxes":} [\\
\hangindent=2.2em \qquad\quad \{\textcolor{pTx!72}{"role":} \textbf{"hand"}, \textcolor{pTx!72}{"box":} [591, 481, 915, 1000]\},\\
\hangindent=2.2em \qquad\quad \{\textcolor{pTx!72}{"role":} \textbf{"tool"}, \textcolor{pTx!72}{"box":} [542, 485, 640, 619]\},\\
\hangindent=2.2em \qquad\quad \textcolor{qBlue}{\{\textcolor{pTx!72}{"role":} \textbf{"target"}, \textcolor{pTx!72}{"box":} [494, 423, 620, 652]\}}]\}]\}],\\
\hangindent=2.2em \textcolor{cHow}{\textbf{"relations"}}: [\{\textcolor{pTx!72}{"source":} \textbf{"E1"}, \textcolor{pTx!72}{"target":} \textbf{"E2"}, \textcolor{pTx!72}{"type":} "enables"\}, \textcolor{qBlue}{\{\textcolor{pTx!72}{"source":} \textbf{"E1"}, \textcolor{pTx!72}{"target":} \textbf{"E3"}, \textcolor{pTx!72}{"type":} "enables"\}}, \{\textcolor{pTx!72}{"source":} \textbf{"E2"}, \textcolor{pTx!72}{"target":} \textbf{"E3"}, \textcolor{pTx!72}{"type":} "enables"\}],\\
\hangindent=2.2em \textcolor{cAns}{\textbf{"answer"}}: \textcolor{qBlue}{"Without the mug being placed in the microwave, it would not be heated, and the contents would remain cold. This would prevent the later stirring and mixing of the contents inside the mug."}\}
\end{minipage}}
\end{minipage}
\caption{\textbf{Outputs of Qwen3.6-27B and \method (2B) on one gold-labeled question} (\cref{tab:main}), one of the 15 questions on which \method (2B) passes every grounding check of GCR$_{STD}$ and answers correctly while Qwen3.6-27B answers correctly but fails the key-event or interaction check (\method (2B) passes every check on 25 of the \EvGold{} questions). Frames: the moments of the predicted interactions, with the boxes of Qwen3.6-27B (red), of \method (blue) and of the ground truth (white dashed). Outputs: the reply of Qwen3.6-27B, and the predicted evidence and answer of \method written in the same format (no track identifiers). Red: elements of Qwen3.6-27B that do not match the ground truth. Blue: elements of \method that match it. Boxes as $[x_{\min},y_{\min},x_{\max},y_{\max}]$ normalized to 0--1000.}
\label{fig:supp_qual}
\end{figure*}

\paragraph{(1) Key events.}
\method localizes the key events more precisely than any other system.
Its F1@.3 is .443, against .099 for its backbone Qwen3.5-2B, .256 for joint SFT, at most .258 for the video and causal reasoning specialists and .399 for GPT-6.1 Sol, and it stays the highest on all evaluation questions (.428).
It also recovers more key events than joint SFT and every specialist at each number of key events, 32.2\% on questions with $\geq$3 key events against 27.4\% and at most 21.1\% (\cref{tab:supp_eventrec}).
\method thus addresses the partial localization of the key events.
The remaining problem is completeness on questions with many key events: \method predicts 1.6 events per question, recovers fewer key events there than GPT-6.1 Sol and Gemini 3.8 Flash (49.9\% and 40.4\%), and its recall decreases with the number of key events as for the other systems (\cref{tab:supp_eventctrl}, \cref{sec:supp_events}).

\paragraph{(2) Participants, roles and association.}
Training on our training split reduces role confusion and association errors, and \method{} adds mainly the localization of the participants.
\method{} gives another role to 11.2\% of the localized tools and targets (8.4\% for the 9B model, the lowest of all systems), against 14.4\% for joint SFT, 9.7--13.9\% for the proprietary models and 18.6--37.3\% for the open-source MLLMs (\cref{tab:supp_where_all}, \cref{sec:supp_roles}).
It places 3.7\% of its correctly localized participants in another interaction (3.5\% for the 9B model), more than joint SFT (2.3\%) and Gemma~4 (3.1\%) and fewer than the other systems (4.7--7.4\%, \cref{sec:supp_inter}).
Given a localized participant, \method{} and joint SFT assign the correct role (.914 and .908) and the correct interaction (.982 and .980) at nearly the same rate (\cref{tab:ablation}), so the rise of \Lthree{} from .129 to .184 comes mainly from localizing more participants (\Lone{} from .145 to .205).
Its \Lthree{} is far above LLaVA-ST (.012) and Cosmos-Reason2 (.005), second only to GPT-6.1 Sol (.213) and the highest on all evaluation questions (.163). In the ablations, the role-specific slots and the self-attention among the queries each raise \Lthree{} (\cref{sec:supp_ablation}).
The remaining problem is localization: \method localizes 19.2\% of the ground-truth participants against 31.4\% for GPT-6.1 Sol, and its \Lthree{} rises to .442 when the ground-truth events are given (\cref{sec:supp_ablation}).

\paragraph{(3) Dependencies.}
\method recovers the annotated dependencies best, and on the pairs it localizes it separates dependent from dependency-negative pairs above chance.
Its Edge F1 is .116, against .012 for its backbone, .030 for joint SFT, at most .014 for the causal reasoning models and at most .058 for the proprietary models, and it is also the highest on all evaluation questions (.077).
Its Macro-F1 (.101) is below that of Qwen3.6-27B (.215) and GPT-6.1 Sol (.235) because Macro-F1 also scores the \textit{before} label and counts every ground-truth pair with an unlocalized event as an error, and \method{} localizes both events of 144 of the 1,180 pairs against 335 and 356. Edge F1 scores only the \textit{causes} and \textit{enables} edges.
On the pairs it localizes (12\% of the ground-truth pairs), it predicts an edge for 69\% of the dependent and 46\% of the dependency-negative pairs (balanced accuracy 62\%). Qwen3.6-27B and GLM-4.6V-Flash also separate them on their own localized pairs (61\% and 59\%), whereas Causal-VLM and MECD+ do not (50\% and 51\%). Since each system is scored on the pairs it localizes, these rates describe each system and do not rank them (\cref{tab:supp_dep_diag}b, \cref{sec:supp_dep_disc}).
\method thus improves on the dependency failure of Finding~4.
The 9B model scores lower on dependencies (Edge F1 .081) largely because it localizes both events of fewer ground-truth pairs (111 against 144). With the END bias chosen for the 2B model, it predicts at least two events for 645 questions against 915, and its dependency accuracy on the localized pairs is similar (.56 against .60).
The remaining problem is the edges it still predicts for dependency-negative pairs, so the separation is not yet complete.

\paragraph{(4) Grounded answers.}
The correct answers of \method are the least often spurious.
Its event-level spurious share SS$_T$ is .54, with the 9B model the lowest of all systems, against .64 for GPT-6.1 Sol, .68 for joint SFT and at least .81 for the specialists (\cref{tab:supp_main_full}), and on all evaluation questions it has the lowest spurious share at every level (SS$_T$ .55, SS$_{ST}$ .81, SS$_{STD}$ .87).
Task training raises GCR$_T$ from .027 to .056 and GCR$_{ST}$ from .001 to .025 ($+$.029 [$+$.016, $+$.043] and $+$.024 [$+$.017, $+$.032] against its backbone), above every specialist (at most .032 and .002, \cref{tab:supp_main_full}). Against joint SFT (.044 and .021), the differences are not significant ($+$.012 [$-$.001, $+$.026] and $+$.004 [$-$.005, $+$.014]), and the 9B model (.084 and .033) is comparable to Gemini 3.8 Flash (.085 and .035).
Its answers also change when its predicted evidence is replaced (\cref{sec:exp_answer}).
\method thus raises the grounded correct rates over its backbone, but not significantly over joint SFT.
The remaining problem is answer accuracy: \method answers fewer questions correctly than GPT-6.1 Sol (.123 and .184 against .330), so its grounded correct rates stay below those of GPT-6.1 Sol, and its GCR$_{STD}$ (.015) is close to that of joint SFT (.018, \cref{sec:exp_learning}).
Its accuracy is lowest on hypothetical questions, none of which it answers correctly (the 9B model answers 10 of 115). All of them come from one EgoTaskQA template that asks for the state of an object after a hypothetical action, which does not occur in the training split, and the answering module answers them with an action (for example, ``open drawer'') instead of a state.

\subsection{Ablations of the grounding module}
\label{sec:supp_ablation}
\Cref{tab:ablation} changes one component at a time, with the same data, schedule and loss weights as the full module, on the \HVGold{} gold-labeled evaluation questions of \cref{tab:main}. Differences are paired, with 95\% bootstrap CIs over source videos.
\emph{Joint training} gives the largest \Lthree{} gain of the tested components. Training the interaction head on the frozen backbone after the event and dependency losses (staged) leaves mTIoU nearly unchanged but lowers every interaction measure.
Against staged training, joint training reaches \Lthree{} 0.184 against 0.120 ($+0.064$ $[+0.053,+0.076]$), localizes more participants (\Lone{} $+0.056$ $[+0.045,+0.068]$), and raises the conditional role accuracy from .848 to .914 ($+0.066$ $[+0.042,+0.093]$) and the association accuracy from .949 to .982 ($+0.033$ $[+0.017,+0.050]$). Its mTIoU is 0.271 against 0.269 (difference $[-0.011,+0.014]$).
\emph{Query self-attention} improves the interactions and leaves mTIoU unchanged.
It raises \Lthree{} from 0.162 to 0.184 ($+0.022$ $[+0.012,+0.033]$), \Lone{} by $+0.017$ $[+0.006,+0.027]$, the conditional role accuracy from .891 to .914 ($+0.023$ $[+0.004,+0.041]$) and the association accuracy from .963 to .982 ($+0.019$ $[+0.008,+0.030]$), with mTIoU $-0.002$ $[-0.014,+0.009]$.
\emph{Role-specific slots} improve the localization and the roles of the participants but not their association.
Against the same number of shared slots with a three-way role classifier, they raise \Lthree{} from 0.153 to 0.184 ($+0.031$ $[+0.021,+0.043]$), \Lone{} by $+0.029$ $[+0.017,+0.040]$ and the conditional role accuracy from .890 to .914 ($+0.024$ $[+0.003,+0.045]$), while the association accuracy changes little ($+0.009$ $[-0.002,+0.021]$).
All three components also raise the dependency Macro-F1, by 0.034 to 0.064 with each CI above zero.
These components share the LoRA of the backbone with the dependency head, so changing them also changes the dependency predictions. Each variant is trained once, and the CIs cover the sampling of the test questions but not the variation between training runs, so differences of this size may partly reflect training variance.

\paragraph{Ground-truth events.}
Given the ground-truth events, the same heads more than double their interaction score, with role and association accuracy similar to those with the retrieved events.
\Lthree{} reaches 0.442, 2.4 times its value with the retrieved events ($+0.258$ $[+0.235,+0.284]$), and the dependency Macro-F1 0.410 (\cref{tab:ablation}, bottom), and about 90\% and 98\% of the localized participants keep the right role and interaction. The 9B module reaches \Lthree{} 0.442 and Macro-F1 0.384.

\subsection{Answer sensitivity to evidence replacement}
\label{sec:exp_answer}
The answering module of \method sees only the question and the evidence predicted by the grounding module. To test whether its answers depend on this evidence, we keep the question and the trained modules fixed and replace the evidence at test time (\cref{tab:answer}).
\emph{Other source video}: the predicted evidence of a gold question cut from a different source video (questions are asked about clips, and several clips can come from one source video), chosen deterministically as the one with the closest number of predicted events, then the closest total event duration. \emph{Same dataset, other source video}: the same rule restricted to questions of the same source dataset, so that the evidence shows a similar kind of activity. \emph{No evidence}: the question with an empty evidence package.
Replacing the evidence lowers accuracy by 2.9 points $[1.6, 4.3]$ (other source video) and 2.7 points $[1.5, 4.0]$ (same dataset), and removing it by 4.5 points $[3.0, 6.2]$. With replaced evidence the module keeps its answer on only a quarter of the questions (24.3\% and 25.6\%), and it almost never gives the answer it gives to the donor question (0.5\% and 1.3\%), so the question still conditions the answer. The drop is largest on EgoPlan-Bench2 (8.3 points $[4.8, 12.2]$), whose questions ask for the next step after the observed actions, and not significant on EgoTempo and ProcObject-10K.
The drop is small in absolute terms because the module answers only 201 questions correctly, but with the evidence of another source video 99 of these 201 answers become wrong.
The answers of \method therefore depend on its predicted evidence.

\begin{table}[t]
\centering
\footnotesize
\setlength{\tabcolsep}{2.5pt}
\resizebox{\linewidth}{!}{%
\begin{tabular}{@{}l c c c c@{}}
\toprule
\begin{tabular}[b]{@{}l@{}}Evidence given to\\the answering module\end{tabular} & Acc & $\Delta$ [95\% CI] & \begin{tabular}[b]{@{}c@{}}Same\\answer\end{tabular} & \begin{tabular}[b]{@{}c@{}}Flips\\($+$/$-$)\end{tabular} \\
\midrule
Own predicted evidence (\cref{tab:main}) & .123 & -- & -- & -- \\
Other source video & .093 & $-.029$ {\scriptsize$[-.043,-.016]$} & .243 & 51 / 99 \\
Same dataset, other source video & .096 & $-.027$ {\scriptsize$[-.040,-.015]$} & .256 & 45 / 89 \\
No evidence & .078 & $-.045$ {\scriptsize$[-.062,-.030]$} & .089 & 59 / 133 \\
\bottomrule
\end{tabular}}
\caption{\textbf{Answer sensitivity to evidence replacement.} The answering module of \method (2B) with its own predicted evidence and with the evidence replaced at test time. The question and both modules are unchanged. \EvGold{} gold-labeled questions from 627 source videos. $\Delta$: paired difference to the answers of \cref{tab:main} with the own evidence, bootstrap over source videos 95\% CI. Same answer: share of questions whose answer is unchanged. Flips: wrong answers that become correct / correct answers that become wrong. GPT-OSS-120B judge.}
\label{tab:answer}
\end{table}

\begin{table}[t]
\centering
\footnotesize
\setlength{\tabcolsep}{2.5pt}
\resizebox{\linewidth}{!}{%
\begin{tabular}{@{}l cc ccc c@{}}
\toprule
& \multicolumn{2}{c}{Events} & \multicolumn{3}{c}{Interactions} & Dep. \\
\cmidrule(lr){2-3}\cmidrule(lr){4-6}\cmidrule(lr){7-7}
Grounding module (2B) & mTIoU & F1@.3 & \Lthree & Role & Assoc. & Macro-F1 \\
\midrule
\method & \underline{\TM{A}{mTIoU}} & \textbf{\TM{A}{F1_30}} & \textbf{\TM{A}{L3}} & \textbf{.914} & \textbf{.982} & \textbf{\TM{A}{MacroF1}} \\
w/o query self-attention & \textbf{.273} & \underline{.428} & \underline{.162} & .891 & .963 & \underline{.067} \\
shared role slots & .254 & .398 & .153 & .890 & .973 & .037 \\
staged training & .269 & .421 & .120 & .848 & .949 & .042 \\
joint SFT & \TM{M1F}{mTIoU} & \TM{M1F}{F1_30} & \TM{M1F}{L3} & \underline{.908} & \underline{.980} & \TM{M1F}{MacroF1} \\
\midrule
\multicolumn{7}{@{}l}{\textit{Given the ground-truth events}} \\
\method (2B) & -- & -- & .442 & .904 & .984 & .410 \\
\method (9B) & -- & -- & .442 & .911 & .959 & .384 \\
\bottomrule
\end{tabular}}
\caption{\textbf{Ablations} on the \HVGold{} gold-labeled evaluation questions (\cref{tab:main} protocol; the \method and joint SFT rows are those of \cref{tab:main}). One change at a time, same data, schedule and loss weights. Role: share of the localized participants with the right role ($\mathrm{TP}_2/\mathrm{TP}_1$); Assoc.: share of those also associated with the right interaction ($\mathrm{TP}_3/\mathrm{TP}_2$). Shared role slots: the same number of participant slots without role specialization, each with a three-way role classifier, matched jointly across roles. Staged training: the LoRA is trained with the event (segmentation, retrieval and description) and dependency losses only, then the interaction head on the frozen backbone. Joint SFT: the same backbone with LoRA on the same modules, trained for the same number of epochs on the same questions to write the events, the interactions with role boxes, the dependencies and the answer as one JSON object from the video (the JSON prompt of \cref{tab:main}). Bottom: the heads read the ground-truth key events instead of the retrieved ones (not ranked). Best in \textbf{bold}, second \underline{underlined}.}
\label{tab:ablation}
\end{table}

\subsection{Discussion on cost}
\label{sec:supp_cost}
We report what \method costs at inference and in training.
We compare it with one system of each kind in \cref{tab:main}: its backbone without task training (Qwen3.5-2B), the joint SFT baseline trained on the same data with the same backbone, the most accurate open-source MLLM (Qwen3.6-27B) and the most accurate video reasoning specialist (EgoThinker).

\paragraph{Inference.}
We time 150 gold-labeled questions, sampled before timing and stratified by source dataset and video duration (4.9--244.9\,s, median 33.8\,s as in the gold set), on NVIDIA H100 80GB GPUs. Every system is loaded once, warmed up on five other questions, and run with the settings of \cref{tab:main} at batch size 1. Latency is end-to-end per question, from video decoding to the answer; for \method it includes both modules: the event segmentation, key-event retrieval, event descriptions and interaction and dependency heads of the grounding module, the rendering of the evidence, and the answering module. GPU-seconds multiply the latency by the number of GPUs. No question failed.

\begin{table}[t]
\centering
\footnotesize
\setlength{\tabcolsep}{2.4pt}
\resizebox{\linewidth}{!}{%
\begin{tabular}{@{}l c cc c c ccc c@{}}
\toprule
& & \multicolumn{2}{c}{Latency (s)} & GPU-s & Train & & & & \\
\cmidrule(lr){3-4}
System & GPUs & median & p90 & per q. & GPU-h & Acc & GCR$_{ST}$ & GCR$_{STD}$ & \Lthree \\
\midrule
Qwen3.5-2B & 1 & 4.8 & 11.2 & 6.6 & -- & \TM{Base2B}{Acc} & \TM{Base2B}{GCR_ST} & \TM{Base2B}{GCR_STD} & \TM{Base2B}{L3} \\
Joint SFT (2B) & 1 & 4.1 & 6.4 & 4.8 & 60.5 & \TM{M1F}{Acc} & \TM{M1F}{GCR_ST} & \TM{M1F}{GCR_STD} & \TM{M1F}{L3} \\
\method (2B) & 1 & 22.7 & 43.4 & 27.8 & 269.3 & \TM{A}{Acc} & \TM{A}{GCR_ST} & \TM{A}{GCR_STD} & \TM{A}{L3} \\
\quad grounding module & & 17.7 & 39.5 & 23.1 & 244.1 & & & & \\
\quad answering module & & 3.5 & 7.9 & 4.7 & 25.2 & & & & \\
\midrule
Qwen3.6-27B & 2 & 18.8 & 28.0 & 40.7 & -- & \TM{Q36}{Acc} & \TM{Q36}{GCR_ST} & \TM{Q36}{GCR_STD} & \TM{Q36}{L3} \\
EgoThinker (7B) & 1 & 5.3 & 10.8 & 7.1 & -- & \TM{EgoThinker}{Acc} & -- & -- & -- \\
\bottomrule
\end{tabular}}
\caption{\textbf{Cost and performance.} Inference on 150 gold-labeled questions (H100 80GB, batch size 1, settings of \cref{tab:main}); GPU-s per question: mean of GPUs $\times$ latency. Train GPU-h: task-specific training on NVIDIA GH200 120GB GPUs (Slurm allocation $\times$ wall time); --: no EgoSPARK-specific training. For \method, the grounding-module training row includes the inference of the grounding module on the training questions of the answering module (53.6 GPU-h). Performance on the \EvGold{} gold-labeled questions (\cref{tab:main}). \method runs in eager PyTorch with Hugging Face Transformers, the other systems in vLLM.}
\label{tab:cost}
\end{table}

\method improves grounding over the joint SFT baseline at a higher inference cost, and it uses less GPU time than Qwen3.6-27B (\cref{tab:cost}).
It needs 27.8 GPU-seconds per question, 5.8 times the joint SFT baseline and 0.7 times Qwen3.6-27B.
Against joint SFT, \Lthree{} rises from \TM{M1F}{L3} to \TM{A}{L3}, Edge F1 from \TM{M1F}{EdgeF1} to \TM{A}{EdgeF1} and F1@0.3 from \TM{M1F}{F1_30} to \TM{A}{F1_30} at a similar accuracy (\TM{M1F}{Acc} against \TM{A}{Acc}).
Against Qwen3.6-27B, \method answers fewer questions correctly (\TM{A}{Acc} against \TM{Q36}{Acc}) but more of them with grounded evidence (GCR$_{ST}$ \TM{A}{GCR_ST} against \TM{Q36}{GCR_ST}), on one GPU instead of two (28\,GB peak memory).
The additional computation of \method thus goes into grounding the evidence.

Two factors affect the latency. \method runs in eager PyTorch, whereas the other systems use the vLLM engine, so part of its latency comes from the engine. Most of its time is spent in the grounding module, on writing the event descriptions (13.4 of 23.1\,s) and on reading the video three times, once each for event segmentation, key-event retrieval and the interaction and dependency heads. It therefore processes 47k visual tokens per question, against 15k for the systems that read the video once.

\paragraph{Training.}
All task-specific training ran on the same GPU type (NVIDIA GH200 120GB). \method needs 269.3 GPU-hours: 190.5 for the grounding module (16 GPUs, 11.9\,h), 53.6 to predict the evidence of the training questions of the answering module (16 GPUs, 3.4\,h) and 25.2 for the answering module (4 GPUs, 6.3\,h). %
The joint SFT baseline needs 60.5 GPU-hours (8 GPUs, 7.6\,h), 4.5 times less.

\section{Dataset documentation}
\label{sec:supp_datasheet}
We document EgoSPARK following the questions of datasheets for datasets~\cite{gebru2021datasheets}.

\paragraph{Motivation and intended use.}
EgoSPARK evaluates whether a model answers a causal question about an egocentric video \emph{and} grounds its answer in the observed evidence: the key events, the interactions with their participants and boxes, and the dependencies between the events (\cref{sec:bench_task}). It is intended for the evaluation and analysis of video MLLMs and grounding methods, and the automatically labeled training split supports the training of such methods. The labels describe hand--object manipulations only and are not intended for identifying or tracking people.

\paragraph{Composition.}
The evaluation split has \EvQ{} questions on \EvVideos{} source videos from \EvDatasets{} public datasets (EgoTaskQA, ProcObject-10K, EgoTempo and EgoPlan-Bench2), \EvGold{} of them with human-verified evidence (\cref{tab:bench_stats}). The training split has 25,894 questions on 4,624 source videos from 11 datasets, with automatic labels and no source video shared with the evaluation split. Each question has a capability label (\cref{sec:supp_capability}), and each label set has key events (interval and caption), the participants of each key event as named tracks with roles and boxes on a 0.5\,s grid, the interactions parsed from the captions, and the dependencies between the key events.

\paragraph{What is released.}
We release the annotations only: for each question, the source dataset, the identifier of the source question and the clip boundaries in the source video, the capability, the split and gold-label membership, and the evidence labels above, together with the evaluation code. %
The videos are not redistributed. They are obtained from the original datasets under their own terms, and a script cuts the clips from the source videos.
The annotations and the evaluation code will be publicly released upon publication. The annotations will be released under CC BY-NC 4.0, and the videos and source questions remain under the licenses of the original datasets.

\paragraph{Collection process.}
The labels are produced by the semi-automatic pipeline of \cref{sec:bench_pipeline}, with automatic stages 1--5 for every question (\cref{sec:supp_bench}) and human verification for the evaluation questions (\cref{sec:supp_hv}). The integrity checks of \cref{sec:bench_integrity} decide which questions enter the evaluation split (\cref{sec:supp_integrity}).

\paragraph{Annotators and annotation cost.}
Human verification was done in a web tool by trained annotators following the written guideline of \cref{tab:supp_hv_guide} after pilot reviews with feedback.
The final reviews of the \HVReviewed{} questions were made by 12 annotators. The annotators were undergraduate and graduate students at the authors' institution, majoring in computer science, statistics and related fields, and we used no crowdsourcing platform. Eight of them were recruited for this work and paid for their time; the other four are authors of this paper. All twelve used the same guideline and tool, which showed no output of the evaluated systems. A review took a median of 180\,s (interquartile range 69--349\,s): 64\,s for a pass, 201\,s for an edited question and 74--134\,s for the verdicts that remove a question (QA wrong, no evidence and cannot judge, \cref{sec:supp_hv}). All saved reviews, including the second reviews, amount to 282 annotator-hours.

\paragraph{Privacy and ethics.}
All videos come from public research datasets that were collected and released under their own consent and privacy procedures (\eg, the de-identification of Ego4D videos), and we record no new video. %
Our annotations add no personal information: a hand label marks a hand that takes part in a manipulation, without identifying the person, and the object names, captions and boxes describe manipulated objects.

\paragraph{Maintenance.}
The dataset will be hosted in a public repository with a contact address, given upon publication. Corrections of the labels will be released as new versions, and evaluation results will state the version they use.

\begin{figure*}[tbp]\centering
\includegraphics[width=\textwidth]{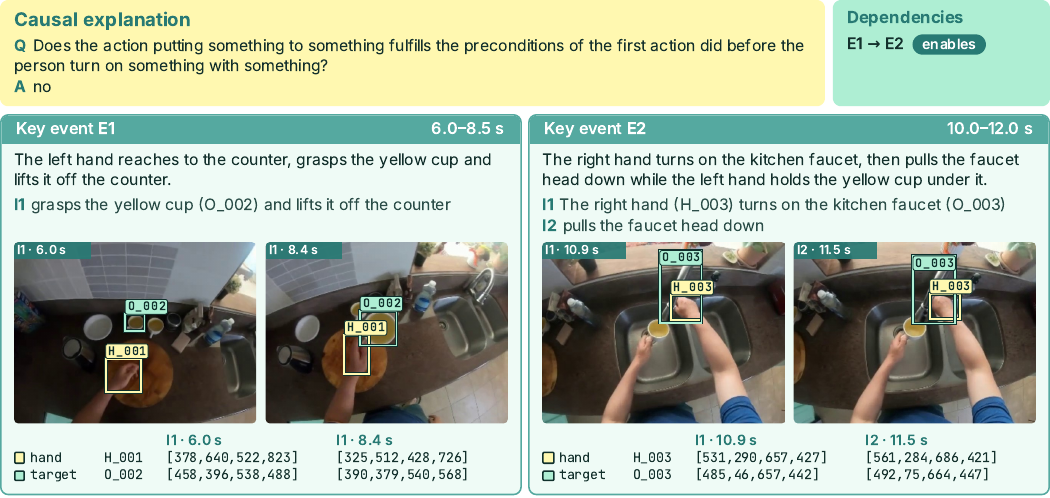}\\[7pt]
\includegraphics[width=\textwidth]{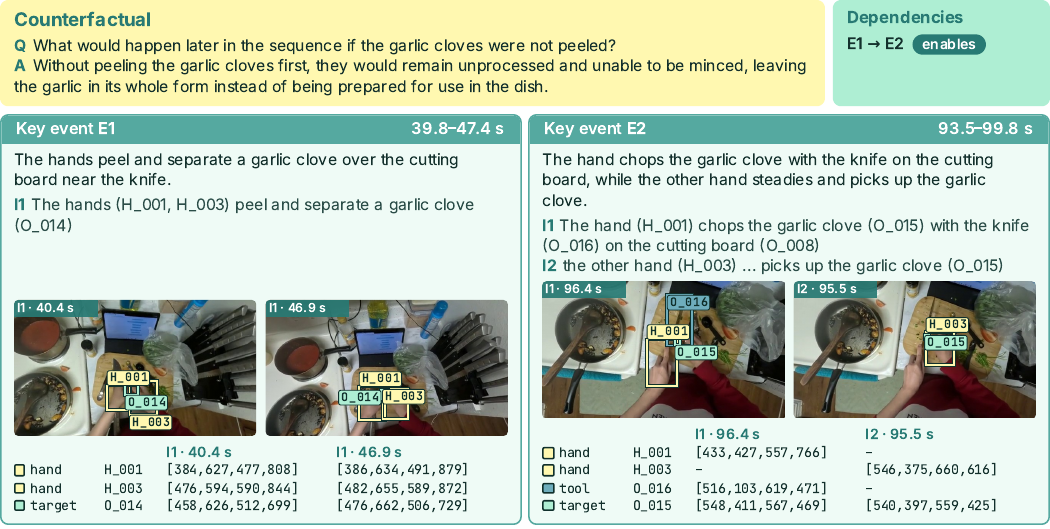}
\caption{\textbf{Gold-labeled examples by capability (1/3)}: causal explanation (top) and counterfactual (bottom). Each example shows the question (Q) and answer (A), the dependencies between its key events and one block per key event with its interval, description and interactions (I1, I2). The frames show moments of these interactions with the boxes of their participants (yellow: hand, blue: tool, green: target), which the table lists in the normalized coordinates of the annotation ($[0,1000]$, --: not a participant of the shown interaction).}
\label{fig:supp_cap_examples1}
\end{figure*}

\begin{figure*}[tbp]\centering
\includegraphics[width=\textwidth]{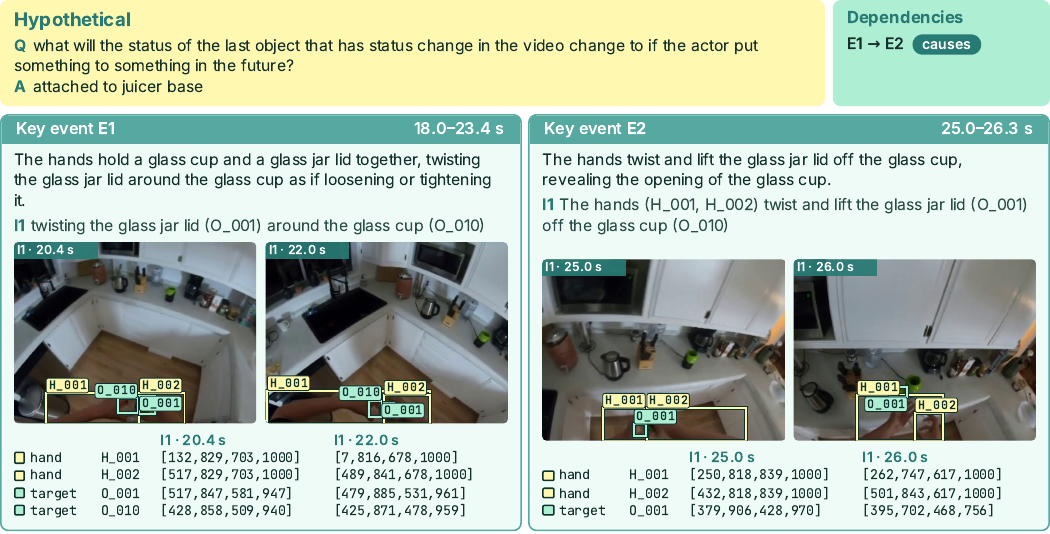}\\[7pt]
\includegraphics[width=\textwidth]{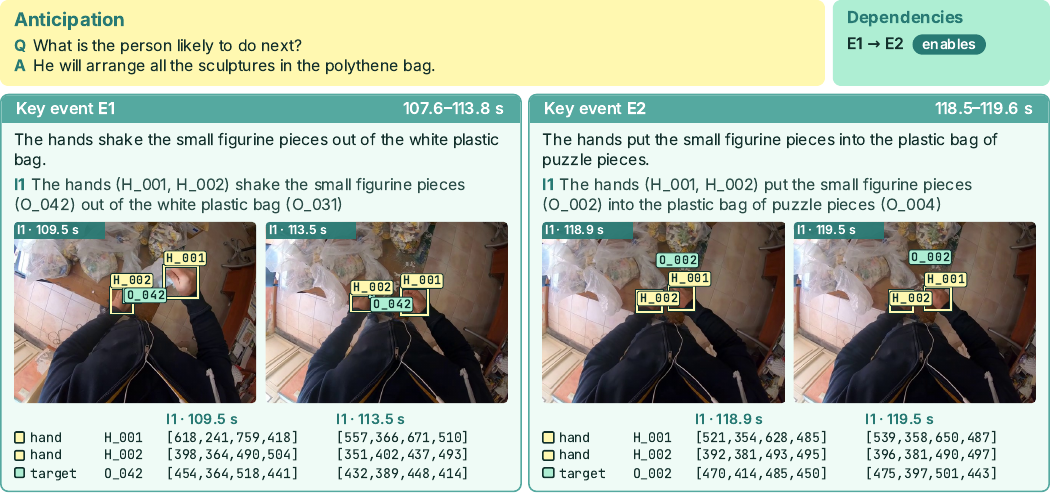}
\caption{\textbf{Gold-labeled examples by capability (2/3)}: hypothetical (top) and anticipation (bottom), shown as in \cref{fig:supp_cap_examples1}.}
\label{fig:supp_cap_examples2}
\end{figure*}

\begin{figure*}[tbp]\centering
\includegraphics[width=\textwidth]{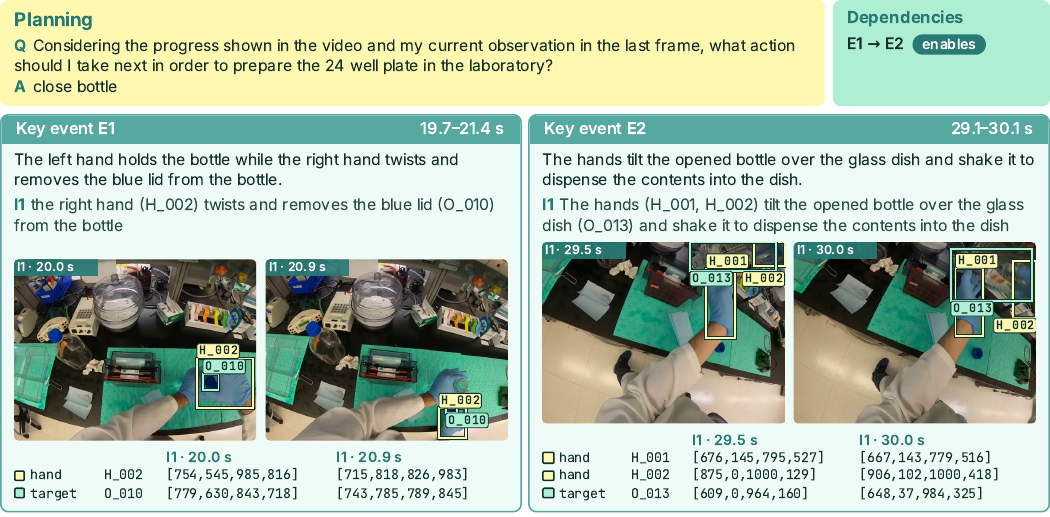}\\[7pt]
\includegraphics[width=\textwidth]{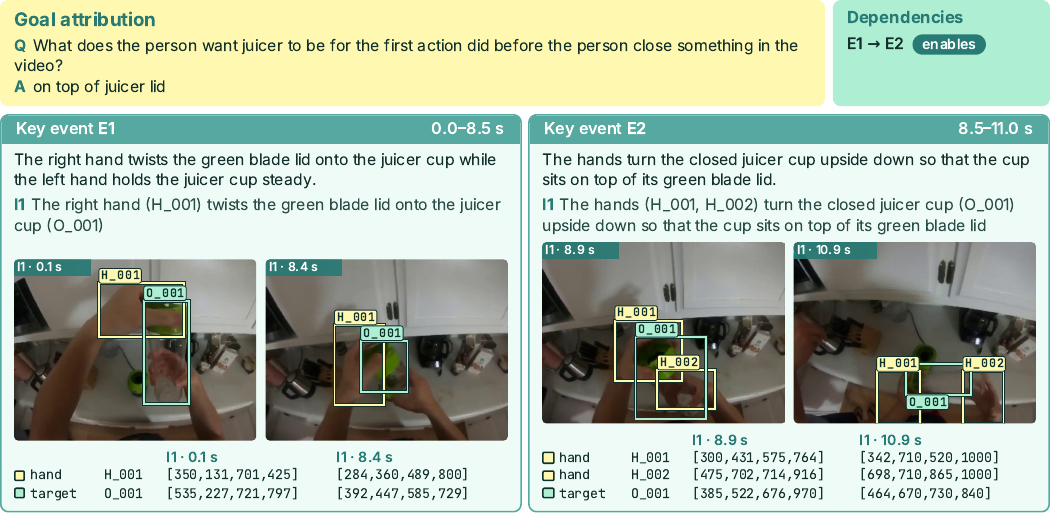}
\caption{\textbf{Gold-labeled examples by capability (3/3)}: planning (top) and goal attribution (bottom), shown as in \cref{fig:supp_cap_examples1}.}
\label{fig:supp_cap_examples3}
\end{figure*}

\paragraph{Examples by capability.}
\Cref{fig:supp_cap_examples1,fig:supp_cap_examples2,fig:supp_cap_examples3} show one gold-labeled question per capability with its verified labels as given in the released annotation.
For each capability in the order of \cref{sec:bench_filter}, the gold-labeled questions with a dependency are shuffled with seed 0, and the first question from the source dataset drawn least often so far is taken.
Each key event is shown at one or two annotated times of its interactions, those with the most participants boxed.

\clearpage

\FloatBarrier
\section*{LLM Usage Statement (supplementary material)}

This statement adds to the LLM Usage Statement of the main paper two uses of generative models in the construction of the training split (\cref{app:bench}) and clarifies which model judges the answers in C1.
\emph{Question construction.} Five training sources provide procedural annotations but no questions (HoloAssist, EgoPER, Ego-Exo4D, CaptainCook4D and Ego4D Goal-Step). Their questions and answers are built from the annotations with templates, and 3,731 of these questions are in the training split (\cref{tab:supp_train_sources}). GPT-5.4-mini assists for EgoPER, CaptainCook4D and Ego-Exo4D: it labels which dependencies of a task graph are necessary, extracts the purpose of a step from expert commentary, and selects, writes and checks the distractors of the multiple-choice questions. For Ego4D Goal-Step, Qwen3.6-27B estimates the prerequisite relations between the annotated substeps that determine the correct answer and the distractors. \bench{} asks these questions without the options, with the correct answer as the reference. The HoloAssist questions are built by rules alone. No evaluation question is built in this way.
\emph{Capability assignment.} For training question types that do not fix a capability (in EgoTools, CFG-Bench and HoloAssist), Qwen3.6-27B judges each question on four criteria, and fixed rules over these criteria give its capability. The capabilities of the evaluation questions are assigned per template or category by hand (\cref{sec:supp_capability}).
\emph{Judging in the integrity checks.} As in the main statement, GPT-OSS-120B judges the answers of C2 and C4. The answers of C1 are judged by GPT-5.4-mini, which also gives them (\cref{sec:supp_integrity}).

\end{document}